# ROGET'S THESAURUS
# AS A LEXICAL RESOURCE
# FOR NATURAL LANGUAGE PROCESSING

Mario Jarmasz

Thesis

submitted to the Faculty of Graduate and Postdoctoral

Studies

in partial fulfillment of the requirements

for the degree of Master of Computer Science

July, 2003

Ottawa-Carleton Institute for Computer Science

School of Information Technology and Engineering

University of Ottawa

Ottawa, Ontario, Canada

# ABSTRACT

*WordNet* proved that it is possible to construct a large-scale electronic lexical database on the principles of lexical semantics. It has been accepted and used extensively by computational linguists ever since it was released. Some of its applications include information retrieval, language generation, question answering, text categorization, text classification and word sense disambiguation. Inspired by *WordNet's* success, we propose as an alternative a similar resource, based on the 1987 Penguin edition of *Roget's Thesaurus of English Words and Phrases*.

Peter Mark Roget published his first *Thesaurus* over 150 years ago. Countless writers, orators and students of the English language have used it. Computational linguists have employed *Roget's* for almost 50 years in Natural Language Processing. Some of the tasks they have used it for include machine translation, computing lexical cohesion in texts and constructing databases that can infer common sense knowledge. This dissertation presents *Roget's* merits by explaining what it really is and how it has been used, while comparing its applications to those of *WordNet*. The NLP community has hesitated in accepting *Roget's Thesaurus* because a proper machine-tractable version was not available.

This dissertation presents an implementation of a machine-tractable version of the 1987 Penguin edition of *Roget's Thesaurus* – the first implementation of its kind to use an entire current edition. It explains the steps necessary for taking a machine-readable file and transforming it into a tractable system. This involves converting the lexical material into a format that can be more easily exploited, identifying data structures and designing classes to computerize the *Thesaurus*. *Roget's* organization is studied in detail and contrasted with *WordNet's*.

We show two applications of the computerized *Thesaurus*: computing semantic similarity between words and phrases, and building lexical chains in a text. The experiments are performed using well-known benchmarks and the results are compared to those of other systems that use *Roget's*, *WordNet* and statistical techniques. *Roget's* has turned out to be an excellent resource for measuring semantic similarity; lexical chains are easily built but more difficult to evaluate. We also explain ways in which *Roget's Thesaurus* and *WordNet* can be combined.

*To my parents, who are my most valued treasure.*

# TABLE OF CONTENTS

# TABLES

# FIGURES

# ACKNOWLEDGEMENTS

I would like to acknowledge the help that I have received in preparing this thesis.

Grateful thanks to:

- Dr. Stan Szpakowicz, for having been my guide and mentor on this quest for the holy grail of computational lexicography.
- Steve Crowdy, Denise McKeough and Martin Toseland from Pearson Education who helped us obtain the electronic copy of *Roget's Thesaurus* and answered our many questions.
- Dr. Peter Turney for having given me the idea to evaluate semantic similarity using synonymy questions and for his great insights on my research.
- Vivi Nastase for always being available to discuss my work, and her contribution to combining *Roget's Thesaurus* and *WordNet*.
- Terry Copeck for his kind encouragement and having prepared the stop list used in building lexical chains.
- Ted Pedersen and Siddharth Patwardhan for promptly developing and edge counting module for *WordNet* when requested.
- Tad Stach for collecting 200 *Reader's Digest Word Power* questions and helping to develop software to answer them using *Roget's Thesaurus*.
- Pierre Chrétien and Gilles Roy for implementing the graphical user interface to the electronic *Roget's Thesaurus*.
- Dr. Caroline Barrière, Dr. Ken Barker, Dr. Sylvain Delisle and Dr. Nathalie Japkowicz for their comments and feedback at the various stages of my thesis.

I would also like to thank the members of my committee:

- Dr. Jean-Pierre Corriveau, Carleton University
- Dr. Stan Matwin, University of Ottawa
- Dr. Stan Szpakowicz (thesis supervisor), University of Ottawa

# Peter Mark Roget and his Thesaurus
## 1779 – 1869

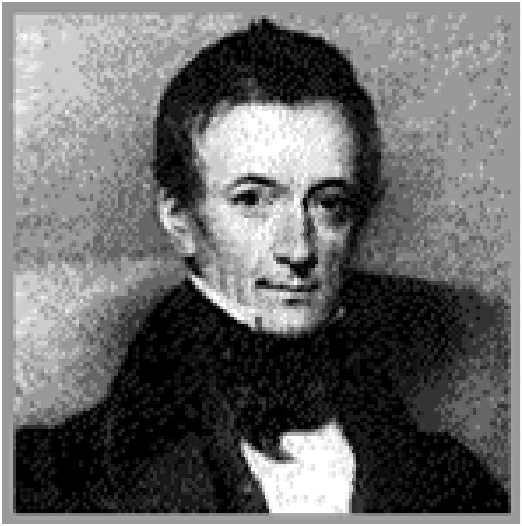

When Peter Mark Roget published his first *Thesaurus* over 150 years ago in 1852, he could not imagine that his work would be used to further research in human language technologies. He was born in London in 1779, the son of Jean Roget, a Genevan protestant pastor, and Catherine, the granddaughter of a French Huguenot who had fled to London after the revocation of the Edict of Nantes. Peter Mark's whole life prepared him to become the author of the *Treasury of Words*. This physician's preferences were anatomy and physiology, subjects that involve dissection and classification. As expressed by Kent (Chapman, 1992, p. vii): "A lifetime of secretaryships for several learned societies had thoroughly familiarized him with the need for clarity and forcefulness of expression. […] It was Roget's meticulous, precise way of looking at order, at plan and interdependence in animal economy that would eventually find expression in his unique and practical lexicographic experiment." As a professor, Roget prepared a notebook of lists of related words and phrases in various orders to help him express himself in the best possible way. Kirkpatrick (Kirkpatrick, 1998, p. xi) describes the manner in which Roget prepared the *Thesaurus* once he retired from professional life: "Now, in his seventies, he was able to draw on a lifetime's experience of lecturing, writing and editing to make these lists into a coherent system available for others to use. It took him four years, longer than he had thought, and required all his organizational skills and the meticulous attention to detail that had characterized his editing work. Not only did the *Thesaurus* utilize all Roget's competences, it also fulfilled a need for him: the need, in a society changing with frightening speed, where the old moral and religious order was increasingly in question, to reaffirm order, stability and unity, and through them the purpose of a universal, supernatural authority." Roget supervised about twenty–five editions and printings of the *Thesaurus* until he died at the age of ninety.[1]

---

[1] Image of Peter Mark Roget: http://www.toadshow.com.au/anne/images/roget_peter.gif

# 1   Introduction

## 1.1   Lexical Resources for Natural Language Processing

Natural Language Processing (NLP) applications need access to vast numbers of words and phrases. There are now many ways of obtaining large-scale lexicons: query words on the Internet (Turney, 2001), use large corpora such as the Wall Street Journal or the British National Corpus, as well as extract information from machine-readable dictionaries (Wilks *et al.*, 1996) or use electronic lexical databases such as *WordNet* (Fellbaum, 1998). Which of these methods is best? Our intuition presented in this dissertation is that computational linguists should extend and computerize the work of lexicographers, professionals who take concrete decisions about words, their senses and how they should be arranged. Computer programmers have been reckless in imagining that they can do without linguists. To investigate this conjecture I have implemented a large-scale electronic lexical knowledge base designed on the 1987 Penguin edition of *Roget's Thesaurus* (Kirkpatrick, 1987).

The first use of *Roget's Thesaurus* in NLP dates back to 1957 (Masterman); *WordNet*, a kind of thesaurus, has been available since 1991. These lexical resources have been used, among others, for the following applications:

- Machine Translation

    - proposing synonyms to improve word for word translations

- Information Retrieval

    - performing word sense disambiguation

    - expanding query terms by proposing synonyms

    - detecting the expected answer types of questions

- Information Extraction and Text Summarization

    - measuring semantic similarity between words

    - building lexical chains

Computational linguists have given presentations on the following subjects in two workshops about *WordNet* organized by the Association for Computational Linguistics (Harabagiu, 1998; Moldovan and Peters, 2001): Information Retrieval, Language Generation, Question Answering,

Text Categorization, Text Classification and Word Sense Disambiguation. Researchers displayed techniques for combining *WordNet* and Roget's at both workshops. I discuss some of these applications further in Chapter 2.

## 1.2   Electronic Lexical Knowledge Bases

Let's begin by defining the term Electronic Lexical Knowledge Base (*ELKB*). It is a model for a lexical resource, implemented in software, for classifying, indexing, storing and retrieving words with their senses and the connections that exist between them. It relies on a rich data repository to do so. This model defines explicit semantic relationships between words and word groups. It maps out an automatic process for building an electronic lexicon. It is electronic not only because it is encoded in a digital format, but rather because it is computer-usable, or tractable. The process for creating an *ELKB* presents all the steps involved in this task: from the preparation and acquisition of the lexical material, to defining the allowable operations on the various words and phrases. The use of a defined systematic approach to building an *ELKB* should reduce the irregularities usually contained in handcrafted lexicons. My thesis is a way of verifying this intuition.

My *ELKB* has been created from the machine readable text files with the contents of the 1987 *Penguin's Roget's Thesaurus*. It must maintain the information available in the printed *Thesaurus* while it is put in a tractable format.  Going from readable to tractable involves cleaning up and re-formatting the original files, deciding what services the *ELKB* should offer and implementing those services.

## 1.3   An Introduction to *Roget's Thesaurus*

*Roget's Thesaurus* is sometimes described as a dictionary in reverse (Wilks *et al.*, 1996, p. 65). According to Roget, it is "… a collection of the words it [the English language] contains and of the idiomatic combinations peculiar to it, arranged, not in alphabetical order as they are in a Dictionary, but according to the *ideas* which they express" (Roget, 1852). The *Thesaurus* is a catalogue of semantically similar words and phrases, divided into nouns, verbs, adjectives, adverbs and interjections. A phrase in *Roget's* is not one in the grammatical sense, but rather a collocation or an idiom, for example: `fatal gift`, `poisoned apple`, or `have kissed the Blarney Stone`. The reader perceives implicit semantic relations between groups of similar words. This resembles how Miller describes *WordNet* as lexical information organized by word meanings, rather than word forms (Miller, 1990). In *WordNet*, English nouns, verbs, adjectives and adverbs are organized into sets of near synonyms, called synsets, each representing a lexicalized concept. Semantic relations serve as links between the synsets.

A major strength of *Roget's Thesaurus* is its unique system of classification "of the ideas which are expressible by language" (Roget, 1852). The original system was organized in six classes: *Abstract relations*, *Space*, *Material World*, *Intellect*, *Volition, Sentient and Moral Powers*. Roget devised this system in the following way: "I have accordingly adopted such principles of arrangement as appeared to me to be the simplest and most natural, and which would not require, either for their comprehension or application, any disciplined acumen, or depth of metaphysical or antiquarian lore" (*ibid.*). Within these six classes are sections, and under the sections are almost 1,000 heads. This system of classification has withstood the test of time remarkably well, hardly changing in 150 years, a tribute to the robustness of *Roget's* design.

*Roget's Thesaurus* is the creative wordsmith's instrument, helping clarify and give shape to one's thoughts. "The assistance it gives is that of furnishing on every topic a copious store of words and phrases, adapted to express all the recognizable shades and modifications of the general idea under which those words and phrases are arranged" (*ibid.*). *Roget's* is indeed, as its name's Greek etymology indicates, a vast treasure house of English words and phrases.

### 1.3.1   The *Roget's Electronic Lexical Knowledge Base*

The objective of this thesis is to produce a machine-tractable version of the 1987 Penguin edition of *Roget's Thesaurus* (Kirkpatrick, 1987) — the first implementation of an *ELKB* that uses an entire current edition. The *FACTOTUM* semantic network (Cassidy, 1996, 2000) is the first implementation of a knowledge base derived from a version of *Roget's Thesaurus*, using the 1911 edition that is publicly available from the Project Gutenberg Web Site (Hart, 1991). A good *ELKB* should have a large, modern vocabulary, a simple way of identifying the different word senses, a clear classification system, and usage frequencies, several ways of grouping words and phrases to represent concepts, explicit links between the various units of meaning and an index of all words and phrases in the resource. It should also contain a lexicon of idiomatic expressions and proper nouns. Definitions of the words and phrases, as well as subcategorization information, akin to what can be found in a learner's dictionary, would also be beneficial. The *ELKB* constructed in the course of this research is not ideal, but it is sufficient to demonstrate that *Roget's* is a useful and interesting resource.

## 1.4   Goals of this Thesis

The goal of this thesis is to investigate the usefulness of *Roget's Thesaurus* for NLP. I expect that it will be an effective alternative to *WordNet*. To achieve this goal, *Roget's* must be first computerized, evaluated and applied to some interesting tasks. The quantitative and qualitative evaluation constantly uses *WordNet 1.7.1* as a benchmark. Other more cursory comparisons are made with other lexical resources.

This treasure of the English language is exploited to create a new resource for computational linguists. The goal is to computerize the *Thesaurus*: create a machine-tractable lexical knowledge base, and represent in it the explicit, and some of the implicit, relationships between words. I have performed experiments using the system for measuring the semantic similarity between words and building lexical chains. This dissertation also presents steps for combining *Roget's* and *WordNet*.

## 1.5   Organization of the Thesis

Chapter 2 of the thesis gives an overview of the role of thesauri in NLP, discussing how such resources benefit research. The various versions of *Roget's Thesaurus* and *WordNet* demand and receive special attention.

Chapter 3 presents the details of the design and implementation of the *ELKB*. It explains all of the necessary steps for transforming the *Roget's* text files into a machine tractable format. These steps discuss the necessary functionality of such a system. The construction of the *ELKB* fulfills the first goal of this thesis.

Chapter 4 presents a measure of semantic similarity between words and phrases using the *Roget's ELKB*. It presents a semantic distance measure and evaluates it using a few typical tests. I perform a comparison to *WordNet* based measures and other statistical techniques.

Chapter 5 presents an implementation of lexical chain construction using the *Roget's ELKB*. It discusses in detail every design decision and includes a comparison to lexical chains built by hand using *WordNet*.

Chapter 6 discusses how to combine *Roget's Thesaurus* and *WordNet*. It presents steps on how to link the senses of the words and phrases included in both resources as well as how to add explicit semantic relations to the *Thesaurus*.

Chapter 7 gives a summary of the thesis, discusses problems, and presents future work to be done in improving the *Roget ELKB* and avenues for further applications.

The appendices contain detailed information regarding the following topics:

- The design and implementation of the *ELKB*. Appendices A through C present the basic functions and use cases of the *ELKB*, the documentation as well as the graphical and command line interfaces to the system.

- The programs developed for this thesis and the preparation of the lexical material. Appendices D through F list the programs developed for this thesis, state the manner in which they must be used to prepare the lexical material for the *ELKB*, give a detailed account of the conversion of the Pearson source files into the format used by the system, and present some errors found in the source files.

- The word lists used by the *ELKB*. Appendices G and H show the 646 American and British spelling variations and the 980-element stop list used by the *ELKB*.

- Experiments performed in this thesis. Appendices I through L present results of the semantic similarity and lexical chain building experiments.

- Appendix M shows the first two levels of the *WordNet 1.7.1* noun hierarchy.

### 1.5.1   Paper Map

Parts of the dissertation have already been subject of papers:

| *Topic* | *dissertation sections* | *previously described in* |
| --- | --- | --- |
| The use of Thesauri in NLP | 2.1-2.3.2, 2.4 | Jarmasz and Szpakowicz (2001a), Jarmasz and Szpakowicz (2001c) |
| The Design of the *ELKB* | 2.5, 3.1-3.4 | Jarmasz and Szpakowicz (2001a), Jarmasz and Szpakowicz (2001b), Jarmasz and Szpakowicz (2001c) |
| The Implementation of the *ELKB* | 3.5, 7.2 | Jarmasz and Szpakowicz (2001b), Jarmasz and Szpakowicz (2001b) |
| Using *Roget's* to Measure Semantic Similarity | 4 | Jarmasz and Szpakowicz (2001c), Jarmasz and Szpakowicz (2003b) |
| Lexical Chain Construction Using *Roget's* | 5 | Jarmasz and Szpakowicz (2003a) |
| Combining *Roget's* and *WordNet* | 2.3.3, 6.1-6.4, 7.3 | Jarmasz and Szpakowicz (2001a), Jarmasz and Szpakowicz (2001b), Jarmasz and Szpakowicz (2001c) |

# 2 The Use of Thesauri in Natural Language Processing

This chapter presents the way in which this research field has used the two most celebrated thesauri in NLP, *Roget's* and *WordNet*. It illustrates the history of both lexical resources, explains their conception and original purpose. I analyze some of the various versions of *Roget's* and elucidate the decision to use Penguin's *Roget's Thesaurus of English Words and Phrases* as the source for my *ELKB*. This chapter further shows the manner in which researchers have used the *Thesaurus* and *WordNet* in NLP and discusses the trend towards merging lexical resources. Finally, I present the desideratum for an *ELKB* based on *Roget's* and outline its evaluation procedure.

## 2.1 The Role of Thesauri in NLP

Computational linguists have used dictionaries and thesauri in NLP ever since they first addressed the problem of language understanding. Ide and Véronis (1998) explain that machine-readable dictionaries (MRDs) became a popular source of knowledge for language processing during the 1980s. Much research activity focused on automatic knowledge extraction from MRDs to construct large knowledge bases. Thesauri using controlled vocabularies, for example the *Medical Subject Headings* thesaurus (*Medical Subject Headings*, 1983), the *Educational Resources Information Centre* thesaurus (Houston, 1984) and the *IEE Inspec* thesaurus which contains technical literature in domains related to engineering (*Inspec Thesaurus*, 1985), have proven effective in information retrieval (Lesk, 1995). George Miller and his team constructed manually *WordNet*, the only broad coverage, freely available lexical resource of its kind. I propose an *ELKB* similar in scope and function to *WordNet* and construct it automatically from the most celebrated thesaurus.

## 2.2 An Overview of *Roget's Thesaurus*

*Roget's Thesaurus*, a collection of words and phrases arranged according to the ideas they express, presents a solid framework for a lexical knowledge base. Its explicit ontology offers a classification system for all concepts that can be expressed by English words; its rich semantic groups are a large resource that this thesis shows to be beneficial for NLP experiments. Yet this resource must be studied carefully before an *ELKB* can be devised.

### 2.2.1 The Many Versions of *Roget's*

The name "Roget's" has become synonymous with the *Thesaurus*, yet most thesauri are not based on the original classification system. Roget published the first edition of his *Thesaurus of*

*English Words and Phrases* in May 1852. Already by 1854, Reverend B. Sears had copied it in the United States, removing all of the phrases and placing all words and expressions borrowed from a foreign language in an Appendix (Kirkpatrick, 1998). Today, a quick search for *Roget's Thesaurus* at *Amazon.com* reveals well over 100 results with such titles as *Roget's 21st Century Thesaurus*, *Roget International Thesaurus*, *Roget's II: The New Thesaurus*, *Roget's Children's Thesaurus*, *Bartlett's Roget's Thesaurus*, *Roget's Super Thesaurus* and *Roget's Thesaurus of the Bible*. Thesauri are now commonplace in written reference libraries or electronic formats, found on the Internet (Lexico, 2001), in word processors or prepared for NLP research like the *ELKB*.

Which thesaurus is the best? A study of the publishers' descriptions of their works suggests that there are many excellent thesauri. The introduction to *Roget's International Thesaurus* describes it as "a more efficient word-finder because it has a structure especially designed to stimulate thought and help you organize your ideas" (Chapman, 1992). *Roget's II: The New Thesaurus* gives itself a clear mandate as "a book devoted entirely to meaning" (Master, 1995). Penguin's 1998 edition of *Roget's Thesaurus of English Words and Phrases* (Kirkpatrick) is the present day thesaurus most similar to the original: "The unique classification system which was devised by Peter Mark Roget and is described fully in the Introduction, has withstood the test of time remarkably well. … it is eminently capable of absorbing new concepts and vocabulary and of reflecting what is happening in the English language as time goes by."

The various thesauri boast anywhere from 200,000 to 300,000 words but size alone is not what matters. *Roget's International Thesaurus* (Chapman, 1992) lists close to 180 different kinds of trees from the `acacia` to the `zebrawood`. This enumeration is not nearly enough to be exhaustive, nor does it really help to describe all aspects of the concept `310 Plants`. If the size of a thesaurus determined a "winner", one could simply publish a list of plants, animals, and so on. This thesis demonstrates that the classification system of the 1987 edition of Penguin's *Roget's Thesaurus* is its great strength. It is interesting to note that Roget's source of inspiration for the *Thesaurus* indeed was plant taxonomy, so it is much more than a mere catalogue. Every word or phrase is carefully placed in the hierarchy. While a good thesaurus must contain many words, it should above all classify them methodically, according to the ideas which they express.

### 2.2.2 A Comparison of Potential Candidates for Building an *ELKB*

In this sea of thesauri, only four candidates remain as contenders for building an electronic lexical knowledge base: the 1911 Project Gutenberg (Hart, 1991) edition, Patrick Cassidy's *FACTOTUM* semantic network (Cassidy, 1996, 2000), HarperCollins' *Roget's International Thesaurus* and Penguin's *Roget's Thesaurus of English Words and Phrases*. The first two resources are electronic versions of an early *Roget's Thesaurus*, the latter two printed versions.

The 1911 Project Gutenberg (Hart, 1991) edition is a text file derived from the 1911 version of *Roget's Thesaurus*. MICRA, INC. prepared it in May 1991. This is a public domain version of the *Thesaurus*. This version consists of six classes, 1035 major subject headings and roughly 41,000 words, 1,000 of them added by MICRA Inc. The original classification system has been preserved. This 1911 version is the foundation for *FACTOTUM*, which is described as a semantic network organized very similarly to *Roget's* but with a more explicit hierarchy and 400 semantic relations that link the individual words to one of the 1035 heads (Cassidy, 2000)

The fifth edition of *Roget's International Thesaurus* is "the most up-to-date and definitive thesaurus" according to its editor, Robert L. Chapman (1992). It consists of fifteen classes, 1073 headwords and more than 325,000 words. Previous printed editions of *Roget's International Thesaurus* have been successfully used in NLP for word sense disambiguation, information retrieval and computing lexical cohesion in texts. We present examples of such experiments in section 2.3.1. As far as we know, no electronic versions of *Roget's International Thesaurus* have been made available to the public at large.

Penguin's *Roget's Thesaurus of English Words and Phrases*, edited by Betty Kirkpatrick (1998) is "a vast treasure-house according to ideas and meanings" as advertised by Penguin Books. It consists of six classes, 990 headwords and more than 250,000 words. It is has maintained a classification system similar to that of the original edition, and the vocabulary has been updated to reflect the changes since the mid-19th century.

Two criteria for choosing the starting point for the electronic lexical knowledge base stand out: an extensive and up-to-date vocabulary and a classification system very similar to that of the original *Thesaurus*. The constraint on the classification system allows investigating how well it has stood the test of time. Only the 1911 version and the Penguin edition have kept the original classification system, but the latter is more complete. For this reasons I have chosen *Roget's Thesaurus of English Words and Phrases* as the foundation of my electronic lexical knowledge base.

## 2.3   NLP Applications of *Roget's Thesaurus* and *WordNet*

It is commonly accepted that a lexical resource should not be prepared if there is no specific task for it. I have developed what is often referred to as a "vanilla flavor" lexicon – a resource that has a broad, general coverage of the English language (Wilks *et al.*, 1996). The fear in creating such a lexical resource is that by trying to be suitable for all applications, it ends up being useful for none. *WordNet* is also such an instrument and it has proven to be invaluable to the NLP community.

Both *Roget's Thesaurus* and *WordNet* were not initially intended for NLP. The first was planned for writers and orators, for those who are "painfully groping their way and struggling with the difficulties of composition" (Kirkpatrick, 1998), the latter as a model for psycholinguists, devised as "an on-line representation of a major part of the English lexicon that aims to be psychologically realistic" (Beckwith et al., 1991). Like penicillin, *WordNet* is now considered a panacea. *Roget's Thesaurus*, whose potential I intend to demonstrate, should become equally effective. The following sections describe how both resources have been used in NLP.

### 2.3.1   Using *Roget's Thesaurus* in NLP

*Roget's Thesaurus* has been used sporadically in NLP since about 1950 when it was first put into a machine-readable form (Masterman, 1957). The most notable applications include machine translation (Masterman, 1957, Sparck Jones, 1964), information retrieval (Driscoll, 1992, Mandala *et al.*, 1999), computing lexical cohesion in texts (Morris and Hirst, 1991) and word sense disambiguation (Yarowsky, 1992).

Masterman (*ibid.*) used a version on punched cards to improve word-for-word machine translation. She demonstrated how the *Thesaurus* could improve an initially unsatisfactory translation. As an example, Masterman explains that the Italian phrase "`tale problema si presenta particolarmente interressante`", translated word-for-word as "`such problems self-present particularly interesting`", can be retranslated as "`such problems strike one as, [or prove] particularly interesting`". The essence of this "thesaurus procedure" is to build, for every significant word of the initial translation, a list of Heads under which they appear, and to look for intersections between these lists. Replacements for inaccurately translated words or phrases are selected from the Head that contains the most words from the initial translation. Alternatives can be manually selected by choosing a better word from this Head. Masterman notes that the sense of a word, as used in a sentence, can be uniquely identified by knowing to which Head this sense belongs.

Sparck Jones (1964) realized that the *Thesaurus* in its present form had to be improved for it to be effective for machine translation. She therefore set out to create the ideal machine translation dictionary that "… has to be a dictionary in the ordinary sense: it must give definitions or descriptions of the meanings of words. It must also, however, give some indication of the kinds of contexts in which the words are used, that is, must be a 'semantic classification' as well as a dictionary" (*ibid.*). Sparck Jones believed that by classifying dictionary definitions using *Roget's* Heads she could construct the resource that she required.

Under *Roget's* headwords there are groups of closely semantically related words, located in the same paragraph and separated by semicolons. Sparck Jones (1964) built "rows" consisting of close semi-synonyms, using the *Oxford English Dictionary*. She then attempted to classify the rows according to the common membership in headwords. Here are some rows that have been classified as belonging to the Head `activity`:

```
activity  animation
activity  liveliness animation
activity  animation movement
activity  action work
activity  energy vigour
```

Sparck Jones later used these techniques for information retrieval. Other people (Driscoll, 1992 and Mandala *et al.*, 1999) have also used the *Thesaurus* for this purpose, but this time to expand the initial queries.

Halliday and Hasan (1976) explain that a cohesive text is identified by the presence of strong semantic relations between the words that it is made up of. Morris and Hirst (1991) calculated lexical cohesion, which they call "the result of chains of related words that contribute to the continuity of lexical meaning" within texts. The fourth edition of *Roget's International Thesaurus* (Chapman, 1977) was used to compute manually lexical chains, which are indicators of lexical cohesion. Stairmand (1994) automated this process using the 1911 edition of *Roget's Thesaurus* but did not obtain good results, as this version of the *Thesaurus* contains limited and antiquated vocabulary. Ellman (2000) once again used the 1911 edition to build lexical chains so as to construct a representation of a text's meaning from its content. An implementation of a lexical chain building system that uses the *ELKB* is presented in Chapter 5.

Word sense disambiguation must be the most popular use of *Roget's Thesaurus* in NLP. Yarowsky (1992) defines the sense of a word as "the categories listed for that word in *Roget's International Thesaurus*" The 1000 headwords allow to partition the major senses of a word quite accurately. To perform sense disambiguation, one must determine under which headword the given sense belongs. This can be determined by using the context of a polysemous word and the words of a given class. Other people who have used *Roget's* for word sense disambiguation include Bryan (1973, 1974), Patrick (1985), Sedelow and Mooney (1988), Kwong (2001b).

*Roget's Thesaurus* has also been used to measure semantic similarity with extremely good correlation with human judgments. It was first accomplished by McHale (1998) using the taxonomy of *Roget's International Thesaurus*, third edition (Berrey and Carruth, 1962). Another implementation, using the *ELKB,* is discussed in Chapter 4.

Scientists have also attempted to build databases that can infer common sense knowledge, for example that "blind men cannot see", using *Roget's*. Some implementations include Cassidy's *FACTOTUM* semantic network (Cassidy, 2000) and the work of Sedelow and Sedelow (1992).

### 2.3.2   Using *WordNet* in NLP

George Miller first thought of *WordNet* in the mid-1960s. The *WordNet* project started in 1985 and seven different versions have been released since version 1.0 in June of 1991 (Miller, 1998a). Although it is an electronic lexical database based on psycholinguistic principles, it has been used almost exclusively in NLP. For numerous research groups around the world it is now a generic resource. A tribute to its success are the Coling-ACL '98 workshop entitled *Usage of WordNet in Natural Language Processing Systems* (Harabagiu, 1998), the NAACL 2001 *WordNet and Other Lexical Resources* workshop (Moldovan and Peters, 2001), the *1st International WordNet Conference* (Fellbaum, 2002) and the close to 300 references in the *WordNet Bibliography* (Mihalcea, 2003). Some of the issues discussed at the *WordNet* workshops and conference include determining for which applications *WordNet* is a valuable resource, evaluating if *WordNet* can be used to develop high performance word sense disambiguation algorithms and extending *WordNet* for specific tasks. The semantic relations in *WordNet* have often been studied, even exploited for a variety of applications, for example measuring semantic similarity (Budanitsky and Hirst, 2001), and encoding models for answers types in open-domain question answering systems (Pasca and Harabagiu, 2001). Using *WordNet* as a blueprint, various multilingual lexical databases have been implemented, the first one being *EuroWordNet* (Vossen, 1998), a multilingual electronic lexical database for Dutch, Italian, Spanish, German, French and Estonian.

Due to the limitations of the printed version of *Roget's Thesaurus*, many researchers have opted for *WordNet* when attempting to extend their algorithms beyond toy problems. Systems that perform word sense disambiguation have been implemented using this electronic lexical database (Sussna, 1993; Okumura and Honda, 1994; Li *et al.*, 1995; Mihalcea and Moldovan, 1998; Kwong, 2001b; Fellbaum *et al.*2001). *WordNet*'s taxonomy has been exploited to measure semantic similarity – for a survey of these metrics see (Budanitsky and Hirst, 2001). Lexical chains (Morris and Hirst, 1991) were first built by hand using *Roget's International Thesaurus*. Hirst and St-Onge (1998) later implemented them using *WordNet*. Lexical chains built using *WordNet* have been applied in text summarization by Barzilay and Elhadad (1997), Brunn *et al.* (2001) as well as Silber and McCoy (2000, 2002). It is impossible to evaluate how many people who have used *WordNet* would have used *Roget's Thesaurus* were it in a machine-tractable form and free.

### 2.3.3   Combining *Roget's* Thesaurus and *WordNet* in NLP

A current trend in NLP is towards combining lexical resources to attempt to overcome their individual weaknesses. *Roget's* taxonomy has been used to compute the semantic similarity between words and the results have been compared to those of the same experiments using *WordNet* (Mc Hale, 1998). Although the two resources were not actually combined, and it is not clear which system produces better results, it is an interesting investigation, which reiterates the fact that both can be used for the same applications. This experiment is repeated in Chapter 4 using the *ELKB*. Others have attempted to enrich *WordNet* with *Roget's Thesaurus* to supplement the lack of relations between part of speech and proper nouns by those available in the *Thesaurus* (Mandala *et al.*, 1999). Kwong (1998, 2001a) presents an algorithm for aligning the word senses of the nouns in the 1987 edition of *Roget's Thesaurus of English Words and Phrases* an those in *WordNet 1.6*. A limited number of noun senses has been mapped. Nastase and Szpakowicz (2001) perform a similar experiment on a smaller set of words but use a larger set of parts-of-speech: nouns, adjectives and adverbs. An implementation of the mapping of word senses in the *ELKB* and *WordNet 1.7* is discussed in Chapter 6. This is of importance, since as Kwong states (1998): "In general we cannot expect that a single resource will be sufficient for any NLP applications. *WordNet* is no exception, but we can nevertheless enhance its utility". The analogy can be drawn with *Roget's*: alone it cannot serve all tasks; combined with *WordNet* the *Thesaurus* will be enriched.

## 2.4   *Roget's Thesaurus* as a Resource for NLP

### 2.4.1   Why Have People Used *Roget's* for NLP?

This chapter has presented examples of NLP applications that used *Roget's Thesaurus*. What were the incentives? The structure based on the hierarchy of categories is very simple to computerize and use, as was demonstrated by Masterman (1957) and Sparck Jones (1964). No "reverse engineering" is required to access this lattice of concepts, as it would have to be if one were building it from a dictionary. *Roget's* has a long established tradition and is believed to be the best thesaurus. It is, however, not machine tractable in the way *WordNet* is. To quote McHale (1998): "*Roget's* remains, though, an attractive lexical resource for those with access to it. Its wide, shallow hierarchy is densely populated with nearly 200,000 words and phrases. The relationships among the words are also much richer than *WordNet's* IS-A or HAS-PART links. The price paid for this richness is a somewhat unwieldy tool with ambiguous links". Indeed, the extreme difficulty of exploiting implicit semantic relations is one of the reasons why the *Thesaurus* has been considered but discarded by many researchers.

### 2.4.2    Why Do People Not Use *Roget's* More for NLP?

It is difficult for a computer to use a resource prepared for humans. *WordNet* is simply easier to use, as explained by Hirst and St-Onge (1995): "Morris and Hirst were never able to implement their algorithm for finding lexical chains with *Roget's* because no on-line copy of the thesaurus was available to them. However, the subsequent development of *WordNet* raises the possibility that, with a suitable modification of the algorithm, *WordNet* could be used in place of *Roget's*". An electronic version of the 1911 edition of *Roget's Thesaurus* has been available since 1991. This edition also proves inadequate for NLP, as Hirst and St-Onge (1995) describe: "Recent editions of *Roget's* could not be licensed. The on-line version of the 1911 edition was available, but it does not include the index that is crucial to the algorithm. Moreover, because of its age, it lacks much of the vocabulary necessary for processing many contemporary texts, especially newspaper and magazine articles and technical papers." Stairmand (1994) confirmed that it is not possible to implement a lexical chainer using the on-line 1911 version.

The literature shows that only Penguin's *Roget's Thesaurus of English Words and Phrases, HarperCollins' Roget's International Thesaurus* as well as the 1911 edition have been used for NLP research. Choosing the concept hierarchy of one or the other does not ensure a definitive advantage, as Yarowsky (1992) states: "Note that this edition of *Roget's Thesaurus* [Chapman, 1977] is much more extensive than the 1911 version, though somewhat more difficult to obtain in electronic form. One could use other concept hierarchies, such as *WordNet* (Miller, 1990) or the *LDOCE* subject codes (Slator, 1992). All that is necessary is a set of semantic categories and a list of the words in each category." *Roget's* is more than a concept hierarchy, but the elements that are most easily accessed using a printed version are the classification system and the index. For this reason, computational linguists have limited their experiments to computerizing and manipulating the index.

The availability of the lexical material of a current edition of *Roget's Thesaurus* is the major hindrance for using this resource in NLP. The publishers of *Roget's* do not make it easy to obtain an electronic copy.

### 2.4.3    A Machine-tractable Version of *Roget's* extended With *WordNet* Relations

*Roget's Thesaurus* has many undeniable advantages. It is based on a well-constructed concept classification, and its entries were written by professional lexicographers. It contains around 250,000 words compared to *WordNet's* almost 200,000. *Roget's* employs a rich set of semantic relations, most of them implicit (Cassidy, 2000). These relationships are one of the most interesting qualities. Morris and Hirst (1991) say: "A thesaurus simply groups related words without attempting to explicitly name each relationship. In a traditional computer database, a systematic semantic relationship can be represented by a slot value for a frame, or by a named link in a semantic network. If it is hard to classify a relationship in a systematic semantic way, it will be hard to represent the relationship in a traditional frame or semantic network formalism". A machine-tractable thesaurus will possibly present a better way of organizing semantic relations, although the challenge will be to label them explicitly.

*Roget's Thesaurus* does not have some of *WordNet's* shortcomings, such as the lack of links between parts of speech and the absence of topical groupings. The clusters of closely related words are obviously not the same in both resources. *WordNet* relies on a set of about 15 semantic relations, which I present in Chapter 3. Search in this lexical database requires a word and a semantic relation; for every word some (but never all) of 15 relations can be used in search. It is impossible to express a relationship that involves more than one of the 15 relations: it cannot be stored in *WordNet*. The *Thesaurus* can link the noun `bank`, the business that provides financial services, and the verb `invest`, to give money to a bank to get a profit, as used in the following sentences, by placing them in a common head **784** `Lending`.

    1. Mary went to the `bank` yesterday.
    2. She `invested` $5,000.00 in mutual funds.

This notion cannot be described using *WordNet's* semantic relations. While an English speaker can identify a relation not provided by *WordNet*, for example that one invests money in a bank, this is not sufficient for a computer system. The main challenge is to label such relations explicitly. I expect to be able to identify a good number of implicit semantic relations in *Roget's* by combining it with *WordNet* .This process is described in Chapter 6.

*WordNet* was built using different linguistic sources. They include the *Brown Corpus* (Francis and Kucera, 1982), the *Basic Book of Synonyms and Antonyms* (Urdang, 1978a), *The Synonym Finder* (Urdang, 1978b), the 4[th] edition of *Roget's International Thesaurus* (Chapman, 1977) and Ralph Grishman's *COMLEX* (Macleod, *et al.*, 1994). Many of the lexical files were written by graduate students hired part-time. Penguin's *Roget's Thesaurus of English Words and Phrases* is prepared by professional lexicographers and validated using data from the *Longman Corpus Network* of many millions words. The carefully prepared *Thesaurus* is therefore more consistent than *WordNet* as is shown by comparing the `soldier` paragraph in the *Thesaurus* Head **722** `Combattant. Army. Navy. Air Force` and the kinds of `soldier` listed in *WordNet 1.7.1*.

*soldier*, army man, pongo; military man, long-term soldier, regular; soldiery, troops, *see armed force*; campaigner, old campaigner, conquistador; old soldier, veteran, Chelsea pensioner; fighting man, warrior, brave, myrmidon; man-at-arms, redcoat, legionary, legionnaire, centurion; vexillary, standard-bearer, colour escort, colour sergeant, ensign, cornet; heavy-armed soldier, hoplite; light-armed soldier, peltast; velites, skirmishers; sharpshooter, sniper, franc-tireur, 287 *shooter*; auxiliary, territorial, Home Guard, militiaman, fencible; yeomanry, yeoman; irregular, irregular troops, moss-trooper, cateran, kern, gallowglass, rapparee, bashi-bazouk; raider, tip-and-run run; guerrilla, partisan, freedom fighter, fedayeen; resistance fighter, underground fighter, Maquis; picked troops, 644 *elite*; guards, housecarls, 660 *protector*, *see armed force*; effective, enlisted man; reservist; volunteer; mercenary; pressed man; conscript, recruit, rookie; serviceman, Tommy, Tommy Atkins, Jock, GI, doughboy, Aussie, Anzac, poilu, sepoy, Gurkha, askari; woman soldier, female warrior, Amazon, Boadicea; battlemaid, valkyrie; Wren, WRAF, WRAC.

**Figure 2.1:** The *Roget's Thesaurus* Paragraph `soldier 722 n.`

*soldier* => cannon fodder, fresh fish; cavalryman, trooper; cavalryman, trooper; color bearer, standard-bearer; Confederate soldier; redcoat, lobsterback; flanker; goldbrick; Green Beret; guardsman; Highlander; infantryman, marcher, foot soldier, footslogger; Janissary; legionnaire, legionary; man-at-arms; militiaman; orderly; paratrooper, para; peacekeeper; pistoleer; point man; ranker; regular; reservist; rifleman; Section Eight; tanker, tank driver; territorial; Unknown Soldier; Allen, Ethan Allen; Bayard, Seigneur de Bayard, Chevalier de Bayard, Pierre Terrail, Pierre de Terrail; Borgia, Cesare Borgia; Higginson, Thomas Higginson, Thomas Wentworth Storrow Higginson; Kosciusko, Thaddeus Kosciusko, Kosciuszko, Tadeusz Andrzej Bonawentura Kosciuszko; Lafayette, La Fayette, Marie Joseph Paul Yves Roch Gilbert du Motier, Marquis de Lafayette; Lawrence, T. E. Lawrence, Thomas Edward Lawrence, Lawrence of Arabia; Lee, Henry Lee, Lighthorse Harry Lee; Mohammed Ali, Mehemet Ali, Muhammad Ali; Morgan, Daniel Morgan; Percy, Sir Henry Percy, Hotspur, Harry Hotspur; Peron, Juan Domingo Peron; Smuts, Jan Christian Smuts; Tancred.

**Figure 2.2:** The kinds of `soldier` in *WordNet 1.7.1*

The *Roget's* paragraph does not contain any proper nouns, which at first may seem as a weakness compared to *WordNet*, but is the most rational decision for such a lexical resource, as *WordNet's* list contains but an infinitely small number of the world's great soldiers. Although *WordNet* lists `Allen`, `Bayard`, `Borgia`, `Higginson`, `Kosciusko`, `Lafayette`, `Lawrence`, `Lee`, `Mohamed Ali`, `Morgan`, `Percy`, `Peron`, `Smuts` and `Tancred` other arguably even greater, such as

`Alexander`, `Caesar`, `Charlemagne`, `Timur`, `Genghis Khan`, `Napoleon`, `Nelson` are absent. This example is not sufficient to prove that the *Thesaurus* is a more carefully crafted resource than *WordNet*, but is enough to indicate that *Roget's* is a very good foundation for an electronic lexical knowledge base and that *WordNet* is not perfect. Extending *Roget's* with *WordNet* can only make it better, as the combined information makes for a richer semantic network. Although *WordNet* has become the *de facto* standard electronic lexical knowledge base for NLP, there is no reason why it should be the only one. The *ELKB* is built from *Roget's Thesaurus* and can be combined with *WordNet*. This results in an interesting alternative for solving NLP problems.

## 2.5   The Evaluation of a Thesaurus Designed for NLP

The evaluation of the *ELKB* must be functional, quantitative and qualitative. It is functional in the sense that the *ELKB* must allow the same manipulations as the printed *Thesaurus*: word and phrase lookup, browsing via the hierarchy, random browsing and following links. Experiments that have been previously done by hand, for example calculating the distance between words or phrases by counting their relative separation in *Roget's*, must be automated. Chapter 3 presents the various use scenarios and discusses how they are implemented in the *ELKB*. The evaluation is quantitative in the sense that the *ELKB* should have a comparable number of word senses as *WordNet*. This evaluation is performed in Chapter 3. It finally is qualitative in the sense that the words and phrases contained in *Roget's* should perform a wide variety of NLP applications. The *Thesaurus* is put to the test by calculating semantic similarity between words and phrases, explained in Chapter 4, and in the task of building lexical chains, described in Chapter 5. The experiments involving mapping *Roget's* senses onto *WordNet* senses in Chapter 6 expose the differences in lexical material between both resources.

# 3 The Design and Implementation of the *ELKB*

The preliminary step in evaluating the usefulness of *Roget's Thesaurus* for NLP is the implementation of the *ELKB*. This chapter describes the steps involved in computerizing the *Thesaurus*, from the details of how this resource is organized, to a Java implementation of the *ELKB*. I explain *Roget's* structure, behavior and function as well as present the way in which the *ELKB* works. This chapter shows the steps involved in transforming the source material into a format that is adequate for further processing, and discusses the required data structures for the system. It finally illustrates some scenarios of how the *ELKB* is to be used. I perform a comparison to *WordNet*, the *de facto* standard for electronic lexical databases, at the various stages of the design and implementation of this electronic resource.

## 3.1 General organization of *Roget's* and *WordNet*

Ontologies have been used in Artificial Intelligence since the 1950s. Researchers agree that they are extremely useful for a wide variety of applications but do not agree on their contents and structure (Lehmann, 1995). Roget proposed a classification system that is essentially a taxonomy of ideas that can be expressed in the English language. His system has a very Victorian bias to it, but this thesis demonstrates that it is useful to modern day researchers nonetheless. Let's examine its properties and compare it to the ontology of nouns implicitly present in *WordNet*.

*Roget's* ontology is headed by six Classes. The first three Classes cover the external world: *Abstract Relations* deals with such ideas as number, order and time; *Space* is concerned with movement, shapes and sizes, while *Matter* covers the physical world and humankind's perception of it by means of five senses. The remaining Classes deal with the internal world of human beings: the mind (*Intellect*), the will (*Volition*), the heart and soul (*Emotion, Religion and Morality*). There is a logical progression from abstract concepts, through the material universe, to mankind itself, culminating in what Roget saw as mankind's highest achievements: morality and religion (Kirkpatrick, 1998). Class Four, *Intellect*, is divided into *Formation of ideas* and *Communication of ideas*, and Class Five, *Volition*, into *Individual volition* and *Social volition*. In practice, therefore, the *Thesaurus* is headed by eight Classes. This is the structure that has been adopted for the *ELKB*.

A path in *Roget's* ontology always begins with one of the Classes. It branches to one of the 39 Sections, then to one of the 79 Sub-Sections, then to one of the 596 Head Groups and finally to one of the 990 Heads. Each Head is divided into paragraphs grouped by parts of speech: nouns, adjectives, verbs and adverbs. According to Kirkpatrick (1998) "Not all Heads have a full complement of parts of speech, nor are the labels themselves applied too strictly, words and phrases being allocated to the part-of-speech which most closely describes their function". Much is left to the lexicographers' intuitions, which makes it hard to use *Roget's* as it is for NLP applications. Finally a paragraph is divided into semicolon groups of semantically closely related words. These paths create a graph in the *Thesaurus* since they are interconnected at various points. An example of a Head in Roget's is **864** *Wonder*. I show it here with the first paragraph for every part-of-speech as well as its path in the ontology:

**Class six:** Emotion, religion and morality

**Section two:** Personal emotion

**Sub-section:** Contemplative

**Head Group:** 864 Wonder – 865 Lack of wonder

**Head:** 864 Wonder

**N.** *wonder*, state of wonder, wonderment, raptness; admiration, hero worship, 887 *love*; awe, fascination; cry of wonder, gasp of admiration, whistle, wolf wolf, exclamation, exclamation mark; shocked silence, 399 *silence*; open mouth, popping eyes, eyes on stalks; shock, surprise, surprisal, 508 *lack of expectation*; astonishment, astoundment, amazement; stupor, stupefaction; bewilderment, bafflement, 474 *uncertainty*; consternation, 854 *fear*.

…

**Adj.** *wondering*, marvelling, admiring, etc. vb.; awed, awestruck, fascinated, spellbound, 818 *impressed*; surprised, 508 *inexpectant*; astonished, amazed, astounded; in wonderment, rapt, lost in wonder, lost in amazement, unable to believe one's eyes *or* senses; wide-eyed, round-eyed, pop-eyed, with one's eyes starting out of one's head, with eyes on stalks; open-mouthed, agape, gaping; dazzled, blinded; dumbfounded, dumb, struck dumb, inarticulate, speechless, breathless, wordless, left without words, silenced, 399 *silent*; bowled over, struck all of a heap, thunderstruck; transfixed, rooted to the spot; dazed, stupefied, bewildered, 517 *puzzled*; aghast, flabbergasted; shocked, scandalized, 924 *disapproving*.

…

**Vb.** *wonder*, marvel, admire, whistle; hold one's breath, gasp, gasp with admiration; hero-worship, 887 *love*; stare, gaze and gaze, goggle at, gawk, open one's eyes wide, rub one's eyes, not believe one's eyes; gape, gawp, open one's mouth, stand in amazement, look aghast, 508 *not expect*; be awestruck, be overwhelmed, 854 *fear*; have no words to express, not know what to say, be reduced to silence, be struck dumb, 399 *be silent*.

...

**Adv.** *wonderfully*, marvellously, remarkably, splendidly, fearfully;
wondrous strange, strange to say, wonderful to relate, mirabile
dictu, to the wonder of all.

**Int.** amazing! incredible! I don't believe it! go on! well Inever!
blow me down! did you ever! gosh! wow! how about that! bless my
soul! 'pon my word! goodness gracious! whatever next! never!

**Figure 3.1:** The *Roget's Thesaurus* Head **864** *Wonder*.

Miller took a different approach to constructing an ontology for *WordNet*. Only nouns are clearly
organized into a hierarchy. Adjectives, verbs and adverbs are organized individually into various
webs that are difficult to untangle. This decision has been based on pragmatic reasons more than
on theories of lexical semantics, as Miller (1998b) admits: "Partitioning the nouns has one
important practical advantage: it reduces the size of the files that lexicographers must work with
and makes it possible to assign the writing and editing of the different files to different people."
Indeed, organizing *WordNet's* more than 100,000 nouns must have required a fair amount of
planning.

In *WordNet* version 1.7.1 noun hierarchies are organized around nine *unique beginners*. A
unique beginner is a synset which is found at the top of the noun ontology. Most synsets are
accompanied with a gloss which is a short definition of the synonym set. The following are the
unique beginners:

**entity, physical thing** (that which is perceived or known or
inferred to have its own physical existence (living or nonliving))
**psychological feature**, (a feature of the mental life of a living
organism)
**abstraction**, (a general concept formed by extracting common
features from specific examples)
**state**, (the way something is with respect to its main attributes;
"the current state of knowledge"; "his state of health"; "in a
weak financial state")
**event**, (something that happens at a given place and time)
**act, human action, human activity**, (something that people do or
cause to happen)
**group, grouping**, (any number of entities (members) considered as a
unit)
**possession**, (anything owned or possessed)
**phenomenon**, (any state or process known through the senses rather
than by intuition or reasoning)

**Figure 3.2:** The *WordNet 1.7.1* unique beginners.

All of the other nouns can eventually be traced back to these nine synsets. If these are considered analogous to *Roget's* Classes, the next level of nouns can be considered as the Sections. In all, the unique beginners have 161 noun synsets directly linked to them (Appendix M). The number of nouns that are two levels away from the top of the noun hierarchies have not been identified, but if even a quarter of the *WordNet* nouns can be found here, they would represent close to 37,000 words.

Miller (1998b) mentions that *WordNet's* noun ontology is relatively shallow in the sense that it seems to have a limited number of levels of specialization. In theory, of course, there is no limit to the number of levels an inheritance system can have. Lexical inheritance systems, however, seldom go more than 10 or 12 levels deep, and the deepest examples usually contain technical distinctions that are not part of the everyday vocabulary. For example, a Shetland pony is a pony, a horse, an equid, an odd-toe ungulate, a placental mammal, a mammal, a vertebrate, a chordate, an animal, an organism, an object and an entity: 13 levels, half of them technical (*ibid.*).

The IS-A relations connect *WordNet's* noun hierarchy in a vertical fashion, whereas the IS-PART, IS-SUBSTANCE, IS-MEMBER and the HAS-PART, HAS-SUBSTANCE, HAS-MEMBER relations allow for horizontal connections. This allows interconnecting various word nets, represented by the synsets, into a large web.

A simple quantitative comparison of the two ontologies is difficult. Roget (1852) claims that organizing words hierarchically is very useful: "In constructing the following system of classification of the ideas which are expressible by language, my chief aim has been to obtain the greatest amount of practical utility." Miller, on the other hand, feels that it is basically impossible to create a hierarchy for all words, since: "these abstract generic concepts [which make up the top levels of the ontology] carry so little semantic information; it is doubtful that people could agree on appropriate words to express them." (1998b) The *Tabular synopsis of categories*, which represents the concept hierarchy, is presented at the beginning of the *Thesaurus*. On the other hand, in *WordNet* only the unique beginners are listed, and *only* in the documentation. This shows that much more value was attributed to the ontology in *Roget's*. Tables 3.1 and 3.2 show the portions of *Roget's* and *WordNet's* ontologies that classify the various abstract relations. Both tables use the Class, Section and Head notation from *Roget's Thesaurus*. The Section `Time` has been expanded to present the underlying Heads. The glosses that accompany each *WordNet* noun synset are not included in the table so as to compare them to semicolon groups, for which the *Thesaurus* does not give definitions.

### Class One: Abstract Relations

| | |
|---|---|
| *1 Existence* | |
| *2 Relation* | |
| *3 Quantity* | |
| *4 Order* | |
| *5 Number* | |

*6 Time*

| | | | |
|---|---|---|---|
| **Absolute:** (definite/indefinite) | 108 Time | 109 Neverness | |
| | 110 Period | 111 Course | |
| | 112 Contingent duration | | |
| | 113 Long duration | 114 Transience | |
| | 115 Perpetuity | 116 Instantaneousness | |
| | 117 Chronometry | 118 Anachronism | |
| **Relative:** (to succession) | 119 Priority | 120 Posterity | |
| | 121 Present time | 122 Different time | |
| | 123 Synchronism | | |
| (to a period) | 124 Futurity | 125 Past time | |
| | 126 Newness | 127 Oldness | |
| | 128 Morning | 129 Evening | |
| | 130 Youth | 131 Age | |
| | 132 Young person | 133 Old person | |
| | 134 Adultness | | |
| (to an effect or purpose) | 135 Earliness | 136 Lateness | |
| | 137 Occasion | 138 Untimeliness | |
| **Recurrent:** | 139 Frequency | 140 Infrequency | |
| | 141 Periodicity | 142 Fitfulness | |

| | |
|---|---|
| *7 Change* | |
| *8 Causation* | |

**Table 3.1:** The hierarchical structure of *Abstract Relations* in *Roget's Thesaurus*.

The Heads in the printed *Roget's Thesaurus* are placed in two distinct columns to express opposing ideas such as **128** *Morning* and **129** *Evening*. Sometimes there is an intermediate idea, for example:

> **132** *Young person*     **133** *Old person*
>                **134** *Adultness*

The visual representation of the hypernym tree for *WordNet's* `Time` Section has been chosen so as to facilitate the comparison with *Roget's*. The order of the synsets in the table is the one given by *WordNet*. There does not seem to be as clear an underlying structure as the one presented by the *Thesaurus*.

**Unique Beginner Three: Abstraction**

*1 Time*

```
1  Geological time, geologic time
2  Biological time
3  Cosmic time
4  Civil time, standard time, local time
5  Daylight-saving time, daylight-savings time, daylight
   saving, daylight savings
6  Present, nowadays
7  Past, past times, yesteryear, yore
8  Future, hereafter, futurity, time to come
9  Musical time
10 Continuum
11 Greenwich Mean Time, Greenwich Time, GMT, universal
   time, UT, UT1
12 Duration, continuance
13 Eternity, infinity, forever
```

*2 Space*

*3 Attribute*

*4 Relation*

*5 Measure,*
*quantity,*
*amount, quantum*

*6 Set*

**Table 3.2:** The hierarchical structure of *Abstraction* in *WordNet*

## 3.2 The Counts of Words and Phrases in *Roget's* and *WordNet*

The simplest way to compare *Roget's Thesaurus* and *WordNet* is to count strings. Table 3.3 shows the word and phrase counts for the 1987 *Roget's*, divided among parts of speech. A sense is defined as the occurrence of a word or phrase within a unique semicolon group, for example the `slope` in `{rising ground, rise, bank, ben, brae, slope, climb, incline}`. Table 3.4 presents the different counts for *WordNet 1.7.1* and the strings in common with *Roget's*. Here a sense is the occurrence of a string within a unique synset, for example *slope* in `{slope, incline, side}`.

The absolute sizes are similar. The surprisingly low 32% overlap may be due to the fact that *WordNet's* vocabulary dates to 1990, while *Roget's* contains a vocabulary that spans 150 years, since many words have been added to the original 1852 edition, but few have been removed. It is also rich in idioms: "The present Work is intended to supply, with respect to the English language, a desideratum hitherto unsupplied in any language; namely a collection of words it contains and of the idiomatic combinations peculiar to it …" (Roget, 1852). Fellbaum (1998b) admits that *WordNet* contains little figurative language. She explains that idioms must appear in an *ELKB* if it is to serve NLP applications that deal with real texts where idiomatic language is pervasive.

| POS | Unique Strings | Paragraphs | Semicolon Groups | Senses |
|---|---|---|---|---|
| Noun | 56307 | 2876 | 31133 | 114052 |
| Verb | 24724 | 1497 | 13968 | 55647 |
| Adjective | 21665 | 1500 | 12889 | 48712 |
| Adverb | 4140 | 498 | 1822 | 5708 |
| Interjection | 372 | 61 | 65 | 406 |
| *Totals* | 107208 | 6432 | 59877 | 224525 |

**Table 3.3:** 1987 *Roget's Thesaurus* statistics.

| POS | Unique Strings | Synsets | Senses | Common with Roget's | % of common strings |
|---|---|---|---|---|---|
| Noun | 109195 | 75804 | 134716 | 27118 | 24.83 |
| Verb | 11088 | 13214 | 24169 | 7231 | 65.21 |
| Adjective | 21460 | 18576 | 31184 | 10465 | 48.76 |
| Adverb | 4607 | 3629 | 5748 | 1585 | 34.40 |
| Interjection | 0 | 0 | 0 | 0 | 0.00 |
| *Totals* | 146350 | 111223 | 195817 | 46399 | 31.70 |

**Table 3.4:** *WordNet 1.7.1* statistics. *Common* refers to strings both in *WordNet* and *Roget's*.

## 3.3   The semantic relations of *Roget's* and *WordNet*

Cassidy (2000) has identified 400 kinds of semantic relations in the FACTOTUM semantic network which is based on the 1911 edition of *Roget's Thesaurus*. This suggests that the 1987 Penguin edition of *Roget's* has a rich set of implicit semantic relations. To build a useful electronic lexical knowledge base from the *Thesaurus*, these relations must be made available explicitly. Some semantic relations are present already within the *Tabular synopsis of categories*, as Kirkpatrick (1998) explains: "Most Heads are in pairs, representing the positive and negative aspects of an idea, e.g. **852** *Hope*, **853** *Hoplessness*." This antonymy relationship that is present for Heads does not necessarily translate into a relation of opposition for the words contained under each Heads. This is due to the fact that Heads and words that belong to it represent two different types of concepts. The Head represents a general concept, whereas the words and phrases represent all of the various aspects of this concept. Thus, under the Head **539** *School* can be found such notions as `;college, lycée, gymnasium, senior secondary school;` … `;lecture room, lecture hall, auditorium, amphitheatre;` and `;platform, stage, podium, estrade;`.

Two types of explicit relationships are present at the word level: Cross-reference and See. Cross-reference is a link between Heads via the syntactic form of a word. For example, the Heads **373** *Female* and **169** *Parentage* are linked by the Cross-reference `169 maternity`. The word `maternity` is present within the group `mother, grandmother 169 maternity` in the Head **373** *Female* and is the first word of a paragraph in the head **169** *Parentage*. According to Kirkpatrick (1998), the See relationship is used to refer the reader to another paragraph within the same Head, where the idea under consideration is dealt with more thoroughly. An example of this is when a general paragraph such as *killing* in Head **362** *Killing: destruction of life* is followed by more specific paragraphs `homicide` and `slaughter`. The relationship appears in the following manner in the text: `murder, assassination, bumping off (`**see** `homicide)`.

It is a common misconception that the *Thesaurus* is simply a book of synonyms. Roget (1852) admits in fact that "it is hardly possible to find two words having in all respect the same meaning, and being therefore interchangeable; that is, admitting of being employed indiscriminately, the one or the other, in all applications". According to Kirkpatrick (1998), the intention is to offer words that express every aspect of an idea, rather than to list synonyms. The groups of words found under a Head follow one another in a logical sequence. Systematic semantic relations, such as IS-A and PART-OF, are not required between the semicolon groups and the Head. For example, both `restaurant` and `have brunch` are found under the same Head **301** *Food: eating and drinking*. Although the native English speaker can identify various relations between `food`, `restaurant` and `have brunch`, it is not an easy thing to discover automatically. This is a major challenge; some possible algorithms for automatic labeling of semantic relations are presented in Chapter 6.

*WordNet* is based on about fifteen semantic relations, the most important of which is synonymy. Every part-of-speech in *WordNet* has a different set of semantic relations. It is important to note that synonymy is the only relation between words. All others are between synsets. For example, the synsets `car, auto, automobile, machine, motorcar -- (4-wheeled motor vehicle; usually propelled by an internal combustion engine; "he needs a car to get to work")` and `accelerator, accelerator pedal, gas pedal, gas, throttle,` gun -- (a pedal that controls the throttle valve; "he stepped on the gas") are linked by the meronym (has part) relation, whereas the nouns `car` and `auto` are linked by synonymy. Table 3.5 summarizes the semantic relations. All of the examples are taken from *WordNet* version 1.7.1. and *{...}* represents a synset.

| Semantic relation | Description | Part-of-speech | | | | Example |
|---|---|---|---|---|---|---|
| | | **N** | **V** | **Adj** | **Adv** | |
| *Synonym* | A concept that means exactly or nearly the same as another. *WordNet* considers immediate hypernyms to be synonyms. | × | × | × | × | *{ sofa, couch, lounge }* are all synonyms of one another. *{ seat }* is the immediate hypernym of the synset. |
| *Antonym* | A concept opposite in meaning to another. | × | × | × | × | *{ love }* is the antonym of *{ hate, detest }.* |
| *Hypernym* | A concept whose meaning denotes a superordinate. | × | × | | | A *{ feline, felid }* is a hypernym of *{ cat, true cat }.* |
| *Hyponym* | A concept whose meaning denotes a subordinate. | × | × | | | A *{ wildcat }* is a hyponym of *{ cat, true cat }.* |
| *Substance meronym* | A concept that is a substance of another concept. | × | | | | A *{ snowflake, flake }* is substance of *{ snow }.* |
| *Part meronym* | A concept that is a part of another concept. | × | | | | A *{ crystal, watch crystal, watch glass }* is a part of a *{ watch, ticker }.* |
| *Member meronym* | A concept that is a member of another concept. | × | | | | An *{ associate }* is a member of an *{ association }.* |
| *Substance of holonym* | A concept that has another concept as a substance. | × | | | | A *{ tear, teardrop }* has *{ water, H20 }* as a substance. |
| *Part of holonym* | A concept that has another concept as a part. | × | | | | A *{ school system }* has a *{ school, schoolhouse }* as a part. |
| *Member of holonym* | A concept that has another concept as a member. | × | | | | *{ organized crime, gangland, gangdom }* has *{ gang, pack, ring, mob }* as a member. |
| *Cause to* | A verb that is the cause of a result. | | × | | | *{ give }* is the cause of the result *{ have, have got, hold }* |
| *Entailment* | A verb that involves unavoidably a result. | | × | | | To *{ die, decease, perish, go, exit, pass away, expire }* involves unavoidably to *{ leave, leave behind }.* |
| *Troponym* | A verb that is a particular way to do another. | | × | | | To *{ samba }* is a particular way to *{ dance, trip the light fantastic }.* |
| *Pertainym* | An adjective or adverb that relates to a noun. | | | × | × | *{ criminal }* relates to *{ crime }.* |
| *Attribute* | An adjective that is the value of a noun. | × | | | | *{ fast (vs. slow) }* is a value of *{ speed, swiftness, fastness}* |
| *Value* | A noun that has an adjective for a value. | | | × | | *{ weight }* has *{ light (vs. heavy)}* as a value. |

**Table 3.5:** The semantic relations in *WordNet*.

*WordNet's* semantic relations are discussed in detail in the *International Journal of Lexicography 3(4)* and *WordNet: An Electronic Lexical Database* (Fellbaum, 1998).

## 3.4   Accessing *Roget's* and *WordNet*

It is very important to have adequate methods of accessing an electronic lexical knowledge base. These access methods should be designed in a computationally efficient manner, since this resource is to be machine-tractable, and faithfully reproduce how the printed version is used. For the task of computerizing *Roget's Thesaurus*, a study of its manual use can offer good suggestions. *WordNet* can also be a source of more ideas. *Roget's* provides an Index of the words and phrases in the *Thesaurus*. For every item a list of keywords, with their Head numbers and part-of-speech, indicates in what Paragraph a word can be found. The different Keywords give an indication of the various senses of a word. The combination `keyword, head number, part-of-speech` represents a unique key in the *Thesaurus*.

The following is an example of an Index entry:

> **daily**
>> *often* 139 adv.
>> *seasonal* 141 adj.
>> *peridodically* 141 adv.
>> *journal* 528 n.
>> *the press* 528 n.
>> *usual* 620 adj.
>> *cleaner* 648 n.
>> *servant* 742 n.

The Paragraph pointed to by the key `journal` 528 n. is the following:

> ***528*** *Publication*
>
> **N.** ...
>
>> *journal*, review magazine, glossy m., specialist m., women's m., male-interest m., pulp m.; part-work, periodical, serial, daily, weekly, monthly, quarterly, annual; gazette, trade journal, house magazine, trade publication 589 *reading matter*.

A *Roget's* Paragraph is made up of a Keyword and a sequence of Semicolon Groups. The Keyword, an italicized word at the beginning of a Paragraph, is not intended to be a synonym of the words that follow it, but is rather a concept that generalizes the whole Paragraph. It also allows to identify the position of other words in the Index and to locate Cross-references (Kirkpatrick, 1998). A Semicolon Group is a list of closely related words and phrases, for example: `; part-work, periodical, serial, daily, weekly, monthly, quarterly, annual;` Such lists are separated by semicolons. This is the smallest unit above single words and phrases in *Roget's*.

Most people use the Index when looking up a word in the *Thesaurus*, but Roget (1852) intended his classification system to also serve this purpose: "By the aid of this table the reader will, with a little practice, readily discover the place which the peculiar topic he is in search of occupies in the series; and on turning to the page in the body of the Work which contains it, he will find the group of expressions he requires, out of which he may cull those that are the most appropriate to his purpose". Searching the *Thesaurus* like this allows looking at all of the words found under a Head, regardless of the part-of-speech. In this manner all of the concepts that express every aspect of a given idea can be found.

For the human user, the Index is the most practical means of looking up a word. For the computer, the classification system is extremely practical, as it has been shown, for example, by Yarowsky's (1992) word sense disambiguation experiment or the semantic similarity metric presented in Chapter 4. It is important to be able to locate a word within its semicolon group, and from there to look at the other words in the same Paragraph, the same Head, knowing at all times in which place the word is found in the classification system. Using the *ELKB* it must be possible to follow the different paths built from the parts of speech, semantic relations and the ontology.

Graphical and command line interfaces exist for *WordNet*. Both work essentially in the same way. After selecting a word, all of its senses appear, listed within the synsets to which they belong, ordered by frequency and part-of-speech. For example, the search for the word `daily` returns the following:

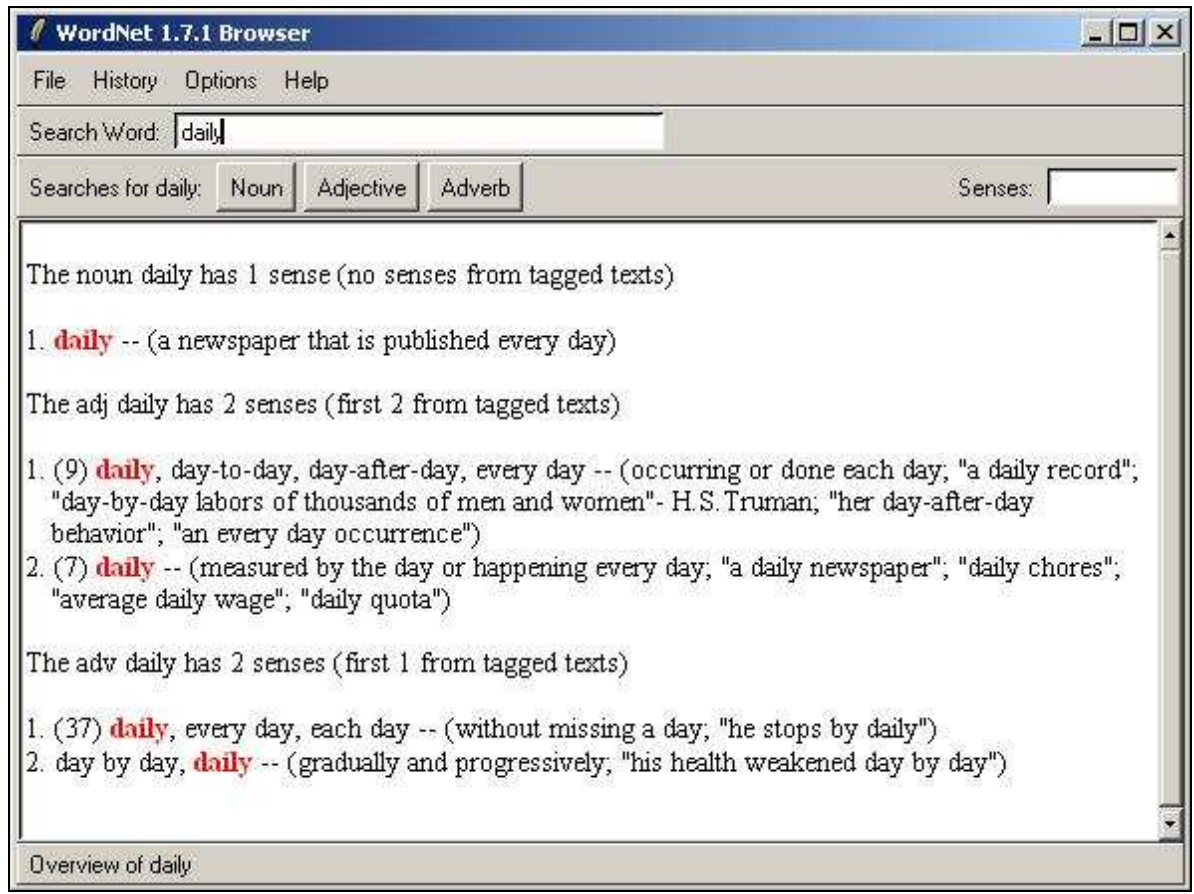

**Figure 3.3:** Overview of `daily` in *WordNet 1.7.1*.

At this point, the user can decide to continue his search of the database by using one of the semantic relations. The system does not show the exact location in the ontology where the search results are originating from, nor can all the concepts describing an idea be easily extracted. The number of times the word senses occur in Semcor (Landes *et al.*, 1998), a semantic concordance based on the Brown Corpus (Francis and Kucera, 1982), is displayed by *WordNet*.

All the methods to access *Roget's* that the printed version offers have been implemented in the *ELKB*. *WordNet* provides an interface to its lexical material in a manner that is similar to only using *Roget's* index. The *ELKB* allows performing this type of search as well as using the classification system. Appendix A presents the use cases and explains the basic functions of the *ELKB*.

## 3.5 The preparation of the Lexical Material

We have licensed the source of the 1987 *Roget's* from Pearson Education. It is divided into files with the text of the *Thesaurus* and files with its index. The Text file and Index file, both about 4

MB in size, are marked up using codes devised by the owners of the resource. Appendix E presents the steps for converting the codes into HTML-like tags. Appendix D lists the Perl scripts used for transforming the lexical material into a format that is suitable for the *ELKB* along with their accompanying documentation. The *ELKB* is created using only the Text file; the Index is constructed using the words and phrases loaded in the knowledge base. This is a necessary step, as the supplied Index file does not contain entries for all of the words contained in the *Thesaurus*.

Certain space-saving conventions are used in the source data. Where consecutive expressions use the same word, repetitions may be avoided using "`or`", as in "`drop a brick or clanger`", "`countryman or -woman`". The repeated word may also be indicated by its first letter, followed by a full stop: "`weasel word, loan w., nonce w.,`" (Kirkpatrick, 1998). All such abbreviations must be expanded before the lexical material is loaded into the *ELKB*. A Perl script was written to do this as well as to replace the Pearson codes by HTML-like tags, easier to process automatically. Other programs validate the expansion errors mostly due to noise in the original material.

### 3.5.1 Errors and Exceptions in the Source Files

The original text files supplied by Pearson Education contain some errors. There are 8 occurrences of lines that include the string "`Bad Character`", for example: `Err{\pbf\ Bad Character: \char`\?\char`\ \ }`. There are some phrases where spaces are missing between the words, for example "`creativeaccounting`" instead of "`creative accounting`". Other words are split which seem to be spelling mistakes at first but a closer look reveals that the missing letters are separated from the words by a space, for example: "`incommunicativeness`". The code `#1$:#5` is frequently inserted in the file but does not mean anything. Appendix F shows 179 instances of errors where spaces are missing and 26 instances of errors that contain an extra space, as well as specifying the original file in which they can be found.

## 3.6 The Java implementation of the *ELKB*

The entire functionality of the printed version of *Roget's* has been implemented in Java. The *ELKB*, which is comprised of eighteen classes, is organized around four major ones: the *RogetELKB* class which is the main entry point into the system, the *Category* class which models the taxonomy and has methods to traverse it, the *RogetText* class which represents the 990 *Heads* as well as the words and phrases stored under them and the *Index* class which contains the references to all of the words and phrases in the *Thesaurus*. These four classes, as well as their relation to the other fourteen, are described in the following sections. The class diagram of the *ELKB* is shown in Figure 3.4. Appendix B presents a detailed documentation of the system.

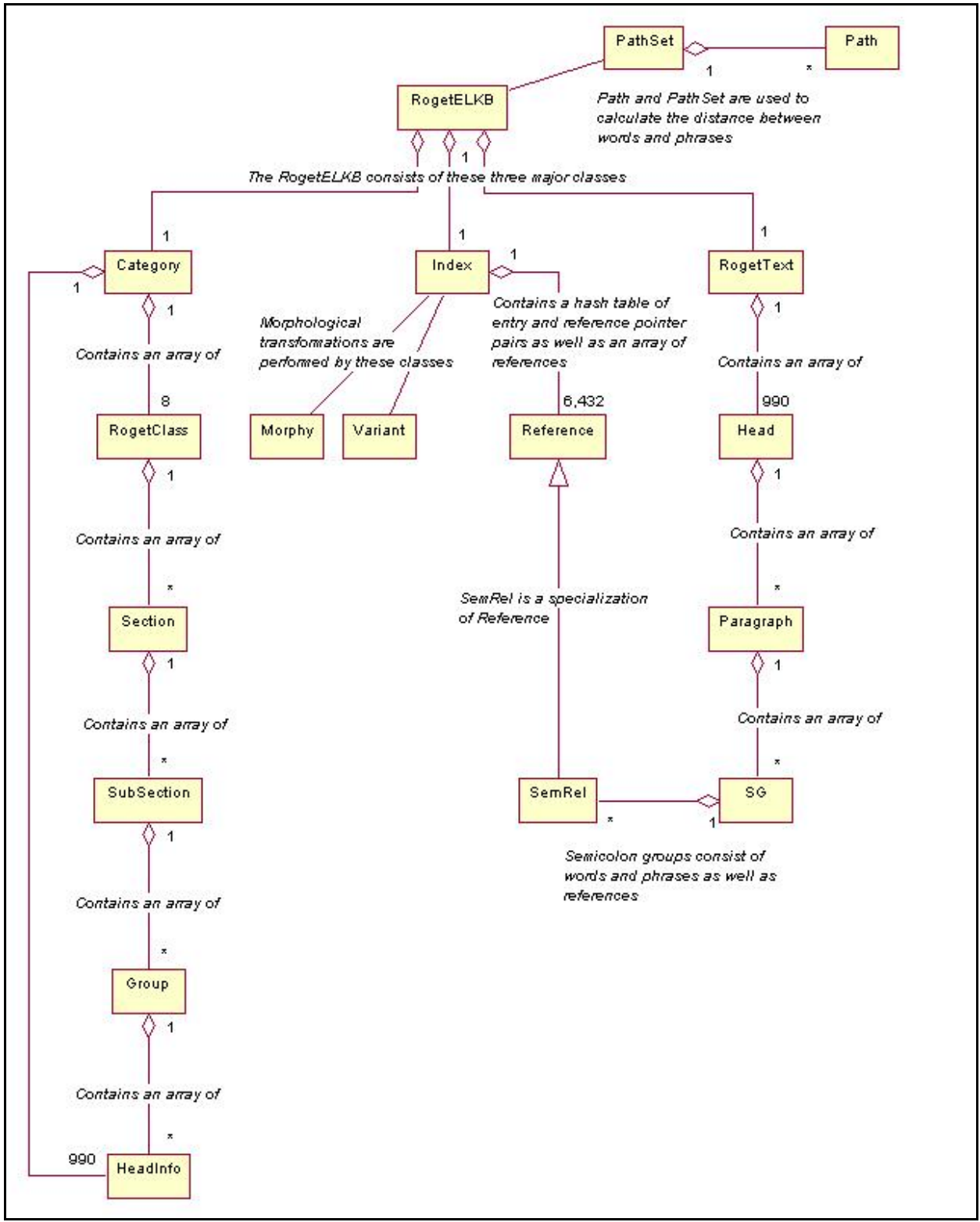

**Figure 3.4:** Class Diagram of the *ELKB*.

Correctly reproducing the printed *Roget's Thesaurus*, creating an application programming interface (API) that is both efficient an easy to use, performance, memory and the availability of the *ELKB* software to the largest possible audience have been the main concerns of this implementation. The library of Java API is used for all of the experiments included in this dissertation: evaluating the semantic similarity of words and phrases, the construction of lexical chains, and investigating algorithms for mapping the *Thesaurus* onto *WordNet*, presented in chapters 4, 5 and 6 respectively. The *ELKB* can be accessed by making direct call to the API, as well as by using the command line and graphical user interface (GUI). A first version of the GUI was created by Pierre Chrétien and Gilles Roy to fulfill the requirements of their fourth-year Honors Software Engineering project. Appendix C presents the GUI and the command line interface.

### 3.6.1   The *ELKBRoget* and Related Classes

The *ELKBRoget* class is the one that contains all others in the system. It has methods to perform manipulations on the *ELKB*, from looking up a word or phrase, calculating the distance betweentwo words or phrases, to identifying their relative position in the taxonomy. Instances of the *Index* and *Category* classes are loaded into memory to allow for rapid access. The Heads, contained in the *RogetText* object, are read from files when required. This configuration is

The best compromise between performance and memory usage. Two objects, *Path* which calculates and stores the path between two references, and *PathSet* which contains all paths between a pair of words and phrases, are accessed from *ELKBRoget*. They allow calculating the shortest path between two words or phrases.

### 3.6.2   The *Category* and Related Classes

The *Category* class models the *Tabular synopsis of categories* described in Section 3.1. It is made up of two arrays; the first contains *Category* and the second *HeadInfo* objects. The *Category* objects are comprised of an array of *Section* objects, which in turn contain an array of *SubSection* objects, which have an array of *Group* objects which are finally made up of an array of *HeadInfo* objects. The *HeadInfo* objects describe a Head entirely with respect to its location in *Roget's* taxonomy. It is defined uniquely by the Class number, Section number, Sub-section name, Head Group, as well as the name and number of the Head. It requires little memory as it does not contain any of the words or phrases. The second array of the *Category* class contains the 990 *HeadInfo* objects that describe the 990 Heads of the *Thesaurus*. Thus, the taxonomy can be traversed depth-wise, starting with the *Class* objects, or breadth-wise by accessing the *HeadInfo* objects. It is often more interesting to access the array of Heads, as this can be done via random access using a Head number.

### 3.6.3   The *RogetText* and Related Classes

The *RogetText* class represents everything that is contained in the Text of the *Thesaurus*. It contains the *Head*, *Paragraph*, *SG* and *SemRel* classes. The *SG* objects contain Semicolon groups. The *SemRel* class is used to model *Roget's* explicit semantic relations, Cross-references and See-references, discussed in Section 3.3. A *Head* object contains five arrays of *Paragraphs*, one for each of nouns, adjectives, verbs, adverbs and interjections, which are in turn made up of an array of *SG* objects. The *RogetText* is stored as 990 files, one for each *Head*. They are loaded as required. A word or phrase is looked up using a reference, for example: `contempt 922 int.` when searching for the phrase `in your face`. The reference specifies the *Head* number, `922` in this example, which allows the specified *Head* to be retrieved in constant time. The correct *Paragraph* is retrieved by finding the index of the *Paragraph* that corresponds to the reference, and looking up this *Paragraph*. The search is done in linear time and the look up in constant time. To find the word or phrase in the Semicolon group, its membership must be first identified in the array of *SG* objects, and then the Semicolon group is searched sequentially. The *ELKB* contains methods to retrieve the *Paragraphs* and *SG* objects in constant time if the references are translated into absolute addresses, for example the reference `contempt 922 int.` could be changed to `922.5.1.1.3.`, which would represent Head `922`, part-of-speech label `interjection`, 1st Paragraph, 1st Semicolon group and third word. As the lexical material does not generally change, these references are not required to be calculated often, but the transformation procedure has not been implemented in this version of the *ELKB*. There are 990 *Head*, 6,432 *Paragraph*, and 59,877 *SG* objects in this computerized *Roget's*.

### 3.6.4   The *Index* and Related Classes

The *Index* contains all of the words and phrases found in the 990 Heads. It consists of a hash table of *Index* entries, which represent the words and phrases of the *Thesaurus*, with pointers to their corresponding references, as well as an array of all distinct references in the *ELKB*. The same reference is used for all of the words or phrases in a Paragraph. It is more economical to store this reference once and to have the entries maintain a pointer to it, rather than to store it every time it is used. The size of the *Index* object is a major concern, as it must be stored in memory to ensure rapid access. Finding an entry in the *Index* hash table and looking up its corresponding references is performed in constant time. The references are stored in *Reference* objects and their pointers are encoded as Strings, which represent their addresses in the references array, separated by semicolons. For example, the Index words `know-how`, `stealth`, `diplomacy` and `intrigue` all have the reference *cunning 698 n.* as do all the other words that are located in this *Roget's* Paragraph. Instead of storing this reference with every entry that it belongs to, only the location in an array of unique references is kept. Since an entry may have several references, all the addresses are kept, for example `345:456:2045:12374`. The *Morphy* and *Variant* classes are used by the *Index* object to perform transformations that allow retrieving words and phrases written in forms that are not contained in the *Thesaurus*. These classes and the transformations are described in the following section. There are 104,333 *Index* entries and 223,219 total References, and 6,432 unique *Reference* objects in the *ELKB*.

### 3.6.5 Morphological Transformations

The ability to perform morphological transformations is essential for a good lexical knowledge base. The lemma of a word or phrase written in British English is generally stored in the Index, as this is the form in which it also appears in the 1987 edition of Penguin's *Roget's Thesaurus*. If an inflected word is passed to the *ELKB* it must be transformed so as to retrieve the appropriate References. Three tools are available to execute the passage of an input string into a recognized form: a file containing pairs of strings in American and British spelling; rules for detaching inflectional endings to obtain base forms; and exception list files for nouns, verbs, adjectives and adverbs in which inflected forms can be searched and base forms found. Appendix G presents the American and British spelling word list used by the *ELKB*. It contains 646 pairs of spelling variations and has been compiled from various lists freely available on the web. The rules, presented in Table 3.6, and the exception lists were taken from *WordNet 1.7.1*. The exception lists are quite extensive: 5992 pairs in the `noun.exc` file, 5285 in the `verb.exc`, 1486 in `adj.exc` and 7 in the `adv.exc` file. These transformations are performed in the *Index* class using the *Variant* and *Morphy* classes described previously. There is no method for identifying the part-of-speech of an input string in the *ELKB*. For this reason all the detachment rules and exception files are applied and searched. Although this is not the best implementation, it allows a good recall of the References stored in the Index.

Retrieving phrases from the Index is problematic for the *ELKB*. There are many in *Roget's*, a lot of which are specific to British English, for example: `man on the Clapham omnibus`, `drunk as David's sow` or come from other languages, for example: `faute de mieux`, `Alea jacta est`. Once again, the exact string must be entered for the corresponding references to be returned. Verbs will not be found if they are preceded by `to` or `be` for example: `to offer`, `be disorderly`. This problem can be easily circumvented, but has not been implemented in the *ELKB* as it is very difficult to conceive rules that deal with all possible verb phrases that contain prepositions, for example: `ask for it`. There are several possible solutions to the problem of phrases. They could be indexed under every word in the phrase or a method that retrieves all the phrases in the Index that contain specific words could be implemented, but none has been so far. This is detrimental to the *ELKB*, as *Roget's Thesaurus* has a very rich collection of phrases, but has not been a hindrance for obtaining good results in NLP applications, as is presented in Chapter 4 and 5.

| *Part-of-speech* | *Suffix* | *Transformation* |
|---|---|---|
| Noun | s | |
| Noun | ses | s |
| Noun | xes | x |
| Noun | zes | z |
| Noun | ches | ch |
| Noun | shes | sh |
| Noun | men | man |
| Noun | ies | y |
| Verb | s | |
| Verb | ies | y |
| Verb | es | e |
| Verb | es | |
| Verb | ed | e |
| Verb | ed | |
| Verb | ing | e |
| Verb | ing | e |
| Adjective | er | |
| Adjective | est | |
| Adjective | er | e |
| Adjective | est | e |

**Table 3.6:** Transformation rules for the various parts-of-speech.

### 3.6.6   Basic Operations of the *ELKB*

The basic operations of the *ELKB* are:

- finding the references for a given word or phrase in the Index

- looking up a given reference in the Text

- traversing the taxonomy to calculate the distance between words and phrases

- identifying the type of relationship that exists between words and phrases as defined by their location in the hierarchy

A host of NLP application can be implemented using these four basic operations. Three are presented in this dissertation: measuring semantic similarity, building lexical chains, and mapping *Roget's Thesaurus* onto *WordNet*. The first uses the taxonomy to measure the distances, the second relies on properties of the *Thesaurus* to identify specific relations, and the third exploits word and phrase look-up as well as *Roget's* hierarchy. In the current implementation of the ELKB, the slowest operation is the word and phrase look up. This is due to the fact that Heads are read from a file, and sequential searches performed on the *Paragraph* and *SG* arrays to find the location of the symbolic references. Performance can be greatly increased by loading the 990 Heads into memory and using absolute addresses for the references. Both of these are realistic improvements.

# 4   Using *Roget's Thesaurus* to Measure Semantic Similarity

Measuring semantic similarity with the *ELKB* allows us to present a first application of the system as well as to perform a qualitative evaluation. In this chapter, we define the notions of synonymy and semantic similarity and explain a metric for calculating similarity based on *Roget's* taxonomy. We evaluate it using a few typical tests. The experiments in this chapter compare the synonymy judgments of the system to gold standards established by Rubenstein and Goodenough (1965), Miller and Charles (1991) as well as Finkelstein *et al.* (2002; Gabrilovich 2002) for assessing the similarity of pairs of words. We further evaluate the metric by using the system to answer Test of English as a Foreign Language [*TOEFL*] (Landauer and Dumais, 1997) and English as a Second Language tests [*ESL*] (Turney, 2001), as well as the Reader's Digest Word Power Game [*RDWP*] (Lewis, 2000-2001) questions where a correct synonym must be chosen amongst four target words. We compare the results to six other *WordNet*-based metrics and two statistical methods.

## 4.1   The notions of synonymy and semantic similarity

People identify synonyms — strictly speaking, near-synonyms (Edmonds and Hirst, 2002) — such as `angel - cherub`, without being able to define synonymy properly. The term tends to be used loosely, even in the crucially synonymy-oriented *WordNet* with the synset as the basic semantic unit (Fellbaum, 1998, p. 23). Miller and Charles (1991) restate a formal, and linguistically quite inaccurate, definition of synonymy usually attributed to Leibniz: "two words are said to be synonyms if one can be used in a statement in place of the other without changing the meaning of the statement". With this strict definition there may be no perfect synonyms in natural language (Edmonds and Hirst, *ibid.*). Computational linguists often find it more useful to establish the degree of synonymy between two words, referred to as semantic similarity.

Miller and Charles' semantic similarity is a continuous variable that describes the degree of synonymy between two words (*ibid.*). They argue that native speakers can order pairs of words by semantic similarity, for example `ship - vessel`, `ship - watercraft`, `ship - riverboat`, `ship - sail`, `ship - house`, `ship - dog`, `ship - sun`. The concept can be usefully extended to quantify relations between non-synonymous but closely related words, for example `airplane - wing`.

Rubenstein and Goodenough (1965) investigated the validity of the assumption that "... pairs of words which have many contexts in common are semantically closely related". This led them to establish *synonymy judgments* for 65 pairs of nouns with the help of human experts. Miller and Charles (*ibid.*) selected 30 of those pairs, and studied semantic similarity as a function of the contexts in which words are used. Others have calculated similarity using semantic nets (Rada *et al.*, 1989), in particular *WordNet* (Resnik, 1995; Jiang and Conrath, 1997; Lin, 1998; Hirst and St-Onge, 1998; Leacock and Chodorow, 1998) and *Roget's Thesaurus* (McHale, 1998), or statistical methods (Landauer and Dumais, 1997; Turney, 2001). Terra and Clarke (2003) present a survey of statistical methods. This leads naturally to combined approaches that rely on statistical methods enhanced with information contained in *WordNet* (Finkelstein *et al.*, 2002) and methods that merge the results of various statistical systems (Bigham *et al.*, 2003).

The objective is to test the intuition that *Roget's Thesaurus*, sometimes treated as a book of synonyms, allows to measure semantic similarity effectively. We propose a measure of *semantic distance*, the inverse of semantic similarity (Budanitsky and Hirst, 2001) based on *Roget's* taxonomy. We convert it into a semantic similarity measure, and empirically compare it to human judgments and to those of NLP systems. We evaluate the measure by performing the task of assigning a similarity value to pairs of nouns and choosing the correct synonym of a problem word given the choice of four target words. This chapter explains in detail the measures and the experiments, and draws a few conclusions.

## 4.2   Edge counting as a metric for calculating synonymy

*Roget's* structure provides an easy mechanism for calculating the semantic distance using edge counting. Given two words, the system looks up the corresponding references in the index, and then calculates all paths between the references using the taxonomy. The distance value is equal to the number of edges in the shortest path as indicated in Table 4.1. For example, the distance between `feline` and `lynx` is 2. It can be calculated as follows:

```
The word feline has the following references in Roget's:
   1) animal 365 ADJ.
   2) cat 365 N.
   3) cunning 698 ADJ.

The word lynx has the following references in Roget's:
   1. cat 365 N.
   2. eye 438 N.
```

These six paths are obtained:

```
Path between feline (cat 365 N.) and lynx (cat 365 N.)    [ length = 2 ]
feline → cat ← lynx
```

| Distance value | Shortest path | Example |
|---|---|---|
| 0 | same semicolon group | *journey's end – terminus* |
| | | In head **295 Arrival**, **N.**, *goal* |
| | | *journey's end*, final point, point of no return, *terminus* 69 *extremity* |
| 2 | same paragraph | *devotion – abnormal affection* |
| | | In head **887 Love**, **N.**, *love* |
| 4 | same part of speech | *popular misconception – glaring error* |
| | | In head **495 Error**, **N.**, *error* |
| | | In head **495 Error**, **N.**, *mistake* |
| 6 | same head | *individual – lonely* |
| | | In head **88 Unity**, **N.**, *unit* |
| | | In head **88 Unity**, **Adj.**, *alone* |
| 8 | same head group | *finance – apply for a loan* |
| | | In head group **784 Lending** – **785 Borrowing** |
| | | In head **784 Lending**, **Vb.**, *lend* |
| | | In head **785 Borrowing**, **Vb.**, *borrow* |
| 10 | same sub-section | *life expectancy – herbalize* |
| | | In sub-section **Vitality** |
| | | In head **360 Life**, **N.**, *life* |
| | | In head **368 Botany**, **Vb.**, *botanize* |
| 12 | same section | *Creirwy (love) – inspired* |
| | | In section *5 Religion* |
| | | In head **967 Pantheon**, **N.**, *Celtic deities* |
| | | In head **979 Piety**, **Adj.**, *pietistic* |
| 14 | same class | *translucid – blind eye* |
| | | In class *3 Matter* |
| | | In head **422 Transparency**, **Adj.**, *transparent* |
| | | In head **439 Blindness**, **N.**, *blindness* |
| 16 | in the *Thesaurus* | *nag – like greased lightning* |
| | | In head **891 Resentment**, **Vb.**, *enrage* |
| | | In head **277 Velocity**, **Adv.**, *swiftly* |

**Table 4.1:** Distance values attributed to the various path lengths in the *Thesaurus*.

Path between **feline (animal 365 ADJ.)** and **lynx (cat 365 N.)** **[ length = 6 ]**
*feline → animal → ADJ. → 365. Animality. Animal ← N. ← cat ← lynx*

Path between **feline (animal 365 ADJ.)** and **lynx (eye 438 N.)** **[ length = 12 ]**
*feline → animal → ADJ. → 365. Animality. Animal → [365, 366] → Vitality → Section three : Organic matter ← Sensation ← [438, 439, 440] ← 438. Vision ← N. ← eye ← lynx*

```
Path between feline (cat 365 N.) and lynx (eye 438 N.)  [ length = 12 ]
feline → cat → N. → 365. Animality. Animal → [365, 366] → Vitality →
Section three : Organic matter → ← Sensation → [438, 439, 440] ← 438. Vision
← N. ← eye ← lynx
```

```
Path between feline (cunning 698 ADJ.) and lynx (cat 365 N.)  [ length = 16 ]
feline → cunning → ADJ. → 698. Cunning → [698, 699] → Complex → Section
three : Voluntary action → Class six : Volition: individual volition → T ←
Class three : Matter ← Section three : Organic matter ← Vitality ← [365,
366] ← 365. Animality. Animal ← N. ← cat ← lynx
```

```
Path between feline (cunning 698 ADJ.) and lynx (eye 438 N.)  [ length = 16 ]
feline → cunning → ADJ. → 698. Cunning → [698, 699] → Complex → Section
three : Voluntary action → Class six : Volition: individual volition → T ←
Class three : Matter ← Section three : Organic matter ← Sensation ← [438,
439, 440] ← 438. Vision ← N. ← eye ← lynx
```

**Figure 4.1:** All the paths between `feline` and `lynx` in *Roget's Thesaurus*.

McHale (1998) has also used the *Third Edition of Roget's International Thesaurus* (Berrey and Carruth, 1962) to measure semantic similarity. He calculated the semantic distance between nouns using four metrics: counting the number of edges, the absolute number of words and phrases between two target nouns, and by using measures first presented by Resnik (1995) as well as Jiang and Conrath (1997) for *WordNet*-based systems. McHale finds that edge counting is the best of the implemented *Roget's*-based measures and correlates extremely well with human judges. He calculates semantic similarity using the Miller and Charles (1991) set in which the pairs `cemetery – woodland` and `shore – woodland` have been removed, as the noun `woodland` is not present in *WordNet* version 1.4, the resource to which the results are being compared. McHale obtains a correlation with the gold standard of *r*=.88, which is quite close to *r*=.90 obtained by Resnik (*ibid.*) who repeated the experiment using human judges on the 28 pairs of nouns. Although the publishers of *Roget's International Thesaurus* have not made it publicly available in a machine-tractable format, it does suggest that we can obtain equally good results using the *ELKB*.

Rada et al. (1989) explain that a distance measure in a taxonomy should satisfy the properties of a metric. A function *f(x,y)* is a metric if the following properties are satisfied:

1) *f(x, x)* = 0, zero property,

2) *f(x, y)* = *f(y, x)*, symmetric property

3) *f(x, y)* ≥ 0, positive property, and

4) *f(x, y)* + *f(y, z)* ≥ *f(x, z)*, triangular inequality.

The proposed semantic distance measure using *Roget's* taxonomy is in fact a metric, as it satisfies the four properies:

1) *zero property*: the shortest distance between a word and itself is always zero as it belongs to a semicolon group.

2) *symmetric property*: the shortest distance between two words is equal to the least number of edges between them. Order is not important, and therefore this property holds.

3) *positive property*: the distance value between two words is an integer between 0 and 16.

4) *triangular inequality*: if *x* and *z* belong to the same semicolon group, this property is true as *f(x, z)* = 0, and the sum of any other distance measure will be at least equal to 0. If *x*, *y* and *z* are all in different classes, then *f(x,z)* = 16 and *f(x,y)* + *f(y,z)* = 32. The shortest path between *x* and *z* going through *y* will always be greater or equal to the shortest path between *x* and *z* as the word *y* introduces the extra distance in the taxonomy towards the first common node.

For the purpose of comparing to other experiments, the semantic distance must be transformed into a semantic similarity measure. The literature proposes two formulas to perform this transformation. The first is to subtract the path length from the maximum possible path length (Resnik, 1995):

$$\text{sim}_1 (w_1, w_2) = 16 - [\text{min distance}(r_1, r_2)] \qquad (1)$$

The second is to take the inverse of the distance value plus one (Lin, 1998):

$$\text{sim}_2 (w_1, w_2) = 1 / (1 + [\text{min distance}(r_1, r_2)] ) \qquad (2)$$

In both formulas $r_1$ and $r_2$ are the sets of references for the words or phrases $w_1$ and $w_2$. As the maximum distance in the *Thesaurus* is 16, the values for $\text{sim}_1$ range from 0 to 16 and for $\text{sim}_2$ from 0.059 to 1.000. In both formulas, the more related the two words or phrases are, the larger the score. As the distances are quite small, the second formula can never reach a value close to 0.

The constant 1 which is added to the divider is also quite arbitrary. The first formula is best suited to edge counting as it maintains the same distribution of values as the distance metric. We use it for the experiments presented in this chapter.

## 4.3   An Evaluation Based on Human Judgments

### 4.3.1   The Experiment

Rubenstein and Goodenough (1965) established synonymy judgments for 65 pairs of nouns. They invited 51 judges who assigned to every pair a score between 4.0 and 0.0 indicating semantic similarity. They chose words from non-technical every day English. They felt that, since the phenomenon under investigation was a general property of language, it was not necessary to study technical vocabulary. Miller and Charles (1991) repeated the experiment restricting themselves to 30 pairs of nouns selected from Rubentein and Goodenough's list, divided equally amongst words with high, intermediate and low similarity. More recently, Finkelstein *et al.* (2002) have prepared the *WordSimilarity – 353 Test Collection* (Gabrilovich, 2002) which contains 353 English word pairs along with similarity judgments performed by humans. The set also contains proper nouns and verbs. It is discussed in more detail in the next section.

The three experiments have been repeated using the *Roget's Thesaurus* system. The results are compared to six other similarity measures that rely on *WordNet*. We use Pedersen's *Semantic Distance* software package (2003) with *WordNet 1.7.1* to obtain the results. The first *WordNet* measure used is edge counting. It serves as a baseline, as it is the simplest and most intuitive measure. The next measure, from Hirst and St-Onge (1998), relies on the path length as well as the number of changes of direction in the path; they define these changes in function of *WordNet* semantic relations. Jiang and Conrath (1997) propose a combined approach based on edge counting enhanced by the node-based approach of the information content calculation proposed by Resnik (1995). Leacock and Chodorow (1998) count the path length in nodes rather than links, and adjust it to take into account the maximum depth of the taxonomy. Lin (1998) calculates semantic similarity using a formula derived from information theory. Resnik (1995) calculates the information content of the concepts that subsume them in the taxonomy. We calculate the Pearson product-moment correlation coefficient for the human judgments with the values achieved by the systems. The correlation is significant to at the 0.01 level. These similarity measures appear in Tables 4.2, 4.3 and 4.4.

### 4.3.2 The Results

We begin the analysis with the results obtained by *Roget's*. The Miller and Charles data in Table 4.2 show that pairs of words with a semantic similarity value of 16 have high similarity, those with a score of 12 to 14 have intermediate similarity, and those with a score below 10 are of low similarity. This is intuitively correct, as words or phrases that are in the same Semicolon Group will have a similarity score of 16, those that are in the same Paragraph, part-of-speech or Head will have a score of 10 to 14, and words that cannot be found in the same Head, therefore do not belong to the same concept, will have a score between 0 and 8. *Roget's* results correlate very well with human judgment for the Miller and Charles list ($r$=.878), almost attaining the upper bound ($r$=.885) set by human judges (Resnik, 1995) despite the outlier `crane - implement`, two words that are not related in the *Thesaurus*.

| Noun Pair | Miller Charles | Penguin Roget | WordNet Edges | Hirst St.Onge | Jiang Conrath | Leacock Chodorow | Lin | Resnik |
|---|---|---|---|---|---|---|---|---|
| *car – automobile* | 3.920 | 16.000 | 30.000 | 16.000 | 1.000 | 3.466 | 1.000 | 6.340 |
| *gem – jewel* | 3.840 | 16.000 | 30.000 | 16.000 | 1.000 | 3.466 | 1.000 | 12.886 |
| *journey – voyage* | 3.840 | 16.000 | 29.000 | 4.000 | 0.169 | 2.773 | 0.699 | 6.057 |
| *boy – lad* | 3.760 | 16.000 | 29.000 | 5.000 | 0.231 | 2.773 | 0.824 | 7.769 |
| *coast – shore* | 3.700 | 16.000 | 29.000 | 4.000 | 0.647 | 2.773 | 0.971 | 8.974 |
| *asylum – madhouse* | 3.610 | 16.000 | 29.000 | 4.000 | 0.662 | 2.773 | 0.978 | 11.277 |
| *magician – wizard* | 3.500 | 14.000 | 30.000 | 16.000 | 1.000 | 3.466 | 1.000 | 9.708 |
| *midday – noon* | 3.420 | 16.000 | 30.000 | 16.000 | 1.000 | 3.466 | 1.000 | 10.584 |
| *furnace – stove* | 3.110 | 14.000 | 23.000 | 5.000 | 0.060 | 1.386 | 0.238 | 2.426 |
| *food – fruit* | 3.080 | 12.000 | 23.000 | 0.000 | 0.088 | 1.386 | 0.119 | 0.699 |
| *bird – cock* | 3.050 | 12.000 | 29.000 | 6.000 | 0.159 | 2.773 | 0.693 | 5.980 |
| *bird – crane* | 2.970 | 14.000 | 27.000 | 5.000 | 0.139 | 2.079 | 0.658 | 5.980 |
| *tool – implement* | 2.950 | 16.000 | 29.000 | 4.000 | 0.546 | 2.773 | 0.935 | 5.998 |
| *brother – monk* | 2.820 | 14.000 | 29.000 | 4.000 | 0.294 | 2.773 | 0.897 | 10.489 |
| *lad – brother* | 1.660 | 14.000 | 26.000 | 3.000 | 0.071 | 1.856 | 0.273 | 2.455 |
| *crane – implement* | 1.680 | 0.000 | 26.000 | 3.000 | 0.086 | 1.856 | 0.394 | 3.443 |
| *journey – car* | 1.160 | 12.000 | 17.000 | 0.000 | 0.075 | 0.827 | 0.000 | 0.000 |
| *monk – oracle* | 1.100 | 12.000 | 23.000 | 0.000 | 0.058 | 1.386 | 0.233 | 2.455 |
| *cemetery – woodland* | 0.950 | 6.000 | 21.000 | 0.000 | 0.049 | 1.163 | 0.067 | 0.699 |
| *food – rooster* | 0.890 | 6.000 | 17.000 | 0.000 | 0.063 | 0.827 | 0.086 | 0.699 |
| *coast – hill* | 0.870 | 4.000 | 26.000 | 2.000 | 0.148 | 1.856 | 0.689 | 6.378 |
| *forest – graveyard* | 0.840 | 6.000 | 21.000 | 0.000 | 0.050 | 1.163 | 0.067 | 0.699 |
| *shore – woodland* | 0.630 | 2.000 | 25.000 | 2.000 | 0.056 | 1.674 | 0.124 | 1.183 |
| *monk – slave* | 0.550 | 6.000 | 26.000 | 3.000 | 0.063 | 1.856 | 0.247 | 2.455 |
| *coast – forest* | 0.420 | 6.000 | 24.000 | 0.000 | 0.055 | 1.520 | 0.121 | 1.183 |
| *lad – wizard* | 0.420 | 4.000 | 26.000 | 3.000 | 0.068 | 1.856 | 0.265 | 2.455 |
| *chord – smile* | 0.130 | 0.000 | 20.000 | 0.000 | 0.066 | 1.068 | 0.289 | 2.888 |
| *glass – magician* | 0.110 | 2.000 | 23.000 | 0.000 | 0.056 | 1.386 | 0.123 | 1.183 |
| *rooster – voyage* | 0.080 | 2.000 | 11.000 | 0.000 | 0.044 | 0.470 | 0.000 | 0.000 |
| *noon – string* | 0.080 | 6.000 | 19.000 | 0.000 | 0.052 | 0.981 | 0.000 | 0.000 |
| *Correlation* | 1.000 | 0.878 | 0.732 | 0.689 | 0.695 | 0.821 | 0.823 | 0.775 |

**Table 4.2:** Comparison of semantic similarity measures using the Miller and Charles data.

The correlation between human judges and *Roget's* for the Rubenstein and Goodenough data is also very good ($r$=.818) as shown in Table 4.3. Appendix I presents the 65 pairs of nouns. The outliers merit discussion. *Roget's* deems five pairs of low similarity words to be of intermediate similarity, all with the semantic distance value of 12. We therefore find these pairs of words all under the same Head and belonging to noun groups. The *Thesaurus* makes correct associations but not the most intuitive ones: `glass - jewel` is assigned a value of 1.78 by the human judges but can be found under the Head **844** `Ornamentation`, `car - journey` is assigned 1.55 and is found under the Head **267** `Land travel`, `monk - oracle` 0.91 found under Head **986** `Clergy`, `boy - rooster` 0.44 under Head **372** `Male`, and `fruit - furnace` 0.05 under Head 301 `Food: eating and drinking`.

|  | **Rubenstein Goodenough** | **Penguin Roget** | **WordNet Edges** | **Hirst St.Onge** | **Jiang Conrath** | **Leacock Chodorow** | **Lin** | **Resnik** |
|---|---|---|---|---|---|---|---|---|
| *Correlation* | 1.000 | 0.818 | 0.787 | 0.732 | 0.731 | 0.852 | 0.834 | 0.800 |

**Table 4.3:** Comparison of semantic similarity measures using the Rubenstein and Goodenough data.

We have also performed the same experiment on the *WordSimilarity – 353 Test Collection*. The correlation of *Roget's* measure with human judges is $r$=.539, which seems quite low, but is still better than the best *WordNet* based measure, $r$=.375, obtained using Resnik's function and comparable to Finkelstein *et al.*'s combined metric which obtains a score of $r$=.550. Table 4.4 summarizes these results and Appendix J presents the entire 353 word pair list. We cannot simply attribute the low scores to the measures not scaling up to larger data sets. The Finkelstein *et al.* list contains pairs that are associated but not similar in the semantic sense, for example: *liquid – water*. The list also contains many culturally biased pairs, for example: *Arafat – terror* and verbs. Table 4.5 presents all of the pairs for which at least one word is not present in *Roget's*. These can be placed in five categories: proper nouns, verbs, new words that were not in widespread use in 1986, words for which the plural is present in *Roget's* but not its singular form, and words that are simply not in the *Thesaurus*. The authors of the list describe it as representing various degrees of similarity and write that they employed 16 subjects to rate the semantic similarity on a scale from 0 to 10, 0 representing totally unrelated words and 10 very much related or identical words (Finkelstein *et al.*, 2002). They do not explain the methodology used for preparing this list. Human subjects find it more difficult to use a scale from 0 to 10 rather than 0 to 4. These issues cast a doubt on the validity of this list, and we therefore do not consider it as a suitable benchmark for performing experiments on semantic similarity.

| | *Finkelstein et al.* | *Penguin Roget* | *WordNet Edges* | *Hirst St.Onge* | *Jiang Conrath* | *Leacock Chodorow* | *Lin* | *Resnik* |
|---|---|---|---|---|---|---|---|---|
| *Correlation* | 1.000 | 0.539 | 0.276 | 0.344 | 0.357 | 0.361 | 0.368 | 0.375 |
| *Number of word pairs not found* | | 20 | 8 | 4 | 8 | 8 | 8 | 8 |

**Table 4.4:** Comparison of semantic similarity measures using the Finkelstein et al. data.

| *Word pair* | *Words not found in Roget's Thesaurus* |
|---|---|
| **Maradona – football** | **Maradona** is not in *Roget's*. |
| **Jerusalem – Israel** | **Israel** is not in *Roget's*. |
| **Harvard – Yale** | **Harvard** and **Yale** are not in *Roget's*. |
| **Jerusalem – Palestinian** | **Palestinian** is not in *Roget's*. |
| **Arafat – terror** | **Arafat** is not in *Roget's*. |
| **Arafat – peace** | **Arafat** is not in *Roget's*. |
| **Arafat – Jackson** | **Arafat** and **Jackson** are not in *Roget's*. |
| **Psychology – Freud** | **Freud** is not in *Roget's*, but **Freudian psychology** is [reference: *psychology 477 n.*]. |
| **Mexico – Brazil** | **Mexico** and **Brazil** are not in *Roget's*, but **Brazil nut** is [reference: *fruit 301 n.*]. |
| **Japanese – American** | **Japanese** and **American** are not *Roget's*, but **un-American** [reference: *extraneous 59 adj.*], **American mustard** [reference: *condiment 389 n.*] and **American organ** [reference: *organ 414 n.*] are all in the *Thesaurus*. |
| **Drink – eat** | **eat** can be a verb or an interjection in *Roget's*. |
| **money – laundering** | **laundering** is a verb in *Roget's*. |
| **fuck – sex** | **fuck** is a verb in *Roget's* [reference: *unite with 45 VB.*] but **sex** and **fucking** appear in the same semicolon group [reference: *coition 45 N.*]. **fuck**, **fucking** and **sex** all appear under the same Head, *45 Union*. |
| **hundred – percent** | **percent** is not in the index, but the phrase **hundred per cent** is [reference: *hundred 99 N.*] as well as **per cent** [reference: *ratio 85 N.*]. |
| **video – archive** | **archive** is not in *Roget's*, but **archives** is [references: *record 548 N., collection 632 N., title deed 767 N.* ]. |
| **grocery – money** | **grocery** is not in *Roget's* but **groceries** is [reference: *provisions 301 N.*]. |
| **computer – internet** | **internet** is not in *Roget's*. |
| **Stock – CD** | **CD** is not in *Roget's*. |
| **aluminum – metal** | **aluminum** as well as **aluminium** are not in *Roget's*. |
| **cup – tableware** | tableware is not in *Roget's*. |

**Table 4.5:** Finkelstein et al. word pairs not found in *Roget's Thesaurus*.

Resnik (1995) argues that edge counting using *WordNet* 1.4 is not a good measure of semantic similarity as it relies on the notion that links in the taxonomy represent uniform distances. Tables 4.2 and 4.3 show that this measure performs well for *WordNet 1.7.1* . It is most probable that George Miller's team has much improved the lexical databases' taxonomy and that the distances between words are more uniform, but the goal of this dissertation is not to investigate the improvements made to *WordNet*. Table 4.6 shows that it is difficult to replicate accurately experiments using *WordNet*-based measures. Budanitsky and Hirst (2001) repeated the Miller and Charles experiment using the *WordNet* similarity measures of Hirst and St-Onge (1998), Jiang and Conrath (1997), Leacock and Chodorow (1998), Lin (1998) and Resnik (1995). They claim that the discrepancies in the results can be explained by minor differences in implementation, different versions of *WordNet*, and differences in the corpora used to obtain the frequency data used by the similarity measures. Pedersen's software (2003) does not yield the exact results either. We concur with Budanitsky and Hirst, pointing out that the Resnik, Leacock and Chodorow as well as the Lin experiments were performed not using the entire Miller and Charles set, but a 28 noun-pair subset discussed previously.

| | *Resnik* | *Jiang Conrath* | *Lin* | *Leacock Chodorow* | *Hirst St-Onge* |
|---|---|---|---|---|---|
| *Original results* | 0.791 | 0.828 | 0.834 | N./A. | N./A. |
| *Budanitsky Hirst ( 28 pairs )* | 0.774 | 0.850 | 0.829 | 0.816 | 0.744 |
| *Distance 0.11 ( 28 pairs )* | 0.778 | 0.687 | 0.841 | 0.831 | 0.682 |
| *Distance 0.11 ( 30 pairs )* | 0.787 | 0.696 | 0.846 | 0.832 | 0.689 |

**Table 4.6:** Comparison of correlation values for the different measures using the Miller and Charles data.

## 4.4 An Evaluation Based on Synonymy Problems

### 4.4.1 The Experiment

Another method of evaluating semantic similarity metrics is to see how well the different measures can score on a standardized synonymy test. Such tests have questions where the correct synonym is one of four possible choices. *TOEFL* (Landauer and Dumais, 1997), *ESL* (Turney, 2001), and *RDWP* (Lewis, 2000-2001) contain these kinds of questions. Although this evaluation method is not widespread in NLP, researchers have used it in Psychology (Landauer and Dumais, *ibid.*) and Machine Learning (Turney, *ibid.*). The experimental question set consists of 80 *TOEFL* questions provided by the Educational Testing Service via Thomas Landauer, 50 *ESL* questions created by Donna Tatsuki for Japanese *ESL* students (Tatsuki, 1998), 100 *RDWP* questions gathered by Peter Turney and 200 *RDWP* questions gathered from 2000 – 2001 issues of the Canadian edition of Reader's Digest (Lewis, *ibid.*) by Tad Stach.

A *RDWP* question is presented like this: "`Check the word or phrase you believe is nearest in meaning. ode – A: heavy debt. B: poem. C: sweet smell. D: surprise.`" (Lewis, 2001, n. 938). The *ELKB* calculates the semantic distance between the problem word and each choice word or phrase. The choice word with the shortest semantic distance becomes the solution. Choosing the word or phrase that has the most paths with the shortest distance breaks ties. Phrases that cannot be found in the *Thesaurus* present a special problem. The distance between each word in the choice phrase and the problem word is calculated; we ignore the conjunction `and`, the preposition `to`, and the verb `be`. The system considers the shortest distance between the individual words of the phrase and the problem word as the semantic distance for the phrase. This technique, although simplistic, can deal with phrases like `rise and fall`; `to urge`; and `be joyous` that may not be found in the *Thesaurus*. The *ELKB* is not restricted to nouns when finding the shortest path – it considers nouns, adjectives, verbs and adverbs. Using the previous *RDWP* example, the system would output the following:

- `ode` N. to `heavy debt` N., length = 12, 42 path(s) of this length

- `ode` N. to `poem` N., length = 2, 2 path(s) of this length

- `ode` N. to `sweet smell` N., length = 16, 6 path(s) of this length

- `ode` N. to `surprise` VB., length = 12, 18 path(s) of this length

  ⇒ `Roget thinks that ode means poem: CORRECT`

**Figure 4.2:** Solution to a *RDWP* question using the *ELKB*.

We put the *WordNet* semantic similarity measures to the same task of answering the synonymy questions. The purpose of this experiment is not to improve the measures, but to use them as a comparison for the *ELKB*. The answer is the choice word that has the largest semantic similarity value with the problem word, except for edge-counting where the system picks the smallest value, which represents the shortest distance. When ties occur, a partial score is given; .5 if two words are tied for the highest similarity value, .33 if three, and .25 if four. The results appear in Tables 4.7 to 4.9. Appendix K presents the output of the *ELKB* and the systems using *WordNet*-based measures implemented using the *Semantic Distance* software package (Pedersen, 2003). We have not tailored the *WordNet* measures to the task of answering these questions. All of them, except Hirst and St-Onge, rely on the IS-A hierarchy to calculate the path between words. This implies that these measures have been limited to finding similarities between nouns, as the *WordNet* hyponym tree only exists for nouns and verbs; there are hardly any links between parts of speech. We have not implemented special techniques to deal with phrases. It is therefore quite probable that the *WordNet*-based similarity measures can be improved for the task of answering synonymy questions.

This experiment also compares the results to those achieved by state-of-the-art statistical techniques. Latent Semantic Analysis (*LSA*) is a general theory of acquired similarity and knowledge representation (Landauer and Dumais, 1997). It was used to answer the 80 *TOEFL* questions. The algorithm, called *PMI-IR* (Turney, 2001), uses Pointwise Mutual Information (PMI) and Information Retrieval (IR) to measure the similarity of pairs of words. Turney has evaluated it using the *TOEFL* and *ESL* questions. Researchers have determined the best statistical methods (Terra and Clarke, 2003; Bigham et al., 2003) and evaluated them using the same 80 *TOEFL* problems.

### 4.4.2 The Results

The *ELKB* answers 78.75% of the *TOEFL* questions (Table 4.7). The two next best systems are Hirst St-Onge and *PMI-IR*, which answer 77.91% and 73.75% of the questions respectively. *LSA* is not too far behind, with 64.38%. Terra and Clarke (2003) obtained a score of 81.25% using a statistical technique similar to Turney's. The discrepancies in results are most probably due to differences in the corpora used to measure the probabilities. By combining the results of four statistical methods, including *LSA* and *PMI-IR*, Bigham *et al.* (2003) obtain a score of 97.50%. They further declare the problem of this *TOEFL* set to be "solved". All the other *WordNet*-based measures perform poorly, with accuracy not surpassing 25.0%. According to Landauer and Dumais (*ibid.*), a large sample of applicants to US colleges from non-English speaking countries took the *TOEFL* tests containing these items. Those people averaged 64.5%, considered an adequate score for admission to many US universities.

| | Penguin Roget | WordNet Edges | Hirst St.Onge | Jiang Conrath | Leacock Chodorow | Lin | Resnik | PMI-IR | LSA |
|---|---|---|---|---|---|---|---|---|---|
| *Correct* | 63 | 17 | 57 | 20 | 17 | 19 | 15 | 59 | 50 |
| *Questions with ties* | 0 | 1 | 18 | 0 | 1 | 1 | 3 | 0 | 6 |
| *Score* | 63 | 17.5 | 62.33 | 20 | 17.5 | 19.25 | 16.25 | 59 | 51.5 |
| *Percent* | **78.75** | **21.88** | **77.91** | **25.00** | **21.88** | **24.06** | **20.31** | **73.75** | **64.38** |
| *Questions not found* | 4 | 53 | 2 | 53 | 53 | 53 | 53 | 0 | 0 |
| *Other words not found* | 22 | 24 | 2 | 24 | 24 | 24 | 24 | 0 | 0 |

**Table 4.7:** Comparison of the similarity measures for answering the 80 *TOEFL* questions.

| | Penguin Roget | WordNet Edges | Hirst St.Onge | Jiang Conrath | Leacock Chodorow | Lin | Resnik | PMI-IR |
|---|---|---|---|---|---|---|---|---|
| *Correct* | 41 | 16 | 29 | 18 | 16 | 18 | 15 | 37 |
| *Questions with ties* | 0 | 4 | 5 | 0 | 4 | 0 | 3 | 0 |
| *Score* | 41 | 18 | 31 | 18 | 18 | 18 | 16.33 | 37 |
| *Percent* | **82.00** | **36.00** | **62.00** | **36.00** | **36.00** | **36.00** | **32.66** | **74.00** |
| *Questions not found* | 0 | 11 | 0 | 11 | 11 | 11 | 11 | 0 |
| *Other words not found* | 2 | 23 | 2 | 23 | 23 | 23 | 23 | 0 |

**Table 4.8:** Comparison of the similarity measures for answering the 50 *ESL* questions.

| | Penguin Roget | WordNet Edges | Hirst St.Onge | Jiang Conrath | Leacock Chodorow | Lin | Resnik |
|---|---|---|---|---|---|---|---|
| *Correct* | 223 | 68 | 123 | 68 | 68 | 63 | 59 |
| *Questions with ties* | 0 | 3 | 44 | 1 | 3 | 9 | 14 |
| *Score* | 223 | 69.33 | 136.92 | 68.5 | 69.33 | 66.17 | 64 |
| *Percent* | **74.33** | **23.11** | **45.64** | **22.83** | **23.11** | **22.06** | **21.33** |
| *Questions not found* | 21 | 114 | 6 | 114 | 114 | 114 | 114 |
| *Other words not found* | 18 | 340 | 377 | 340 | 340 | 340 | 340 |

**Table 4.9:** Comparison of the similarity measures for answering the 300 *RDWP* questions.

The *ESL* experiment (Table 4.8) presents similar results. Once again, the *Roget's system* is best, answering 82% of the questions correctly. The two next best systems, *PMI-IR* and Hirst and St-Onge fall behind, with scores of 74% and 62% respectively. All other *WordNet* measures give very poor results, not answering more than 36% of the questions. The *Roget's* similarity measure is clearly superior to the *WordNet* ones for the *RDWP* questions (Table 4.9). Roget's answers 74.33% of the questions, which is almost equal to a "Good" vocabulary rating according to Reader's Digest (Lewis, 2000-2001), where the next best *WordNet* measure, Hirst and St-Onge, answers only 45.65% correctly. All others do not surpass 25%.

### 4.4.3   The Impact of Nouns on Semantic Similarity Measures

The *TOEFL*, *ESL* and *RDWP* experiments give a clear advantage to measures that can evaluate the similarity between words of different parts-of-speech. This is the case for *Roget's*, Hirst and St-Onge, and the statistical measures. To be fair to the other *WordNet*-based systems, the experiments have been repeated using subsets of the questions that contain only nouns. The results are presented in Tables 4.10 to 4.12.

| | Penguin Roget | WordNet Edges | Hirst St.Onge | Jiang Conrath | Leacock Chodorow | Lin | Resnik |
|---|---|---|---|---|---|---|---|
| *Correct* | 17 | 14 | 12 | 17 | 14 | 15 | 11 |
| *Questions with ties* | 0 | 0 | 4 | 0 | 0 | 1 | 3 |
| *Score* | 17 | 14 | 13.5 | 17 | 14 | 15.25 | 12.25 |
| *Percent* | **94.44** | **77.78** | **75.00** | **94.44** | **77.78** | **84.72** | **68.06** |
| *Questions not found* | 0 | 1 | 1 | 1 | 1 | 1 | 1 |
| *Other words not found* | 0 | 2 | 2 | 2 | 2 | 2 | 2 |

**Table 4.10:** Comparison of the measures for answering the 18 *TOEFL* questions that contain only nouns.

| | Penguin Roget | WordNet Edges | Hirst St.Onge | Jiang Conrath | Leacock Chodorow | Lin | Resnik |
|---|---|---|---|---|---|---|---|
| *Correct* | 19 | 13 | 16 | 15 | 13 | 15 | 13 |
| *Questions with ties* | 0 | 4 | 2 | 0 | 4 | 0 | 2 |
| *Score* | 19 | 15 | 16.75 | 15 | 15 | 15 | 13.83 |
| *Percent* | **76.00** | **60.00** | **67.00** | **60.00** | **60.00** | **60.00** | **55.32** |
| *Questions not found* | 0 | 0 | 0 | 0 | 0 | 0 | 0 |
| *Other words not found* | 1 | 0 | 0 | 0 | 0 | 0 | 0 |

**Table 4.11:** Comparison of the measures for answering the 25 *ESL* questions that contain only nouns.

The *WordNet* measures perform much more uniformly and yield better results, but the *Roget's* system is still best. The performance of the *ELKB* has increased for the *TOEFL* questions, decreased for the *ESL* and remained about the same for *RDWP*. Although this is not an exhaustive manner of evaluating the efficiency of edge counting as a measure of semantic similarity for various parts-of-speech, it does show that it is effective for nouns, as well as adjectives, verbs and adverbs. Most of the nouns not found in *WordNet* are phrases. For example, the *RDWP* problem "`swatch – A: sample of cloth. B: quick blow. C: petty theft. D: repair of clothing.`" cannot be answered using *WordNet*. The phrases `sample of cloth`; `quick blow`; `petty theft` and `repair of clothing` are simply not in the lexical database. The *ELKB* finds the correct answer by using the technique presented in section 4.4.1. If a tailored method were used to deal with phrases in *WordNet*, the scores of the systems using this resource would definitely improve but this research is beyond the scope of this dissertation as the goal of this thesis is to investigate the usefulness of *Roget's Thesaurus* for NLP.

| | *Penguin Roget* | *WordNet Edges* | *Hirst St.Onge* | *Jiang Conrath* | *Leacock Chodorow* | *Lin* | *Resnik* |
|---|---|---|---|---|---|---|---|
| *Correct* | 115 | 61 | 55 | 62 | 62 | 57 | 53 |
| *Questions with ties* | 0 | 3 | 20 | 1 | 3 | 8 | 13 |
| *Score* | 115 | 62.33 | 61.5 | 62.5 | 63.33 | 59.83 | 57.67 |
| *Percent* | **74.68** | **40.47** | **39.94** | **40.58** | **41.12** | **38.85** | **37.45** |
| *Questions not found* | 13 | 3 | 3 | 3 | 3 | 3 | 3 |
| *Other words not found* | 5 | 235 | 232 | 235 | 235 | 235 | 235 |

**Table 4.12:** Comparison of the measures for answering the 154 *RDWP* questions that contain only nouns.

### 4.4.4 Analysis of results obtained by the *ELKB* for *RDWP* questions

Twenty *RDWP* questions are presented in every issue of Reader's Digest (Lewis, 2000-2001). These questions generally belong to a specific topic, for example: `nature`, `Canadian Forces peace keeping` or `Food preparation, serving and eating`. The results per topic are presented in Table 4.13. *Reader's Digest* gives the following *Vocabulary Ratings* for the human who plays the game:

- **Fair:** 10 – 14 (50% – 70%)

- **Good:** 15 – 17 (75% – 85%)

- **Excellent:** 18 – 20 (90% – 100%)

The issue per issue analysis allows identifying some of the topics that are well represented and some that are not in the *Roget's*. The *ELKB* performs extremely well, obtaining a rating of `Excellent`, for the questions pertaining to `Greek rooted words` and `manners`. This can be attributed to the fact that the first edition of the *Thesaurus* was prepared during the Victorian era by a doctor who was well accustomed to Greek words and good manners. Words are generally not removed from subsequent editions of *Roget's*, but the process of adding new words is a more arduous one, in particular technical terms, as is demonstrated by the low rating of `Fair` obtained the `financial term` set.

| Month | Description | Number of questions | Correct | Percent | Questions not found | Other words not found |
|---|---|---|---|---|---|---|
| *Jan-00* | `Nature` | 20 | 15 | **75.00** | 0 | 2 |
| *Mar-00* | `Words from recent issues of RD` | 20 | 17 | **85.00** | 1 | 1 |
| *Apr-00* | `Financial terms` | 20 | 12 | **60.00** | 5 | 1 |
| *May-00* | `Canadian Forces peace keeping` | 20 | 15 | **75.00** | 1 | 0 |
| *Jun-00* | `Seaside vacation` | 20 | 12 | **60.00** | 4 | 0 |
| *Jul-00* | `Greek rooted words` | 20 | 18 | **90.00** | 0 | 2 |
| *Aug-00* | `Food preparation, serving and eating` | 20 | 14 | **70.00** | 1 | 0 |
| *Sep-00* | `Areas of study` | 20 | 13 | **65.00** | 4 | 0 |
| *Jan-01* | `Manners` | 20 | 18 | **90.00** | 0 | 2 |
| *May-01* | `Character traits` | 20 | 17 | **85.00** | 0 | 2 |
| *Web Set* | `Questions taken RD web site` | 100 | 72 | **72.00** | 5 | 8 |

**Table 4.13:** Score of the *ELKB* for the *RDWP* questions per category.

## 4.5 Summary of results

This chapter has shown that the electronic version of the *ELKB* is as good as, if not better than, *WordNet* for measuring semantic similarity. The distance measure used, often called edge counting, can be calculated quickly and performs extremely well on a series of standard synonymy tests. Table 4.14 summarizes the results for the *Roget's* and *WordNet*-based measures. Out of 8 experiments, the *ELKB* is first every time, except on the Rubenstein and Goodenough list of 65 noun pairs. Combined statistical methods that use the Internet as a corpus perform better, but they access many more words than are contained in either lexical resource.

| Experiment | Penguin Roget | WordNet Edges | Hirst St.Onge | Jiang Conrath | Leacock Chodorow | Lin | Resnik |
|---|---|---|---|---|---|---|---|
| *Miller Charles* | 1 | 5 | 7 | 6 | 3 | 2 | 4 |
| *Rubenstein Goodenough* | 3 | 5 | 6 | 7 | 1 | 2 | 4 |
| *Finkelstein et al.* | 1 | 7 | 6 | 5 | 4 | 3 | 2 |
| *TOEFL* | 1 | 5 | 2 | 3 | 5 | 4 | 7 |
| *ESL* | 1 | 3 | 2 | 3 | 3 | 3 | 7 |
| *Reader's Digest* | 1 | 3 | 2 | 5 | 3 | 6 | 7 |
| *TOEFL - Nouns* | 1 | 4 | 5 | 2 | 4 | 3 | 6 |
| *ESL - Nouns* | 1 | 3 | 2 | 3 | 3 | 3 | 7 |
| *Reader's Digest - Nouns* | 1 | 4 | 5 | 3 | 2 | 6 | 7 |

**Table 4.14:** Summary of results – ranking of similarity measures for the experiments.

Most of the *WordNet*-based systems perform poorly at the task of answering synonym questions. This is due in part to the fact that the similarity measures can only by calculated between nouns, because they rely on the hierarchical structure that is almost only present for nouns in *WordNet*. These systems also suffer from not being able to deal with many phrases. A system that is tailored to evaluate synonymy between pairs of words and phrases might perform much better than what has been presented here.

The *Roget's Thesaurus* similarity measures correlate well with human judges, and perform similarly to the *WordNet*-based measures at assigning synonymy judgments to pairs of nouns. *Roget's* shines at answering standard synonym tests. This result was expected, but remains impressive: the semantic distance measure is extremely simple and no context is taken into account, and the system does not perform word sense disambiguation when answering the questions. Standardized language tests appear quite helpful in evaluating NLP systems, as they focus on specific linguistic phenomena and offer an inexpensive alternative to human evaluation.

# 5   Automating the Construction of Lexical Chains using *Roget's*

Morris and Hirst (1991) present a method of linking significant words that are about the same topic. The resulting lexical chains are a means of identifying cohesive regions in a text, with applications in many natural language processing tasks, including text summarization. Morris and Hirst constructed the first lexical chains manually using *Roget's International Thesaurus*. They wrote that automation would be straightforward given an electronic thesaurus. Most applications so far have used *WordNet* to produce lexical chains, perhaps because adequate electronic versions of *Roget's* were not available until recently. This chapter discusses the building of lexical chains using the electronic version of *Roget's Thesaurus*, the second application of the *ELKB*. We implement a variant of the original algorithm. We explain the necessary design decisions and include a comparison with other implementations. Computational linguists have proposed several evaluation methods, in particular one where they construct lexical chains for a variety of documents and then compare them to gold standard summaries. This chapter discusses related research on the topic of lexical chains.

## 5.1   Previous Work on Lexical Chains

Lexical chains (Morris and Hirst, *ibid.*) are sequences of words in a text that represent the same topic. The original implementation was inspired by the notion of cohesion in discourse (Halliday and Hasan, 1976). An electronic system requires a sufficiently rich and subtle lexical resource to decide on the semantic proximity of words.

Computational linguists have used lexical chains in a variety of tasks, from text segmentation (Morris and Hirst, 1991; Okumura and Honda, 1994), to summarization (Barzilay, 1997; Barzilay and Elhadad, 1997; Brunn, Chali and Pinchak, 2001; Silber and McCoy, 2000, 2002), detection of malapropisms (Hirst and St-Onge, 1998), the building of hypertext links within and between texts (Green, 1999), analysis of the structure of texts to compute their similarity (Ellman, 2000), topic detection (Chali, 2001), and even a form of word sense disambiguation (Barzilay, 1997; Okumura and Honda, 1994). Most of the systems use *WordNet* to build lexical chains, perhaps in part because it is readily available. Building lexical chains is a natural task for *Roget's Thesaurus* as they were conceived using this resource. Ellman (*ibid.*) has used the 1911 edition of *Roget's* and the 1987 edition of Longman's *Original Roget's Thesaurus of English Words and Phrases*. The lexical chain construction process is computationally expensive but the price seems worth paying if lexical semantics can be incorporated in natural language systems.

Our implementation builds the lexical chains using the *ELKB*. The original lexical chain algorithm (Morris and Hirst, *ibid.*) exploits certain organizational properties of *Roget's Thesaurus*. *WordNet*-based implementations cannot take advantage of *Roget's* relations. They also usually only link nouns, as relations between parts-of-speech are limited in *WordNet*. Morris and Hirst wrote: "Given a copy [of a machine readable thesaurus], implementation [of lexical chains] would clearly be straightforward". The goal of this experiment is to test this statement in practice. This work is guided by the efforts of those who originally conceived lexical chains, as well as Barzilay and Elhadad (1997), the first to use a *WordNet*-based implementation for text summarization, and Silber and McCoy (2002), the authors of the most efficient *WordNet*-based implementation.

## 5.2  Lexical Chain Building Algorithms

Algorithms that build lexical chains consider one by one words for inclusion in the chains constructed so far. Important parameters to consider are the lexical resource used, which determines the lexicon and the possible relations between the words, called *thesaural relations* by Morris and Hirst (1991), the thesaural relations themselves, the transitivity of word relations and the distance — measured in sentences — allowed between words in a chain (Morris and Hirst, *ibid.*).

Our lexical chain building process builds *proto-chains*, a set of words linked via thesaural relations. Our implementation refines the proto-chains to obtain the final lexical chains. We summarize the lexical chain building process with these five high levels steps:

1. Choose a set of thesaural relations;
2. Select a set of candidate words;
3. Build all proto-chains for each candidate word;
4. Select the best proto-chains for each candidate word;
5. Select the lexical chains.

### 5.2.1  Step 1: Choose a Set of Thesaural Relations

Halliday and Hasan (1976) have identified five basic classes of dependency relationships between words that allow classifying lexical cohesion. Identifying these relationships in a text is the first step towards constructing lexical chains. These five classes are:

1. Reiteration with identity of reference:
   a. Mary bit into a *peach*.
   b. Unfortunately the *peach* wasn't ripe.

2. Reiteration without identity of reference:

   a. `Mary ate some` *`peaches`*`.`

   b. `She likes` *`peaches`* `very much.`

3. Reiteration by means of a superordinate:

   a. `Mary ate a` *`peach`*`.`

   b. `She likes` *`fruit`*`.`

4. Systematic semantic relation (systematically classifiable):

   a. `Mary likes` *`green`* `apples.`

   b. `She does not like` *`red`* `ones.`

5. Nonsystematic semantic relation (not systematically classifiable):

   a. `Mary spent three hours in the` *`garden`* `yesterday.`

   b. `She was` *`digging`* `potatoes.`

Of the five basic classes of dependency relationships, the first two are easy to identify, the next two are identifiable using a resource such as *Roget's* or *WordNet*. Morris and Hirst identify five types of *thesaural* relations that suggest the inclusion of a candidate word in a chain (1991). Although the fourth edition of *Roget's International Thesaurus* (Chapman, 1977) is used, the relations can be described according to the structure of Penguin's *Roget's Thesaurus*, which has been presented in Chapter 3. The five thesaural relations used are:

1. `Inclusion in the same Head.`

2. `Inclusion in two different Heads linked by a Cross-reference.`

3. `Inclusion in References of the same Index Entry.`

4. `Inclusion in the same Head Group.`

5. `Inclusion in two different Heads linked to a common third Head by a Cross-reference.`

Morris and Hirst state that although these five relations are used "the first two are by far the most prevalent, constituting over 90% of the lexical relationships."

In our implementation, the decision has been made to adopt only a refinement of the first thesaural relation, as it is the most frequent relation, can be computed rapidly and consists of a large set of closely related words. The use of the second relation is computationally expensive and not intuitive. A Cross-reference in *Roget's Thesaurus* belongs to a Semicolon Group and points to another Paragraph in a specific Head. For example, the Cross-reference `137` *`timely`* in the Semicolon Group `;in loco, well-timed, auspicious, opportune, 137` *`timely;`* points from this Semicolon Group in the adjective Paragraph with keyword *`advisable`* in the Head `642` `Good Policy` to the adjective paragraph with keyword `timely` in Head `137` `Occasion: timeliness`. The Cross-reference is therefore a relation from a Semicolon Group to a Paragraph and does not link all words and phrases in a Paragraph to those of the Paragraph to which it points. It is clearly not symmetric and does not link comparable concepts. There are about 10 times more words and phrases in the *Thesaurus* than Cross-references, which suggests that the first relation should be at least 10 times more frequent than the second one.

In conjunction with the first relation, simple term repetition is exploited. All other presented by Morris and Hirst are discarded. The two relations used for the implementation of lexical chains using the *ELKB* are:

1. Repetition of the same word, for example: *Rome, Rome*.

2. Inclusion in the same Paragraph.

Chapter 3 discusses the manner in which words and phrases found under the same Paragraph are related. A large number of them are near-synonyms, or are related by the IS-A and PART-OF relations, as is experimentally shown in Chapter 6.

For the sake of comparison, here are *WordNet* relations that Silber and McCoy (2002) use in their implementation of lexical chains:

1. Two noun instances are identical, and are used in the same sense.

2. Two noun instances are synonyms.

3. The senses of two noun instances are linked by the hypernym / hyponym relation.

4. The senses of two noun instances are siblings in the hypernym / hyponym tree.

The first three relations used by Silber and McCoy have counterparts in the *Roget's Thesaurus* implementation. A sense of a word or phrase can be uniquely identified in the *ELKB* by its location in the taxonomy. The fourth relationship is used to link all of the words and phrases that are hypernyms or hyponyms of a synset. In this manner, `train` and `railroad train` are related as they belong to the same synset, they are related to `boat train` as it is hyponym of `train`. `Car train`, `freight train`, `rattler`, `hospital train`, `mail train`, `passenger train`, `streamliner` and `subway train` are in turn all related to `boat train` as they are hyponyms of `train`. A counterpart of this relation cannot be explicitly found in the *Thesaurus* as it is dependent on *WordNet's* structure, although the synsets that are grouped by these four relations are comparable to the Semicolon Groups that make up a Paragraph, as is discussed in Chapter 6.

Morphological processing must be automated to assess the relation between words. This is done both by *WordNet* and the *ELKB*. A resource that contains proper names and world knowledge, such as the layout of streets in the city of Ottawa, or who is the Prime Minister of Canada, would be a great asset for the construction of lexical chains. This information is not found in *Roget's* or *WordNet*, but could be added in some simplified form, using gazetteers and other knowledge sources, like the *World Gazetteer* (World Gazetter, 2003) or the *Central Intelligence Agency World Factbook* (CIA Factbook, 2002). We have not incorporated this kind of information into the *ELKB*.

### 5.2.2    Step 2: Select a Set of Candidate Words

The building process does not consider repeated occurrences of closed-class words and high frequency words (Morris and Hirst, 1991). Our system removes the words that should not appear in lexical chains using a 980-element stop list, union of five publicly-available lists: *Oracle 8 ConText*, *SMART*, *Hyperwave*, and lists from the *University of Kansas* and *Ohio State University*. The stop list is presented in Appendix H. After eliminating these high frequency words it would be beneficial to identify nominal compounds and proper nouns. Most of the known *WordNet*-based implementations of lexical chains consider only nouns. This may be due to limitations in *WordNet*, in particular the fact that the IS-A hierarchy, essential to most systems, is only developed extensively for nouns. *Roget's* allows building lexical chains using nouns, adjectives, verb, adverbs and interjections. Our implementation considers the five parts-of-speech. Nominal compounds can be crucial in building correct lexical chains, as argued by Barzilay (1997); considering the words *crystal* and *ball* independently is not at all the same thing as considering the phrase *crystal ball*. *Roget's* has a very large number of phrases, but this is not exploited. We have not developed a method for tagging phrases in a text in conjunction with the *ELKB*. *Roget's* contains around 100 000 unique words and phrases, but very few are technical or proper nouns. Any word or phrase that is not in the *Thesaurus* can only be included in a chain via simple repetition.

### 5.2.3    Step 3: Build all Proto-chains for Each Candidate Word

Inclusion in a proto-chain requires a relation between the candidate word and the chain. This is an essential step, open to interpretation. Should all word in the proto-chain be related via a thesaural relation, or is it enough to link adjacent words in the chain? An example of a chain is {`cow, sheep, wool, scarf, boots, hat, snow`} (Morris and Hirst, 1991). Should all of the words in the chain be directly related to one another? This would mean that `cow` and `snow` should not appear in the same chain. Should only specific senses of a word be included in a chain? Should a chain be built on an entire text, or only segments of it? Barzilay (1997) performs word sense disambiguation as well segmentation before building lexical chains. In theory, chains should disambiguate individual senses of words and segment the text in which they are found; in practice this is difficult to achieve. What should be the distance between two words in a chain? These issues are discussed by Morris and Hirst (*ibid.*) but not definitively answered by any implementation. These are serious considerations, as it easy to generate spurious chains.

Silber and McCoy (2002) build all possible proto-chains for the candidate words. All of the words in a chain must be related to one another. The best intermediate chains are kept and become the output to the system. This implementation adopts a similar methodology. All possible proto-chains are built for the set of candidate words. All words in a chain must be related via the two proposed thesaural relations. For example, all the words in the chain `{driving, exciting, hating, setting, set, setting, hated, drive, driving, driven, drove, drove, drove, cut}` can be found in the Head `46` `Disunion` under the following Paragraph once morphological transformations have been applied:

> **Adj.** `set apart`, put aside, set aside, 632 `store`; conserve, 666 `preserve`; mark out, tick off, distinguish, 15 `differentiate`, 463 `discriminate`; single out, pick out, 605 `select`; except, exempt, leave out, 57 `exclude`; boycott, send to Coventry, 620 `avoid`; taboo, black, blacklist, 757 `prohibit`; insulate, isolate, cut off, 235 `enclose`; zone, compartmentalize, screen off, declare a no-go area, 232 `circumscribe`; segregate, ghettoize, sequester, quarantine, maroon, 883 `seclude`; keep apart, hold apart, drive apart.

**Figure 5.1:** The *Roget's Thesaurus* Paragraph `set apart` **46** adj.

Forcing all words to be related allows building cohesive chains. Transitive relations that would allow two words to be related via a third one, for example `sheep` and `scarf` related through `wool`, are not allowed in this implementation. The proto-chains are scored using the procedure described in the next section and the best ones are kept. The text is not segmented; rather the distance in sentences between words in a proto-chain is taken into account by the scoring system.

### 5.2.4 Step 4: Select the Best Proto-chains for Each Candidate Word

As a word or phrase may have several senses, it may also have several proto-chains, but the system must only keep one. Morris and Hirst (1991) identify three factors for evaluating strength of a lexical chain: reiteration, density, defined in terms of the types of thesaural relations that are contained in the chain, and length. The more repetitious, denser and longer the chain, the stronger it is. This notion has been generally accepted by the other implementations of lexical chains, with the addition of taking into account the type of relations used in the chain when scoring its strength. The values in Table 5.1 are used to score the meta-chains. This is similar to Silber and McCoy's (2002) scoring system.

| *thesaural relation* | *sentence window* *1-3 sentences* | *3 – 5 sentences* | *more than* *5 sentences* |
|---|---|---|---|
| *Repetition of the same word* | 1.00 | 1.00 | 1.00 |
| *Inclusion in the same Head* | 1.00 | 0.75 | 0.50 |

**Table 5.1:** Scores attributed to thesaural relations in the meta-chains.

The rationale for these scores is that the repetition of the same term anywhere in a text represents a strong relation. Good writing style encourages the use of synonyms to convey the same idea. These can be found in the same *Roget's* Paragraph as the other words in the chain, but since a Paragraph does not only contain synonyms, this relation is not as strong as reiteration. Since the further two words are in a text, the less chance they have of discussing the same topic, unless it is a reference to previous idea, the relation based on inclusion in the Paragraph decreases in strength as the distance increases. The scores attributed to each relation have been chosen on an ad hoc basis. These values can be refined in conjunction with an accurate evaluation method.

### 5.2.5 Step 5: Select the Lexical Chains

Our system selects the lexical chains from the best proto-chains. In Sibler and McCoy's implementation (2002) a word belongs to only one lexical chain. Most implementations have adopted this strategy. We have as well so as to compare our lexical chains to those of other systems. A word belongs in the chain to which it contributes the most, which means the proto-chain with the highest score. The word is removed from all other proto-chains and their scores are adjusted accordingly. The lexical chain building procedure stops once the best proto-chain is selected for each word.

## 5.3 Step-by-Step Example of Lexical Chain Construction

Ellman (2000) analyses the following quotation, attributed to Einstein, for the purpose of building lexical chains. The words in bold are the candidate words retained by this implementation that uses the *ELKB* after the stop list has been applied.

> We **suppose** a very long **train travelling** along the **rails** with a **constant velocity** v and in the **direction** indicated in **Figure 1**. **People travelling** in this **train** will with **advantage** use the **train** as a rigid **reference-body**; they **regard** all **events** in **reference** to the **train**. Then every **event** which **takes** place along the **line** also **takes** place at a particular point of the **train**. Also, the **definition** of **simultaneity** can be given **relative** to the train in exactly the same way as with **respect** to the **embankment**.

**Figure 5.2:** The Einstein quotation for which lexical chains are built.

Our system builds all possible proto-chains, consisting of at least two words, for each candidate word, proceeding forward through the text. Since most words have multiple senses, they also have multiple proto-chains. For this example, there are 9 proto-chains for the word `suppose`, 167 for `train`, 29 for `travelling`, 1 for `rails`, 2 for `constant`, 7 for `direction`, 3 for `advantage`, 11 for `regard`, 15 for `events`, 131 for `takes` and 2 for `line`. These proto-chains are presented in Appendix L. The chain building procedure considers only the candidate words found between the current location and the end of the file. The number of meta-chains is a function of the number of senses of a word and the number of remaining candidate words to be considered for the chain. The best meta-chains retained for each word by the system ordered by their score with the sense number (which corresponds to the Head in which the word can be found) and line numbers of the first word are:

```
1. train, rails, train, train, train, line, train, train, embankment
   [score: 9.0, sense: 624, line: 1]

2. direction, regard, reference, respect [score: 4.0, sense: 9, line: 1]

3. travelling, travelling, takes, takes [score: 4.0, sense: 981, line: 1]

4. suppose, regard, takes, takes [score: 4.0, sense: 485, line: 1]

5. regard, takes, takes [score: 3.0, sense: 438, line: 2]

6. advantage, takes, takes [score: 3.0, sense: 916, line: 2]

7. takes, takes, respect [score: 3.0, sense: 851, line: 3]

8. constant, rigid [score: 2.0, sense: 494, line: 1]

9. events, event [score: 2.0, sense: 725, line: 2]

10.line, relative [score: 2.0, sense: 27, line: 3]

11.rails, respect [score: 1.75, sense: 924, line: 1]
```

Once it is determined to which meta-chain a word contributes the most, the final lexical chains generated by the system are:

```
1. train, rails, train, train, train, line, train, train, embankment
   [score: 9.0, sense: 624, line: 1]

2. suppose, regard, takes, takes [score: 4.0, sense: 485, line: 1]

3. direction, reference, respect [score: 3.0, sense: 9, line: 1]

4. travelling, travelling [score: 2.0, sense: 981, line: 1]

5. constant, rigid [score: 2.0, sense: 494, line: 1]
```

```
6. events, event [score: 2.0, sense: 725, line: 2]
```

As a comparison, these eight lexical chains are obtained by Ellman (2002):

```
1. train, rails, train, line, train, train, embankment
```

```
2. direction, people, direction
```

```
3. reference, regard, relative-to, respect
```

```
4. travelling, velocity, travelling, rigid
```

```
5. suppose, reference-to, place, place
```

```
6. advantage, events, event
```

```
7. long, constant
```

```
8. figure, body
```

The Einstein quotation was first studied by St-Onge (1995) who obtained the following nine lexical chains using his *WordNet*-based system:

```
1. train, velocity, direction, train, train, train, advantage, reference,
   reference-to, train, train, respect-to, simultaneity
```

```
2. travelling, travelling
```

```
3. rails, line
```

```
4. constant, given
```

```
5. figure, people, body
```

```
6. regard, particular, point
```

```
7. events, event, place, place
```

```
8. definition
```

```
9. embankment
```

The *ELKB* does not generate as many chains as Ellman or St-Onge, but the chains seem to adequately represent the paragraph. The best lexical chains generated by the ELKB *{train, rails, train, train, train, line, train, train, embankment}* and Ellman *{train, rails, train, line, train, train, embankment}* are almost identical. This is to be expected, as they both use *Roget's Thesaurus*. The only difference is the number of repetitions of the nouns *train*, which is an indication that Ellman's implementation is not as rigorous as it should be. It is surprising that the remaining chains are so different, especially since certain words are not even related in the *ELKB*, for example `direction` and `people`, or `advantage` and `event`, as in the chains `{direction, people, direction}` and `{advantage, events, event}`. This is a clear indication that the versions of *Roget's* used by the systems are quite different. Ellman's second chain `{direction, people, direction}` is clearly erroneous since the word `direction` only appears once in the paragraph. St-Onge generates chains that are hard to quantify as coherent compared the ones the two other systems build. In the chain `{train, velocity, direction, train, train, train, advantage, reference, reference-to, train, train, respect-to, simultaneity}` there is no intuitive relation between `velocity` and `respect-to`, although it is possible to consider a transitive relation using other words in the chain. It is odd that `rails` and `line` are not in the same chain as `train`, since these concepts are very closely related. The singleton chains `{definition}` and `{embankment}` are also listed. Lexical chain building systems generally do not consider these as they are too short to represent a cohesive region in a text. This subjective comparison does not allow determining which system is best. An objective way is required for evaluating lexical chains, which we discuss in Section 5.6.

## 5.4   A Comparison to the Original Implementation

Morris' and Hirst's (1991) demonstrate their manual lexical chain procedure on the first section of an article in *Toronto* magazine, December 1987, by Jay Teitel, entitled "Outland". This section presents the text, where the candidate words are highlighted, and compares the lexical chains generated by the *ELKB* to those of the original algorithm.

> *I **spent** the first 19 years of my **life** in the **suburbs**, the **initial** 14 or so relatively **contented**, the last four or five wanting mainly to be elsewhere. The **final** two I **remember** vividly: I **passed** them **driving** to and from the **University** of **Toronto** in a **red** 1962 Volkswagen 1500 **afflicted** with **night blindness**. The **car's lights** never worked - every **dusk** turned into a **kind** of **medieval race** against **darkness**, a **panicky**, **mournful rush north**, away from everything I **knew** was **exciting**, toward everything I **knew** was **deadly**. I **remember** looking through the **windows** at the **commuters mired** in **traffic** beside me and **actively hating** them for their **passivity**. I actually **punched holes** in the **white vinyl ceiling** of the Volks and then, by way of **penance**, wrote beside them the **names** and **phone** numbers of the **girls** I would **call** when I had my own **apartment** in the **city**. One thing I **swore** to myself: I would never **live** in the **suburbs** again.*
>
> *My **aversion** was as much a **matter** of **environment** as it was **traffic** - one particular **piece** of the suburban **setting**: "the **cruel sun**." **Growing** up in the **suburbs** you can get used to a **surprising** number of things - the **relentless** "**residentialness**" of your **surroundings**, the **weird certainty** you have that everything will **stay** vaguely **new-looking** and **immune** to **historic soul** no **matter** how many years **pass**. You don't **notice** the eerie **silence** that **descends** each **weekday** when every **sound** is **drained** out of your **neighbourhood** along with all the **people who've** gone to work. I got used to **pizza**, and **cars**, and the fact that the **cultural hub** of my **community** was the **collective TV*** *set. But once a **week** I would **step** outside as **dusk** was about to **fall** and be **absolutely** bowled over by the **setting sun**, **slanting huge** and **cold** across the **untreed front lawns**, **reminding** me not just how **barren** and **sterile**, but how **undefended life** could be. As much as I **hated** the suburban **drive** to **school**, I wanted to get away from the **cruel suburban sun**.*

*When I was **married** a few years later, my **attitude** hadn't **changed**. My **wife** was a **city girl** herself, and although her **reaction** to the **suburbs** was less **intense** than mine, we **lived** in a **series** of **apartments** safely **straddling Bloor Street**. But four years **ago**, we had a second **child**, and **simultaneously** the **school** my **wife taught** at moved to **Bathurst Street north** of **Finch Avenue**. She was now **driving** 45 **minutes** north to work every **morning**, along a **route** that was **perversely identical** to the one I'd **driven** in **college**.*

*We **started** looking for a **house**. Our first **limit** was **St. Clair** - we would go no **farther north**. When we took a **closer** look at the **price tags** in the area though, we **conceded** that maybe we'd have to go to **Eglinton** - but that was definitely it. But the **streets** whose **names** had once been **magical barriers**, **latitudes** of **tolerance**, **quickly changed** to something else as the **Sundays passed**. **Eglinton** became **Lawrence**, which became **Wilson**, which became **Sheppard**. One **windswept day** in May I **found** myself **sitting** in a **town-house development north** of **Steeles Avenue** called **Shakespeare Estates**. It wasn't until we **stepped** outside, and the **sun**, **blazing unopposed** over a **country club**, **smacked** me in the **eyes**, that I came to. It was the **cruel sun**. We got into the **car** and **drove** back to the **Danforth** and **porches** as **fast** as we could, **grateful** to have been **reprieved**.*

*And then one **Sunday** in **June** I **drove north** alone. This **time** I drove up **Bathurst** past my **wife's** new **school**, **hit Steeles**, and kept going, beyond **Centre Street** and **past Highway** 7 as well. I **passed farms**, a man **selling lobsters** out of his **trunk** on the **shoulder** of the **road**, a **chronic care hospital**, a **country club** and what **looked** like a **mosque**. I **reached** a **light** and turned right. I saw a **sign** that said **Houses** and turned right again.*

*In **front** of me **lay** a **virgin crescent cut** out of **pine bush**. A **dozen houses** were going up, in various **stages** of **construction**, **surrounded** by **hummocks** of **dry earth** and **stands** of **precariously tall trees nude halfway** up their **trunks**. They were the **kind** of **trees** you might see in the **mountains**. A **couple** was **walking hand-in-hand** up the **dusty dirt roadway**, **wearing matching blue track suits**. On a "**front lawn**" beyond them, several little **girls** with **hair** exactly the same **colour** of **blond** as my **daughter's** were **whispering** and **laughing** together. The **air smelled** of **sawdust** and **sun**.*

*It was a **suburb**, but somehow different from any **suburb** I knew. It felt **warm**.*

*It was **Casa Drive**.*

*In 1976 there were 2,124,291 **people** in **Metropolitan Toronto**, an area **bordered** by **Steeles Avenue** to the **north**, **Etobicoke Creek** on the **west**, and the **Rouge River** to the **east**. In 1986, the same area **contained** 2,192,721 **people**, an **increase** of 3 **percent**, all but **negligible** on an **urban scale**. In the same **span** of **time** the three **outlying regions stretching** across the **top** of **Metro** - **Peel**, **Durham**, and **York** - **increased** in **population** by 55 **percent**, from 814,000 to some 1,262,000. **Half** a million **people** had **poured** into the **crescent north** of **Toronto** in the **space** of a **decade**, during which **time** the **population** of the **City** of **Toronto** actually **declined** as did the **population** of the "old" **suburbs** with the **exception** of **Etobicoke** and **Scarborough**. If the*

> ***sprawling agglomeration*** *of* **people** *known as* **Toronto** *has* **boomed** *in the* **past** *10 years it has*
> **boomed** *outside the* **traditional city confines** *in a totally new* **city***, a new* **suburbia** *containing one*
> *and a* **quarter** *million* **people***.*

**Figure 5.3:** The first section of the Outland article.

The *ELKB* generates 110 lexical chains. The first 9 are presented here, the remaining are in
Appendix L.

1. *suburbs, commuters, city, suburbs, suburbs, community, city, suburbs,*
   *closer, streets, road, crescent, houses, suburb, suburb, urban,*
   *crescent, suburbs, sprawling, city, city, suburbia [score: 20.0, sense:*
   *192, line: 1]*

2. *life, lights, rush, notice, weekday, week, fall, life, minutes, morning,*
   *day, time, light, span, time, stretching, decade, time, quarter [score:*
   *17.0, sense: 110, line: 1]*

3. *driving, exciting, hating, setting, set, setting, hated, drive, driving,*
   *driven, drove, drove, drove, cut [score: 12.75, sense: 46, line: 2]*

4. *north, north, north, limit, north, north, north, north, top, north*
   *[score: 10.0, sense: 213, line: 3]*

5. *girls, people, girl, virgin, girls, people, people, people, people,*
   *people [score: 8.75, sense: 132, line: 5]*

6. *spent, passed, pass, moved, passed, fast, past, past, passed, wearing,*
   *past [score: 8.5, sense: 111, line: 1]*

7. *final, night, call, house, called, hit, stages, construction, stands,*
   *whispering [score: 8.25, sense: 594, line: 2]*

8. *sun, sun, sun, sun, sun, air, sun, space [score: 7.0, sense: 383, line:*
   *7]*

Morris and Hirst identified the 9 following lexical chains:

1. *suburbs, driving, Volkswagen, car's, lights, commuters, traffic, Volks,*
   *apartment, city, suburbs, traffic, suburban, suburbs, residentialness,*
   *neighbourhood, community, suburban, drive, suburban, city, suburbs,*
   *apartments, Bloor St., Bathurst St., Finch St., driving, route, driven,*
   *house, St. Clair, Eglinton, streets, Eglinton, Lawrence, Wilson,*
   *Sheppard, town-house, Steeles, car, drove, Danforth, porches, drove,*
   *drove, Bathurst, Steeles, Centre St., Highway 7, trunk, road, light,*
   *turned, houses, turned, houses, roadway, lawn, suburb, suburb, people,*
   *Metropolitan Toronto, Steeles, people, urban, Metro, Peel, Durham, York,*
   *population, people, Toronto, population, city, Toronto, population,*
   *suburbs, Etobicoke, Scarborough, people, Toronto, city, suburbia,*
   *people.*

2. *afflicted, darkness, panicky, mournful, exciting, deadly, hating,*
   *aversion, cruel, relentless, weird, eerie, cold, barren, sterile, hated,*
   *cruel, perversely, cruel*

3. `married, wife, wife, wife`

4. `conceded, tolerance`

5. `virgin, pine, bush, trees, trunks, trees`

6. `hand-in-hand, matching, whispering, laughing, warm`

7. `first, initial, final`

8. `night, dusk, darkness`

9. `environment, setting, surrounding`

The lexical chains produced by Morris and Hirst can be quite long. The first has 84 and the second 19 words. Specific knowledge of Toronto is used to build the first chain, something that cannot be reproduced by an automated implementation based only on *Roget's* or *WordNet*. Both sets of chains identify suburbia as the main topic of the text. The *ELKB* hints that driving is a hated activity in the 3[rd] lexical chain. As with the *Einstein* example, a subjective comparison of lexical chains is not very conclusive.

## 5.5   Complexity of the Lexical Chain Building Algorithm

The most computationally expensive part of the lexical chain building process is the construction of all possible meta-chains as described in *Step 3: Build All Proto-chains for Each Candidate Word*. The complexity of the other components of the implementation is negligible compared to this one. *Step 3* can be described by the following pseudo-code:

```
for each ( ucw(i) in the set of unique_candidate_words)

   for each ( sense(j) of ucw(i) )

      for each ( cw(k) in the set of candidate_words)

           if there exists thesaural_relation ( ucw(i) and cw(k) )

         then

            add cw(k) to meta_chain
```

**Figure 5.4:** Algorithm for building all proto-chains.

Given that there are `n` candidate words in a text, and each word has on average 2.14 senses in the *ELKB* and that in the worst case there are as many unique candidate words as there are total candidate words in a text, the complexity of *Step 3* is *n* * 2.14 * *n* which is $O(n^2)$. We use heuristics to improve performance, for example a sense, identified by the triple Head number, Paragraph key and part-of-speech, is only considered once during the meta-chain building process, and the list of candidate words is reduced once all chains have been built for a given unique candidate word, but the computational complexity of the chain building procedure remains $O(n^2)$.

Silber and McCoy (2002) propose a linear time algorithm for the implementation of lexical chains. Their system can process a 40,000 word document in 11 seconds using a *Sparc Ultra 10 Creator*. As a manner of comparison, the *ELKB* implementation requires 5 seconds to process the 89 word *Einstein* text and 51 seconds for the 964 word *Outlands* document using an *Intel Pentium 4*, 2.40 GHz processor with 256 MB of RAM. The 4 seconds that it takes to load the *ELKB* into memory is included in these times. This implementation is clearly much slower than Silber and McCoy's although it has not been refined for this task. One of the goals of their *WordNet*-based implementation is to optimize the process so as to construct chains for extremely large documents. It is the fastest known lexical chain building system.

## 5.6   Evaluating Lexical Chains

Morris and Hirst (1991) evaluate their lexical chains by comparing them to the heading structure of a text assigned by the author. This evaluation is adequate if the goal of lexical chains is to segment a text into distinct regions according to their topic. It assumes that the author has presented the only possible correct partitioning of the text. Lexical chains are generally not used for this purpose, Barzilay (1997) has even segmented the text before building the chains, and authors do not always assign subject headings to identify the various ideas. For these reasons, that evaluation procedure cannot be used to evaluate and objectively compare the lexical chains created by various systems.

Hirst and St-Onge (1998) propose the task of malapropism detection to evaluate lexical chains. A malapropism is defined as "the confounding of an intended word with another word of similar sound or similar spelling that has a quite different and malapropos meaning, for example, *an ingenuous* [for *ingenious*] *machine for peeling oranges*." (Fellbaum, 1998, p. 304). This task is not very common and the evaluation procedure requires a corpus of malapropisms, a resource that is not readily available.

Silber and McCoy (2002) evaluate their implementation by comparing their lexical chains to summaries of a document collection. Their evaluation method is inspired by those used in text summarization. Their corpus is made up of scientific documents with abstracts and chapters from University textbooks that contain chapter summaries. Marcu (1999) has argued that abstracts of articles can be accepted as reasonable summaries. This procedure involves comparing the senses of the words in the lexical chains to the senses of the words in the abstract. This is necessary, as the summaries many not contain the same words as the texts, and therefore the lexical chains.

This evaluation procedure is interesting, since large amounts of documents with their summaries are produced for the *Document Understanding Conferences* (DUC, 2001, 2002). For this evaluation procedure to work with *Roget's*, it is necessary to tag the senses of the words and phrases in the texts and summaries using those found in the *Thesaurus*. We have not performed or implemented any manual or automatic procedure to do so. Although promising, we have not evaluated our implementation of lexical chains using this procedure.

Lexical chains can also be evaluated by assessing the quality of the summaries that are produced by them (Barzilay and Elhadad, 1997; Brunn, Chali and Pinchak, 2001) but the investigation of this task is beyond the scope of this dissertation.

## 5.7   About the Straightforwardness of Implementing Lexical Chains

The experiment shows that it is possible to create lexical chains using our electronic version of *Roget's Thesaurus*, but that it is not as straightforward as it was originally claimed. *Roget's* has a very rich structure that can be exploited for lexical chain construction. Using the *ELKB*, many more thesaural relations can be used than in this implementation, but they come with a computational cost. *WordNet* implementations have access to a different set of relations and lexical material. Although there is a consensus on the high-level algorithm, there are significant differences in implementations. The major criticism of lexical chains is that there is no adequate evaluation of their quality. Until it is established, it will be hard to compare implementations of lexical chain construction algorithms. This experiment demonstrates that the *ELKB* and *WordNet* can be used effectively for the same task.

# 6 Finding the Hidden Treasures in the *Thesaurus*

The experiments presented in Chapters 4 and 5 show that *Roget's Thesaurus* is a valuable resource for NLP, yet these applications exploit only a fraction of this abundant lexical knowledge base. A current trend in NLP is to combine lexical resources to overcome their individual weaknesses. This chapter discusses the correspondence between *Roget's* and *WordNet*. We show a method for disambiguating *Roget's* paragraphs by mean of groups of synsets. This procedure exposes *WordNet's* semantic relations that are present in the *Thesaurus*. The fact that the *ELKB* does not label semantic relations explicitly is a major hindrance for NLP applications: "*Roget's* remains … an attractive lexical resource for those with access to it. Its wide, shallow hierarchy is densely populated with nearly 200,000 words and phrases. The relationships among the words are also much richer than *WordNet's* IS-A or HAS-PART links. The price paid for this richness is a somewhat unwieldy tool with ambiguous links" (McHale, 1998). Machine learning techniques can label these relations given sufficient training data. This chapter concludes with a study of avenues to improve the *ELKB* using *Longman's Dictionary of Contemporary English* (*LDOCE*) (Procter, 1978).

## 6.1 A Quantitative Comparison of *Roget's* and *WordNet*

Chapter 3 describes the similarities between *Roget's* and *WordNet*. This section presents a detailed examination of the portions that contain the most and least overlap in lexical content. *Roget's* ontology can be divided into classes that describe the external world (`Abstract Relations`, `Space`, `Matter`) and ones that describe the internal world of the human. These subjects almost evenly dived the lexical material, 446 headwords belong to the external world, 544 to the internal world. *WordNet* seems to favor the external world, with only two of the nine unique beginners *{psychological feature}* and *{act, human action, human activity}* describing the internal world of the human. Intuition suggests that *Roget's* and *WordNet* should have a big overlap in lexical material pertaining to the material world. Experiments have identified a list of over 45,000 strings that can be found in *WordNet 1.7.1* and the 1987 *Roget's*. Table 6.1 presents the distribution of words and phrases within the *ELKB* ordered by class number. % of c.h., % of c.k. and % of c.s. in Table 6.1 indicate the percentage of heads, keywords and strings that can be found in this common word and phrase list.

Both lexical resources are similar in absolute size, containing about 200,000 word-sense pairs. 53% of all words in *Roget's* are nouns, 20% adjectives, 23% verbs, 4% adverbs and less than 1% are interjections. 74% of *WordNet* are nouns, 15% adjectives, 8% verbs and 3% adverbs. There are no interjections in *WordNet*. Intuition suggested a large overlap between both resources, but the 46,399 common words and phrases only represent about 43% of the unique words and phrases in the *Thesaurus* and 32% of *WordNet*. All of the occurrences of common strings make up 63% of *Roget's* total lexical content. The equivalent calculation has not been performed for *WordNet*. The top-level of the ontologies hinted that the overlap would be concentrated in the first three *Roget* classes, but the results in Table 6.1 shows the common strings distributed pretty evenly across the whole resource. A head-per-head analysis shows that 78% of head names as well as 75% of paragraph keywords can be found in *WordNet*. 86% of heads have at least 50% of their words in common with those of *WordNet* and 93% of heads have at least 50% of keywords in common. Table 6.2 shows the 10 heads with the highest and lowest percentage of common strings. H in WN indicates that the head name can be found in *WordNet*.

| Class # | # of sections | # of heads | # of paragraphs | # of SGs | # of strings | % of c.h. | % of c.k. | % of c.s. |
|---------|---------------|------------|-----------------|----------|--------------|-----------|-----------|-----------|
| 1 | 8 | 182 | 1146 | 10479 | 38029 | 63.74 | 75.48 | 65.75 |
| 2 | 4 | 136 | 992 | 8904 | 32634 | 71.32 | 80.04 | 63.77 |
| 3 | 3 | 128 | 714 | 6429 | 22734 | 72.66 | 79.97 | 64.31 |
| 4 | 7 | 67 | 460 | 4226 | 15471 | 79.10 | 75.87 | 56.65 |
| 5 | 3 | 81 | 495 | 4953 | 18401 | 88.89 | 78.18 | 59.33 |
| 6 | 5 | 138 | 927 | 9616 | 36525 | 88.41 | 72.49 | 59.53 |
| 7 | 4 | 84 | 447 | 4125 | 15873 | 86.90 | 71.14 | 55.88 |
| 8 | 5 | 174 | 1251 | 11149 | 44858 | 83.91 | 67.63 | 55.04 |
| ***Total***: | 39 | 990 | 6432 | 59877 | 224525 | 77.98 | 74.66 | 60.31 |

**Table 6.1:** Distribution of words and phrases within *Roget's Thesaurus* ordered by class number.

Identifying the areas where *WordNet* and the *Thesaurus* overlap are of interest since this should be a good indicator of where the two resources can be combined. The comparison of the Head `567 Perspicuity` has with the *WordNet* synsets that are synonymous to `perspicuity` and `perspicuous` shows that the semicolon groups and the synsets organized around these two words can be quite similar. The content of the Head is the following:

**567** *Perspicuity*

N. **perspicuity**, perspicuousness, clearness, clarity, lucidity, limpidity, 422 *transparency*; limpid style, lucid prose, 516 *intelligibility*; directness, 573 *plainness*; definition, definiteness, exactness, 494 *accuracy*.

ADJ. **perspicuous**, lucid, limpid, 422 *transparent*; clear, unambiguous, 516 *intelligible*; explicit, clear-cut, 80 *definite*; exact, *accurate*; uninvolved, direct, 573 *plain*.

**Figure 6.1:** The *Roget's Thesaurus* Head **567** *Perspicuity*.

| % of c.s. | Class # | Head | H. in WN | # of paragraphs | # of SGs | # of strings | % of c.k. |
|---|---|---|---|---|---|---|---|
| 97 | 1 | *42: Decrement- thing deducted* | No | 1 | 10 | 36 | 100 |
| 94 | 1 | *77: Class* | Yes | 5 | 28 | 121 | 100 |
| 94 | 5 | *567: Perspicuity* | Yes | 2 | 9 | 31 | 100 |
| 91 | 5 | *571: Vigour* | Yes | 3 | 28 | 140 | 67 |
| 87 | 5 | *576: Inelegance* | Yes | 2 | 36 | 118 | 100 |
| 86 | 6 | *631: Materials* | No | 3 | 32 | 145 | 67 |
| 85 | 4 | *493: Ignoramus* | Yes | 2 | 10 | 39 | 100 |
| 85 | 5 | *572: Feebleness* | Yes | 2 | 24 | 104 | 50 |
| 83 | 3 | *353: Air pipe* | No | 1 | 12 | 48 | 0 |
| 83 | 5 | *564: Grammar* | Yes | 4 | 37 | 117 | 100 |
| ... | | | | | | | |
| 35 | 5 | *585: Soliloquy* | Yes | 4 | 11 | 29 | 50 |
| 34 | 8 | *907: Gratitude* | Yes | 7 | 27 | 112 | 50 |
| 34 | 8 | *957: Judge* | Yes | 3 | 25 | 80 | 71 |
| 31 | 3 | *361: Death* | Yes | 9 | 116 | 407 | 100 |
| 31 | 8 | *908: Ingratitude* | No | 5 | 17 | 64 | 67 |
| 31 | 4 | *506: Oblivion* | Yes | 6 | 52 | 162 | 40 |
| 30 | 8 | *958: Lawyer* | Yes | 7 | 36 | 123 | 83 |
| 30 | 1 | *109: Neverness* | No | 2 | 13 | 27 | 86 |
| 29 | 8 | *977: Heterodoxy* | Yes | 8 | 50 | 133 | 50 |
| 24 | 8 | *948: Sobriety* | Yes | 5 | 25 | 96 | 63 |

**Table 6.2:** Distribution of words and phrases within *Roget's Thesaurus* ordered by percentage of common strings.

The synsets containing the synonyms of *perspicuity* and *perspicuous* are:

> **perspicuity**, perspicuousness, plainness -- (clarity as a consequence of being perspicuous); clarity, lucidity, pellucidity, clearness, limpidity -- (free from obscurity and easy to understand; the comprehensibility of clear expression)
>
> limpid, lucid, luculent, pellucid, crystal clear, **perspicuous** -- ((of language) transparently clear; easily understandable; "writes in a limpid style"; "lucid directions"; "a luculent oration"- Robert Burton; "pellucid prose"; "a crystal clear explanation"; "a perspicuous argument"); clear (vs. unclear) -- (clear to the mind; "a clear and present danger"; "a clear explanation"; "a clear case of murder"; "a clear indication that she was angry"; "gave us a clear idea of human nature")

**Figure 6.2:** The *WordNet* synsets for `perspicuity` and `perspicuous`.

The synonym synsets do not contain all of the words and phrases in the Head, even though 94% of these are contained in *WordNet*. This is an indication that the semantic relations that link the semicolon groups in a paragraph extend beyond synonymy. This is further discussed in section 6.3.

Both lexical resources are comparable in size, but *WordNet's* 111,223 synsets are almost double the 59,877 semicolon groups in the *ELKB*. Only 1,431 semicolon groups and synsets are identical, 916 consist of one word or phrase, 459 of two, 51 of three, 4 of four and 1 of five words and phrases. The common sets of four or more words are:

- `{compass, grasp, range, reach}`
- `{ease, relaxation, repose, rest}`
- `{escape, leak, leakage, outflow}`
- `{fourfold, quadruple, quadruplex, quadruplicate}`
- `{coronach, dirge, lament, requiem, threnody}`

Semicolon groups contain on average 3.75 words and phrases, synsets 1.76. This may indicate that synsets represent a much more focused concept than semicolon groups even though both are defined as sets of closely related words. The next section investigates a technique for matching semicolon groups to synsets.

## 6.2 Combining *Roget's* and *WordNet*

The semicolon group and the synset represent the smallest independent unit of *Roget's* and *WordNet*. Although not identical, these groups can be compared and linked. Kwong (1998, 2001) proposes an algorithm for aligning *WordNet* noun synsets with their equivalent noun sense in the 1987 edition of Penguin's *Roget's Thesaurus*. A sense in *Roget's* is defined by a noun and its location within a specific semicolon group, paragraph and head. The following steps describe my variant of Kwong's algorithm:

**Step 0:** Take an index item $W$ from *Roget's* index. For example, the word `desk` is an index item in:

**desk**

    `cabinet` 194 n.
    `stand` 218 n.
    `classroom` 539 n.

**Step 1:** In *Roget's*, find all paragraphs $P_m$ such that $W \in P_m$.

**Step 2:** In *WordNet*, build all mini-nets $M_n$ for $W$. A mini-net consists of a synset $S_n$ such that $W \in S_n$ with its corresponding hypernym synsets $Hyp(S_n)$, and coordinate synsets $Co(S_n)$. The coordinate synsets represent the immediate hypernyms of the synset as well as the hypernyms' immediate hyponyms. This is done to compare similar structures in both resources and to ensure enough lexical material to calculate a significant overlap.

**Step 3:** Compute a similarity score matrix *A* for the *WordNet* mini-nets and the *Roget's* paragraphs. A similarity score $A_{jk}$ is computed for the $j^{th}$ *WordNet* mini-net and the $k^{th}$ *Roget's* paragraph, according to the following formula:

$$Ajk = \alpha_1 |S_j \cap P_k| + \alpha_2 |Hyp(S_j) \cap P_k| + \alpha_3 |Co(S_j) \cap P_k|$$

Kwong sets $\alpha_1 = \alpha_2 = \alpha_3 = 1$. These weights are reasonable as it seems that no one relation is more important than another. The procedure uses lemmata when comparing words but does not lemmatize elements of phrases.

**Step 4:** Find the global maximum $max(A_{jk})$ of the matrix *A*. The $j_{th}$ *WordNet* mini-net can be aligned with the $k_{th}$ *Roget's Thesaurus* paragraph found in the maximum intersection.

Kwong (*ibid.*) takes a *WordNet* synset and assigns a *Roget's* sense to it. The system maps 18,000 noun synsets onto 30,000 senses. She gives the following statement regarding accuracy: "Although it had been difficult and impractical to check the mappings … exhaustively given the huge amount of data, extensive sampling of the results showed that over 70% of the mappings are expected to be accurate". We have not verified this precision value.

Nastase and Szpakowicz (2001) have implemented a procedure that links *WordNet* semicolon groups to *Roget's* senses for all parts-of-speech. As the same complement of semantic relations is not available to all parts-of-speech in *WordNet*, different ones must be used for building mini-nets:

- using nouns: synonyms, hyponyms, hypernyms, meronyms and holonyms;

- using adjectives and adverbs: synonyms;

- using words that are derived from another word *w*: the information pertaining to the word *w*, according to its part-of-speech.

The precision of this algorithm is 57% when applied to the various parts-of-speech, comparatively to Kwong's 70% for nouns. We have not implemented much of this algorithm using the *ELKB* due to the large amount of data which makes it very time consuming to evaluate.

The task of aligning semicolon groups with synsets is more complicated as this mapping is many-to-many. Daudé *et al.* (2001) map *WordNet* 1.5 synsets onto *WordNet* 1.6 synsets using relaxation labeling. Their technique is very effective, but it relies on the structure imposed by the semantic relations. This mapping cannot be simply translated to align *Roget's* semicolon groups onto *WordNet* synsets as no explicit semantic relations are given in the *Thesaurus*. Without further experiments, it is difficult to assess the feasibility of this algorithm.

## 6.3 Importing Semantic Relations from *WordNet* into *Roget's*

*Roget's* lacks explicitly labelled semantic relations. Chapter 4 has shown that its structure can be exploited to measure semantic similarity effectively, but having labeled semantic relations allows to further untangling the rich information contained in the paragraphs. This can be illustrated by examining the first paragraph of the Head `276 Aircraft`. The hyponym and meronym relations used by *WordNet* are clearly present, as well as the notions of "`science of aircraft`", "`testing of an aircraft`", as well as "`places where an aircraft can land`":

> **276** `Aircraft`
>
> N. `aircraft`,
>
> **`science of aircraft:`**
>
> 271 `aeronautics;`
>
> **`kinds of aircraft:`**
>
> aerodyne, flying machine; aeroplane, airplane, crate; plane, monoplane, biplane, triplane; amphibian; hydroplane, seaplane, flying boat; airliner, airbus, transport, freighter; warplane, fighter, bomber, 722 *air force;* stratocruiser, jet plane, jet, jumbo jumbo, jump jumbo, supersonic jumbo, turbojet, turboprop, turbofan, propfan; VTOL,STOL, HOTOL;
>
> **`parts of an aircraft:`**
>
> flying instruments, controls, flight recorder, black box, autopilot, automatic pilot, joystick, rudder; aerofoil, fin, tail; flaps, aileron, 271 *wing;* prop, 269 *propeller;* cockpit, flight deck; undercarriage, landing gear; safety belt, life jacket, parachute, ejection seat, 300 *ejector;*
>
> **`testing of an aircraft:`**
>
> test bed, wind tunnel; flight simulator;
>
> **`places where an aircraft can land:`**
>
> aerodrome, airport, 271 *air travel.*

**Figure 6.3:** The first paragraph of the *Roget's Thesaurus* Head **276** `Aircraft`.

Nastase and Szpakowicz (*ibid.*) have found empirically the hypernym relation to be prevalent between the keyword and the other phrases that make up a paragraph. Cassidy (2000) has identified 400 semantic relations in the 1911 edition of *Roget's Thesaurus*. Some of these relations are: `is-caused-by`, `is-performed-by`, `has-consequence`, `is-measured-by`, `is-job-of`. Cassidy's work is done manually. An alternative to this is to align paragraphs and mini-nets and label the relations automatically. The paragraph for the noun `decrement` illustrates the procedure:

```
Head 42 Decrement: thing deducted

N.  decrement,  deduction,  depreciation,  cut  37  diminution;
allowance;  remission;  tare,  drawback,  clawback,  rebate,  810
discount;  refund,  shortage,  slippage,  defect 307  shortfall, 636
insufficiency;  loss,  sacrifice,  forfeit  963  penalty;  leak,
leakage, escape 298 outflow; shrinkage 204 shortening; spoilage,
wastage, consumption 634 waste; subtrahend, rake-off, 786 taking;
toll 809 tax.
```

**Figure 6.4:** The *Roget's Thesaurus* noun paragraph of Head **42** `Decrement`.

The mini-net for the noun `decrement` can be built in the following way:

```
Overview of noun decrement The noun decrement has 2 senses:

    1. decrease, decrement -- (the amount by which something decreases)
    2. decrease, decrement -- (a process of becoming smaller)

Synonyms/Hypernyms of noun decrement:

    Sense 1 - decrease, decrement: {amount}
    Sense 2 - decrease, decrement: {process}

Hyponyms of noun decrement:

    Sense 1 - decrease, decrement: {drop, fall}, {shrinkage}
    Sense 2 - decrease, decrement: {wastage}, {decay, decline}, {slippage},
    {decline, diminution}, {desensitization, de sensitisation}, {narrowing}

Coordinate Terms of noun decrement:

    Sense 1 - decrease, decrement : {amount}, {quantity}, {increase,
    increment}, {decrease, decrement}, {insufficiency,
    inadequacy,deficiency}, {number, figure}
    Sense 2 - decrease, decrement : {process}, {natural process, natural
    action, action,activity}, {photography}, {chelation}, {human process},
    {development, evolution}, {economic process}, {decrease, decrement},
    {increase, increment, growth}, {processing}, {execution},
    {degeneration}, {shaping, defining}, {dealignment}, {uptake}
```

**Figure 6.5:** The *WordNet* mini-net for the noun `decrement`.

By matching semicolon groups and synsets where at least one word or phrase is in common, it is possible to rearrange the *Roget's* paragraph in the following manner:

```
N. decrement

Hyponym: deduction,  depreciation,  cut  37  diminution;  refund,
shortage,  slippage,  defect 307  shortfall,  636  insufficiency;
```

```
shrinkage  204  shortening;  spoilage,  wastage,  consumption  634
waste.

No label : allowance; remission; tare, drawback, clawback, rebate,
810 discount; loss, sacrifice, forfeit 963 penalty; leak, leakage,
escape 298 outflow; subtrahend, rake-off, 786 taking; toll 809
tax.
```

**Figure 6.6:** The *Roget's* noun paragraph of Head **42** `Decrement` labelled with *WordNet* relations.

This algorithm only allows for discovering the *WordNet* relations that are present in *Roget's*. Learning the relations labeled by Cassidy and using machine learning techniques (O'Hara and Wiebe, 2003) would expose more clearly the richness of the *Thesaurus*. This has yet to be attempted using the *ELKB*.

## 6.4 Augmenting *WordNet* with Information contained in *Roget's*

Semicolon groups are organized around subjects in *Roget's* whereas synsets are linked by the closed set of semantic relations in *WordNet*. Fellbaum (1998, p.10) calls this particularity of *WordNet* the *Tennis Problem* and describes it in the following manner: "… *WordNet* does not link `racquet`, `ball`, and `net` in a way that would show that these words, and the concepts behind them, are part of another concept that can be expressed by `court game`." George Miller has promised that this will be corrected in *WordNet* 2.0. *Roget's Thesaurus* can help in this task. It contains the paragraph `ball game` in the Head **837** `Amusement`:

```
N. ball game, pat-ball, bat and ball game; King Willow, cricket,
French cricket; baseball, softball, rounders; tennis, lawn tennis,
real tennis, table tennis, pingpong; badminton, battledore and
shuttlecock; squash, rackets; handball, volleyball; fives, pelota;
netball, basketball; football, Association football, soccer;
rugby, Rugby football, Rugby Union, Rugby League, rugger;
lacrosse, hockey, ice hockey; polo, water polo; croquet, putting,
golf, clock golf, crazy golf; skittles, ninepins, bowls, petanque,
boule, curling; marbles, dibs; quoits, deck quoits, hoop-la;
billiards, snooker, pool; bagatelle, pinball, bar billiards, shove
ha'penny, shovelboard.
```

**Figure 6.7:** The *Roget's Thesaurus* `ball game` 837 n. paragraph.

This paragraph contains `rackets`, `ball game` and `netball`, similar words to `racquet`, `ball` and `net`. A native English speaker can make the connection. This example illustrates that the organization of *Roget's* lexical material is quite different than that of *WordNet's*, that *WordNet* would benefit from *Roget's* topical clustering and that adding such links automatically is not a trivial task. Stevenson (2001) considers that synsets can be linked using a new relationship when a *Roget's* paragraph has a strong overlap with three or more synsets. He links 24,633 *WordNet* synsets to 3,091 *Roget's International Thesaurus* (Chapman, 1977) paragraphs. These figures represent almost 25% of *WordNet* synsets and 50% of *ELKB* paragraphs in terms of absolute numbers, which suggests that augmenting *WordNet* in this manner is a promising avenue of research.

## 6.5   Other Techniques for Improving the *ELKB*

Kwong (1998) has shown that it is possible to obtain a mapping between *LDOCE* and *Roget's Thesaurus* using *WordNet*. Although Kwong only performs this experiment on a small set of 36 nouns, the idea of incorporating information contained in *LDOCE* into *Roget's* is very attractive. *LDOCE* contains definition and frequency information that is very beneficial. Researchers have also proven it to be a valuable resource for *NLP*. The *ELKB* that would contain definitions, frequency information as well as a set of labeled semantic relations is very close to the holy grail of computational lexicography:  "a *neutral, machine-tractable, dictionary*" (Wilks *et al.*, 1996).

# 7   Summary, Discussion, and Future Work

This chapter summarizes the contributions of the thesis and presents known flaws of the *ELKB*, aspects that should be improved and ideas for future applications.

## 7.1   Summary

The goal of this thesis was to establish if *Roget's Thesaurus* can be a realistic alternative to *WordNet*. To achieve this, various sub-goals had to be met. The first is the design and implementation of the *ELKB*; next I performed *NLP* experiments whose results are compared to those of *WordNet*-based systems. The thesis also contains a quantitative comparison of both lexical knowledge bases.

Chapter 1 presents the context, goals and organization of this thesis.

Chapter 2 gives a brief history of how computational linguists have used thesauri in *NLP*. It discusses the various versions of *Roget's* and explains the rationale for choosing the 1987 edition of Penguin's *Roget's Thesaurus of English Words and Phrases* as the source for the *ELKB*. This chapter also discusses several applications of *Roget's Thesaurus* and *WordNet* in NLP.

Chapter 3 discusses the design of the *ELKB* as well as its implementation. It shows the necessary steps to transform the computer-readable Pearson Education files into a tractable form. This involves converting the lexical material into a format that can be more easily exploited, identifying data structures and classes to computerize the *Thesaurus*, indexing all of the words and phrases in the resource and ensuring that they can be retrieved even when the exact string is not supplied. I explain in detail *Roget's* organization and contrast it with *WordNet's*. The implementation verifies the accuracy of the design and ensures that *Roget's* functionality is faithfully reproduced by the *ELKB*.

Chapter 4 explains how *Roget's Thesaurus* can be used to measure semantic distance. Using three well known benchmarks for the evaluation of semantic similarity, I correlate the similarity values calculated by the *ELKB* and six *WordNet*-based measures with those assigned by human judges.  *Roget's* gets scores of over .80 for two of the three benchmarks, quite close to those obtained when the experiments are replicated using humans. The system outperforms the *WordNet*-based measures most of the time. The chapter presents a second class of experiments, where the correct synonym must be selected amongst a group of four words. These are taken from *ESL*, *TOEFL* and *Readers' Digest* questions. The *ELKB* is compared to the same *WordNet*- based measures as well as to statistical methods. The *ELKB* outperforms all systems that do not rely on combined approaches, obtaining scores in the 80% range.

Chapter 5 explains how lexical chains can be built using the *ELKB*. It presents the necessary design decisions for automating the chain building procedure by walking through the algorithm that has been used for all implementations. I compare the lexical chains the *ELKB* constructs to those built manually by the inventors of lexical chains, automatically using a partially computerized *Roget's Thesaurus* and a *WordNet*-based system. This chapter discusses several evaluation procedures, in particular one in which the lexical chains are compared to summaries of texts.

Chapter 6 describes steps for combining *Roget's* and *WordNet*. It shows some of the rich implicit semantic relations that are found in the *Thesaurus*. I explain how *WordNet* can enrich *Roget's* and vice-versa, as well as present an algorithm for aligning both resources.

Chapter 7 presents a summary of the dissertation. It discusses known flaws as of the *ELKB* as well as future extensions and applications.

## 7.2 Conclusions

This dissertation has shown that it is possible to computerize *Roget's Thesaurus* so that it maintains all of the functionality of the printed version and allows for manipulations suitable for NLP applications. I have used the *ELKB* in a few experiments, but these are not enough to determine if it is a credible alternative to *WordNet*. I offer a few ideas for those who intend to use the *ELKB* or want to build a similar knowledge base.

### 7.2.1 Building an *ELKB* from an Existing Lexical Resource

Building an *ELKB* from an existing lexical resource is a very attractive proposition. A computational linguist can save much time by exploiting the structure and lexical material contained in existing dictionaries and thesauri. I have encountered two major problems when implementing the *ELKB*. The lexicographer's directives are not known and it is very tedious to comprehend the organization of paragraphs and semicolon groups without having specific explanations for the underlying decisions. Implementing the *ELKB* would have been much simpler had the editor's instructions for the preparation of the *Roget's Thesaurus* been available. The next problem is that the lexical material must be licensed from the publisher for a considerable price. This hinders the public acceptance of the *ELKB* as most research groups are unwilling to spend money on an unproven resource.

### 7.2.2   Comparison of the *ELKB* to *WordNet*

The *ELKB* is comparable to *WordNet* in many ways. It contains a similar number of words and phrases and this thesis has shown that they both can be used for the same tasks. Although they are similar, this dissertation demonstrates that their organization is quite different. The *ELKB* draws on the 150 years that lexicographers have taken to prepare the *Thesaurus*. Pearson's publishes a new edition roughly every ten years. The *ELKB* lacks the support of the NLP community which *WordNet* has. *WordNet* is slightly more than ten years old, new versions are released about every two years. Version 2.0 promises to correct many flaws that are discussed in this dissertation. Several research groups work independently from George Miller's to enhance this lexical resource. Its prevalence is not only due to its quality but largely also to the fact that it is free.

### 7.2.3   Using the *ELKB* for NLP Experiments

This thesis has used the *ELKB* to measure semantic similarity between words and phrases and to build lexical chains. I have been able to perform these experiments with ease using the Java implementations. These two applications can be integrated into larger systems, for example one that performs Text Summarization or Question Answering tasks. The *ELKB* has also been used in two computer science honors projects. Gilles Roy and Pierre Chrétien created a graphical user interface for the *Thesaurus*, Tad Stach wrote a program to play the *Reader's Digest Word Power Game*. The *ELKB* must be used in more experiments to test the software thoroughly.

### 7.2.4   Known Errors in the *ELKB*

The *ELKB* still contains errors, mostly in its lexical material. These are often due to the mistakes contained in the original Pearson files. I estimate that about 2% - 3% of the words and phrases in the *ELKB* are incorrect. Although this percentage is small, it is enough to be noticed and adversely affect future applications.

### 7.2.5   Improvements to the *ELKB*

The computerized *Roget's Thesaurus* that I have implemented is far from being the perfect lexical knowledge base. Many improvements can be made to the software.

#### 7.2.5.1   Retrieval of Phrases

The *ELKB* does not perform any morphological transformations when looking up a phrase. If a user does not supply the exact string contained in the Index, no result will be returned. For example, the phrase "`sixty four thousand dollar question`" will not be found because the exact string in the *ELKB* is "`the sixty-four-thousand-dollar question`". Giving access to all of *Roget's* phrases is a difficult problem to solve but is one that merits attention. The *Thesaurus* contains many phrases, some of them very peculiar, for example: "`Cheshire cat grin`", "`Homeric laughter`" or "`wisest fool in Christendom`". It is possibly an area where the *ELKB* is superior to *WordNet*. I have not investigated this. An ideal solution would be to integrate in the Index a method that could extract all phrases that contain certain words. Also, morphological transformations would have to be performed on all words in a phrase to find the form contained in the *Thesaurus*. An imperfect solution that I have adopted for the *ELKB* is to index all two word phrases under each of the words. Although this improves the recall of phrases, it introduces many odd references for index entries. For example, the phrase `fish food` is now indexed under `fish` and `food`. The *ELKB* determines that the distance between `food` and `rooster` is 4, meaning that the words are quite similar, when the intuitive association is not that strong. The system is finding the shortest path between all references of `food` and `rooster`, which happens to be between `fish food` and `rooster`, found in two different noun paragraphs of the head **365** `Animality. Animal.` When the phrase `fish food` is ignored, the distance between `food` and `rooster` is 10.

### 7.2.5.2 *Displaying the Semicolon Group Which Contains a Variant of the Search Word*

As described in Chapter 3, when a word is looked up, morphological transformations are performed to find all matching entries that are contained in the Index. For example, when a user enters the word `tire`, the *ELKB* finds the words `tire` and `tyre` in the Index. The reference for `tyre` is `wheel 250 N`. The system finds the Head **250** `Circularity: simple circularity`, and locates the `wheel` noun paragraph. The *ELKB* searches the paragraph sequentially until the word `tire` is found. Since `tyre` is the word contained in the paragraph, the correct semicolon group is not returned. This causes a slight problem when calculating semantic distance. For example, the system determines the distance between the words `hub` and `tire` to be 2 instead of 0 as the words do not appear in the semicolon group `;hub, felloe, felly, tyre;`. This is not difficult to correct, but awkward, so I left it as one of a number of future adjustments.

### 7.2.5.3 *Original vs. New Index*

The *ELKB* uses an Index that is generated from all of the words and phrases that it contains. Pearson Education supplies an Index that is about half the size of the automatically generated one. The system stores the two in separate files that cannot be used at the same time. I would have more faithfully reproduced *Roget's Thesaurus* if the entries in the Index had been flagged as `original` and `new`.

### 7.2.5.4 *Optimization of the ELKB*

This version of the *ELKB* serves as a proof of concept. Future releases must improve memory usage and speed if this resource is to be a viable alternative to *WordNet*. Performance can be improved by loading the text of the 990 Heads into memory and storing absolute references to Paragraphs and semicolon groups, as described in Chapter 3. The current implementation loads in 3 seconds on a Pentium 4, 2.40 GHz with 256 MB of RAM, and requires about 40 MB of RAM.

## 7.3 Future Work

The ultimate goal of this research should be to make the *ELKB* available to any research group that requests it. Beyond the fact that the lexical material must be licensed, future maintainers of the system should thoroughly evaluate the *ELKB* and use it in a wide variety of applications so as to attract the interest of the NLP community. It should also be enhanced to make it more competitive with regards to *WordNet*.

### 7.3.1 More Complete Evaluation of the *ELKB*

This dissertation has performed a partial evaluation of the *ELKB* by comparing it to *WordNet*-based systems and statistical techniques. A comparison to other versions of *Roget's Thesaurus*, namely the 1911 edition, *FACTOTUM* and *Roget's International Thesaurus* should be carried out. Until this is done, I cannot say how good this version of *Roget's* is compared to all others. Future research should perform further benchmark experiments with the *ELKB*, namely Word Sense Disambiguation. This is a problem that has a long history in NLP and for which thesauri have been used (Ide and Véronis, 1998).

### 7.3.2 Extending the Applications Presented in the Thesis

Turney (2002) has used his semantic similarity metric to classify automobile and movie reviews. Bigham *et al.* (2003) use their similarity metric to answer analogy problems. In an analogy problem, the correct pair of words must be chosen amongst four pairs, for example: `cat:meow`:: (a) `mouse:scamper`, (b) `bird:peck`, (c) `dog:bark`, (d) `horse:groom`, (e) `lion:scratch`. To correctly answer `dog:bark`, a system must know that a meow is the sound that a cat makes and a bark the sound that a dog makes. Both of these applications can be implemented with the *ELKB*.

As discussed in Chapter 5, several researchers have used lexical chains for Text Summarization, most notably Barzilay and Elhadad (1997) as well as Silber and McCoy (2000). Since I have implemented a system that can build lexical chains, it would be very interesting to put it to this task.

### 7.3.3   Enhancing the *ELKB*

Chapter 6 describes several enhancements to the *ELKB*. If I had combined *Roget's* with *WordNet*, labeled the implicit semantic relations and included frequency information as well as dictionary definitions from LDOCE, the *ELKB* would be one of the premier lexical resources for NLP.

# Appendix A: The Basic Functions and Use Cases of the *ELKB*

These are the basic functions of the *ELKB*:

1. Look up a Word or Phrase.
2. Browse the Taxonomy.
3. Look up All Words and Phrases in a Head.
4. Calculate the Distance between Two Words or Phrases.
5. Identify the Thesaural Relation between Two Words or Phrases.

These functions can be described by their accompanying use cases.

## 1   Look up a Word or Phrase

1. The user enters a word or phrase.
2. The system performs morphological transformations on the word or phrases.
3. The system searches the index for all entries that contain the transformed search term.
4. The system returns all references for the found index entries.
5. The user chooses a reference from the result list.
6. The system returns the paragraph that contains the reference.
7. The semicolon group that contains the reference is located.

**Alternative:** *The search term is not in the index.*
At step 3, the system fails to find the search term in the index.
Allow the user to re-enter a word or phrase.
Return to primary scenario at step 2.

**Alternative:** *The user cancels the look up.*
At step 1 or 5, the user cancels look up.

## 2   Browse the Taxonomy

1. The system displays the names of the classes.
2. The user chooses a class to expand.
3. The system displays the sections that belong to the selected class.
4. The user chooses a section to expand.
5. The system displays the sub-sections that belong to the selected section.
6. The user chooses a sub-section to expand.
7. The system displays the head groups that belong to the selected sub-section.
8. The user chooses a head group to expand.
9. The system displays the heads that belong to the selected head group.
10. The user chooses a head to expand.
11. The system displays the text of the selected head.

**Alternative:** *The user selects another class, section, sub-section, head group or head.*
At steps 2, 4, 6, 8 or 10 the user can decide to expand another class, section, sub-section, head group or head.
Return to primary scenario at step 3, 5, 7, 9 or 11 depending on what step has been performed.

**Alternative:** *The user collapses a class, section, sub-section, head group or head.*
At step 3, 5, 7, 9 or 11 the user can decide to collapse a class, section, sub-section, head group or head.
The system hides any of the content displayed by the selected class, section, sub-section, head group or head.
Return to primary scenario at step 2, 4, 6, 8 or 10 depending on what steps are possible.

**Alternative:** *The user specifies a head number.*
The user may know the exact head number he wants to look up. The system displays the entire path indicating the class, section, sub-section, head group and head. The system continues at step 11.

## 3   Look up all Words and Phrases in a Head

1. The user selects or enters a head number.
2. The system displays the text of the selected head.

## 4   Calculate the Distance between Two Words or Phrases

1. The user enters two words or phrases.
2. The system performs morphological transformations on each word or phrase.
3. The system looks up each transformed word or phrase in the index.
4. The system finds all paths between each reference of the words or phrases.
5. The system assigns a score to every path: 0 if the two references point to the same semicolon group, 2 if they point to the same paragraph, 4 if the point to the same part-of-speech of the same head, 6 if they point to the same head, 8 if they point to the same head group, 10 if they point to the same sub-section, 12 if they point to the same section, 14 if they point to the same class and 16 if the references are in two different classes of the *ELKB*.
6. The distance is given by the smallest score.

## 5   Identify the Thesaural Relation between Two Words or Phrases

1. The user enters two words or phrases
2. If the same lexicographical string was entered, the thesaural relation is "T0: reiteration". Terminate the procedure.
3. Else, the system performs morphological transformations on each word or phrase.

4.   The system looks up each transformed word or phrase in the index.

5.   The system compares pair wise the references of the index entries.

6.   If two references point to the same paragraph, then the thesaural relation is "T1". Terminate the procedure.

7.   Else, no thesaural relations exist between these two words or phrases.

# Appendix B: The *ELKB* Java Documentation

This appendix presents a summary of the Java documentation for all the classes of the *ELKB*.

## Package ca.site.elkb

| Class Summary | |
|---|---|
| **Category** | Represents the *Roget's Thesaurus Tabular Synopsis of Categories*. |
| **Group** | Represents a *Roget's Thesaurus* Head group. |
| **Head** | Represents a *Roget's Thesaurus* Head. |
| **HeadInfo** | Object used to store the information that defines a Head but not its words and phrases. |
| **Index** | Represents the computer index of the words and phrases of *Roget's Thesaurus*. |
| **Morphy** | Performs morphological transformations using the same rules as *WordNet*. |
| **Paragraph** | Represents a *Roget's Thesaurus* Paragraph. |
| **Path** | Represents a path in *Roget's Thesaurus* between two words or phrases. |
| **PathSet** | A set that contains all of the paths between two words and phrases as well as the number of minimum length paths. |
| **Reference** | Represents a symbolic pointer to a location where a specific word or phrase can be found in *Roget's Thesaurus*. |
| **RogetClass** | Represents the topmost element in *Roget's Thesaurus Tabular Synopsis of Categories*. |
| **RogetELKB** | Main class of the *Roget's Thesaurus Electronic Lexical KnowledgeBase*. |
| **RogetText** | Represents the Text of *Roget's Thesaurus*. |
| **Section** | Represents a *Roget's Thesaurus* Section. |
| **SemRel** | Represents a *Roget's Thesaurus* relation between a word or phrase. |
| **SG** | Represents a *Roget's Thesaurus* Semicolon Group. |
| **SubSection** | Represents a *Roget's Thesaurus* Sub-section. |
| **Variant** | Allows to obtain a variant of an English spelling. |

**ca.site.elkb**
# Class Category

```
java.lang.Object
  |
  +--ca.site.elkb.Category
```

public class **Category**
extends java.lang.Object

Represents the *Roget's Thesaurus Tabular Synopsis of Categories*. The topmost level of this ontology divides the *Thesaurus* into eight Classes:

1. *Abstract Relations*
2. *Space*
3. *Matter*
4. *Intellect: the exercise of the mind (Formation of ideas)*
5. *Intellect: the exercise of the mind (Communication of ideas)*
6. *Volition: the exercise of the will (Individual volition)*
7. *Volition: the exercise of the will (Social volition)*
8. *Emotion, religion and morality*

Classes are further divided into Sections, Sub-sections, Head groups, and Heads.

## Constructor Summary

| |
|---|
| **Category**() |
|     Default constructor. |
| **Category**(java.lang.String filename) |
|     Constructor that builds the `Category` object using the information contained in a file. |

## Method Summary

| | |
|---:|---|
| int | **getClassCount**()<br>    Returns the number of *Roget's* Classes in this ontology. |
| java.util.ArrayList | **getClassList**()<br>    Returns the array of `RogetClass` objects. |
| int | **getHeadCount**()<br>    Returns the number of Heads in this ontology. |
| int | **getHeadGroupCount**()<br>    Returns the number of Head groups in this ontology. |
| java.util.ArrayList | **getHeadList**()<br>    Returns the array of `HeadInfo` objects. |
| ca.site.elkb.RogetClass | **getRogetClass**(int index)<br>    Returns the *Roget's* Class at the specified position in the array of |
| | Classes. |
| `int` | **<u>getSectionCount</u>**`()`<br>        Returns the number of Sections in this ontology. |
| `int` | **<u>getSubSectionCount</u>**`()`<br>        Returns the number of Sub-sections in this ontology. |
| `void` | **<u>printHeadInfo</u>**`()`<br>        Prints the array of `HeadInfo` objects to the standard output. |
| `void` | **<u>printRogetClass</u>**`(int index)`<br>        Prints the *Roget's* Class at the specified position in the array of Classes to the standard output. |
| `java.lang.String` | **<u>toString</u>**`()`<br>        Converts to a string representation the `Category` object. |

**ca.site.elkb**
# Class Group

```
java.lang.Object
  |
  +--ca.site.elkb.Group
```

public class **Group**
extends java.lang.Object

Represents a *Roget's Thesaurus* Head group. For example:

79 Generality    80 Speciality

A `Group` can contain 1,2 or 3 `HeadInfo` objects.

## Constructor Summary

| |
|---|
| **Group**() <br>     Default constructor. |
| **Group**(int start) <br>     Constructor that takes an integer to indicate first Head number of the Group. |

## Method Summary

| | |
|---:|---|
| void | **addHead**(ca.site.elkb.HeadInfo head) <br>     Add a `HeadInfo` object to this Group. |
| int | **getHeadCount**() <br>     Returns the number of Heads in this Group. |
| java.util.ArrayList | **getHeadList**() <br>     Returns the array of `HeadInfo` objects. |
| int | **getHeadStart**() <br>     Returns the number of the first Head in this Group. |
| void | **setHeadStart**(int start) <br>     Sets the number of the first Head in this Group. |
| java.lang.String | **toString**() <br>     Converts to a string representation the `Group` object. |

**ca.site.elkb**
# Class Head

```
java.lang.Object
  |
  +--ca.site.elkb.Head
```

public class **Head**
extends java.lang.Object

Represents a *Roget's Thesaurus* Head. A Head is defined by the following attributes:

- Head number
- Head name
- Class number
- Section num
- list of paragraphs
- number of paragraphs
- number of semicolon groups
- number of words and phrases
- number of cross-references
- number of see references

The relative postons of the noun, adjective verb, adverb and interjection paragraphs in the array of paragarphs is kept by the `nStart`, `adjStart`, `vbStart`, `advStart`, and `intStart` attributes.

---

## Constructor Summary

| |
|---|
| **Head**() |
|     Default constructor. |
| **Head**(int num, java.lang.String name, int clNum, int section) |
|     Constructor which sets the Head number and name, as well as the Class and Section number. |
| **Head**(java.lang.String fname) |
|     Constructor that builds the `Head` object using the information contained in a file. |

---

## Method Summary

| | |
|---|---|
| int | **getAdjCount**()<br>    Returns the number of adjective word and phrases of this Head. |
| int | **getAdjCRefCount**()<br>    Returns the number of adjective cross-references of this Head. |
| int | **getAdjParaCount**()<br>    Returns the number of adjective paragraphs of this Head. |
| int | **getAdjSeeCount**()<br>    Returns the number of adjective references of this Head. |
| int | **getAdjSGCount**`()`<br>Returns the number of adjective semicolon groups of this Head. |
| int | **getAdjStart**`()`<br>Returns the index of the first adjective paragraph in the array of `Pragraph` objects of this Head. |
| int | **getAdvCount**`()`<br>Returns the number of adverb word and phrases of this Head. |
| int | **getAdvCRefCount**`()`<br>Returns the number of adverb cross-references of this Head. |
| int | **getAdvParaCount**`()`<br>Returns the number of adverb paragraphs of this Head. |
| int | **getAdvSeeCount**`()`<br>Returns the number of adverb references of this Head. |
| int | **getAdvSGCount**`()`<br>Returns the number of adverb groups of this Head. |
| int | **getAdvStart**`()`<br>Returns the index of the first adverb paragraph in the array of `Pragraph` objects of this Head. |
| int | **getClassNum**`()`<br>Returns the Class number of this Head. |
| int | **getCRefCount**`()`<br>Returns the number of cross-references of this Head. |
| java.lang.String | **getHeadName**`()`<br>Returns the name of this Head. |
| int | **getHeadNum**`()`<br>Returns the number of this Head. |
| int | **getIntCount**`()`<br>Returns the number of interjection word and phrases of this Head. |
| int | **getIntCRefCount**`()`<br>Returns the number of interjection cross-references of this Head. |
| int | **getIntParaCount**`()`<br>Returns the number of interjection paragraphs of this Head. |
| int | **getIntSeeCount**`()`<br>Returns the number of interjection references of this Head. |
| int | **getIntSGCount**`()`<br>Returns the number of interjection semicolon groups of this Head. |
| int | **getIntStart**`()`<br>Returns the index of the first interjection paragraph in the array of `Pragraph` objects of this Head. |
| int | **getNCount**`()`<br>Returns the number of noun word and phrases of this Head. |
| int | **getNCRefCount**`()`<br>Returns the number of noun cross-references of this Head. |
| int | **getNParaCount**() <br> Returns the number of noun paragraphs of this Head. |
| int | **getNSeeCount**() <br> Returns the number of noun see references of this Head. |
| int | **getNSGCount**() <br> Returns the number of noun semicolon groups of this Head. |
| int | **getNStart**() <br> Returns the index of the first noun paragraph in the array of `Pragraph` objects of this Head. |
| ca.site.elkb.Paragraph | **getPara**(int paraNum, java.lang.String pos) <br> Returns the a `Paragraph` object specified by the paragraph number and part-of-speech. |
| ca.site.elkb.Paragraph | **getPara**(java.lang.String paraKey, java.lang.String pos) <br> Returns the a `Paragraph` object specified by the paragraph key and part-of-speech. |
| int | **getParaCount**() <br> Returns the number of paragraphs of this Head. |
| int | **getSectionNum**() <br> Returns the Section number of this Head. |
| int | **getSeeCount**() <br> Returns the number of see references of this Head. |
| int | **getSGCount**() <br> Returns the number of semicolon groups of this Head. |
| int | **getVbCount**() <br> Returns the number of verb word and phrases of this Head. |
| int | **getVbCRefCount**() <br> Returns the number of verb cross-references of this Head. |
| int | **getVbParaCount**() <br> Returns the number of verb paragraphs of this Head. |
| int | **getVbSeeCount**() <br> Returns the number of verb references of this Head. |
| int | **getVbSGCount**() <br> Returns the number of verb groups of this Head. |
| int | **getVbStart**() <br> Returns the index of the first verb paragraph in the array of `Pragraph` objects of this Head. |
| int | **getWordCount**() <br> Returns the number of words of this Head. |
| void | **print**() <br> Prints the contents of this Head to the standard output. |
| void | **printAllSG**() <br> Prints all the semicolon groups of this Head separated on a separate line to the standard output. |
| void | **printAllWords**()<br>    Prints all the words and phrases of this Head separated on a separate line to the standard output. |
| void | **setClassNum**(int num)<br>    Sets the Class number of this Head. |
| void | **setHeadName**(java.lang.String name)<br>    Sets the name of this Head. |
| void | **setHeadNum**(int num)<br>    Sets the number of this Head. |
| void | **setSectionNum**(int num)<br>    Sets the Section number of this Head. |
| java.lang.String | **toString**()<br>    Converts to a string representation the Head object. |

**ca.site.elkb**
# Class HeadInfo

```
java.lang.Object
  |
  +--ca.site.elkb.HeadInfo
```

public class **HeadInfo**
extends java.lang.Object

Object used to store the information that defines a Head but not its words and phrases. It contains the following attributes:

- Head number
- Head name
- Class number
- Section number
- Sub-section name
- Head group, defined as a list of `HeadInfo` objects

## Constructor Summary

| |
|---|
| **HeadInfo**() <br>     Default constructor. |
| **HeadInfo**(int number, java.lang.String name, int cn, int sn, <br> java.lang.String subName, java.util.ArrayList groupList) <br>     Constructor which sets the Head number and name, as well as the Class and Section number, Sub-section name and Head group list. |
| **HeadInfo**(java.lang.String sInfo, int cn, int sn, <br> java.lang.String subSectInfo, java.lang.String sGroupInfo) <br>     Constructor which sets the Head number and name, as well as the Class and Section number, Sub-section name and Head group list. |

## Method Summary

| | |
|---:|---|
| int | **getClassNum**() <br>     Returns the Class number of this Head. |
| java.util.ArrayList | **getHeadGroup**() <br>     Returns the array of `HeadGroup` objects of this Head. |
| java.lang.String | **getHeadName**() <br>     Returns the name of this Head. |
| int | **getHeadNum**() <br>     Returns the number of this Head. |
| int | **getSectNum**() <br>     Returns the Section number of this Head. |
| java.lang.String | **getSubSectName**() |

| | | |
|---|---|---|
| | | Returns the Sub-section name of this Head. |
| `void` | **<u>setClassNum</u>**`(int num)` | Sets the number of this Head. |
| `void` | **<u>setHeadGroup</u>**`(java.util.ArrayList group)` | Sets the array of `HeadGroup` objects of this Head. |
| `void` | **<u>setHeadName</u>**`(java.lang.String name)` | Sets the name of this Head. |
| `void` | **<u>setHeadNum</u>**`(int num)` | Sets the number of this Head. |
| `void` | **<u>setSectNum</u>**`(int num)` | Sets the Section number of this Head. |
| `void` | **<u>setSubSectName</u>**`(java.lang.String name)` | Sets the Section name of this Head. |
| `java.lang.String` | **<u>toString</u>**`()` | Converts to a string representation the `HeadInfo` object. |

**ca.site.elkb**
# Class Index

```
java.lang.Object
  |
  +--ca.site.elkb.Index
```

**All Implemented Interfaces:**
> java.io.Serializable

---

public class **Index**
extends java.lang.Object
implements java.io.Serializable

Represents the computer index of the words and phrases of *Roget's Thesaurus.* According to Kirkpatrick (1998) "The index consists of a list of items, each of which is followed by one or more references to the text. These references consist of a Head number, a *keyword* in italics, and a part of speech label (n. for nouns, adj. for adjectives, vb. for verbs, adv. for adverbs, and int. for interjections). The *keyword* is given to identify the paragraph which contains the word you have looked up; it also gives and indication of the ideas contained in that paragraph, so it can be used as a clue where a word has several meanings and therefire several references." An example of an Index Entry is:

> **stork**
>> *obstetrics* 167 n.
>> *bird* 365 n.

In this example **stork** is an Index Item and *obstetrics* 167 n. is a Reference. This `Index` object consists of a hashtable of Index Entries, hashed on the String value of the Index Item. For every key (Index Item) the value is a list of Reference objects. The hashtable is implemented using a HashMap.

**See Also:**
> Serialized Form

---

## Constructor Summary

| |
|---|
| **Index**() <br>      Default constructor. |
| **Index**(java.lang.String filename) <br>      Constructor that builds the `Index` object using the information contained in a file. |
| **Index**(java.lang.String fileName, int size) <br>      Constructor that builds the `Index` object using the information contained in a file and sets the initial size of the index hashtable. |

---

## Method Summary

| | |
|---|---|
| boolean | **containsEntry**(java.lang.String key) <br>      Returns `true` if the specified entry is contained in this index. |
| java.util.TreeSet | **getEntry**(java.lang.String key) <br>      Returns all references for a given word or phrase in the index. |
| `java.util.ArrayList` | **getEntryList**(`java.lang.String key`)<br>Returns the list of references for a given word or phrase in the index. |
| `java.util.ArrayList` | **getEntryList**(`java.lang.String key, int itemNo`)<br>Returns the list of references for a given word or phrase in the index preceded by a number to identify the reference. |
| `java.util.TreeSet` | **getHeadNumbers**(`java.lang.String key`)<br>Returns a set of head numbers in which a word or phrase can be found. |
| `int` | **getItemCount**()<br>Returns the number of entries in this index. |
| `int` | **getItemsMapSize**()<br>Returns the number of items contained in the hash map of this index. |
| `int` | **getRefCount**()<br>Returns the number of references in this index. |
| `java.util.ArrayList` | **getRefObjList**(`java.lang.String key`)<br>Returns an array of `Reference` objects. |
| `java.lang.String` | **getRefPOS**(`java.lang.String key`)<br>Returns a string containing the part-of-speech of the references for a given index entry. |
| `java.lang.String` | **getStrRef**(`java.lang.String strIndex`)<br>Returns a reference in String format as printed in *Roget's Thesaurus*. |
| `java.util.ArrayList` | **getStrRefList**(`java.lang.String key`)<br>Returns a list of references in string format instead of pointers. |
| `int` | **getUniqRefCount**()<br>Returns the number of unique references in this index. |
| `void` | **printEntry**(`java.lang.String key`)<br>Prints the index entry along with its references to the standard output. |
| `void` | **printEntry**(`java.lang.String key, int itemNo`)<br>Prints the index entry along with its numbered references to the standard output. |

**ca.site.elkb**
# Class Morphy

```
java.lang.Object
   |
   +--ca.site.elkb.Morphy
```

**All Implemented Interfaces:**
>    java.io.Serializable

---

public class **Morphy**
extends java.lang.Object
implements java.io.Serializable

Performs morphological transformations using the same rules as *WordNet*.

The following suffix substitutions are done for:

- **nouns:**
    1. "s" -> ""
    2. "ses" -> "s"
    3. "xes" -> "x"
    4. "zes" -> "z"
    5. "ches" -> "ch"
    6. "shes" -> "sh"
    7. "men" -> "man"
- **adjectives:**
    1. "er" -> ""
    2. "est" -> ""
    3. "er" -> "e"
    4. "est" -> "e"
- **verbs:**
    1. "s" -> ""
    2. "ies" -> "y"
    3. "es" -> "e"
    4. "es" -> ""
    5. "ed" -> "e"
    6. "ed" -> ""
    7. "ing" -> "e"
    8. "ing" -> ""

The `noun.exc`, `adj.exc`, `verb.exc` and `adv.exc` exception files, located in the `$HOME/roget_elkb` directory, are searched before applying the rules of detachment.

**See Also:**
>    Serialized Form

---

## Field Summary

| | |
|---|---|
| static java.lang.String | **ADJ_EXC**<br>Location of the `adj.exc` file. |
| static java.lang.String | **ADV_EXC**<br>Location of the `adv.exc` file. |
| static java.lang.String | **ELKB_PATH**<br>Location of the *ELKB* data directory. |
| static java.lang.String | **NOUN_EXC**<br>Location of the `noun.exc` file. |
| static java.lang.String | **USER_HOME**<br>Location of user's `Home` directory. |
| static java.lang.String | **VERB_EXC**<br>Location of the `verb.exc` file. |

## Constructor Summary

| | |
|---|---|
| **Morphy**()<br>Default constructor. | |

## Method Summary

| | |
|---|---|
| java.util.HashSet | **getBaseForm**(java.lang.String words)<br>Reruns all the base forms for a given word. |
| static void | **main**(java.lang.String[] args)<br>Allows the `Morphy` class to be used via the command line. |

**ca.site.elkb**
# Class Paragraph

```
java.lang.Object
  |
  +--ca.site.elkb.Paragraph
```

public class **Paragraph**
extends java.lang.Object

Represents a *Roget's Thesaurus* Paragraph. A Paragraph is defined by the following attributes:

- Head number
- Paragraph name
- Paragraph keyword
- Part-of-speech
- list of Semicolon Groups
- number of Semicolon Groups
- number of words and phrases
- number of Cross-references
- number of See references

## Constructor Summary

| |
| --- |
| **Paragraph**() <br>     Default constructor. |
| **Paragraph**(int head, int para, java.lang.String p) <br>     Constructor which sets the Head number, Paragraph number and part-of-speech. |
| **Paragraph**(int head, int para, java.lang.String key, java.lang.String p) <br>     Constructor which sets the Head number, Paragraph number, keyword, and part-of-speech. |

## Method Summary

| | |
| --- | --- |
| void | **addSG**(java.lang.String sg) <br>     Adds a Semicolon Group, repreented as a string, to the Paragraph. |
| boolean | **equals**(java.lang.Object anObject) <br>     Compares this paragraph to the specified object. |
| java.lang.String | **format**() <br>     Converts to a string representation, similar to the printed format, the Paragraph object. |
| java.util.ArrayList | **getAllWordList**() <br>     Returns all of the words and phrases in a paragraph. |
| int | **getCRefCount**() <br>     Returns the number of Cross-references in this Paragraph. |
| int | **getHeadNum**() |
| | Returns the Head number of this Paragraph. |
| `java.lang.String` | **`getParaKey`**`()`<br>Returns the keyword of this Paragraph. |
| `int` | **`getParaNum`**`()`<br>Returns the number of this Paragraph. |
| `java.lang.String` | **`getPOS`**`()`<br>Returns the part-of-speech of this Paragraph. |
| `int` | **`getSeeCount`**`()`<br>Returns the number of See references in this Paragraph. |
| `ca.site.elkb.SG` | **`getSG`**`(int index)`<br>Returns the Semicolon Group at the specified position in the array of Semicolon Groups. |
| `ca.site.elkb.SG` | **`getSG`**`(java.lang.String word)`<br>Returns the the first Semicolon Group in this Paragraph which contains the given word. |
| `int` | **`getSGCount`**`()`<br>Returns the number of Semicolon Groups in this Paragraph. |
| `java.util.ArrayList` | **`getSGList`**`()`<br>Returns the array of Semicolon Groups of this Paragraph. |
| `int` | **`getWordCount`**`()`<br>Returns the number of words in this Paragraph. |
| `java.lang.String` | **`parseParaKey`**`(java.lang.String line)`<br>Extracts the keyword from a Semicolon Group represented as a string. |
| `void` | **`print`**`()`<br>Prints the contents of this Paragraph to the standard output. |
| `void` | **`printAllSG`**`()`<br>Prints all the contents of all Semicolon Groups, including references, without any special formatting. |
| `void` | **`printAllWords`**`()`<br>Prints all of the words and phrases in the Paragraph on a separate line to the standard output. |
| `void` | **`setHeadNum`**`(int num)`<br>Sets the Head number of this Paragraph. |
| `void` | **`setParaKey`**`(java.lang.String key)`<br>Sets the keyword of this Paragraph. |
| `void` | **`setParaNum`**`(int num)`<br>Sets the number of this Paragraph. |
| `void` | **`setPOS`**`(java.lang.String p)`<br>Sets the part-of-speech of this Paragraph. |
| `java.lang.String` | **`toString`**`()`<br>Converts to a string representation the `Paragraph` object. |

**ca.site.elkb**
# Class Path

```
java.lang.Object
  |
  +--ca.site.elkb.Path
```

**All Implemented Interfaces:**
> java.lang.Comparable

---

public class **Path**
extends java.lang.Object
implements java.lang.Comparable

Represents a path in *Roget's Thesaurus* between two words or phrases.

---

## Constructor Summary

| |
|---|
| **Path**() |
|     Default constructor. |
| **Path**(java.util.ArrayList path) |
|     Constructor that initialized this `Path` object with a Path. |

## Method Summary

| | |
|---:|---|
| int | **compareTo**(java.lang.Object other) |
| |     Compares two paths. |
| java.lang.String | **getKeyWord1**() |
| |     Returns the keyword of the the first word or phrase in this Path. |
| java.lang.String | **getKeyWord2**() |
| |     Returns the keyword of the the second word or phrase in this Path. |
| java.lang.String | **getPath**() |
| |     Returns the path between the first and second word or phrase. |
| java.lang.String | **getPathInfo1**() |
| |     Returns the location in the ontology of the first word or phrase in this Path. |
| java.lang.String | **getPathInfo2**() |
| |     Returns the location in the ontology of the second word or phrase in this Path. |
| java.lang.String | **getPos1**() |
| |     Returns the part-of-speech of the the first word or phrase in this Path. |
| java.lang.String | **getPos2**() |
| |     Returns the part-of-speech of the the second word or phrase in this Path. |
| java.lang.String | **getWord1**() |
| |     Returns the first word or phrase in this Path. |
| java.lang.String | **getWord2**() |
| |     Returns the second word or phrase in this Path. |

| int | **length**`()`<br>Returns the number of elements in this Path. |
|---:|---|
| int | **size**`()`<br>Returns the length in this Path. |
| `java.lang.String` | **toString**`()`<br>Converts to a string representation the `Path` object. |

**ca.site.elkb**
# Class PathSet

```
java.lang.Object
   |
   +--ca.site.elkb.PathSet
```

**All Implemented Interfaces:**
> java.lang.Comparable

---

public class **PathSet**
extends java.lang.Object
implements java.lang.Comparable

A set that contains all of the paths between two words and phrases as well as the number of minimum length paths. This class is used to measure semantic distance.

The PathSet also contains the original strings before any morphological transformations of modifications of phrases These are contained in `origWord1` and `origWord2`.

---

## Constructor Summary

| |
|---|
| **PathSet**() |
|     Default constructor. |
| **PathSet**(java.util.TreeSet pathSet) |
|     Constructor that initialized this `PathSet` object with a PathSet. |

## Method Summary

| | |
|---:|---|
| int | **compareTo**(java.lang.Object other)<br>    Compares two PathSets according to the length of the shortest path. |
| java.util.TreeSet | **getAllPaths**()<br>    Returns all Paths in this PathSet. |
| int | **getMinLength**()<br>    Returns the length of the shortest Path in this PathSet. |
| int | **getMinPathCount**()<br>    Returns the number of minimum length Paths in this PathSet. |
| java.lang.String | **getOrigWord1**()<br>    Returns the original form of the first word or phrase in this PathSet. |
| java.lang.String | **getOrigWord2**()<br>    Returns the original form of the second word or phrase in this PathSet. |
| java.lang.String | **getPos1**()<br>    Returns the part-of-speech of the first word or phrase in this PathSet. |
| java.lang.String | **getPos2**()<br>    Returns the part-of-speech of the second word or phrase in this PathSet. |
| `java.lang.String` | **<u>getWord1</u>**`()`<br>    Returns the first word or phrase after the morphological transformations are applied in this PathSet. |
| `java.lang.String` | **<u>getWord2</u>**`()`<br>    Returns the second word or phrase after the morphological transformations are applied in this PathSet. |
| `java.lang.String` | **<u>getWordPair</u>**`()`<br>    Converts to a string representation the `PathSet` object - used for debugging. |
| `void` | **<u>setOrigWord1</u>**`(java.lang.String word)`<br>    Sets the original form of the first word or phrase in this PathSet. |
| `void` | **<u>setOrigWord2</u>**`(java.lang.String word)`<br>    Sets the original form of the second word or phrase in this PathSet. |
| `java.lang.String` | **<u>toString</u>**`()`<br>    Converts to a string representation the `PathSet` object. |

**ca.site.elkb**
# Class Reference

```
java.lang.Object
   |
   +--ca.site.elkb.Reference
```

**All Implemented Interfaces:**
>        java.io.Serializable

**Direct Known Subclasses:**
>        SemRel

---

public class **Reference**
extends java.lang.Object
implements java.io.Serializable

Represents a symbolic pointer to a location where a specific word or phrase can be found in *Roget's Thesaurus*. A reference is identified by a keyword, head number and part of speech sequence.

An example of a Reference is: *obstetrics* 167 n. This instance of a `Reference` is represented as:

- **Reference name**: obstetrics
- **Head number**: 167
- **Part-of-speech**: N.

A Reference is always liked to an index entry, for example: *stork*.

**See Also:**
>        Serialized Form

---

## Constructor Summary

| |
|---|
| **Reference**`()` |
|     Default constructor. |
| **Reference**`(java.lang.String ref)` |
|     Constructor that creates a `Reference` object by parsing a string. |
| **Reference**`(java.lang.String name, int head, java.lang.String p)` |
|     Constructor which sets the reference name, Head number and part-of-speech. |
| **Reference**`(java.lang.String name, int head, java.lang.String p, java.lang.String entry)` |
|     Constructor which sets the referebnce name, Head number, part-of-speech, and Index entry. |

## Method Summary

| | |
|---:|---|
| int | **getHeadNum**`()` |
| |     Returns the Head number of this Reference. |
| java.lang.String | **getIndexEntry**`()` |
| |     Returns the Index entry of this Reference. |
| `java.lang.String` | **getPos**`()`<br>      Returns the part-of-speech of this Reference. |
| `java.lang.String` | **getRefName**`()`<br>      Returns the name of this Reference. |
| `void` | **print**`()`<br>      Prints this Reference to the standard output. |
| `void` | **setHeadNum**`(int head)`<br>      Sets the Head number of this Reference. |
| `void` | **setIndexEntry**`(java.lang.String entry)`<br>      Sets the Index entry of this Reference. |
| `void` | **setPos**`(java.lang.String p)`<br>      Sets the part-of-speech of this Reference. |
| `void` | **setRefName**`(java.lang.String name)`<br>      Sets the name of this Reference. |
| `java.lang.String` | **toString**`()`<br>      Converts to a string representation the `Reference` object. |

**ca.site.elkb**
# Class RogetClass

```
java.lang.Object
   |
   +--ca.site.elkb.RogetClass
```

public class **RogetClass**
extends java.lang.Object

Represents the topmost element in *Roget's Thesaurus Tabular Synopsis of Categories*. It is represented by its number, name, subclass name if it is a subclass of an original Roget Class, and range of Sections that it contains. For example, Class *4. Intellect: the exercise of the mind (Formation of ideas)* is represented as:

- **Class number**: 4
- **Class number in string format**: Class four
- **Class Name**: Intellect: the exercise of the mind
- **First section**: 16
- **Last section**: 22

---

## Constructor Summary

**RogetClass**()
    Default constructor.

**RogetClass**(int num, java.lang.String name)
    Constructor which sets the Class number and name.

**RogetClass**(int num, java.lang.String name, int start, int end)
    Constructor which sets the Class number and name, as well as the first and last Section number.

**RogetClass**(int num, java.lang.String strClassNum,
java.lang.String strClassName)
    Constructor which sets the Class number, Class number in string format and Class name, while parsing the strings for the Class number and name.

**RogetClass**(int num, java.lang.String snum, java.lang.String name, int start,
int end)
    Constructor which sets the Class number, Class number in string format, Class name, as well as the first and last Section number.

**RogetClass**(int num, java.lang.String snum, java.lang.String name,
java.lang.String subClass)
    Constructor which sets the Class number, Class number in string format, Class and Sub-class name.

**RogetClass**(int num, java.lang.String snum, java.lang.String name,
java.lang.String subClass, int start, int end)
    Constructor which sets the Class number, Class number in string format, Class name, Sub-class name as well as the first and last Section number.

---

## Method Summary

| | |
|---:|:---|
| void | **addSection**(`ca.site.elkb.Section section`)<br>Adds a Section to this RogetClass. |
| java.lang.String | **getClassName**()<br>Returns the name of this RogetClass. |
| int | **getClassNum**()<br>Returns the number of this RogetClass. |
| int | **getSectionEnd**()<br>Returns the number of the last section of this RogetClass. |
| java.util.ArrayList | **getSectionList**()<br>Returns the array of `Section` objects in this RogetClass. |
| int | **getSectionStart**()<br>Returns the number of the first section of this RogetClass. |
| java.lang.String | **getStrClassNum**()<br>Returns the number of this RogetClass in string format. |
| java.lang.String | **getSubClassName**()<br>Returns the Sub-class name of this RogetClass. |
| int | **headCount**()<br>Returns the number of Heads of this RogetClass. |
| void | **print**()<br>Prints the contents of this RogetClass to the standard output. |
| int | **sectionCount**()<br>Returns the number of Sections of this RogetClass. |
| void | **setClassName**(`java.lang.String name`)<br>Sets the name of this RogetClass. |
| void | **setClassNum**(`int num`)<br>Sets the number of this RogetClass. |
| void | **setSectionEnd**(`int end`)<br>Sets the number of the last section of this RogetClass. |
| void | **setSectionStart**(`int start`)<br>Sets the number of the first section of this RogetClass. |
| void | **setStrClassNum**(`java.lang.String snum`)<br>Sets the number of this RogetClass in string format. |
| void | **setSubClassName**(`java.lang.String subClass`)<br>Sets the Sub-class name of this RogetClass. |
| java.lang.String | **toString**()<br>Converts to a string representation the `RogetClass` object. |

**ca.site.elkb**
# Class RogetELKB

```
java.lang.Object
   |
   +--ca.site.elkb.RogetELKB
```

public class **RogetELKB**
extends java.lang.Object

Main class of the *Roget's Thesaurus Electronic Lexical KnowledgeBase*. It is made up of three major components:

- the Index
- the Tabular Synopsis of Categories
- the Text

Required files:

- `elkbIndex.dat`: The Index in binary file format.
- `rogetMap.rt`: The *Tabular Synopsis of Categories*.
- `./heads/head*`: The 990 heads
- `AmBr.lst`: The American to British spelling word list.
- `noun.exc`, `adj.exc`, `verb.exc`, `adv.exc`: exception lists used for the morphological transformations.

These files are found in the `$HOME/roget_elkb` directory.

## Field Summary

| | |
|---|---|
| static java.lang.String | **CATEG**<br>Location of the *ELKB Tabular Synopsis of Categories*. |
| ca.site.elkb.Category | **category**<br>The *ELKB Tabular Synopisis of Categories*. |
| static java.lang.String | **ELKB_PATH**<br>Location of the *ELKB* data directory. |
| static java.lang.String | **HEADS**<br>Location of the Heads. |
| ca.site.elkb.Index | **index**<br>The *ELKB* Index. |
| static java.lang.String | **INDEX**<br>Location of the *ELKB* Index. |
| ca.site.elkb.RogetText | **text**<br>The *ELKB* Text. |
| static java.lang.String | **USER_HOME**<br>Location of user's Home directory. |

## Constructor Summary

**RogetELKB**()
    Default constructor.

## Method Summary

| | |
|---:|---|
| java.util.TreeSet | **getAllPaths**(java.lang.String strWord1, java.lang.String strWord2)<br>    Returns all the paths between two words or phrases. |
| java.util.TreeSet | **getAllPaths**(java.lang.String strWord1, java.lang.String strWord2, java.lang.String POS)<br>    Returns all the paths between two words or phrases of a given part-of-speech. |
| static void | **main**(java.lang.String[] args)<br>    Allows the *ELKB* to be used via the command line. |
| ca.site.elkb.Path | **path**(java.lang.String strWord1, java.lang.String strRef1, java.lang.String strWord2, java.lang.String strRef2)<br>    Calculates the path between two senses of words or phrases. |
| java.lang.String | **t1Relation**(java.lang.String strWord1, int iHeadNum1, java.lang.String sRefName1, java.lang.String sPos1, java.lang.String strWord2)<br>    Determines the thesaural relation that exists between a specific sense of a words or phrases and another word or phrase. |
| java.lang.String | **t1Relation**(java.lang.String strWord1, java.lang.String strWord2)<br>    Determines the thesaural relation that exists between two words or phrases. |

**ca.site.elkb**
# Class RogetText

```
java.lang.Object
   |
   +--ca.site.elkb.RogetText
```

**All Implemented Interfaces:**
> java.io.Serializable

---

public class **RogetText**
extends java.lang.Object
implements java.io.Serializable

Represents the Text of *Roget's Thesaurus*. The following information is maintained for the Text:

- number of Heads
- number of Paragraphs
- number of words and phrases
- number of Semicolon Groups
- number of Cross-references
- number of See references

This information is also kept for all nouns, adjectives, verbs, adverbs and interjections.

**See Also:**
> <u>Serialized Form</u>

---

## Constructor Summary

| |
|---|
| **RogetText**() |
|     Default constructor. |
| **RogetText**(int capacity) |
|     Constructor which specifies the number of Heads contained in this RogetText. |
| **RogetText**(int capacity, java.lang.String fileName) |
|     Constructor that builds the RogetText object by specifying the number of Heads and using the information contained files which end with .txt. |
| **RogetText**(int capacity, java.lang.String fileName, java.lang.String extension) |
|     Constructor that builds the RogetText object by specifying the number of Heads and using the information contained files which end with the given extension. |
| **RogetText**(java.lang.String path) |
|     Constructor which specifies the directory in which the Heads are found. |

## Method Summary

| | |
|---:|:---|
| void | **addHead**(`ca.site.elkb.Head headObj`)<br>Adds a Head object to this RogetText. |
| void | **addHead**(`java.lang.String fileName`)<br>Adds a Head which is contained in the specified file to this RogetText. |
| int | **getAdjCount**()<br>Returns the number of adjectives in this RogetText. |
| int | **getAdjCRefCount**()<br>Returns the number of adjective Cross-references in this RogetText. |
| int | **getAdjParaCount**()<br>Returns the number of adjective Paragraphs in this RogetText. |
| int | **getAdjSeeCount**()<br>Returns the number of adjective See referencs in this RogetText. |
| int | **getAdjSGCount**()<br>Returns the number of ajective Semicolon Groups in this RogetText. |
| int | **getAdvCount**()<br>Returns the number of adverbs in this RogetText. |
| int | **getAdvCRefCount**()<br>Returns the number of adverb Cross-references in this RogetText. |
| int | **getAdvParaCount**()<br>Returns the number of adverb Paragraphs in this RogetText. |
| int | **getAdvSeeCount**()<br>Returns the number of adverb See referencs in this RogetText. |
| int | **getAdvSGCount**()<br>Returns the number of adverb Semicolon Groups in this RogetText. |
| int | **getCRefCount**()<br>Returns the number of Cross-references in this RogetText. |
| ca.site.elkb.Head | **getHead**(`int headNum`)<br>Returns the Head with the specified number. |
| int | **getHeadCount**()<br>Returns the number of Heads in this RogetText. |
| int | **getIntCount**()<br>Returns the number of interjections in this RogetText. |
| int | **getIntCRefCount**()<br>Returns the number of interjection Cross-references in this RogetText. |
| int | **getIntParaCount**()<br>Returns the number of interjection Paragraphs in this RogetText. |
| int | **getIntSeeCount**()<br>Returns the number of interjection See referencs in this RogetText. |
| int | **getIntSGCount**()<br>Returns the number of interjection Semicolon Groups in this RogetText. |
| int | **getNCount**() |

| | | |
|---|---|---|
| | | Returns the number of nouns in this RogetText. |
| | int | **getNCRefCount**() <br> Returns the number of noun Cross-references in this RogetText. |
| | int | **getNParaCount**() <br> Returns the number of noun Paragraphs in this RogetText. |
| | int | **getNSeeCount**() <br> Returns the number of noun See referencs in this RogetText. |
| | int | **getNSGCount**() <br> Returns the number of noun Semicolon Groups in this RogetText. |
| | int | **getParaCount**() <br> Returns the number of Paragraphs in this RogetText. |
| | int | **getSeeCount**() <br> Returns the number of See referencs in this RogetText. |
| | int | **getSGCount**() <br> Returns the number of Semicolon Groups in this RogetText. |
| | int | **getVbCount**() <br> Returns the number of verbs in this RogetText. |
| | int | **getVbCRefCount**() <br> Returns the number of verb Cross-references in this RogetText. |
| | int | **getVbParaCount**() <br> Returns the number of verb Paragraphs in this RogetText. |
| | int | **getVbSeeCount**() <br> Returns the number of verb See referencs in this RogetText. |
| | int | **getVbSGCount**() <br> Returns the number of verb Semicolon Groups in this RogetText. |
| | int | **getWordCount**() <br> Returns the number of words and phrases in this RogetText. |
| | void | **printHead**(int headNum) <br> Prints the contents of a Head specified by its number to the standard output. |
| | java.lang.String | **toString**() <br> Converts to a string representation the RogetText object. |

**ca.site.elkb**
# Class Section

```
java.lang.Object
   |
   +--ca.site.elkb.Section
```

public class **Section**
extends java.lang.Object

Represents a *Roget's Thesaurus* Section. A Section is defined by the following attributes:

- Section number
- Section number in string format
- Section name
- number of the first Head
- number of the last Head
- array of Heads

A Section can contain `Head` or `HeadInfo` objects, depending on the use.

## Constructor Summary

| |
|---|
| **Section**() <br>     Default constructor. |
| **Section**(int number, java.lang.String name) <br>     Constructor which sets the Section number and name. |
| **Section**(int number, java.lang.String name, int start, int end) <br>     Constructor which sets the Section number and name, as well as the number of the first and last Head. |
| **Section**(int number, java.lang.String strNum, java.lang.String strName) <br>     Constructor which sets the Section number, name, and Section number in string format and Class name, while parsing the strings for the Section number and name. |

## Method Summary

| | |
|---:|---|
| void | **addHeadInfo**(ca.site.elkb.HeadInfo head) <br>     Adds a `HeadInfo` object to this Section. |
| int | **getHeadEnd**() <br>     Returns the number of the last Head of this Section. |
| java.util.ArrayList | **getHeadInfoList**() <br>     Returns the array of `HeadInfo` objects of this Section. |
| int | **getHeadStart**() <br>     Returns the number of the first Head of this Section. |
| java.lang.String | **getSectionName**() <br>     Returns the name of this Section. |
| int | **getSectionNum**`()`<br>Returns the number of this Section. |
| java.lang.String | **getStrSectionNum**`()`<br>Returns the number of this Section in string format. |
| int | **headCount**`()`<br>Returns the number of Heads in this Section. |
| void | **print**`()`<br>Prints the content of this Section to the standard output. |
| void | **printHeadInfo**`()`<br>Prints the information regarding the Heads contained in this Section to the standard output. |
| void | **setHeadEnd**`(int end)`<br>Sets the number of the last Head of this Section. |
| void | **setHeadStart**`(int start)`<br>Sets the number of the first Head of this Section. |
| void | **setSectionName**`(java.lang.String name)`<br>Sets the number of this Section in string format. |
| void | **setSectionNum**`(int num)`<br>Sets the number of this Section. |
| void | **setStrSectionNum**`(java.lang.String snum)`<br>Sets the number of this Section in string format. |
| java.lang.String | **toString**`()`<br>Converts to a string representation the `Section` object. |

**ca.site.elkb**
# Class SemRel

```
java.lang.Object
  |
  +--ca.site.elkb.Reference
        |
        +--ca.site.elkb.SemRel
```

**All Implemented Interfaces:**
> java.io.Serializable

---

public class **SemRel**
extends Reference

Represents a *Roget's Thesaurus* relation between a word or phrase. This can be a Cross-reference or a See reference. For example:

- See *drug taking*
- 646 *perfect*

Relation types currently used by the *ELKB* are `cref` and `see`.

**See Also:**
> Serialized Form

---

## Constructor Summary

| |
|---|
| **SemRel**() <br>     Default constructor. |
| **SemRel**(java.lang.String t, int headNum, java.lang.String refName) <br>     Constructor which sets the relation type, Head number and Reference name. |

---

## Method Summary

| | |
|---|---|
| java.lang.String | **getType**() <br>     Returns the relation type. |
| void | **print**() <br>     Prints this relation to the standard output. |
| void | **setType**(java.lang.String t) <br>     Sets the relation type. |
| java.lang.String | **toString**() <br>     Converts to a string representation the SemRel object. |

---

## Methods inherited from class ca.site.elkb.Reference

getHeadNum, getIndexEntry, getPos, getRefName, setHeadNum, setIndexEntry,

`setPos`, `setRefName`

**ca.site.elkb**
# Class SG

```
java.lang.Object
  |
  +--ca.site.elkb.SG
```

public class **SG**
extends java.lang.Object

Represents a *Roget's Thesaurus* Semicolon Group. For example:

- zeal, ardour, ernestness, seriousness;

A Semicolon Group is defined by the following attributes:

- Head number
- Paragraph number
- Paragraph keyword
- Part-of-speech
- Semicolon Group number
- number of Cross-references
- number of See references
- number of See references
- list of word and phrases
- list of special tags for the words and phrases
- list of references

## Constructor Summary

**SG**()
    Default constructor.

**SG**(int numSG, int numP, int numH, java.lang.String text, java.lang.String p)
    Constructor that sets the Semicolon Group number, Paragraph number, Head number, the words and phases of the Semicolon Group and the part-of-speech.

**SG**(int num, java.lang.String text)
    Constructor that sets the Semicolon Group number and the words and phases that it contains.

## Method Summary

| | |
|---:|---|
| void | **addSemRel**(ca.site.elkb.SemRel rel)<br>    Adds a relation to this Semicolon Group |
| void | **addWord**(java.lang.String word)<br>    Adds a word or phrase to this Semicolon Group. |
| void | **addWord**(java.lang.String word,      java.lang.String tag)<br>    Adds a word or phrase and its style tag to this Semicolon Group. |
| `java.lang.String` | **format**`()`<br>Returns this Semicolon Group formatted in a string, including references, style tags and punctuation. |
| `java.util.ArrayList` | **getAllWordList**`()`<br>Returns the list of words and phrases, including the references, contained in this Semicolon Group. |
| `int` | **getCRefCount**`()`<br>Returns the number Cross-references in this Semicolon Group. |
| `java.lang.String` | **getGroup**`()`<br>Returns a string containing all of the words and phrases in the Semicolon Group minus the references. |
| `int` | **getHeadNum**`()`<br>Returns the Head number of this Semicolon Group. |
| `java.lang.String` | **getOffset**`()`<br>Returns a symbolic adress of this Semicolon Group. |
| `java.util.ArrayList` | **getOffsetList**`()`<br>Returns a list of Semicolon Groups with their symbolic adresses. |
| `java.lang.String` | **getParaKey**`()`<br>Returns the Paragraph keyword of this Semicolon Group. |
| `int` | **getParaNum**`()`<br>Returns the Paragraph number of this Semicolon Group. |
| `java.lang.String` | **getPOS**`()`<br>Returns the part-of-speech of this Semicolon Group. |
| `java.lang.String` | **getReference**`()`<br>Returns a string containing only the references of this Semicolon Group. |
| `int` | **getSeeCount**`()`<br>Returns the number See refereces in this Semicolon Group. |
| `java.util.ArrayList` | **getSemRelList**`()`<br>Returns the list of relations of this Semicolon Group. |
| `int` | **getSGNum**`()`<br>Returns the number of this Semicolon Group. |
| `java.util.ArrayList` | **getStyleTagList**`()`<br>Returns the list of style tags of this Semicolon Group. |
| `int` | **getWordCount**`()`<br>Returns the number of words and phrases in this Semicolon Group. |
| `java.util.ArrayList` | **getWordList**`()`<br>Returns the list of words and phrases, minus the references, contained in this Semicolon Group. |
| `void` | **print**`()`<br>Prints this Semicolon Group to the standard output. |
| `void` | **setHeadNum**`(int num)`<br>Sets the Head number of this Semicolon Group. |
| `void` | **setParaKey**`(java.lang.String key)` |

| | | |
|---|---|---|
| | | Sets the Paragraph keyword of this Semicolon Group. |
| | void | **setParaNum**(int num) <br> Sets the Paragraph number of this Semicolon Group. |
| | void | **setPOS**(java.lang.String p) <br> Sets the part-of-speech of this Semicolon Group. |
| | void | **setSGNum**(int num) <br> Sets the number of this Semicolon Group. |
| | void | **setText**(java.lang.String text) <br> Sets the words and phrases used in this Semicolon Group. |
| | java.lang.String | **toString**() <br> Converts to a string representation the SG object. |

**ca.site.elkb**
# Class SubSection

```
java.lang.Object
   |
   +--ca.site.elkb.SubSection
```

---

public class **SubSection**
extends java.lang.Object

Represents a *Roget's Thesaurus* Sub-section. A Sub-section may or may not exist. Here is an example:

- **Class one**: Abstract Relations
- **Section one**: Existence
- **Sub-section title**: Abstract
- **Head group**:1 Existence - 2 Nonexistence

Sub-sections may contain several Head groups.

---

## Constructor Summary

| |
|---|
| **SubSection**() |
|     Default constructor. |
| **SubSection**(int start) |
|     Constructor which sets the number of the first Head. |
| **SubSection**(int start, java.lang.String sInfo) |
|     Constructor which sets the number of the first Head and the name of the Section supplied as a string to be parsed. |
| **SubSection**(java.lang.String sInfo) |
|     Constructor which sets the name of the Section by parsing a string. |

## Method Summary

| | |
|---|---|
| void | **addGroup**(ca.site.elkb.Group group)<br>    Adds a Head Group to this Sub-section. |
| int | **getGroupCount**()<br>    Returns the number of Head groups in this Sub-section. |
| java.util.ArrayList | **getGroupList**()<br>    Returns the list of Head groups in this Sub-section. |
| int | **getHeadCount**()<br>    Returns the number of Heads in this Sub-section. |
| int | **getHeadStart**()<br>    Returns the number of the first Head in this Sub-section. |
| void | **print**()<br>    Displays the content of a Sub-section in a similar way to *Roget's Thesaurus* |
| | *Tabular Synopisis of Categories* to the standard output. |
| void | **<u>setHeadStart</u>**(int start)<br>    Sets the number of the first Head in this Sub-section. |
| java.lang.String | **<u>toString</u>**()<br>    Converts to a string representation the `SubSection` object. |

**ca.site.elkb**
# Class Variant

```
java.lang.Object
  |
  +--ca.site.elkb.Variant
```

**All Implemented Interfaces:**
>     java.io.Serializable

---

public class **Variant**
extends java.lang.Object
implements java.io.Serializable

Allows to obtain a variant of an English spelling. A British spelling variant can be obtained form an American spelling and vice-versa.

The default American and British word list is `AmBr.lst` contained in the `$HOME/roget_elkb` directory. It is loaded by the default constructor.

**See Also:**
>     Serialized Form

---

## Field Summary

| | |
|---|---|
| static java.lang.String | **AMBR_FILE** |
| | Location of the default American and British spelling word list. |
| static java.lang.String | **ELKB_PATH** |
| | Location of the *ELKB* data directory. |
| static java.lang.String | **USER_HOME** |
| | Location of user's `Home` directory. |

## Constructor Summary

| |
|---|
| **Variant**() |
| Default constructor. |
| **Variant**(java.lang.String filename) |
| Constructor that builds the `Variant` object using the information contained in the specified file. |

## Method Summary

| | |
|---|---|
| java.lang.String | **amToBr**(java.lang.String american) |
| | Returns the British spelling of a word, or `null` if the word cannot be found. |
| java.lang.String | **brToAm**(java.lang.String british) |
| | Returns the American spelling of a word, or `null` if the word cannot be found. |

# Appendix C: The *ELKB* Graphical and Command Line Interfaces

This appendix presents the graphical and command line interfaces to the *ELKB* along with their related documentation.

## 1    The Graphical User Interface

The Graphical User Interface (GUI) is a configurable mechanism for querying the *ELKB*. The GUI is designed to be as versatile, intuitive and informative as the printed version of the *Thesaurus*. To use it, a user supplies a word or phrase that is looked up *ELKB*'s index. The interface returns a list of references if the given word or phrase is found. The user must select one to display the paragraph in which the word or phrase is contained. An example using the word `please` is shown Figure C1.

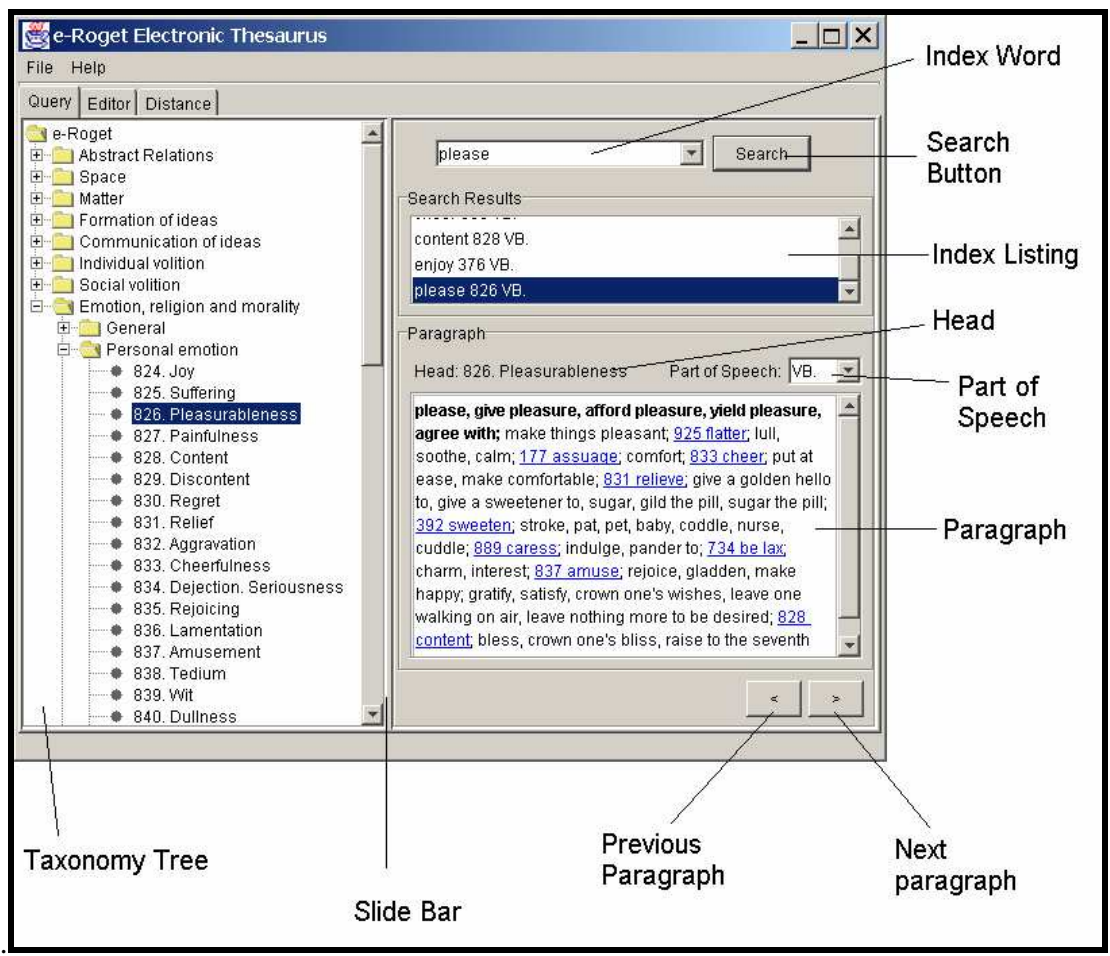

**Figure C1:** Screenshot of the GUI

Figure C1 shows the following parts of the GUI:

- **Index Word**: the word or phrase to be looked up. By hitting `Enter` or the clicking on the `Search Button` in the `Search Results` box. A history of queried words and phrases is maintained by the GUI.

- **Search Button**: searches the index for the word or phrase found in the `Index Word` box.

- **Search Results**: the result of the search is displayed in this box, also labeled as `Index Listings` in Figure C1. The user clicks on the desired reference to display the matching paragraph.

- **Paragraph**: a paragraph is displayed in this text box when the user clicks on a result in the `Search Results`, or when the user clicks on a head in the `Taxonomy Tree`. The GUI displays the semicolon group containing the `Index Word` in bold, and references to other heads in the *Thesaurus* as blue underlined text. A word or phrase in the `Paragraph` display can be selected by holding down the left mouse button while moving the mouse. If a word or phrase is selected, a menu appears with an option to perform a query on the selected text. If a user right clicks on a reference a popup menu appears with an option to follow the link. If the user clicks on the follow link menu item, the referenced paragraph appears in the `Paragraph` window. When text is selected, it can be copied to the system clipboard by pressing CTRL-C.

- **Taxonomy Tree**: An alternative way to use the *Thesaurus* is to browse the words and phrases using the classification system. A user can expand a node in the tree by double clicking on it, or by clicking on the "+" beside the node. If a user double clicks on a head, the first paragraph of the head appears in the `Paragraph` window. A node collapses when the "-" beside it is clicked, hiding any sub nodes.

- **Side Bar**: This bar can be moved left or right to modify the size of the `Taxonomy Tree` and the other half of the GUI.

- **Previous Paragraph**: this button displays the paragraph that precedes the one currently shown in the `Paragraph` window as ordered in *Roget's Thesaurus*.

- **Next Paragraph**: this button displays the paragraph that follows the one currently shown in the `Paragraph` window as ordered in *Roget's Thesaurus*.

- **Part of Speech:** each paragraph belongs to a part of speech. The part of speech of the currently displayed paragraph appears in a drop down list box. A user can display all of the different parts of speech that exist in the current head by clicking the down arrow of the list box. Clicking on one of the choices displays the first paragraph of the selection in the `Paragraph` window.

- **Head:** The head name and number of the currently displayed paragraph is shown here.

## 2    The Command Line Interface

The command line interface allows looking up a word or phrase, or calculating the distance between two words or phrases. Figure C2 shows the possible options of the interface, Figures C3 and C4 the steps for looking up the word `please`, and Figure C5 the distance between words `God` and `Yahweh`.

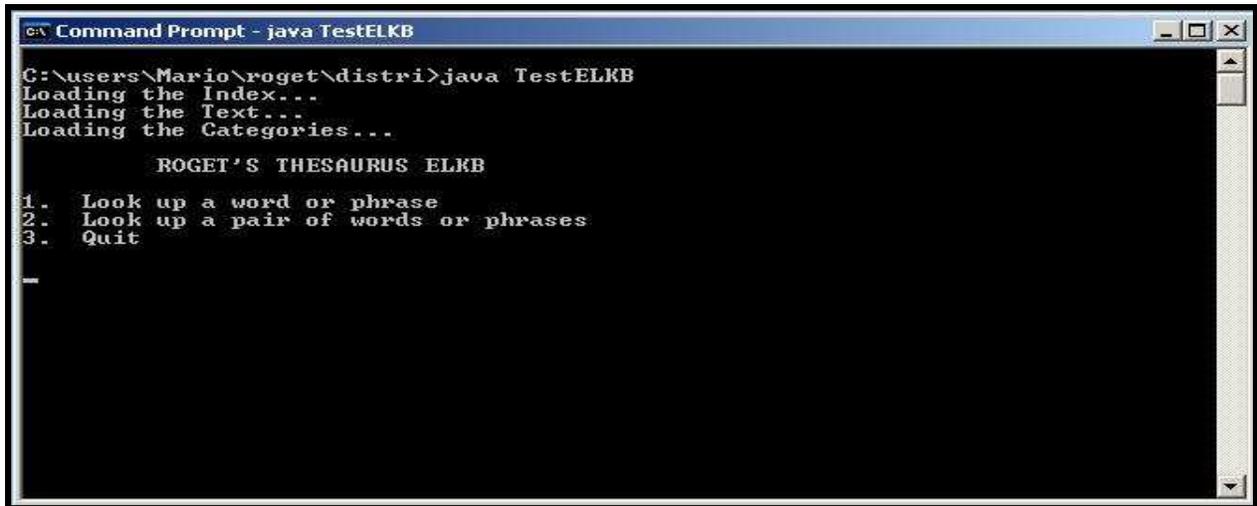

**Figure C2:** Screenshot of the command line interface

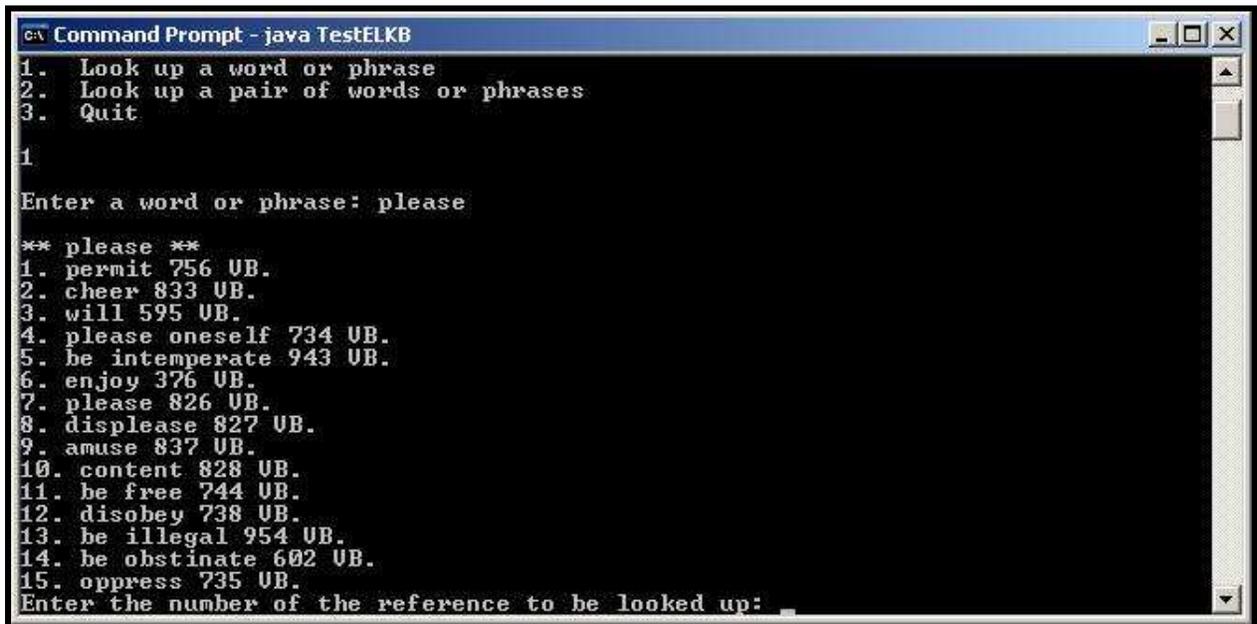

**Figure C3:** The references of the word `please`

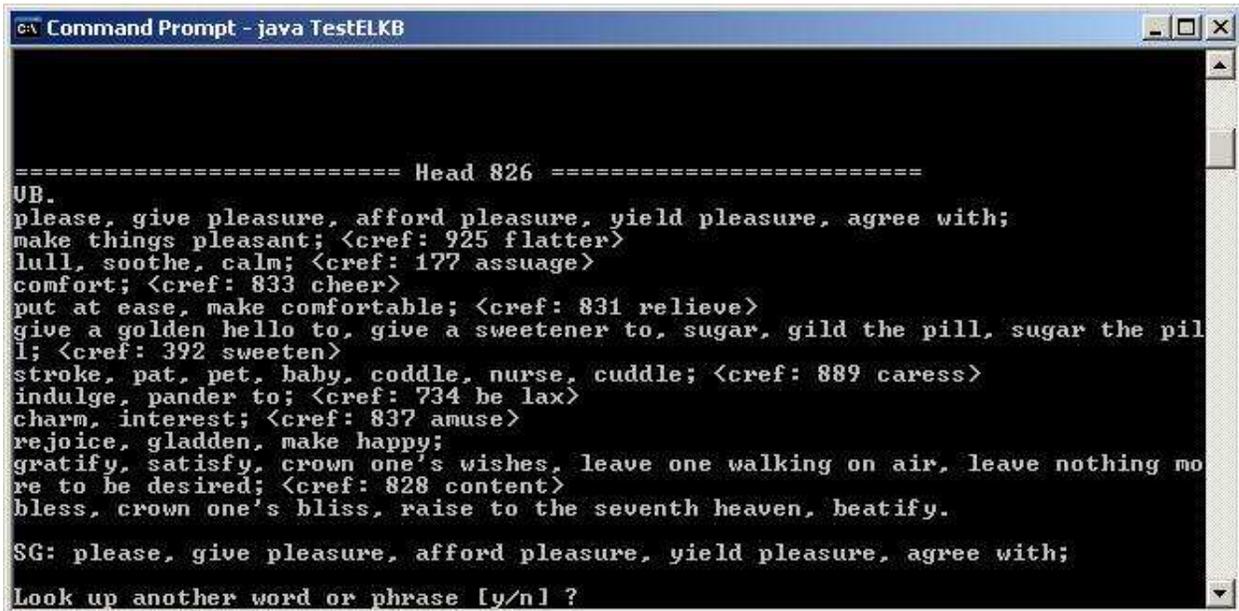

**Figure C4:** The paragraph of the reference `7. please 826 VB`.

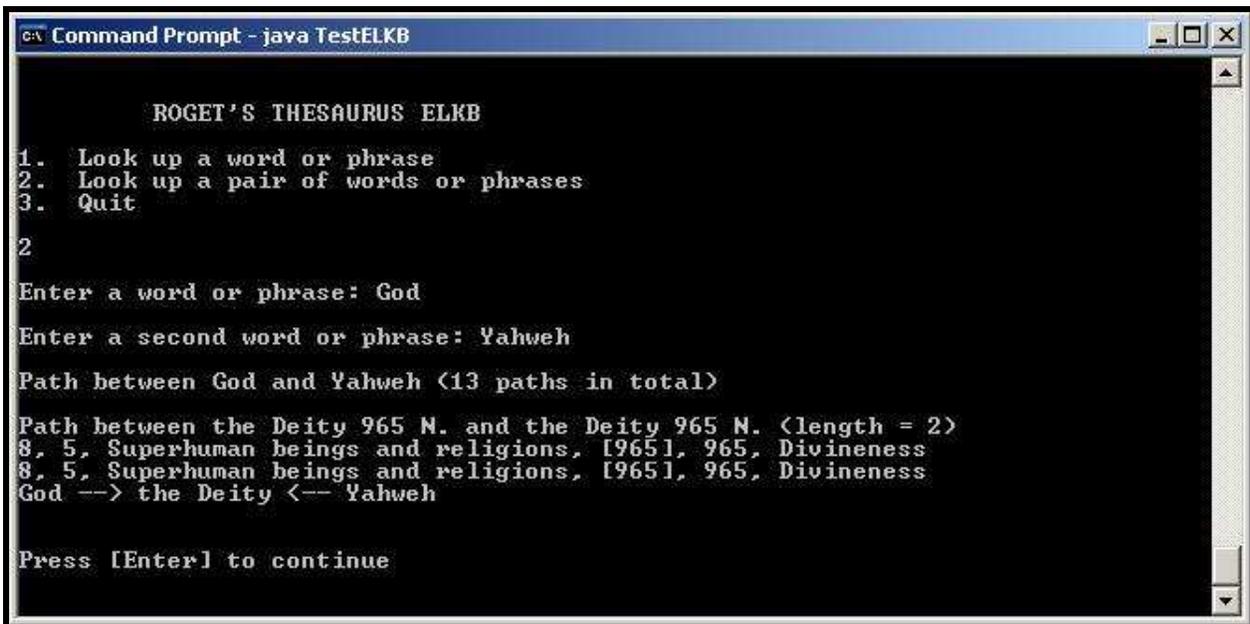

**Figure C5:** The distance between words `God` and `Yahweh`

# Appendix D: The Programs Developed for the Thesis

This appendix lists the programs developed for the thesis. The programs can be classified in three categories: preparation of the lexical material for the *ELKB*, testing and use of the *ELKB* and experiments. The programs implemented for this thesis are:

- Preparation of the lexical material:

  - o **`format:`** Perl program that converts the Pearson source files into a format recognizable by the ELKB.

  - o **`getHeads.pl:`** Perl program that takes Pearson Text file converted by the `format` program and separates it into 990 files, one for each head.

  - o **`ELKBWords:`** Java Program that lists all of the words and phrases found in the *ELKB* as well as their paragraph keyword, head number and part-of-speech.

  - o **`createIndex.pl:`** Perl program that takes the output of `ELKBWords` and converts it into an Index file to be used by the *ELKB*.

  - o **`index2.pl:`** Perl program that removes errors from the output of `createIndex.pl`.

  - o **`MakeBinIndex:`** Java program that takes the output of `index2.pl` and transforms the Index file in a binary format to be used by the *ELKB*.

- Testing of the *ELKB*:

  - o **`Driver:`** Java program that tests the various methods of the *ELKB*.

  - o **`TestELKB:`** Java program that implements the command-line interface of the *ELKB*.

  - o **`CERoget:`** Java program that implements the graphical user interface of the *ELKB*.

- Experiments:

  - o **`Similarity:`** Java program used for the experiments on semantic similarity.

  - o **`similarity.pl:`** Perl program used for the experiments on semantic similarity.

  - o **`LexicalChain:`** Java program that builds lexical chains using the *ELKB*.

It is enough to run most of these programs to know how they should be used. The preparation of the lexical material requires special attention. I have supplied here the documentation to perform this task.

## The Preparation of the Lexical Material for use by the *ELKB*.

The `format`, `getHeads.pl`, `ELKBWords`, `createIndex.pl`, `index2.pl` and `MakeBinIndex` programs must be used in the following manner to convert the Pearson source files. This procedure will create the Text and Index files to be used by the *ELKB*.

1. Concatenate all of the `rogetXXX.txt` files supplies by Pearson Education.
2. Run the `format` script. Usage: `format -t input_file output_file`. The `format` program can be used to convert the format of the Pearson Index files by using the `-i` flag.
3. Run the `getHeads.pl` script on the resulting file. This creates a `heads` directory that contains the 990 Heads.
4. Copy the heads directory to the user's home directory.
5. Run `ELKBWords` and re-direct the output to a file
6. Sort the file using `sort`.
7. Remove duplicate entries using `uniq`.
8. Remove errors by hand, these include:
   - the first few hundred lines
   - lines specified by `Error:`
   - lines containing `At line:`
9. Run the `createIndex.pl` program on the resulting file.
10. Run the `index2.pl` script on the output of `createIndex.pl`

# Appendix E: Converting the Pearson Codes into HTML-like Tags

The text files supplied by Pearson's Education are not easy to read, and they use their own specific codes. Even though these codes are explained in their documentation, it is preferable to use easily understood HTML-like tags. Here are the first few lines of the Text file:

```
#t#6Class one
#L#6Abstract Relations
#U#5Section one
#V#1Existence
#H#3[001] 1 Existence
#S#1N.
#H#3[001] 1 Existence
#S#1N.
#D#10        /    /............./../..../...../././../.............../
#T#6existence, #5being, entity; absolute being, the
absolute 965#6divineness#5; aseity, self-existence; monad, a being,
an entity, ens, essence, #1$:#5quiddity; Platonic idea, universal;
subsistence 360 #6life#5; survival, eternity 115 #6perpetuity#5; preexistence
119 #6priority#5; this life 121 #6present time#5; existence in space,
prevalence 189 #6presence#5; entelechy, realization, becoming, evolution
147 #6conversion#5; creation 164 #6production#5; potentiality 469
#6possibility#5; ontology, metaphysics; realism, materialism, idealism,
existentialism 449 #6philosophy#5.
```

The *format* Perl script performs the following sixteen steps to convert the Text file:

[1]   Replace the `#t#6` codes indicating a Class number and italics by `<classNumber>#<i>`. A closing `</i>#</classNumber>` tag is added and the Class number is separated by `#`.

example: **#t#6Class one** → **<classNumber>#<i>#Class one #</i>#</classNumber>**

[2]   Replace the `#L#6` codes indicating a Class title and italics by `<classTitle>#<i>`. A closing `</i>#</classTitle>` tag is added and the Class title is separated by `#`.

example: **#L#6Abstract Relations** →
         **<classTitle>#<i>#Abstract Relations #</i>#</classTitle>**

[3]   Replace the `#I#1` codes indicating a Sub-class title and small bold by `<subClassTitle>#<size=-1>#<b>`. A closing tag is added.

example: **#I#14.1 Formation of ideas** →
         **<subClassTitle>#<size=-1>#<b>**
                 **#4.1 Formation of ideas #**
         **</b>#</size>#</subClassTile>**

[4]   Replace the `#U#5` codes indicating a Section number and roman by `<sectionNumber>`. A closing `</sectionNumber>` tag is added.

example: **#U#5Section one** ⟶ **<sectionNumber>#Section one #</sectionNumber>**

[5]   Replace the `#V#1` codes indicating a Section title and small bold by `<sectionTitle>#` `<size=-1>`. Closing `</size>#</sectionTitle>` tags are added.

example: **#V#1Existence** ⟶
**<sectionTitle>#<size=-1>#<b>#Existence #</b>#</size>#</sectionTitle>**

[6]   Replace the `#H#3` codes indicating a Headword and large bold by `<headword>#<b>`. A closing `</b>#</headword>` tag is added and every field is separated by a `#`.

example: **#H#3[001] 1 Existence** ⟶
        **<headword>#<b>#[001] #1# Existence #</b>#</headword>**

[7]   Replace the `#S#1` codes indicating the part of speech and small bold symbol by `<pos>#<size=-1>#<b>`. A closing `</b>#</size>#</pos>` tag is added after the part of speech label.

example: **#S#1N.** ⟶ **<pos>#<size=-1>#<b>#N.#</b>#</size>#</pos>**

[8]   Replace the `#D#10 /`
`/............../../..../....../././../.............../` code by `<br>` that indicates a blank line.

example:  **#D#10 / /............../../..../....../././../.............../**
        ⟶ **<br>**

[9]   Replace the `#T` Paragraph code by `<paragraph>` and a new line.

example: **#T** ⟶ **<paragraph>**

[10]  Place every Semicolon Group on an individual line, label it with the `<sg>` `</sg>` tags and replace the `#6` (italic) and `#5` (roman) codes by `<i>` and `</i>`. Separate every word, phrase and final punctuation symbol with a comma.

example: **#6existence, #5being, entity;** ⟶
        **<sg><i>existence, </i>being, entity,;</sg>**

[11] Remove the `#1$:#5` codes which appear often in the files but do not mean anything.

example: **`#1$:#5quiddity` → `quiddity`**

[12] Label the Cross-references, `<number>#6<string>#5`, with the `<cref> </cref>` tags.

example: **`360 #6life#5` → `<cref>360 <i>life</i></cref>`**

[13] Label the See-references, `(#1see #6<string>#5)`, with the `<see> </see>` tags.

example: **`(#1see #6turmoil#5)` → `<see><i>turmoil</i></see>`**

[14] Finish the paragraph with a `</paragraph>` tag.

[15] Expand abbreviations when required.

[16] Additional tags `(derog)`, `(e)`, `(tdmk)`, and `(vulg)`, which respectively indicate words or phrases that are derogatory, of French origin that require a final "e" if applied to a woman, a registered trademark and vulgar, are replaced by `(<derog>)`, `(<e>)`, `(<tdmk>)`, and `(<vulg>)`.

The sample of the Pearson file after the substitutions looks like this:

```
<classNumber>#1#</classNumber>
<sectionNumber>#1#</sectionNumber>
<headword>#<b>#[001] #1# Existence #</b>#</headword>
<pos>#<size=-1>#<b>#N.#</b>#</size>#</pos>
<br>
<paragraph>
<sg><i>existence, </i>being, entity,¡</sg>
<sg>absolute being, the absolute, <cref>965<i>divineness</i></cref>,¡</sg>
<sg>aseity, self-existence,¡</sg>
<sg>monad, a being, an entity, ens, essence, quiddity,¡</sg>
<sg>Platonic idea, universal,¡</sg>
<sg>subsistence, <cref>360 <i>life</i></cref>,¡</sg>
<sg>survival, eternity, <cref>115 <i>perpetuity</i></cref>,¡</sg>
<sg>preexistence, <cref>119 <i>priority</i></cref>,¡</sg>
<sg>this life, <cref>121 <i>present time</i></cref>,¡</sg>
<sg>existence in space, prevalence, <cref>189 <i>presence</i></cref>,¡</sg>
<sg>entelechy, realization, becoming, evolution, <cref> 147<i>conversion</i>
    </cref>,¡</sg>
<sg>creation, <cref>164 <i>production</i></cref>,¡</sg>
<sg>potentiality, <cref>469 <i>possibility</i></cref>,¡</sg>
<sg>ontology, metaphysics,¡</sg>
<sg>realism, materialism, idealism, existentialism, <cref>449 <i>philosophy
    </i></cref>,.</sg>
</paragraph>
```

# Appendix F: Some Errors in the Pearson Source Files

This appendix lists errors that I have identified in the Pearson source files. 179 phrases where a space is missing and 26 words that are cut by a space have been found and corrected.

| File Name | Original string | Corrected string |
|---|---|---|
| roget016.txt | *andneedles* | *and needles* |
| roget036.txt | *andNicolette* | *and Nicolette* |
| roget028.txt | *anotherplace* | *another place* |
| roget029.txt | *antimissilemissile* | *antimissile missile* |
| roget029.txt | *antitankobstacles* | *antitank obstacles* |
| roget029.txt | *armstraffic* | *arms traffic* |
| roget023.txt | *artnouveau* | *art nouveau* |
| roget013.txt | *aslead* | *a slead* |
| roget013.txt | *bambooshoots* | *bamboo shoots* |
| roget009.txt | *becomehorizontal* | *become horizontal* |
| roget005.txt | *beunderpopulated* | *be underpopulated* |
| roget021.txt | *blunderhead* | *blunder head* |
| roget028.txt | *blunderhead* | *blunder head* |
| roget026.txt | *bookcollection* | *book collection* |
| roget020.txt | *bookwoman* | *book woman* |
| roget021.txt | *byallusion* | *by allusion* |
| roget011.txt | *caninetooth* | *canine tooth* |
| roget017.txt | *choirleader* | *choir leader* |
| roget004.txt | *cognizanceof* | *cognizance of* |
| roget021.txt | *commonplacebook* | *commonplace book* |
| roget016.txt | *deathby* | *death by* |
| roget026.txt | *dechoix* | *de choix* |
| roget022.txt | *deskwork* | *desk work* |
| roget011.txt | *dogpaddle* | *dog paddle* |
| roget011.txt | *dogsleigh* | *dog sleigh* |
| roget035.txt | *emptystomach* | *empty stomach* |
| roget033.txt | *excitedfeeling* | *excited feeling* |
| roget035.txt | *fearof* | *fear of* |
| roget024.txt | *fictionalbiography* | *fictional biography* |
| roget020.txt | *fineadjustment* | *fine adjustment* |
| roget020.txt | *flatteringhope* | *flattering hope* |
| roget015.txt | *foodplant* | *food plant* |
| roget021.txt | *goldendream* | *golden dream* |
| roget016.txt | *gothrough* | *go through* |
| roget036.txt | *hardbitten* | *hard bitten* |
| roget014.txt | *hardwater* | *hard water* |
| roget020.txt | *havea* | *have a* |

| File Name | Original string | Corrected string |
|---|---|---|
| roget040.txt | *havemercy* | *have mercy* |
| roget039.txt | *headfor* | *head for* |
| roget040.txt | *hedgepriest* | *hedge priest* |
| roget013.txt | *icecream* | *ice cream* |
| roget011.txt | *inkdrop* | *ink drop* |
| roget024.txt | *inkslinger* | *ink slinger* |
| roget030.txt | *inlast* | *in last* |
| roget020.txt | *interlocutorydecree* | *interlocutory decree* |
| roget035.txt | *inwishful* | *in wishful* |
| roget029.txt | *inwrestling* | *in wrestling* |
| roget034.txt | *ladykiller* | *lady killer* |
| roget039.txt | *lawcourts* | *law courts* |
| roget036.txt | *lawhusband* | *law husband* |
| roget027.txt | *leadpollution* | *lead pollution* |
| roget013.txt | *leapfrogger* | *leap frogger* |
| roget034.txt | *legpull* | *leg pull* |
| roget024.txt | *lightreading* | *light reading* |
| roget002.txt | *lorryload* | *lorry load* |
| roget002.txt | *lorryload* | *lorry load* |
| roget011.txt | *lorryload* | *lorry load* |
| roget021.txt | *lossofreason* | *loss of reason* |
| roget038.txt | *lovepot* | *love pot* |
| roget020.txt | *makeabsolute* | *make absolute* |
| roget026.txt | *mechanicaladvantage* | *mechanical advantage* |
| roget006.txt | *mellowfruitfulness* | *mellow fruitfulness* |
| roget021.txt | *mentaldeficiency* | *mental deficiency* |
| roget007.txt | *mentalweakness* | *mental weakness* |
| roget036.txt | *mixedmarriage* | *mixed marriage* |
| roget011.txt | *mouseproof* | *mouse proof* |
| roget013.txt | *naturalfunctions* | *natural functions* |
| roget029.txt | *needlegun* | *needle gun* |
| roget037.txt | *ofGod* | *of God* |
| roget037.txt | *ofhonour* | *of honour* |
| roget028.txt | *ofParliament* | *of Parliament* |
| roget001.txt | *ofreference* | *of reference* |
| roget016.txt | *ofsmell* | *of smell* |
| roget021.txt | *ofspeaking* | *of speaking* |
| roget007.txt | *ofstrength* | *of strength* |
| roget026.txt | *ofsubstance* | *of substance* |
| roget016.txt | *oftorture* | *of torture* |
| roget021.txt | *ofunsoundmind* | *of unsound mind* |
| roget040.txt | *oncedelivered* | *once delivered* |
| roget035.txt | *one'sbreath* | *one's breath* |

| File Name | Original string | Corrected string |
|---|---|---|
| roget025.txt | one'slot | one's lot |
| roget033.txt | one'smind | one's mind |
| roget016.txt | one'snose | one's nose |
| roget030.txt | one'spockets | one's pockets |
| roget032.txt | onefor | one for |
| roget024.txt | onehander | one hander |
| roget020.txt | onesidedness | one sidedness |
| roget021.txt | onesyllable | one syllable |
| roget023.txt | onesyllable | one syllable |
| roget037.txt | onlyoneself | only oneself |
| roget016.txt | onthe | on the |
| roget008.txt | outof | out of |
| roget012.txt | paddlewheel | paddle wheel |
| roget016.txt | painfulaftermath | painful aftermath |
| roget038.txt | pamperedappetite | pampered appetite |
| roget016.txt | pipeof | pipe of |
| roget016.txt | pipetobacco | pipe tobacco |
| roget008.txt | pistolshot | pistol shot |
| roget038.txt | profitmaking | profit making |
| roget008.txt | puddingbasin | pudding basin |
| roget021.txt | puddinghead | pudding head |
| roget035.txt | racialprejudice | racial prejudice |
| roget010.txt | rainhat | rain hat |
| roget032.txt | remainderman | remainder man |
| roget009.txt | rubbingshoulders | rubbing shoulders |
| roget024.txt | runthrough | run through |
| roget026.txt | safeconduct | safe conduct |
| roget021.txt | setbefore | set before |
| roget035.txt | setdown | set down |
| roget020.txt | sexprejudice | sex prejudice |
| roget028.txt | shopfloor | shop floor |
| roget013.txt | shortcrust | short crust |
| roget022.txt | situationcomedy | situation comedy |
| roget010.txt | skiboots | ski boots |
| roget036.txt | slangwhang | slang whang |
| roget005.txt | soonafter | soon after |
| roget029.txt | staffwork | staff work |
| roget011.txt | stationwaggon | station wagon |
| roget011.txt | swallowhole | swallow hole |
| roget011.txt | swordpoint | sword point |
| roget029.txt | swordstick | sword stick |
| roget004.txt | systemsanalyst | systems analyst |
| roget026.txt | tablemat | table mat |

| File Name | Original string | Corrected string |
|---|---|---|
| roget036.txt | *takeoffence* | *take offence* |
| roget011.txt | *talkdown* | *talk down* |
| roget024.txt | *talknineteen* | *talk nineteen* |
| roget018.txt | *thatyou* | *that you* |
| roget016.txt | *theagony* | *the agony* |
| roget030.txt | *theascendant* | *the ascendant* |
| roget021.txt | *thebend* | *the bend* |
| roget040.txt | *thechurch* | *the church* |
| roget012.txt | *theclappers* | *the clappers* |
| roget022.txt | *theeducationally* | *the educationally* |
| roget018.txt | *theeyes* | *the eyes* |
| roget015.txt | *thefallen* | *the fallen* |
| roget019.txt | *thehouse* | *the house* |
| roget039.txt | *thelaw* | *the law* |
| roget015.txt | *theLongKnives* | *the Long Knives* |
| roget020.txt | *thematter* | *the matter* |
| roget030.txt | *themoney* | *the money* |
| roget029.txt | *theoffensive* | *the offensive* |
| roget039.txt | *therap* | *the rap* |
| roget030.txt | *therose* | *the rose* |
| roget013.txt | *thescales* | *the scales* |
| roget025.txt | *thescent* | *the scent* |
| roget013.txt | *theshakes* | *the shakes* |
| roget007.txt | *thespout* | *the spout* |
| roget022.txt | *thetrumpets* | *the trumpets* |
| roget038.txt | *theup* | *the up* |
| roget013.txt | *theweight* | *the weight* |
| roget016.txt | *ticklingsensation* | *tickling sensation* |
| roget002.txt | *tieup* | *tie up* |
| roget018.txt | *tiger'seye* | *tiger's eye* |
| roget003.txt | *timeslip* | *time slip* |
| roget005.txt | *timeslip* | *time slip* |
| roget005.txt | *timewarp* | *time warp* |
| roget016.txt | *tobaccochewer* | *tobacco chewer* |
| roget017.txt | *tomtom* | *tom tom* |
| roget007.txt | *topcondition* | *top condition* |
| roget006.txt | *tothe* | *to the* |
| roget007.txt | *toughguy* | *tough guy* |
| roget008.txt | *townsperson* | *towns person* |
| roget013.txt | *trencherwoman* | *trencher woman* |
| roget013.txt | *turnthe* | *turn the* |
| roget034.txt | *twicetold* | *twice told* |
| roget024.txt | *typefoundry* | *type foundry* |

| File Name | Original string | Corrected string |
|---|---|---|
| `roget022.txt` | *tyremark* | *tyre mark* |
| `roget023.txt` | *undDrang* | *und Drang* |
| `roget034.txt` | *underone's* | *under one's* |
| `roget007.txt` | *vicelike* | *vice like* |
| `roget020.txt` | *voxpopuli* | *vox populi* |
| `roget015.txt` | *wastepipe* | *waste pipe* |
| `roget026.txt` | *wastepipe* | *waste pipe* |
| `roget040.txt` | *watchnight* | *watch night* |
| `roget007.txt` | *weakas* | *weak as* |
| `roget029.txt` | *wholehogging* | *whole hogging* |
| `roget031.txt` | *withdrawpermission* | *withdraw permission* |
| `roget009.txt` | *withinside* | *with inside* |
| `roget014.txt` | *withrain* | *with rain* |

**Table F1:** 179 phrases where a space is missing in the Pearson source files

| *File Name* | *Original string* | *Corrected string* |
|---|---|---|
| `roget030.txt` | *decentralizatio n* | *decentralization* |
| `roget007.txt` | *destruct ion* | *destruct ion* |
| `roget024.txt` | *editio n* | *edition* |
| `roget009.txt` | *extraterritoriali ty* | *extraterritoriality* |
| `roget006.txt` | *fatherf.* | *father figure.* |
| `roget013.txt` | *featherwei ght* | *featherweight* |
| `roget036.txt` | *glorificatio n* | *glorification* |
| `roget019.txt` | *impracticabilit y* | *impracticability* |
| `roget027.txt` | *misappropriatio n* | *misappropriation* |
| `roget003.txt` | *overfulfil ment* | *overfulfilment* |
| `roget026.txt` | *overfulfil ment* | *overfulfilment* |
| `roget013.txt` | *geog raphical* | *geographical* |
| `roget040.txt` | *a rchdeacon* | *archdeacon* |
| `roget010.txt` | *di sequilibrium* | *disequilibrium* |
| `roget027.txt` | *ince ssant* | *ince ssant* |
| `roget022.txt` | *suggestio n* | *suggestion* |
| `roget022.txt` | *suggestio n* | *suggestion* |
| `roget007.txt` | *superfecundatio n* | *superfecundation* |
| `roget022.txt` | *suppressio n* | *suppression* |
| `roget022.txt` | *suppressio n* | *suppression* |
| `roget021.txt` | *technica l* | *technical* |
| `roget024.txt` | *televisi on* | *television* |
| `roget021.txt` | *telltal e* | *telltale* |
| `roget038.txt` | *unconscientiousn ess* | *unconscientiousness* |
| `roget019.txt` | *unpredictabilit y* | *unpredictabilit y* |
| `roget029.txt` | *withdr awal* | *withdrawal* |

**Table F2:** 26 words and phrases with a space in the wrong place from the Pearson files

# Appendix G: The 646 American and British Spelling Variations

This appendix shows the 646 American and British spelling variations used by the *ELKB*. It is a union of three publicly available word lists: *The American British – British American Dictionary* (Smith, 2003), *Delphion's American/British Patent Term* (Derwent, 2001) and *XPNDC – American and British Spelling Variations* (XPNDC, 2003).

| American | British |
|----------|---------|
| abridgment | abridgement |
| accouterment | accoutrement |
| acknowledgment | acknowledgement |
| adapter | adaptor |
| advertisement | advertizement |
| advisor | adviser |
| adz | adze |
| aerospace plane | aerospaceplane |
| afterward | afterwards |
| aging | ageing |
| airily | aerify |
| airplane | aeroplane |
| airy | aery |
| alluvium | alluvion |
| alright | allright |
| aluminum | aluminium |
| ameba | amoeba |
| Americanize | Americanise |
| amid | amidst |
| among | amongst |
| amphitheater | amphitheatre |
| analog | analogue |
| analyze | analyse |
| anemia | anaemia |
| anemic | anaemic |
| anesthesia | anaesthesia |
| anesthetic | anaesthetic |
| anesthetist | anaesthetist |
| annex | annexe |
| antiaircraft | anti-aircraft |
| apologize | apologise |
| apothegm | apophthegm |
| appall | appal |
| apprise | apprise |
| arbor | arbour |

| American | British |
|----------|---------|
| archeology | archaeology |
| ardor | ardour |
| armor | armour |
| armorer | armourer |
| armory | armoury |
| artifact | artefact |
| ashtray | ash-tray |
| asphalt | asphalte |
| ass | arse |
| atchoo | atishoo |
| ax | axe |
| B.S. | B.Sc. |
| back scratch | backscratch |
| backward | backwards |
| balk | baulk |
| ball gown | ballgown |
| baloney | boloney |
| baritone | barytone |
| bark | barque |
| barreled | barelled |
| barreled | barrelled |
| barreling | barrelling |
| battle-ax | battleaxe |
| bedeviled | bedevilled |
| behavior | behaviour |
| behavioral | behavioural |
| behoove | behove |
| belabor | belabour |
| bell ringer | bellringer |
| belly flop | bellyflop |
| beside | besides |
| bicolor | bicolour |
| bisulfate | bisulphate |
| bladder wrack | bladderwrack |
| book collection | bookcollection |

| American | British | American | British |
|----------|---------|----------|---------|
| bookkeeper | book-keeper | check | cheque |
| boric | boracic | checker | chequer |
| break dance | breakdance | chili arch | chiliarch |
| brier | briar | chili | chilli |
| buncombe | bunkum | choir stall | choirstall |
| burden | burthen | cigaret | cigarette |
| burglarize | burglarise | citrus | citrous |
| burned | burnt | civilization | civilization |
| by allusion | byallusion | clamor | clamour |
| cachexia | cachexy | clangor | clangour |
| cafe | café | clarinetist | clarinetist |
| caliber | calibre | claw back | clawback |
| caliper | calliper | clearstory | clerestory |
| calipers | callipers | clever stick | cleverstick |
| calisthenics | callisthenics | cloture | closure |
| call girl | callgirl | cogency | coagency |
| canceled | cancelled | colonize | colonize |
| canceling | cancelling | color | colour |
| canceling | cancellling | conjuror | conjurer |
| candor | candour | connection | connexion |
| cantaloupe | cantaloup | cornflower | cornflour |
| capitalize | capitalise | councilor | councilor |
| carburetor | carburettor | counseled | counseled |
| carcass | carcass | counseling | counseling |
| caroler | caroller | counselor | counselor |
| caroling | carolling | cozy | cosy |
| cat slick | catslick | crawfish | crayfish |
| catalog | catalogue | criticize | criticize |
| catalyze | catalyse | curb | kerb |
| categorize | categorize | cutlas | cutlass |
| catsup | ketchup | czar | tsar |
| caviler | caviller | dark fall | darkfall |
| cell phone | cellphone | daydream | day-dream |
| center | centre | deathbed repentance | deathbedrepentance |
| centerboard | centreboard | defense | defence |
| centerfold | centrefold | deflection | deflexion |
| centering | centring | deflexion | deflection |
| centerpiece | centrepiece | demeanor | demeanour |
| centimeter | centimetre | dependent | dependant |
| cesarean | caesarean | deviled | devilled |
| cesarian | caesarian | deviling | devilling |
| cesium | caesium | dialog | dialogue |
| chamomile | camomile | dialyze | cialyse |
| channeled | channelled | diarrhea | ciarrhea |
| characterize | characterise | dieresis | diaeresis |

| American | British |
|----------|---------|
| discolor | discolour |
| disfavor | disfavour |
| disheveled | dishevelled |
| disheveling | dishevelling |
| dishonor | dishonour |
| disk | disc |
| dissention | dissension |
| distill | distil |
| disulfide | disulphide |
| dolor | dolour |
| donut | doughnut |
| doodad | doodah |
| doom watch | doomwatch |
| draft | draught |
| draftsman | draughtsman |
| drafty | draughty |
| dramatize | dramatise |
| dreamed | dreamt |
| driveling | drivelling |
| dryly | drily |
| drypoint | dry-point |
| duelist | duellist |
| duelists | duellists |
| eager | eagre |
| ecology | oecology |
| ecumenical | oecumenical |
| edema | oedema |
| edematous | oedematous |
| elite | élite |
| emphasize | emphasise |
| enameled | enamelled |
| enameling | enamelling |
| enamor | enamour |
| encyclopedia | encyclopaedia |
| encyclopedia | encyclopedia |
| endeavor | endeavour |
| enology | oenology |
| enroll | enrol |
| enrollment | enrolment |
| enthrall | enthral |
| entree | entrée |
| enure | inure |
| envelop | envelope |
| eon | aeon |
| eons | aeons |

| American | British |
|----------|---------|
| epaulet | epaulette |
| epicenter | epicentre |
| epilog | epilogue |
| equaled | equalled |
| equaling | equalling |
| equalize | equalise |
| esophagus | oesophagus |
| esthete | aesthete |
| esthetic | aesthetic |
| estival | aestival |
| estrogen | oestrogen |
| estrus | oestrus |
| ether | aether |
| etiological | aetiological |
| etiology | aetiology |
| eurythmy | eurhythmy |
| fagot | faggot |
| fagoting | faggoting |
| fantasize | fantasise |
| favor | favour |
| favored | favoured |
| favorite | favourite |
| favoritism | favouritism |
| fecal | faecal |
| feces | faeces |
| fervor | fervour |
| fetal | foetal |
| fete | fête |
| fetid | foetid |
| fetor | foetor |
| fetus | foetus |
| fiber | fibre |
| fiberboard | fibreboard |
| fiberglass | fibreglass |
| flakey | flaky |
| flavor | flavour |
| flavored | flavoured |
| floatation | flotation |
| font | fount |
| foregather | forgather |
| forego | forgo |
| form | forme |
| forward | forwards |
| frog march | frogmarch |
| fueled | fuelled |

| American | British |
|----------|---------|
| fueling | fuelling |
| fulfill | fulfil |
| fulfillment | fulfilment |
| furor | furore |
| fuse | fuze |
| galipot | gallipot |
| gallows bird | gallowsbird |
| gantlet | gauntlet |
| garrote | garotte |
| garroted | garotted |
| garroting | garotting |
| gasoline | gasolene |
| gayety | gaiety |
| gel | gell |
| genuflection | genuflexion |
| glamor | glamour |
| glamorize | glamorise |
| goiter | goitre |
| gonorrhea | gonorrhoea |
| good-by | goodbye |
| gram | gramme |
| gray | grey |
| groveled | grovelled |
| groveler | groveller |
| groveling | grovelling |
| grueling | gruelling |
| gynecology | gynaecology |
| gypsy | gipsy |
| hair space | hairspace |
| Halloween | Hallowe'en |
| halyard | halliard |
| harbor | harbour |
| harmonize | harmonise |
| have mercy | havemercy |
| hell hag | hellhag |
| hemoglobin | haemoglobin |
| hemophilia | haemophilia |
| hemorrhage | haemorrhage |
| hemorrhoid | haemorrhoid |
| hold all | holdall |
| homeopath | homoeopath |
| homeostasis | homoeostasis |
| homolog | homologue |
| honor | honour |
| hosteled | hostelled |

| American | British |
|----------|---------|
| hosteler | hosteller |
| hosteling | hostelling |
| hostler | ostler |
| humor | humour |
| ill betide | illbetide |
| immortalize | immortalise |
| impanel | empanel |
| in appetence | inappetence |
| in expectancy | inexpectancy |
| in wrestling | inwrestling |
| incase | encase |
| inclose | enclose |
| indorse | endorse |
| inflection | inflexion |
| inquire | enquire |
| inquiry | enquiry |
| instal | install |
| installment | instalment |
| instill | instil |
| insure | ensure |
| intern | interne |
| jail | gaol |
| jeweler | jeweller |
| jewelry | jewellery |
| jewlry | jewellery |
| jibe | gybe |
| jimmy | jemmy |
| Jr | Jnr. |
| judgment | judgement |
| karat | carat |
| kidnaped | kidnapped |
| kidnaper | kidnapper |
| kidnaping | kidnapping |
| kilometer | kilometre |
| kneeled | knelt |
| knob stick | knobstick |
| know all | knowall |
| labeled | labelled |
| labor | labour |
| lackluster | lacklustre |
| lady killer | ladykiller |
| lave rock | laverock |
| lay stall | laystall |
| lead pollution | leadpollution |
| leaned | leant |

| American | British |
|----------|---------|
| leaped | leapt |
| learned | learnt |
| leg pull | legpull |
| lemongrass | lemon |
| leukemia | leukaemia |
| leveled | levelled |
| leveler | leveller |
| leveler | leveller |
| leveling | levelling |
| libeled | libelled |
| libeling | libelling |
| libelous | libellous |
| license | licence |
| licorice | liquorice |
| light well | lightwell |
| limp back | limpback |
| liter | litre |
| logorrhea | logorrhoea |
| long shore | longshore |
| louver | louvre |
| low fellow | lowfellow |
| Luster | lustre |
| M.S. | M.Sc. |
| malodor | malodour |
| man hour | manhour |
| man oeuvre | manoeuvre |
| maneuver | manoeuvre |
| marshaled | marshalled |
| marveled | marvelled |
| marveling | marvelling |
| marvelous | marvellous |
| marvelously | marvellously |
| matinee | matinée |
| meager | meagre |
| medieval | mediaeval |
| mega there | megathere |
| menorrhea | menorrhoea |
| mental deficiency | mentaldeficiency |
| metaled | metalled |
| metaling | metalling |
| meter | metre |
| mill pool | millpool |
| millimeter | millimetre |
| misbehavior | misbehaviour |
| misdemeanor | misdemeanour |

| American | British |
|----------|---------|
| misjudgment | misjudgement |
| miter | mitre |
| mobilize | mobilise |
| modeled | modelled |
| modeler | modeller |
| modeling | modelling |
| mold | mould |
| molding | moulding |
| mollusk | mollusc |
| molt | moult |
| mom | mum |
| monolog | monologue |
| motorize | motorise |
| mum chance | mumchance |
| mustache | moustache |
| naive | naïve |
| naturalize | naturalise |
| naught | nought |
| neighbor | neighbour |
| neighborhood | neighbourhood |
| neighborly | neighbourly |
| neoclassical | neo-classical |
| net | nett |
| night watch | nightwatch |
| nite | night |
| niter | nitre |
| not respect | notrespect |
| note paper | notepaper |
| ocher | ochre |
| odor | odour |
| offense | offence |
| omelet | omelette |
| organize | organise |
| organized | organised |
| orthopedics | orthopaedics |
| outmaneuver | outmanoeuvre |
| paddy whack | paddywhack |
| pajamas | pyjamas |
| paleobotany | palaeobotany |
| Paleocene | Palaeocene |
| paleoclimatology | palaeoclimatology |
| paleogeography | palaeogeography |
| paleography | palaeography |
| paleolithic | palaeolithic |
| paleomagnetism | palaeomagnetism |

| American | British |
|----------|---------|
| paleontology | palaeontology |
| Paleozoic | Palaeozoic |
| panatela | panatella |
| paneled | panelled |
| paneling | panelling |
| panelist | panellist |
| paralyze | paralyse |
| parameterize | parametrize |
| parlor | parlour |
| pastel list | pastellist |
| pasteurized | pasteurised |
| pavior | paviour |
| pean | paean |
| peas | pease |
| pedagog | pedagogue |
| pedagogy | paedagogy |
| pedaled | pedalled |
| pedaling | pedalling |
| peddler | pedlar |
| pederast | paederast |
| pediatric | paediatric |
| pediatrician | paediatrician |
| pediatrics | paediatrics |
| pedler | pedlar |
| pedophile | paedophile |
| pedophilia | paedophilia |
| penciled | pencilled |
| penciling | pencilling |
| persnickety | pernickety |
| philter | philtre |
| pickaninny | piccaninny |
| picket | piquet |
| pill winks | pilliwinks |
| pillar box | pillarbox |
| pipe tobacco | pipetobacco |
| pjamas | pyjamas |
| plow | plough |
| plowman | ploughman |
| plowshare | ploughshare |
| polyethylene | polythene |
| popularize | popularise |
| port fire | portfire |
| practice | practise |
| pretense | pretence |
| pricey | pricy |

| American | British |
|----------|---------|
| primeval | primaeval |
| program | programme |
| programed | programmed |
| programer | programmer |
| programing | programming |
| prolog | prologue |
| propellent | propellant |
| propellor | propeller |
| pudgy | podgy |
| pull through | pullthrough |
| pummeling | pummelling |
| pupilage | pupillage |
| pygmy | pigmy |
| quarreled | quarrelled |
| quarreler | quarreller |
| quarreling | quarrelling |
| racial prejudice | racialprejudice |
| rancor | rancour |
| raveled | ravelled |
| realize | realise |
| recognizance | recognisance |
| recognize | recognise |
| reconnoiter | reconnoitre |
| remodeling | remodelling |
| retroflection | retroflexion |
| reveled | revelled |
| reveler | reveller |
| reveling | revelling |
| revery | reverie |
| reviviscence | revivescence |
| rigor | rigour |
| rivaling | rivalling |
| role | rôle |
| roll mops | rollmops |
| roller coaster | rollercoaster |
| romanize | romanise |
| ruble | rouble |
| rumor | rumour |
| saber | sabre |
| safe conduct | safeconduct |
| saga more | sagamore |
| sally port | sallyport |
| saltier | saltire |
| saltpeter | saltpetre |
| sanitorium | sanatorium |

| American | British | American | British |
|----------|---------|----------|---------|
| satirize | satirise | spelled | spelt |
| savior | saviour | spilled | spilt |
| savor | savour | spiraling | spiralling |
| savory | savoury | splendor | splendour |
| scalawag | scallywag | spoiled | spoilt |
| scalp lock | scalplock | Sr | Snr |
| scepter | sceptre | stanch | staunch |
| scimitar | scimetar | standardize | standardise |
| septicemia | septicaemia | stenosis | stegnosis |
| sepulcher | sepulchre | story | storey |
| sex prejudice | sexprejudice | stout fellow | stoutfellow |
| sheep track | sheeptrack | succor | succour |
| shooting gallery | shootinggallery | suffix ion | suffixion |
| shoveled | shovelled | sulfate | sulphate |
| show | shew | sulfide | sulphide |
| shrink pack | shrinkpack | sulfur | sulphur |
| shriveled | shrivelled | sulfureted | sulphuretted |
| signaler | signaller | swallow hole | swallowhole |
| signaling | signalling | symbolize | symbolise |
| siphon | syphon | synagog | synagogue |
| siren | syren | syneresis | synaeresis |
| skeptic | sceptic | synesthesia | synaesthesia |
| skeptical | sceptical | syphon | siphon |
| skepticism | scepticism | taffy | toffee |
| skillful | skilful | the fallen | thefallen |
| skillfully | skilfully | theater | theatre |
| skin flick | skinflick | thralldom | thraldom |
| slug | slog | throw stick | throwstick |
| slush | slosh | thru | through |
| smelled | smelt | tidbit | titbit |
| smoke duct | smokeduct | tike | tyke |
| smolder | smoulder | till ant | tillant |
| snail shell | snailshell | tire | tyre |
| snicker | snigger | tiro | tyro |
| sniveled | snivelled | titer | titre |
| sniveler | sniveller | toilet | toilette |
| sniveling | snivelling | tonite | tonight |
| snow pack | snowpack | toward | towards |
| snowplow | snowplough | toweling | towelling |
| soft back | softback | trammeled | trammelled |
| somber | sombre | traveled | travelled |
| soy sauce | soysauce | traveler | traveller |
| specialize | specialise | traveling | travelling |
| specialty | speciality | travelog | travelogue |
| specter | spectre | tricolor | tricolour |

| American | British |
|---|---|
| trisulfate | trisulphate |
| troweled | trowelled |
| troweling | trowelling |
| tumor | tumour |
| tunneling | tunnelling |
| ultrahigh | ultra-high |
| ultramodern | ultra-modern |
| unraveled | unravelled |
| unraveled | untravelled |
| unraveling | unravelling |
| untrammeled | untrammelled |
| valor | valour |
| vapor | vapour |
| vaporize | vaporise |
| vaporware | vapourware |
| veranda | verandah |
| vial | phial |
| video pack | videopack |
| vigor | vigour |
| vise | vice |
| visually challenged | visuallychallenged |
| wagon | waggon |
| watercolor | watercolour |
| weed killer | weedkiller |
| whey face | wheyface |
| while | whilst |
| whiskey | whisky |
| willful | wilful |
| willie | willy |
| woolen | woollen |
| wooly | woolly |
| word stock | wordstock |
| worshiped | worshipped |
| worshiper | worshipper |
| worshiping | worshipping |
| yodeling | yodelling |

**Table G1:** The 646 American and British Spelling Variations

# Appendix H: The 980-element Stop List

This 980-element stop list is a union of five publicly-available lists: *Oracle 8 ConText*, *SMART*, *Hyperwave*, and lists from the *University of Kansas* and *Ohio State University*.

| | | | | | |
|---|---|---|---|---|---|
| 0 | 45 | 81 | along | b | can |
| 1 | 46 | 82 | alpha | back | can't |
| 10 | 47 | 83 | already | backed | cannot |
| 11 | 48 | 84 | also | backing | cant |
| 12 | 49 | 85 | although | backs | caption |
| 13 | 5 | 86 | always | barely | case |
| 14 | 50 | 87 | am | be | cases |
| 15 | 51 | 88 | among | became | cause |
| 16 | 52 | 89 | amongst | because | causes |
| 17 | 53 | 9 | an | become | certain |
| 18 | 54 | 90 | and | becomes | certainly |
| 19 | 55 | 91 | another | becoming | changes |
| 2 | 56 | 92 | any | been | chi |
| 20 | 57 | 93 | anybody | before | circa |
| 21 | 58 | 94 | anyhow | beforehand | clear |
| 22 | 59 | 95 | anyone | began | clearly |
| 23 | 6 | 96 | anything | begin | cm |
| 24 | 60 | 97 | anyway | beginning | co |
| 25 | 61 | 98 | anyways | behind | co. |
| 26 | 62 | 99 | anywhere | being | com |
| 27 | 63 | a | apart | beings | come |
| 28 | 64 | a's | appear | believe | comes |
| 29 | 65 | able | appreciate | below | con |
| 3 | 66 | about | appropriate | beside | concerning |
| 30 | 67 | above | are | besides | consequently |
| 31 | 68 | according | area | best | consider |
| 32 | 69 | accordingly | areas | beta | considering |
| 33 | 7 | across | aren't | better | contain |
| 34 | 70 | actually | around | between | containing |
| 35 | 71 | adj | as | beyond | contains |
| 36 | 72 | after | aside | big | corp |
| 37 | 73 | afterwards | ask | billion | corresponding |
| 38 | 74 | again | asked | both | could |
| 39 | 75 | against | asking | brief | couldn't |
| 4 | 76 | ain't | asks | but | course |
| 40 | 77 | all | associated | by | currently |
| 41 | 78 | allow | at | c | d |
| 42 | 79 | allows | available | c'mon | db |
| 43 | 8 | almost | away | c's | definitely |
| 44 | 80 | alone | awfully | came | delta |
| described | et | further | he'd | inc. | latest |
| despite | eta | furthered | he'll | indeed | latter |
| did | etc | furthering | he's | indicate | latterly |
| didn't | even | furthermore | hello | indicated | lb |
| didst | evenly | furthers | help | indicates | lbs |
| differ | ever | g | hence | inner | least |
| different | every | gamma | henceforth | inside | less |
| differently | everybody | gave | her | insofar | lest |
| do | everyone | general | here | instead | let |
| doer | everything | generally | here's | interest | let's |
| does | everywhere | get | hereafter | interested | lets |
| doesn't | ex | gets | hereby | interesting | like |
| doest | exactly | getting | herein | interests | liked |
| doeth | example | give | hereupon | into | likely |
| doing | except | given | hers | inward | little |
| don't | f | gives | herself | iota | ln |
| done | face | go | hi | is | lo |
| dost | faces | goes | high | isn't | long |
| doth | fact | going | higher | it | longer |
| down | facts | gone | highest | it'd | longest |
| downed | fairly | good | him | it'll | look |
| downing | far | goods | himself | it's | looking |
| downs | felt | got | his | its | looks |
| downwards | few | gotten | hither | itself | ltd |
| during | fewer | great | hopefully | iv | m |
| e | fifteen | greater | how | ix | made |
| each | fifth | greatest | howbeit | j | mainly |
| early | fifty | greetings | however | just | make |
| edu | find | group | hundred | k | makes |
| eg | finds | grouped | hz | kappa | making |
| eight | first | grouping | i | keep | man |
| eighteen | five | groups | i'd | keeps | many |
| eighty | followed | h | i'll | kept | may |
| either | following | had | i'm | kg | maybe |
| eleven | follows | hadn't | i've | km | me |
| else | for | happens | ie | know | mean |
| elsewhere | former | hardly | if | known | meantime |
| end | formerly | has | ignored | knows | meanwhile |
| ended | forth | hasn't | ii | l | member |
| ending | forty | hast | iii | lamda | members |
| ends | found | hath | immediate | large | men |
| enough | four | have | important | largely | merely |
| entirely | fourteen | haven't | in | last | mi |
| epsilon | from | having | inasmuch | lately | might |
| especially | ft | he | inc | later | million |
| mine | noone | ourselves | qv | seventeen | such |
| miss | nor | out | r | seventy | sup |
| ml | normally | outside | rather | several | sure |
| mm | not | over | rd | shall | t |
| more | nothing | overall | re | shalt | t's |
| moreover | novel | own | really | she | take |
| most | now | oz | reasonably | she'd | taken |
| mostly | nowhere | p | recent | she'll | taking |
| mr | nu | part | recently | she's | tau |
| mrs | number | parted | regarding | should | tell |
| ms | numbers | particular | regardless | shouldn't | ten |
| mu | o | particularly | regards | show | tends |
| much | obviously | parting | relatively | showed | th |
| must | of | parts | respectively | showing | than |
| my | off | per | rho | shows | thank |
| myself | often | perhaps | right | sides | thanks |
| mz | oh | phi | room | sigma | thanx |
| n | ok | pi | rooms | simply | that |
| name | okay | place | roughly | since | that'll |
| namely | old | placed | s | six | that's |
| nay | older | places | said | sixteen | that've |
| nd | oldest | please | same | sixty | thats |
| near | omega | plus | saw | small | the |
| nearly | omicron | point | say | smaller | thee |
| necessary | on | pointed | saying | smallest | their |
| need | once | pointing | says | so | theirs |
| needed | one | points | sec | some | them |
| needing | one's | possible | second | somebody | themselves |
| needs | ones | pre | secondly | somehow | then |
| neither | only | present | seconds | someone | thence |
| never | onto | presented | secs | something | there |
| nevertheless | open | presenting | see | sometime | there'd |
| new | opened | presents | seeing | sometimes | there'll |
| newer | opens | presumably | seem | somewhat | there're |
| newest | or | pro | seemed | somewhere | there's |
| next | order | probably | seeming | soon | there've |
| nine | ordered | problem | seems | sorry | thereafter |
| nineteen | ordering | problems | seen | specified | thereby |
| ninety | orders | provides | self | specify | therefore |
| Nm | other | psi | selves | specifying | therein |
| No | others | put | sensible | state | thereof |
| nobody | otherwise | puts | sent | states | thereon |
| non | ought | q | serious | still | theres |
| none | our | que | seriously | stop | thereupon |
| nonetheless | ours | quite | seven | sub | these |

| | | | | |
|---|---|---|---|---|
| theta | turning | we | whole | yea |
| they | turns | we'd | whom | year |
| they'd | twelve | we'll | whomever | years |
| they'll | twenty | we're | whomsoever | yes |
| they're | twice | we've | whose | yet |
| they've | two | welcome | whoso | you |
| thine | u | well | whosoever | you'd |
| thing | un | wells | why | you'll |
| things | under | went | will | you're |
| think | unfortunately | were | willing | you've |
| thinks | unless | weren't | wish | young |
| third | unlike | what | with | younger |
| thirteen | unlikely | what'll | within | youngest |
| thirty | until | what's | without | your |
| this | unto | what've | won't | yours |
| thorough | up | whatever | wonder | yourself |
| thoroughly | upon | whatsoever | work | yourselves |
| those | upsilon | when | worked | z |
| thou | us | whence | working | zero |
| though | use | whenever | works | zeta |
| thought | used | whensoever | would | |
| thoughts | useful | where | wouldn't | |
| thousand | uses | where's | x | |
| three | using | whereafter | xi | |
| through | usually | whereas | xii | |
| throughout | uucp | whereby | xiii | |
| thru | v | wherefore | xiv | |
| thus | value | wherein | xix | |
| thy | various | whereinto | xv | |
| thyself | very | whereof | xvi | |
| to | vi | whereon | xvii | |
| today | via | wheresoever | xviii | |
| together | vii | whereupon | xx | |
| too | viii | wherever | xxi | |
| took | viz | wherewith | xxii | |
| toward | vs | whether | xxiii | |
| towards | w | which | xxiv | |
| tried | want | while | xxix | |
| tries | wanted | whilst | xxv | |
| trillion | wanting | whither | xxvi | |
| truly | wants | who | xxvii | |
| try | was | who'd | xxviii | |
| trying | wasn't | who'll | y | |
| turn | way | who's | yd | |
| turned | ways | whoever | ye | |

# Appendix I: The Rubenstein and Goodenough 65 Noun Pairs

This appendix contains the Rubenstein and Goodenough (1965) 65 noun pairs and the semantic similarity scores for the *ELKB* as well as the *WordNet*-based measures. They are correlated to Rubenstein and Goodenough's results.

| Noun Pair | Rubenstein Goodenough | ELKB | WordNet Edges | Hirst St.Onge | Jiang Conrath | Leacock Chodorow | Lin | Resnik |
|---|---|---|---|---|---|---|---|---|
| gem – jewel | 3.940 | 16.000 | 30.000 | 16.000 | 1.000 | 3.466 | 1.000 | 12.886 |
| midday – noon | 3.940 | 16.000 | 30.000 | 16.000 | 1.000 | 3.466 | 1.000 | 10.584 |
| automobile – car | 3.920 | 16.000 | 30.000 | 16.000 | 1.000 | 3.466 | 1.000 | 6.340 |
| cemetery – graveyard | 3.880 | 16.000 | 30.000 | 16.000 | 1.000 | 3.466 | 1.000 | 10.689 |
| cushion – pillow | 3.840 | 16.000 | 29.000 | 4.000 | 0.662 | 2.773 | 0.975 | 9.891 |
| boy – lad | 3.820 | 16.000 | 29.000 | 5.000 | 0.231 | 2.773 | 0.824 | 7.769 |
| cock – rooster | 3.680 | 16.000 | 30.000 | 16.000 | 1.000 | 3.466 | 1.000 | 11.277 |
| implement – tool | 3.660 | 16.000 | 29.000 | 4.000 | 0.546 | 2.773 | 0.935 | 5.998 |
| forest – woodland | 3.650 | 14.000 | 30.000 | 16.000 | 1.000 | 3.466 | 1.000 | 10.114 |
| coast – shore | 3.600 | 16.000 | 29.000 | 4.000 | 0.647 | 2.773 | 0.971 | 8.974 |
| autograph – signature | 3.590 | 16.000 | 29.000 | 4.000 | 0.325 | 2.773 | 0.912 | 10.807 |
| journey – voyage | 3.580 | 16.000 | 29.000 | 4.000 | 0.169 | 2.773 | 0.699 | 6.057 |
| serf – slave | 3.460 | 16.000 | 27.000 | 5.000 | 0.261 | 2.079 | 0.869 | 9.360 |
| grin – smile | 3.460 | 16.000 | 30.000 | 16.000 | 1.000 | 3.466 | 1.000 | 9.198 |
| glass – tumbler | 3.450 | 16.000 | 29.000 | 6.000 | 0.267 | 2.773 | 0.873 | 9.453 |
| cord – string | 3.410 | 16.000 | 29.000 | 6.000 | 0.297 | 2.773 | 0.874 | 8.214 |
| hill – mound | 3.290 | 12.000 | 30.000 | 16.000 | 1.000 | 3.466 | 1.000 | 11.095 |
| magician – wizard | 3.210 | 14.000 | 30.000 | 16.000 | 1.000 | 3.466 | 1.000 | 9.708 |
| furnace – stove | 3.110 | 14.000 | 23.000 | 5.000 | 0.060 | 1.386 | 0.238 | 2.426 |
| asylum – madhouse | 3.040 | 16.000 | 29.000 | 4.000 | 0.662 | 2.773 | 0.978 | 11.277 |
| brother – monk | 2.740 | 14.000 | 29.000 | 4.000 | 0.294 | 2.773 | 0.897 | 10.489 |
| food – fruit | 2.690 | 12.000 | 23.000 | 0.000 | 0.088 | 1.386 | 0.119 | 0.699 |
| bird – cock | 2.630 | 12.000 | 29.000 | 6.000 | 0.159 | 2.773 | 0.693 | 5.980 |
| bird – crane | 2.630 | 14.000 | 27.000 | 5.000 | 0.139 | 2.079 | 0.658 | 5.980 |
| oracle – sage | 2.610 | 16.000 | 23.000 | 0.000 | 0.057 | 1.386 | 0.226 | 2.455 |
| sage – wizard | 2.460 | 14.000 | 25.000 | 2.000 | 0.060 | 1.674 | 0.236 | 2.455 |
| brother – lad | 2.410 | 14.000 | 26.000 | 3.000 | 0.071 | 1.856 | 0.273 | 2.455 |
| crane – implement | 2.370 | 0.000 | 26.000 | 3.000 | 0.086 | 1.856 | 0.394 | 3.443 |
| magician – oracle | 1.820 | 6.000 | 28.000 | 6.000 | 0.533 | 2.367 | 0.957 | 9.708 |
| glass – jewel | 1.780 | 12.000 | 24.000 | 2.000 | 0.064 | 1.520 | 0.249 | 2.426 |
| cemetery – mound | 1.690 | 0.000 | 20.000 | 0.000 | 0.055 | 1.068 | 0.076 | 0.699 |
| car – journey | 1.550 | 12.000 | 17.000 | 0.000 | 0.075 | 0.827 | 0.000 | 0.000 |
| hill – woodland | 1.480 | 0.000 | 25.000 | 2.000 | 0.060 | 1.674 | 0.132 | 1.183 |
| crane – rooster | 1.410 | 12.000 | 23.000 | 0.000 | 0.080 | 1.386 | 0.510 | 5.980 |
| furnace – implement | 1.370 | 6.000 | 25.000 | 2.000 | 0.081 | 1.674 | 0.299 | 2.426 |

| Noun Pair | Rubenstein Goodenough | ELKB | WordNet Edges | Hirst St.Onge | Jiang Conrath | Leacock Chodorow | Lin | Resnik |
|---|---|---|---|---|---|---|---|---|
| *coast – hill* | 1.260 | 4.000 | 26.000 | 2.000 | 0.148 | 1.856 | 0.689 | 6.378 |
| *bird – woodland* | 1.240 | 8.000 | 22.000 | 0.000 | 0.068 | 1.269 | 0.147 | 1.183 |
| *shore – voyage* | 1.220 | 2.000 | 18.000 | 0.000 | 0.049 | 0.901 | 0.000 | 0.000 |
| *cemetery – woodland* | 1.180 | 6.000 | 21.000 | 0.000 | 0.049 | 1.163 | 0.067 | 0.699 |
| *food – rooster* | 1.090 | 6.000 | 17.000 | 0.000 | 0.063 | 0.827 | 0.086 | 0.699 |
| *forest – graveyard* | 1.000 | 6.000 | 21.000 | 0.000 | 0.050 | 1.163 | 0.067 | 0.699 |
| *lad – wizard* | 0.990 | 4.000 | 26.000 | 3.000 | 0.068 | 1.856 | 0.265 | 2.455 |
| *mound – shore* | 0.970 | 6.000 | 26.000 | 3.000 | 0.126 | 1.856 | 0.649 | 6.378 |
| *automobile – cushion* | 0.970 | 4.000 | 23.000 | 3.000 | 0.084 | 1.386 | 0.386 | 3.443 |
| *boy – sage* | 0.960 | 8.000 | 25.000 | 2.000 | 0.067 | 1.674 | 0.260 | 2.455 |
| *monk – oracle* | 0.910 | 12.000 | 23.000 | 0.000 | 0.058 | 1.386 | 0.233 | 2.455 |
| *shore – woodland* | 0.900 | 2.000 | 25.000 | 2.000 | 0.056 | 1.674 | 0.124 | 1.183 |
| *grin – lad* | 0.880 | 6.000 | 17.000 | 0.000 | 0.053 | 0.827 | 0.000 | 0.000 |
| *coast – forest* | 0.850 | 6.000 | 24.000 | 0.000 | 0.055 | 1.520 | 0.121 | 1.183 |
| *asylum – cemetery* | 0.790 | 0.000 | 19.000 | 0.000 | 0.046 | 0.981 | 0.064 | 0.699 |
| *monk – slave* | 0.570 | 6.000 | 26.000 | 3.000 | 0.063 | 1.856 | 0.247 | 2.455 |
| *cushion – jewel* | 0.450 | 6.000 | 24.000 | 0.000 | 0.062 | 1.520 | 0.243 | 2.426 |
| *boy – rooster* | 0.440 | 12.000 | 19.000 | 0.000 | 0.064 | 0.981 | 0.228 | 2.171 |
| *glass – magician* | 0.440 | 2.000 | 23.000 | 0.000 | 0.056 | 1.386 | 0.123 | 1.183 |
| *graveyard – madhouse* | 0.420 | 4.000 | 16.000 | 0.000 | 0.045 | 0.758 | 0.062 | 0.699 |
| *asylum – monk* | 0.390 | 0.000 | 20.000 | 0.000 | 0.049 | 1.068 | 0.109 | 1.183 |
| *asylum – fruit* | 0.190 | 6.000 | 24.000 | 0.000 | 0.060 | 1.520 | 0.215 | 2.426 |
| *grin – implement* | 0.180 | 0.000 | 17.000 | 0.000 | 0.062 | 0.827 | 0.000 | 0.000 |
| *mound – stove* | 0.140 | 6.000 | 24.000 | 2.000 | 0.071 | 1.520 | 0.296 | 3.443 |
| *automobile – wizard* | 0.110 | 0.000 | 19.000 | 0.000 | 0.068 | 0.981 | 0.147 | 1.183 |
| *autograph – shore* | 0.060 | 0.000 | 18.000 | 0.000 | 0.047 | 0.901 | 0.000 | 0.000 |
| *fruit – furnace* | 0.050 | 12.000 | 24.000 | 0.000 | 0.064 | 1.520 | 0.225 | 2.426 |
| *noon – string* | 0.040 | 6.000 | 19.000 | 0.000 | 0.052 | 0.981 | 0.000 | 0.000 |
| *rooster – voyage* | 0.040 | 2.000 | 11.000 | 0.000 | 0.044 | 0.470 | 0.000 | 0.000 |
| *cord – smile* | 0.020 | 0.000 | 18.000 | 0.000 | 0.054 | 0.901 | 0.165 | 1.821 |
| *Correlation* | 1.000 | 0.818 | 0.787 | 0.732 | 0.731 | 0.852 | 0.834 | 0.800 |

**Table I1:** Comparison of semantic similarity measures using the Rubenstein and Goodenough data

# Appendix J: The WordSimilarity-353 Test Collection

This appendix presents The WordSimilarity-353 Test Collection (Finkelstein *et al.*, 2002; Gabrilovich 2002) and the semantic similarity scores for the *ELKB* as well as the *WordNet*-based measures. They are correlated to Finkelstein *et al.*'s results.

| *Word Pair* | *Gabr.* | *ELKB* | *WN Edges* | *Hirst St.O.* | *Jiang Con.* | *Lea. Chod.* | *Lin* | *Res.* |
|---|---|---|---|---|---|---|---|---|
| *tiger – tiger* | 10.00 | 16.00 | 30.00 | 24.00 | 1.00 | 3.47 | 1.00 | 12.18 |
| *fuck – sex* | 9.44 | | 28.00 | 3.00 | 0.18 | 2.37 | 0.78 | 8.27 |
| *journey – voyage* | 9.29 | 16.00 | 29.00 | 4.00 | 0.17 | 2.77 | 0.70 | 6.05 |
| *midday – noon* | 9.29 | 16.00 | 30.00 | 16.00 | 1.00 | 3.47 | 1.00 | 10.57 |
| *dollar – buck* | 9.22 | 16.00 | 30.00 | 16.00 | 1.00 | 3.47 | 1.00 | 10.31 |
| *money – cash* | 9.15 | 16.00 | 28.00 | 5.00 | 0.19 | 2.37 | 0.74 | 7.14 |
| *coast – shore* | 9.10 | 16.00 | 29.00 | 4.00 | 0.65 | 2.77 | 0.97 | 8.96 |
| *money – cash* | 9.08 | 16.00 | 28.00 | 5.00 | 0.19 | 2.37 | 0.74 | 7.14 |
| *money – currency* | 9.04 | 16.00 | 29.00 | 5.00 | 0.41 | 2.77 | 0.90 | 7.14 |
| *football – soccer* | 9.03 | 16.00 | 29.00 | 4.00 | 0.27 | 2.77 | 0.88 | 10.17 |
| *magician – wizard* | 9.02 | 14.00 | 30.00 | 16.00 | 1.00 | 3.47 | 1.00 | 9.70 |
| *type – kind* | 8.97 | 16.00 | 29.00 | 4.00 | 0.62 | 2.77 | 0.95 | 5.60 |
| *gem – jewel* | 8.96 | 16.00 | 30.00 | 16.00 | 1.00 | 3.47 | 1.00 | 12.87 |
| *car – automobile* | 8.94 | 16.00 | 30.00 | 16.00 | 1.00 | 3.47 | 1.00 | 6.33 |
| *street – avenue* | 8.88 | 16.00 | 29.00 | 4.00 | 0.21 | 2.77 | 0.81 | 8.09 |
| *asylum – madhouse* | 8.87 | 14.00 | 29.00 | 4.00 | 0.66 | 2.77 | 0.98 | 11.26 |
| *boy – lad* | 8.83 | 16.00 | 29.00 | 5.00 | 0.23 | 2.77 | 0.82 | 7.76 |
| *environment – ecology* | 8.81 | 14.00 | 29.00 | 4.00 | 0.17 | 2.77 | 0.74 | 7.14 |
| *furnace – stove* | 8.79 | 14.00 | 23.00 | 5.00 | 0.06 | 1.39 | 0.24 | 2.45 |
| *seafood – lobster* | 8.70 | 16.00 | 28.00 | 5.00 | 0.24 | 2.37 | 0.84 | 8.08 |
| *mile – kilometer* | 8.66 | 14.00 | 27.00 | 4.00 | 0.10 | 2.08 | 0.55 | 5.34 |
| *Maradona – football* | 8.62 | | | | | | | |
| *OPEC – oil* | 8.59 | 4.00 | 17.00 | 0.00 | 0.05 | 0.83 | 0.00 | 0.00 |
| *king – queen* | 8.58 | 16.00 | 28.00 | 5.00 | 0.27 | 2.37 | 0.89 | 11.49 |
| *murder – manslaughter* | 8.53 | 14.00 | 28.00 | 5.00 | 0.17 | 2.37 | 0.76 | 7.84 |
| *money – bank* | 8.50 | 16.00 | 24.00 | 0.00 | 0.10 | 1.52 | 0.47 | 4.11 |
| *computer – software* | 8.50 | 14.00 | 16.00 | 0.00 | 0.06 | 0.76 | 0.00 | 0.00 |
| *Jerusalem – Israel* | 8.46 | | 20.00 | 4.00 | 0.06 | 1.07 | 0.31 | 3.71 |
| *vodka – gin* | 8.46 | 14.00 | 28.00 | 5.00 | 0.12 | 2.37 | 0.70 | 8.43 |
| *planet – star* | 8.45 | 14.00 | 28.00 | 5.00 | 0.35 | 2.37 | 0.88 | 6.84 |
| *calculation – computation* | 8.44 | 16.00 | 30.00 | 16.00 | 1.00 | 3.47 | 1.00 | 8.88 |
| *money – dollar* | 8.42 | 16.00 | 26.00 | 3.00 | 0.18 | 1.86 | 0.73 | 7.14 |
| *law – lawyer* | 8.38 | 12.00 | 21.00 | 0.00 | 0.06 | 1.16 | 0.00 | 0.00 |
| *championship – tournament* | 8.36 | 6.00 | 22.00 | 0.00 | 0.04 | 1.27 | 0.00 | 0.00 |
| *seafood – food* | 8.34 | 14.00 | 29.00 | 16.00 | 0.29 | 2.77 | 0.83 | 5.69 |
| *weather – forecast* | 8.34 | 14.00 | 17.00 | 0.00 | 0.05 | 0.83 | 0.00 | 0.00 |
| *FBI – investigation* | 8.31 | 14.00 | 19.00 | 0.00 | 0.05 | 0.98 | 0.00 | 0.00 |
| *network – hardware* | 8.31 | 6.00 | 27.00 | 4.00 | 0.06 | 2.08 | 0.32 | 3.44 |
| *nature – environment* | 8.31 | 4.00 | 24.00 | 0.00 | 0.06 | 1.52 | 0.07 | 0.71 |

| Word Pair | Gabr. | ELKB | WN Edges | Hirst St.O. | Jiang Con. | Lea. Chod. | Lin | Res. |
|---|---|---|---|---|---|---|---|---|
| *man – woman* | 8.30 | 16.00 | 27.00 | 4.00 | 0.13 | 2.08 | 0.59 | 4.81 |
| *money – wealth* | 8.27 | 16.00 | 29.00 | 4.00 | 0.96 | 2.77 | 1.00 | 8.87 |
| *psychology – Freud* | 8.21 | | 12.00 | 0.00 | 0.04 | 0.52 | 0.00 | 0.00 |
| *news – report* | 8.16 | 16.00 | 29.00 | 5.00 | 0.83 | 2.77 | 0.99 | 6.99 |
| *vodka – brandy* | 8.13 | 14.00 | 28.00 | 5.00 | 0.14 | 2.37 | 0.73 | 8.43 |
| *war – troops* | 8.13 | 12.00 | 22.00 | 0.00 | 0.06 | 1.27 | 0.00 | 0.00 |
| *Harvard – Yale* | 8.13 | | 28.00 | 5.00 | 0.17 | 2.37 | 0.79 | 10.17 |
| *bank – money* | 8.12 | 16.00 | 24.00 | 0.00 | 0.10 | 1.52 | 0.47 | 4.11 |
| *physics – proton* | 8.12 | 12.00 | 13.00 | 0.00 | 0.05 | 0.58 | 0.00 | 0.00 |
| *planet – galaxy* | 8.11 | 12.00 | 23.00 | 4.00 | 0.05 | 1.39 | 0.17 | 2.17 |
| *stock – market* | 8.08 | 16.00 | 24.00 | 4.00 | 0.10 | 1.52 | 0.40 | 3.04 |
| *psychology – psychiatry* | 8.08 | 16.00 | 24.00 | 2.00 | 0.11 | 1.52 | 0.62 | 6.51 |
| *planet – moon* | 8.08 | 16.00 | 27.00 | 4.00 | 0.25 | 2.08 | 0.82 | 6.84 |
| *planet – constellation* | 8.06 | 12.00 | 27.00 | 4.00 | 0.13 | 2.08 | 0.57 | 4.50 |
| *credit – card* | 8.06 | 16.00 | 25.00 | 2.00 | 0.07 | 1.67 | 0.37 | 4.41 |
| *hotel – reservation* | 8.03 | 6.00 | 20.00 | 0.00 | 0.05 | 1.07 | 0.07 | 0.71 |
| *planet – sun* | 8.02 | 12.00 | 27.00 | 4.00 | 0.28 | 2.08 | 0.84 | 6.84 |
| *tiger – jaguar* | 8.00 | 16.00 | 28.00 | 5.00 | 0.21 | 2.37 | 0.84 | 9.74 |
| *tiger – feline* | 8.00 | 14.00 | 28.00 | 6.00 | 0.25 | 2.37 | 0.85 | 8.41 |
| *closet – clothes* | 8.00 | 12.00 | 24.00 | 0.00 | 0.08 | 1.52 | 0.31 | 2.45 |
| *planet – astronomer* | 7.94 | 12.00 | 24.00 | 0.00 | 0.06 | 1.52 | 0.23 | 2.45 |
| *soap – opera* | 7.94 | 16.00 | 20.00 | 0.00 | 0.06 | 1.07 | 0.20 | 1.96 |
| *movie – theater* | 7.92 | 16.00 | 23.00 | 0.00 | 0.06 | 1.39 | 0.00 | 0.00 |
| *planet – space* | 7.92 | 12.00 | 23.00 | 3.00 | 0.07 | 1.39 | 0.19 | 1.96 |
| *treatment – recovery* | 7.91 | 6.00 | 24.00 | 0.00 | 0.08 | 1.52 | 0.28 | 2.25 |
| *liquid – water* | 7.89 | 16.00 | 29.00 | 6.00 | 0.99 | 2.77 | 1.00 | 6.19 |
| *life – death* | 7.88 | 16.00 | 28.00 | 5.00 | 0.23 | 2.37 | 0.81 | 6.95 |
| *baby – mother* | 7.85 | 14.00 | 26.00 | 3.00 | 0.22 | 1.86 | 0.76 | 6.01 |
| *aluminum – metal* | 7.83 | | 29.00 | 4.00 | 0.21 | 2.77 | 0.79 | 7.09 |
| *cell – phone* | 7.81 | 6.00 | 26.00 | 3.00 | 0.13 | 1.86 | 0.68 | 7.21 |
| *lobster – food* | 7.81 | 14.00 | 27.00 | 3.00 | 0.15 | 2.08 | 0.67 | 5.69 |
| *dollar – yen* | 7.78 | 14.00 | 27.00 | 3.00 | 0.11 | 2.08 | 0.63 | 6.85 |
| *wood – forest* | 7.73 | 14.00 | 30.00 | 16.00 | 1.00 | 3.47 | 1.00 | 8.28 |
| *money – deposit* | 7.73 | 16.00 | 28.00 | 6.00 | 0.16 | 2.37 | 0.72 | 6.82 |
| *television – film* | 7.72 | 16.00 | 26.00 | 3.00 | 0.22 | 1.86 | 0.80 | 7.23 |
| *psychology – mind* | 7.69 | 16.00 | 24.00 | 0.00 | 0.09 | 1.52 | 0.41 | 3.39 |
| *game – team* | 7.69 | 12.00 | 23.00 | 0.00 | 0.07 | 1.39 | 0.00 | 0.00 |
| *admission – ticket* | 7.69 | 16.00 | 22.00 | 0.00 | 0.06 | 1.27 | 0.27 | 2.88 |
| *Jerusalem – Palestinian* | 7.65 | | 16.00 | 0.00 | 0.04 | 0.76 | 0.06 | 0.71 |
| *Arafat – terror* | 7.65 | | | | | | | |
| *dividend – payment* | 7.63 | 14.00 | 28.00 | 6.00 | 0.15 | 2.37 | 0.71 | 7.09 |
| *profit – loss* | 7.63 | 14.00 | 25.00 | 3.00 | 0.33 | 1.67 | 0.86 | 6.54 |
| *computer – keyboard* | 7.62 | 14.00 | 27.00 | 2.00 | 0.08 | 2.08 | 0.43 | 4.29 |
| *boxing – round* | 7.61 | 14.00 | 24.00 | 0.00 | 0.13 | 1.52 | 0.67 | 6.85 |
| *century – year* | 7.59 | 14.00 | 28.00 | 4.00 | 0.13 | 2.37 | 0.52 | 3.72 |
| *rock – jazz* | 7.59 | 16.00 | 28.00 | 5.00 | 0.17 | 2.37 | 0.78 | 8.67 |
| *computer – internet* | 7.58 | | 23.00 | 5.00 | 0.06 | 1.39 | 0.31 | 3.44 |

| Word Pair | Gabr. | ELKB | WN Edges | Hirst St.O. | Jiang Con. | Lea. Chod. | Lin | Res. |
|---|---|---|---|---|---|---|---|---|
| *money – property* | 7.57 | 14.00 | 28.00 | 6.00 | 0.34 | 2.37 | 0.88 | 6.96 |
| *tennis – racket* | 7.56 | 6.00 | 22.00 | 0.00 | 0.06 | 1.27 | 0.27 | 3.12 |
| *announcement – news* | 7.56 | 14.00 | 26.00 | 3.00 | 0.10 | 1.86 | 0.47 | 3.84 |
| *canyon – landscape* | 7.53 | 6.00 | 19.00 | 0.00 | 0.05 | 0.98 | 0.17 | 1.96 |
| *day – dawn* | 7.53 | 14.00 | 28.00 | 5.00 | 0.08 | 2.37 | 0.38 | 3.72 |
| *food – fruit* | 7.52 | 12.00 | 24.00 | 0.00 | 0.11 | 1.52 | 0.33 | 1.96 |
| *telephone – communication* | 7.50 | 14.00 | 21.00 | 0.00 | 0.08 | 1.16 | 0.00 | 0.00 |
| *currency – market* | 7.50 | 14.00 | 23.00 | 0.00 | 0.06 | 1.39 | 0.00 | 0.00 |
| *psychology – cognition* | 7.48 | 12.00 | 25.00 | 3.00 | 0.13 | 1.67 | 0.46 | 2.89 |
| *Marathon – sprint* | 7.47 | 14.00 | 20.00 | 0.00 | 0.05 | 1.07 | 0.19 | 2.25 |
| *seafood – sea* | 7.47 | 6.00 | 23.00 | 0.00 | 0.06 | 1.39 | 0.09 | 0.71 |
| *book – paper* | 7.46 | 16.00 | 28.00 | 5.00 | 0.12 | 2.37 | 0.58 | 5.18 |
| *book – library* | 7.46 | 16.00 | 24.00 | 2.00 | 0.07 | 1.52 | 0.28 | 2.45 |
| *Mexico – Brazil* | 7.44 | | 26.00 | 3.00 | 0.08 | 1.86 | 0.53 | 6.05 |
| *media – radio* | 7.42 | 16.00 | 19.00 | 0.00 | 0.04 | 0.98 | 0.00 | 0.00 |
| *psychology – depression* | 7.42 | 12.00 | 21.00 | 0.00 | 0.06 | 1.16 | 0.28 | 2.80 |
| *jaguar – cat* | 7.42 | 14.00 | 29.00 | 4.00 | 0.33 | 2.77 | 0.91 | 9.74 |
| *fighting – defeating* | 7.41 | 4.00 | | 0.00 | | | | |
| *movie – star* | 7.38 | 12.00 | 27.00 | 0.00 | 0.07 | 1.39 | 0.31 | 2.88 |
| *bird – crane* | 7.38 | 14.00 | 18.00 | 5.00 | 0.14 | 2.08 | 0.66 | 5.97 |
| *hundred – percent* | 7.38 | | 20.00 | 0.00 | 0.07 | 0.90 | 0.21 | 1.81 |
| *dollar – profit* | 7.38 | 16.00 | 29.00 | 0.00 | 0.06 | 1.07 | 0.18 | 1.81 |
| *tiger – cat* | 7.35 | 14.00 | 28.00 | 4.00 | 0.36 | 2.77 | 0.92 | 9.74 |
| *physics – chemistry* | 7.35 | 14.00 | 23.00 | 2.00 | 0.23 | 2.37 | 0.81 | 7.04 |
| *country – citizen* | 7.31 | 12.00 | 27.00 | 5.00 | 0.07 | 1.39 | 0.10 | 0.71 |
| *money – possession* | 7.29 | 12.00 | 14.00 | 5.00 | 0.17 | 2.08 | 0.63 | 4.11 |
| *jaguar – car* | 7.27 | 6.00 | 30.00 | 0.00 | 0.06 | 0.63 | 0.08 | 0.71 |
| *cup – drink* | 7.25 | 14.00 | 27.00 | 5.00 | 0.19 | 2.08 | 0.77 | 7.08 |
| *psychology – health* | 7.23 | 12.00 | 20.00 | 0.00 | 0.05 | 1.07 | 0.00 | 0.00 |
| *museum – theater* | 7.19 | 6.00 | 24.00 | 2.00 | 0.06 | 1.52 | 0.24 | 2.45 |
| *summer – drought* | 7.16 | 6.00 | 27.00 | 4.00 | 0.07 | 2.08 | 0.36 | 3.72 |
| *phone – equipment* | 7.13 | 6.00 | 28.00 | 2.00 | 0.24 | 2.37 | 0.80 | 6.04 |
| *investor – earning* | 7.13 | 4.00 | | 0.00 | | | | |
| *bird – cock* | 7.10 | 12.00 | 29.00 | 6.00 | 0.16 | 2.77 | 0.69 | 5.97 |
| *company – stock* | 7.08 | 14.00 | 25.00 | 2.00 | 0.12 | 1.67 | 0.46 | 3.35 |
| *tiger – carnivore* | 7.08 | 14.00 | 27.00 | 5.00 | 0.18 | 2.08 | 0.74 | 6.78 |
| *stroke – hospital* | 7.03 | 12.00 | 20.00 | 0.00 | 0.06 | 1.07 | 0.06 | 0.71 |
| *liability – insurance* | 7.03 | 8.00 | 26.00 | 3.00 | 0.19 | 1.86 | 0.79 | 8.28 |
| *game – victory* | 7.03 | 14.00 | 24.00 | 0.00 | 0.12 | 1.52 | 0.50 | 3.78 |
| *doctor – nurse* | 7.00 | 12.00 | 27.00 | 5.00 | 0.25 | 2.08 | 0.83 | 7.28 |
| *tiger – animal* | 7.00 | 14.00 | 27.00 | 2.00 | 0.12 | 2.08 | 0.55 | 4.32 |
| *psychology – anxiety* | 7.00 | 12.00 | 21.00 | 0.00 | 0.08 | 1.16 | 0.33 | 2.80 |
| *game – defeat* | 6.97 | 14.00 | 24.00 | 0.00 | 0.10 | 1.52 | 0.46 | 3.78 |
| *FBI – fingerprint* | 6.94 | 4.00 | 16.00 | 0.00 | 0.04 | 0.76 | 0.00 | 0.00 |
| *money – withdrawal* | 6.88 | 6.00 | 21.00 | 0.00 | 0.06 | 1.16 | 0.00 | 0.00 |
| *street – block* | 6.88 | 14.00 | 25.00 | 2.00 | 0.08 | 1.67 | 0.30 | 2.46 |
| *opera – performance* | 6.88 | 12.00 | 24.00 | 2.00 | 0.08 | 1.52 | 0.34 | 2.88 |

| Word Pair | Gabr. | ELKB | WN Edges | Hirst St.O. | Jiang Con. | Lea. Chod. | Lin | Res. |
|---|---|---|---|---|---|---|---|---|
| *drink – eat* | 6.87 | | | 5.00 | | | | |
| *drug – abuse* | 6.85 | 14.00 | 22.00 | 0.00 | 0.06 | 1.27 | 0.00 | 0.00 |
| *tiger – mammal* | 6.85 | 14.00 | 25.00 | 3.00 | 0.14 | 1.67 | 0.64 | 5.43 |
| *psychology – fear* | 6.85 | 6.00 | 21.00 | 0.00 | 0.08 | 1.16 | 0.33 | 2.80 |
| *cup – tableware* | 6.85 | | 28.00 | 6.00 | 0.58 | 2.37 | 0.95 | 7.60 |
| *student – professor* | 6.81 | 14.00 | 23.00 | 0.00 | 0.07 | 1.39 | 0.28 | 2.45 |
| *football – basketball* | 6.81 | 14.00 | 28.00 | 5.00 | 0.16 | 2.37 | 0.76 | 8.45 |
| *concert – virtuoso* | 6.81 | 14.00 | 20.00 | 0.00 | 0.05 | 1.07 | 0.00 | 0.00 |
| *computer – laboratory* | 6.78 | 6.00 | 21.00 | 0.00 | 0.05 | 1.16 | 0.08 | 0.71 |
| *love – sex* | 6.77 | 12.00 | 29.00 | 4.00 | 0.18 | 2.77 | 0.78 | 8.27 |
| *television – radio* | 6.77 | 16.00 | 28.00 | 5.00 | 0.33 | 2.37 | 0.90 | 9.51 |
| *Problem – challenge* | 6.75 | 14.00 | 25.00 | 2.00 | 0.08 | 1.67 | 0.37 | 3.84 |
| *Arafat – Peace* | 6.73 | | | | | | | |
| *movie – critic* | 6.73 | 12.00 | 20.00 | 0.00 | 0.06 | 1.07 | 0.00 | 0.00 |
| *bed – closet* | 6.72 | 6.00 | 28.00 | 4.00 | 0.15 | 2.37 | 0.70 | 6.74 |
| *psychology – science* | 6.71 | 14.00 | 29.00 | 4.00 | 0.24 | 2.77 | 0.81 | 6.51 |
| *fertility – egg* | 6.69 | 12.00 | 20.00 | 0.00 | 0.04 | 1.07 | 0.00 | 0.00 |
| *bishop – rabbi* | 6.69 | 12.00 | 26.00 | 3.00 | 0.16 | 1.86 | 0.74 | 7.52 |
| *lawyer – evidence* | 6.69 | 6.00 | 20.00 | 0.00 | 0.06 | 1.07 | 0.00 | 0.00 |
| *precedent – law* | 6.65 | 16.00 | 28.00 | 6.00 | 0.17 | 2.37 | 0.74 | 6.94 |
| *football – tennis* | 6.63 | 16.00 | 25.00 | 2.00 | 0.17 | 1.67 | 0.76 | 8.09 |
| *minister – party* | 6.63 | 16.00 | 25.00 | 2.00 | 0.07 | 1.67 | 0.26 | 2.45 |
| *professor – doctor* | 6.62 | 14.00 | 24.00 | 0.00 | 0.20 | 1.52 | 0.77 | 6.55 |
| *psychology – clinic* | 6.58 | 12.00 | 17.00 | 0.00 | 0.05 | 0.83 | 0.00 | 0.00 |
| *cup – coffee* | 6.58 | 14.00 | 25.00 | 2.00 | 0.11 | 1.67 | 0.61 | 6.12 |
| *water – seepage* | 6.56 | 6.00 | 22.00 | 0.00 | 0.05 | 1.27 | 0.00 | 0.00 |
| *government – crisis* | 6.56 | 6.00 | 23.00 | 0.00 | 0.06 | 1.39 | 0.00 | 0.00 |
| *space – world* | 6.53 | 16.00 | 26.00 | 4.00 | 0.07 | 1.86 | 0.19 | 1.96 |
| *Japanese – American* | 6.50 | | 25.00 | 2.00 | 0.09 | 1.67 | 0.54 | 6.27 |
| *dividend – calculation* | 6.48 | 6.00 | 21.00 | 0.00 | 0.05 | 1.16 | 0.00 | 0.00 |
| *victim – emergency* | 6.47 | 6.00 | 23.00 | 0.00 | 0.05 | 1.39 | 0.07 | 0.71 |
| *luxury – car* | 6.47 | 8.00 | 18.00 | 0.00 | 0.05 | 0.90 | 0.00 | 0.00 |
| *tool – implement* | 6.46 | 16.00 | 29.00 | 4.00 | 0.55 | 2.77 | 0.94 | 5.99 |
| *competition – price* | 6.44 | 2.00 | 23.00 | 0.00 | 0.08 | 1.39 | 0.25 | 2.65 |
| *street – place* | 6.44 | 14.00 | 25.00 | 3.00 | 0.07 | 1.67 | 0.30 | 3.35 |
| *psychology – doctor* | 6.42 | 16.00 | 18.00 | 0.00 | 0.06 | 0.90 | 0.00 | 0.00 |
| *gender – equality* | 6.41 | 2.00 | 23.00 | 0.00 | 0.06 | 1.39 | 0.30 | 3.25 |
| *listing – category* | 6.38 | 4.00 | 23.00 | 0.00 | 0.07 | 1.39 | 0.00 | 0.00 |
| *discovery – space* | 6.34 | 12.00 | 23.00 | 0.00 | 0.06 | 1.39 | 0.00 | 0.00 |
| *oil – stock* | 6.34 | 14.00 | 22.00 | 0.00 | 0.09 | 1.27 | 0.43 | 5.19 |
| *video – archive* | 6.34 | | 23.00 | 0.00 | 0.06 | 1.39 | 0.22 | 2.45 |
| *governor – office* | 6.34 | 17.00 | 22.00 | 0.00 | 0.06 | 1.27 | 0.24 | 2.45 |
| *train – car* | 6.31 | 16.00 | 25.00 | 4.00 | 0.18 | 1.67 | 0.71 | 5.50 |
| *record – number* | 6.31 | 14.00 | 29.00 | 6.00 | 0.32 | 2.77 | 0.84 | 5.87 |
| *shower – thunderstorm* | 6.31 | 16.00 | 24.00 | 0.00 | 0.08 | 1.52 | 0.56 | 6.85 |
| *brother – monk* | 6.27 | 14.00 | 29.00 | 4.00 | 0.29 | 2.77 | 0.90 | 10.48 |
| *nature – man* | 6.25 | 14.00 | 27.00 | 4.00 | 0.08 | 2.08 | 0.30 | 2.40 |

| Word Pair | Gabr. | ELKB | WN Edges | Hirst St.O. | Jiang Con. | Lea. Chod. | Lin | Res. |
|---|---|---|---|---|---|---|---|---|
| *production – crew* | 6.25 | 8.00 | 24.00 | 0.00 | 0.06 | 1.52 | 0.00 | 0.00 |
| *family – planning* | 6.25 | 16.00 | 23.00 | 0.00 | 0.06 | 1.39 | 0.00 | 0.00 |
| *disaster – area* | 6.25 | 16.00 | 22.00 | 0.00 | 0.08 | 1.27 | 0.37 | 3.35 |
| *skin – eye* | 6.22 | 16.00 | 25.00 | 2.00 | 0.10 | 1.67 | 0.50 | 4.21 |
| *food – preparation* | 6.22 | 16.00 | 27.00 | 3.00 | 0.12 | 2.08 | 0.47 | 3.21 |
| *bread – butter* | 6.19 | 12.00 | 27.00 | 5.00 | 0.12 | 2.08 | 0.61 | 5.69 |
| *movie – popcorn* | 6.19 | 0.00 | 18.00 | 0.00 | 0.05 | 0.90 | 0.00 | 0.00 |
| *game – series* | 6.19 | 12.00 | 28.00 | 5.00 | 0.21 | 2.37 | 0.75 | 5.67 |
| *lover – quarrel* | 6.19 | 8.00 | 21.00 | 0.00 | 0.05 | 1.16 | 0.07 | 0.71 |
| *preservation – world* | 6.19 | 12.00 | 24.00 | 0.00 | 0.06 | 1.52 | 0.00 | 0.00 |
| *dollar – loss* | 6.09 | 6.00 | 21.00 | 0.00 | 0.07 | 1.16 | 0.19 | 1.81 |
| *weapon – secret* | 6.06 | 14.00 | 21.00 | 0.00 | 0.06 | 1.16 | 0.00 | 0.00 |
| *precedent – antecedent* | 6.04 | 16.00 | 25.00 | 4.00 | 0.05 | 1.67 | 0.30 | 3.84 |
| *shower – flood* | 6.03 | 16.00 | 26.00 | 3.00 | 0.09 | 1.86 | 0.60 | 7.56 |
| *registration – arrangement* | 6.00 | 4.00 | 25.00 | 2.00 | 0.07 | 1.67 | 0.32 | 3.12 |
| *arrival – hotel* | 6.00 | 2.00 | 21.00 | 0.00 | 0.05 | 1.16 | 0.07 | 0.71 |
| *announcement – warning* | 6.00 | 16.00 | 25.00 | 2.00 | 0.08 | 1.67 | 0.40 | 3.84 |
| *baseball – season* | 5.97 | 4.00 | 17.00 | 0.00 | 0.06 | 0.83 | 0.00 | 0.00 |
| *game – round* | 5.97 | 16.00 | 26.00 | 4.00 | 0.16 | 1.86 | 0.72 | 7.23 |
| *drink – mouth* | 5.96 | 12.00 | 24.00 | 0.00 | 0.07 | 1.52 | 0.25 | 2.40 |
| *energy – crisis* | 5.94 | 14.00 | 24.00 | 0.00 | 0.06 | 1.52 | 0.30 | 3.35 |
| *grocery – money* | 5.94 | | 20.00 | 0.00 | 0.06 | 1.07 | 0.00 | 0.00 |
| *life – lesson* | 5.94 | 6.00 | 21.00 | 0.00 | 0.07 | 1.16 | 0.29 | 2.88 |
| *cucumber – potato* | 5.92 | 14.00 | 27.00 | 4.00 | 0.11 | 2.08 | 0.64 | 7.50 |
| *king – rook* | 5.92 | 16.00 | 28.00 | 5.00 | 0.20 | 2.37 | 0.85 | 10.93 |
| *reason – criterion* | 5.91 | 4.00 | 23.00 | 0.00 | 0.09 | 1.39 | 0.37 | 2.89 |
| *equipment – maker* | 5.91 | 6.00 | 22.00 | 0.00 | 0.07 | 1.27 | 0.10 | 0.71 |
| *cup – liquid* | 5.90 | 12.00 | 25.00 | 3.00 | 0.16 | 1.67 | 0.69 | 5.97 |
| *deployment – withdrawal* | 5.88 | 6.00 | 22.00 | 0.00 | 0.05 | 1.27 | 0.21 | 2.25 |
| *tiger – zoo* | 5.87 | 6.00 | 23.00 | 0.00 | 0.04 | 1.39 | 0.06 | 0.71 |
| *journey – car* | 5.85 | 12.00 | 17.00 | 0.00 | 0.07 | 0.83 | 0.00 | 0.00 |
| *precedent – example* | 5.85 | 16.00 | 29.00 | 4.00 | 0.23 | 2.77 | 0.83 | 7.81 |
| *smart – stupid* | 5.81 | 8.00 | 18.00 | 3.00 | 0.04 | 0.90 | 0.00 | 0.00 |
| *plane – car* | 5.77 | 16.00 | 24.00 | 3.00 | 0.21 | 1.52 | 0.75 | 5.55 |
| *planet – people* | 5.75 | 4.00 | 23.00 | 4.00 | 0.07 | 1.39 | 0.00 | 0.00 |
| *lobster – wine* | 5.70 | 12.00 | 21.00 | 0.00 | 0.07 | 1.16 | 0.33 | 3.21 |
| *money – laundering* | 5.65 | | 21.00 | 0.00 | 0.05 | 1.16 | 0.00 | 0.00 |
| *Mars – scientist* | 5.63 | 4.00 | 20.00 | 0.00 | 0.06 | 1.07 | 0.09 | 0.71 |
| *decoration – valor* | 5.63 | 4.00 | 19.00 | 0.00 | 0.06 | 0.98 | 0.18 | 1.81 |
| *OPEC – country* | 5.63 | 2.00 | 24.00 | 5.00 | 0.08 | 1.52 | 0.35 | 3.34 |
| *summer – nature* | 5.63 | 6.00 | 22.00 | 0.00 | 0.07 | 1.27 | 0.22 | 1.81 |
| *tiger – fauna* | 5.62 | 12.00 | 27.00 | 2.00 | 0.12 | 2.08 | 0.55 | 4.32 |
| *psychology – discipline* | 5.58 | 4.00 | 28.00 | 6.00 | 0.22 | 2.37 | 0.77 | 6.01 |
| *glass – metal* | 5.56 | 16.00 | 26.00 | 3.00 | 0.09 | 1.86 | 0.40 | 3.21 |
| *alcohol – chemistry* | 5.54 | 2.00 | 20.00 | 0.00 | 0.06 | 1.07 | 0.00 | 0.00 |
| *disability – death* | 5.47 | 6.00 | 24.00 | 0.00 | 0.08 | 1.52 | 0.38 | 3.35 |
| *change – attitude* | 5.44 | 16.00 | 25.00 | 2.00 | 0.08 | 1.67 | 0.30 | 3.25 |

| Word Pair | Gabr. | ELKB | WN Edges | Hirst St.O. | Jiang Con. | Lea. Chod. | Lin | Res. |
|---|---|---|---|---|---|---|---|---|
| *arrangement –* | | | | | | | | |
| *accommodation* | 5.41 | 14.00 | 27.00 | 4.00 | 0.08 | 2.08 | 0.43 | 4.92 |
| *territory – surface* | 5.34 | 6.00 | 25.00 | 2.00 | 0.16 | 1.67 | 0.56 | 3.39 |
| *credit – information* | 5.31 | 6.00 | 27.00 | 4.00 | 0.12 | 2.08 | 0.50 | 4.78 |
| *size – prominence* | 5.31 | 14.00 | 25.00 | 2.00 | 0.08 | 1.67 | 0.35 | 3.35 |
| *exhibit – memorabilia* | 5.31 | 4.00 | 21.00 | 0.00 | 0.05 | 1.16 | 0.24 | 2.88 |
| *territory – kilometer* | 5.28 | 4.00 | 22.00 | 0.00 | 0.06 | 1.27 | 0.00 | 0.00 |
| *death – row* | 5.25 | 6.00 | 24.00 | 0.00 | 0.06 | 1.52 | 0.22 | 2.25 |
| *man – governor* | 5.25 | 14.00 | 26.00 | 3.00 | 0.09 | 1.86 | 0.32 | 3.44 |
| *doctor – liability* | 5.19 | 6.00 | 21.00 | 0.00 | 0.06 | 1.16 | 0.00 | 0.00 |
| *impartiality – interest* | 5.16 | 14.00 | 22.00 | 0.00 | 0.05 | 1.27 | 0.00 | 0.00 |
| *energy – laboratory* | 5.09 | 14.00 | 20.00 | 0.00 | 0.06 | 1.07 | 0.00 | 0.00 |
| *secretary – senate* | 5.06 | 6.00 | 18.00 | 0.00 | 0.05 | 0.90 | 0.00 | 0.00 |
| *death – inmate* | 5.03 | 4.00 | 22.00 | 0.00 | 0.05 | 1.27 | 0.00 | 0.00 |
| *monk – oracle* | 5.00 | 12.00 | 23.00 | 0.00 | 0.06 | 1.39 | 0.23 | 2.45 |
| *cup – food* | 5.00 | 14.00 | 25.00 | 3.00 | 0.14 | 1.67 | 0.61 | 4.99 |
| *doctor – personnel* | 5.00 | 14.00 | 21.00 | 0.00 | 0.07 | 1.16 | 0.00 | 0.00 |
| *travel – activity* | 5.00 | 14.00 | 25.00 | 2.00 | 0.19 | 1.67 | 0.51 | 2.25 |
| *journal – association* | 4.97 | 4.00 | 24.00 | 0.00 | 0.05 | 1.52 | 0.22 | 2.65 |
| *car – flight* | 4.94 | 12.00 | 23.00 | 0.00 | 0.07 | 1.39 | 0.28 | 2.45 |
| *street – children* | 4.94 | 12.00 | 22.00 | 0.00 | 0.07 | 1.27 | 0.10 | 0.71 |
| *space – chemistry* | 4.88 | 6.00 | 26.00 | 3.00 | 0.06 | 1.86 | 0.23 | 2.88 |
| *situation – conclusion* | 4.81 | 4.00 | 25.00 | 0.00 | 0.10 | 1.67 | 0.32 | 2.25 |
| *tiger – organism* | 4.77 | 6.00 | 28.00 | 2.00 | 0.10 | 2.37 | 0.32 | 2.17 |
| *peace – plan* | 4.75 | 12.00 | 22.00 | 0.00 | 0.08 | 1.27 | 0.34 | 2.80 |
| *word – similarity* | 4.75 | 14.00 | 22.00 | 0.00 | 0.09 | 1.27 | 0.26 | 1.81 |
| *consumer – energy* | 4.75 | 6.00 | 21.00 | 0.00 | 0.06 | 1.16 | 0.00 | 0.00 |
| *ministry – culture* | 4.69 | 4.00 | 21.00 | 0.00 | 0.06 | 1.16 | 0.29 | 3.04 |
| *hospital – infrastructure* | 4.63 | 6.00 | 18.00 | 2.00 | 0.04 | 0.90 | 0.00 | 0.00 |
| *smart – student* | 4.62 | 6.00 | 18.00 | 0.00 | 0.05 | 0.90 | 0.00 | 0.00 |
| *investigation – effort* | 4.59 | 4.00 | 27.00 | 4.00 | 0.17 | 2.08 | 0.65 | 4.48 |
| *image – surface* | 4.56 | 16.00 | 26.00 | 3.00 | 0.11 | 1.86 | 0.37 | 3.39 |
| *life – term* | 4.50 | 14.00 | 28.00 | 5.00 | 0.11 | 2.37 | 0.48 | 3.72 |
| *computer – news* | 4.47 | 8.00 | 20.00 | 0.00 | 0.06 | 1.07 | 0.00 | 0.00 |
| *board – recommendation* | 4.47 | 6.00 | 17.00 | 0.00 | 0.06 | 0.83 | 0.00 | 0.00 |
| *start – match* | 4.47 | 6.00 | 24.00 | 4.00 | 0.09 | 1.52 | 0.42 | 3.78 |
| *lad – brother* | 4.46 | 14.00 | 26.00 | 3.00 | 0.07 | 1.86 | 0.27 | 2.45 |
| *food – rooster* | 4.42 | 12.00 | 18.00 | 0.00 | 0.06 | 0.90 | 0.09 | 0.71 |
| *coast – hill* | 4.38 | 4.00 | 26.00 | 2.00 | 0.15 | 1.86 | 0.69 | 6.36 |
| *observation – architecture* | 4.38 | 4.00 | 25.00 | 2.00 | 0.06 | 1.67 | 0.29 | 3.12 |
| *attempt – peace* | 4.25 | 8.00 | 24.00 | 0.00 | 0.06 | 1.52 | 0.00 | 0.00 |
| *deployment – departure* | 4.25 | 6.00 | 23.00 | 0.00 | 0.06 | 1.39 | 0.21 | 2.25 |
| *benchmark – index* | 4.25 | 12.00 | 27.00 | 4.00 | 0.07 | 2.08 | 0.50 | 6.15 |
| *consumer – confidence* | 4.13 | 4.00 | 21.00 | 0.00 | 0.05 | 1.16 | 0.00 | 0.00 |
| *start – year* | 4.06 | 6.00 | 27.00 | 4.00 | 0.11 | 2.08 | 0.49 | 3.72 |
| *focus – life* | 4.06 | 14.00 | 25.00 | 2.00 | 0.08 | 1.67 | 0.37 | 3.35 |
| *development – issue* | 3.97 | 12.00 | 27.00 | 4.00 | 0.16 | 2.08 | 0.61 | 4.18 |

| Word Pair | Gabr. | ELKB | WN Edges | Hirst St.O. | Jiang Con. | Lea. Chod. | Lin | Res. |
|---|---|---|---|---|---|---|---|---|
| *day – summer* | 3.94 | 16.00 | 27.00 | 4.00 | 0.11 | 2.08 | 0.48 | 3.72 |
| *theater – history* | 3.91 | 6.00 | 24.00 | 0.00 | 0.06 | 1.52 | 0.00 | 0.00 |
| *situation – isolation* | 3.88 | 6.00 | 26.00 | 3.00 | 0.10 | 1.86 | 0.43 | 3.35 |
| *media – trading* | 3.88 | 16.00 | 23.00 | 0.00 | 0.05 | 1.39 | 0.21 | 2.25 |
| *profit – warning* | 3.88 | 8.00 | 20.00 | 0.00 | 0.06 | 1.07 | 0.18 | 1.81 |
| *chance – credibility* | 3.88 | 14.00 | 25.00 | 2.00 | 0.06 | 1.67 | 0.19 | 1.81 |
| *precedent – information* | 3.85 | 6.00 | 28.00 | 6.00 | 0.14 | 2.37 | 0.63 | 5.21 |
| *architecture – century* | 3.78 | 2.00 | 21.00 | 0.00 | 0.06 | 1.16 | 0.18 | 1.81 |
| *population – development* | 3.75 | 6.00 | 25.00 | 0.00 | 0.11 | 1.67 | 0.57 | 5.60 |
| *stock – live* | 3.73 | 16.00 | | 0.00 | | | | |
| *cup – object* | 3.69 | 14.00 | 26.00 | 4.00 | 0.14 | 1.86 | 0.38 | 1.96 |
| *atmosphere – landscape* | 3.69 | 12.00 | 22.00 | 0.00 | 0.07 | 1.27 | 0.32 | 3.27 |
| *minority – peace* | 3.69 | 14.00 | 25.00 | 2.00 | 0.06 | 1.67 | 0.29 | 3.35 |
| *peace – atmosphere* | 3.69 | 6.00 | 26.00 | 3.00 | 0.08 | 1.86 | 0.41 | 4.26 |
| *morality – marriage* | 3.69 | 8.00 | 24.00 | 0.00 | 0.06 | 1.52 | 0.00 | 0.00 |
| *report – gain* | 3.63 | 4.00 | 22.00 | 0.00 | 0.08 | 1.27 | 0.23 | 1.81 |
| *music – project* | 3.63 | 14.00 | 26.00 | 3.00 | 0.10 | 1.86 | 0.41 | 3.12 |
| *seven – series* | 3.56 | 4.00 | 21.00 | 0.00 | 0.06 | 1.16 | 0.20 | 1.81 |
| *experience – music* | 3.47 | 12.00 | 23.00 | 0.00 | 0.08 | 1.39 | 0.35 | 2.89 |
| *school – center* | 3.44 | 16.00 | 28.00 | 2.00 | 0.12 | 2.37 | 0.59 | 5.33 |
| *announcement – production* | 3.38 | 4.00 | 23.00 | 0.00 | 0.08 | 1.39 | 0.33 | 2.88 |
| *five – month* | 3.38 | 4.00 | 23.00 | 0.00 | 0.08 | 1.39 | 0.35 | 2.97 |
| *money – operation* | 3.31 | 6.00 | 23.00 | 0.00 | 0.08 | 1.39 | 0.00 | 0.00 |
| *delay – news* | 3.31 | 8.00 | 22.00 | 0.00 | 0.07 | 1.27 | 0.21 | 1.81 |
| *morality – importance* | 3.31 | 2.00 | 26.00 | 3.00 | 0.13 | 1.86 | 0.57 | 4.23 |
| *governor – interview* | 3.25 | 4.00 | 18.00 | 0.00 | 0.05 | 0.90 | 0.00 | 0.00 |
| *practice – institution* | 3.19 | 16.00 | 28.00 | 6.00 | 0.21 | 2.37 | 0.82 | 8.52 |
| *century – nation* | 3.16 | 6.00 | 22.00 | 0.00 | 0.07 | 1.27 | 0.00 | 0.00 |
| *coast – forest* | 3.15 | 16.00 | 24.00 | 0.00 | 0.06 | 1.52 | 0.20 | 1.96 |
| *shore – woodland* | 3.08 | 6.00 | 25.00 | 2.00 | 0.06 | 1.67 | 0.21 | 1.96 |
| *drink – car* | 3.04 | 6.00 | 22.00 | 0.00 | 0.10 | 1.27 | 0.31 | 1.96 |
| *president – medal* | 3.00 | 4.00 | 16.00 | 0.00 | 0.05 | 0.76 | 0.00 | 0.00 |
| *prejudice – recognition* | 3.00 | 6.00 | 22.00 | 0.00 | 0.07 | 1.27 | 0.30 | 2.89 |
| *viewer – serial* | 2.97 | 4.00 | 20.00 | 0.00 | 0.06 | 1.07 | 0.23 | 2.45 |
| *Mars – water* | 2.94 | 12.00 | 23.00 | 0.00 | 0.07 | 1.39 | 0.24 | 1.96 |
| *peace – insurance* | 2.94 | 4.00 | 27.00 | 4.00 | 0.15 | 2.08 | 0.74 | 8.28 |
| *cup – artifact* | 2.92 | 0.00 | 27.00 | 5.00 | 0.15 | 2.08 | 0.45 | 2.45 |
| *media – gain* | 2.88 | 6.00 | 22.00 | 0.00 | 0.05 | 1.27 | 0.00 | 0.00 |
| *precedent – cognition* | 2.81 | 4.00 | 27.00 | 5.00 | 0.11 | 2.08 | 0.41 | 2.89 |
| *announcement – effort* | 2.75 | 4.00 | 20.00 | 0.00 | 0.06 | 1.07 | 0.00 | 0.00 |
| *crane – implement* | 2.69 | 0.00 | 26.00 | 3.00 | 0.09 | 1.86 | 0.39 | 3.44 |
| *line – insurance* | 2.69 | 6.00 | 26.00 | 3.00 | 0.11 | 1.86 | 0.51 | 4.79 |
| *drink – mother* | 2.65 | 16.00 | 25.00 | 2.00 | 0.09 | 1.67 | 0.34 | 3.21 |
| *opera – industry* | 2.63 | 6.00 | 18.00 | 0.00 | 0.06 | 0.90 | 0.18 | 1.81 |
| *volunteer – motto* | 2.56 | 0.00 | 17.00 | 0.00 | 0.04 | 0.83 | 0.00 | 0.00 |
| *listing – proximity* | 2.56 | 6.00 | 19.00 | 0.00 | 0.07 | 0.98 | 0.27 | 2.65 |
| *Arafat – Jackson* | 2.50 | | | | | | | |

| Word Pair | Gabr. | ELKB | WN Edges | Hirst St.O. | Jiang Con. | Lea. Chod. | Lin | Res. |
|---|---|---|---|---|---|---|---|---|
| *precedent – collection* | 2.50 | 6.00 | 27.00 | 5.00 | 0.13 | 2.08 | 0.61 | 5.21 |
| *cup – article* | 2.40 | 6.00 | 26.00 | 3.00 | 0.55 | 1.86 | 0.95 | 7.52 |
| *sign – recess* | 2.38 | 8.00 | 26.00 | 5.00 | 0.07 | 1.86 | 0.25 | 2.45 |
| *problem – airport* | 2.38 | 4.00 | 20.00 | 0.00 | 0.05 | 1.07 | 0.00 | 0.00 |
| *reason – hypertension* | 2.31 | 4.00 | 23.00 | 0.00 | 0.06 | 1.39 | 0.37 | 4.26 |
| *direction – combination* | 2.25 | 6.00 | 24.00 | 0.00 | 0.08 | 1.52 | 0.27 | 3.12 |
| *Wednesday – news* | 2.22 | 4.00 | 18.00 | 0.00 | 0.07 | 0.90 | 0.21 | 1.81 |
| *cup – entity* | 2.15 | 0.00 | 25.00 | 3.00 | 0.13 | 1.67 | 0.16 | 0.71 |
| *cemetery – woodland* | 2.08 | 6.00 | 21.00 | 0.00 | 0.05 | 1.16 | 0.07 | 0.71 |
| *glass – magician* | 2.08 | 4.00 | 22.00 | 0.00 | 0.05 | 1.27 | 0.08 | 0.71 |
| *possibility – girl* | 1.94 | 6.00 | 22.00 | 0.00 | 0.06 | 1.27 | 0.00 | 0.00 |
| *cup – substance* | 1.92 | 6.00 | 25.00 | 2.00 | 0.13 | 1.67 | 0.44 | 3.21 |
| *forest – graveyard* | 1.85 | 6.00 | 21.00 | 0.00 | 0.05 | 1.16 | 0.07 | 0.71 |
| *stock – egg* | 1.81 | 14.00 | 24.00 | 2.00 | 0.10 | 1.52 | 0.53 | 4.99 |
| *energy – secretary* | 1.81 | 4.00 | 20.00 | 0.00 | 0.06 | 1.07 | 0.00 | 0.00 |
| *month – hotel* | 1.81 | 0.00 | 20.00 | 0.00 | 0.06 | 1.07 | 0.00 | 0.00 |
| *precedent – group* | 1.77 | 6.00 | 26.00 | 4.00 | 0.10 | 1.86 | 0.35 | 2.46 |
| *production – hike* | 1.75 | 2.00 | 23.00 | 0.00 | 0.07 | 1.39 | 0.24 | 2.25 |
| *stock – phone* | 1.62 | 12.00 | 24.00 | 2.00 | 0.09 | 1.52 | 0.41 | 4.29 |
| *holy – sex* | 1.62 | 6.00 | 22.00 | 0.00 | 0.05 | 1.27 | 0.00 | 0.00 |
| *stock – CD* | 1.31 | | 25.00 | 2.00 | 0.07 | 1.67 | 0.40 | 4.29 |
| *drink – ear* | 1.31 | 6.00 | 23.00 | 0.00 | 0.07 | 1.39 | 0.21 | 1.96 |
| *delay – racism* | 1.19 | 4.00 | 24.00 | 0.00 | 0.05 | 1.52 | 0.21 | 2.25 |
| *stock – jaguar* | 0.92 | 12.00 | 25.00 | 2.00 | 0.09 | 1.67 | 0.52 | 5.47 |
| *stock – life* | 0.92 | 12.00 | 24.00 | 2.00 | 0.09 | 1.52 | 0.39 | 3.35 |
| *monk – slave* | 0.92 | 6.00 | 26.00 | 3.00 | 0.06 | 1.86 | 0.25 | 2.45 |
| *lad – wizard* | 0.92 | 4.00 | 26.00 | 3.00 | 0.07 | 1.86 | 0.27 | 2.45 |
| *sugar – approach* | 0.88 | 6.00 | 23.00 | 0.00 | 0.07 | 1.39 | 0.24 | 1.96 |
| *rooster – voyage* | 0.62 | 2.00 | 13.00 | 0.00 | 0.04 | 0.58 | 0.00 | 0.00 |
| *chord – smile* | 0.54 | 0.00 | 20.00 | 0.00 | 0.07 | 1.07 | 0.29 | 2.88 |
| *noon – string* | 0.54 | 6.00 | 19.00 | 0.00 | 0.05 | 0.98 | 0.00 | 0.00 |
| *professor – cucumber* | 0.31 | 0.00 | 18.00 | 0.00 | 0.05 | 0.90 | 0.19 | 2.17 |
| *king – cabbage* | 0.23 | 12.00 | 21.00 | 0.00 | 0.06 | 1.16 | 0.27 | 3.21 |
| *Correlation* | 1.00 | 0.54 | 0.27 | 0.34 | 0.35 | 0.36 | 0.36 | 0.37 |

**Table J1:** Comparison of semantic similarity measures using the WordSimilarity-353 Test Collection

# Appendix K: *TOEFL*, *ESL* and *RDWP* questions

This appendix presents 80 *TOEFL* (ETS, 2003), 50 *ESL* (Tatsuki, 1998) and 100 *RDWP* questions (Turney, 2001; Lewis 2000-2001) as well as the answers given by my system using the *ELKB* and the *WordNet-based* system which uses the Hirst and St-Onge (1998) measure. The *WordNet-*based system is implemented using the Semantic Distance software package (Pedersen, 2002). Tad Stach collected the other 200 RDWP questions from the following Canadian issues of *Reader's Digest* (Lewis 2000-2001): January, March, April, May, June, August and September 2000; January and May 2001.

## 1    Semantic Distance measured using the *ELKB*

### 1.A.  80 *TOEFL* Questions

```
Question 1
enormously | tremendously | appropriately | uniquely | decidedly
enormously ADV. [enormously] to tremendously ADV. [tremendously], length
   = 4, 1 path(s) of this length
enormously ADV. [enormously] to appropriately ADV. [appropriately],
   length = 14, 1 path(s) of this length
uniquely is NOT IN THE INDEX
enormously ADV. to decidedly ADV., length = 4, 1 path(s) of this length
Roget thinks that enormously means tremendously
CORRECT

Question 2
provisions | stipulations | interrelations | jurisdictions |
   interpretations
provisions N. [provisions] to stipulations N. [stipulations], length = 0,
   1 path(s) of this length
provisions N. [provisions] to interrelations N. [interrelations], length
   = 14, 3 path(s) of this length
provisions N. [provisions] to jurisdictions N. [jurisdictions], length =
   12, 3 path(s) of this length
provisions N. to interpretations N., length = 12, 36 path(s) of this length
Roget thinks that provisions means stipulations
CORRECT

Question 3
haphazardly | randomly | dangerously | densely | linearly
haphazardly ADV. [haphazardly] to randomly ADV. [randomly], length = 0, 1
   path(s) of this length
haphazardly ADV. [haphazardly] to dangerously VB. [dangerously], length =
   12, 1 path(s) of this length
haphazardly ADV. [haphazardly] to densely ADJ. [densely], length = 16, 2
   path(s) of this length
linearly is NOT IN THE INDEX
Roget thinks that haphazardly means randomly
CORRECT

Question 4
prominent | conspicuous | battered | ancient | mysterious
prominent ADJ. [prominent] to conspicuous ADJ. [conspicuous], length = 0,
   4 path(s) of this length
prominent ADJ. [prominent] to battered ADJ. [battered], length = 10, 10
   path(s) of this length
prominent ADJ. [prominent] to ancient ADJ. [ancient], length = 4, 1
   path(s) of this length
prominent ADJ. to mysterious ADJ., length = 8, 2 path(s) of this length
Roget thinks that prominent means conspicuous
CORRECT

Question 5
zenith | pinnacle | completion | outset | decline
zenith N. [zenith] to pinnacle N. [pinnacle], length = 0, 3 path(s) of
```

```
    this length
zenith N. [zenith] to completion N. [completion], length = 0, 4 path(s)
    of this length
zenith N. [zenith] to outset N. [outset], length = 8, 1 path(s) of this length
zenith N. to decline N., length = 8, 3 path(s) of this length
Roget thinks that zenith means completion
TIE LOST
INCORRECT
```

**Question 6**
```
flawed | imperfect | tiny | lustrous | crude
flawed ADJ. [flawed] to imperfect ADJ. [imperfect], length = 0, 5 path(s)
    of this length
flawed N. [flawed] to tiny ADJ. [tiny], length = 12, 2 path(s) of this length
flawed N. [flawed] to lustrous ADJ. [lustrous], length = 12, 3 path(s) of
    this length
flawed ADJ. to crude ADJ., length = 2, 2 path(s) of this length
Roget thinks that flawed means imperfect
CORRECT
```

**Question 7**
```
urgently | desperately | typically | conceivably | tentatively
urgently ADV. [urgently] to desperately ADV. [desperately], length = 16,
    1 path(s) of this length
typically is NOT IN THE INDEX
urgently ADV. [urgently] to conceivably ADV. [conceivably], length = 16,
    1 path(s) of this length
tentatively is NOT IN THE INDEX
Roget thinks that urgently means desperately
CORRECT
```

**Question 8**
```
consumed | eaten | bred | caught | supplied
consumed VB. [consumed] to eaten VB. [eaten], length = 0, 6 path(s) of
    this length
consumed VB. [consumed] to bred VB. [bred], length = 8, 6 path(s) of this length
consumed VB. [consumed] to caught VB. [caught], length = 0, 1 path(s) of
    this length
consumed VB. to supplied N., length = 8, 9 path(s) of this length
Roget thinks that consumed means eaten
TIE BROKEN
CORRECT
```

**Question 9**
```
advent | coming | arrest | financing | stability
advent N. [advent] to coming N. [coming], length = 0, 5 path(s) of this length
advent N. [advent] to arrest N. [arrest], length = 12, 2 path(s) of this length
advent N. [advent] to financing VB. [financing], length = 16, 28 path(s)
    of this length
advent N. to stability N., length = 10, 5 path(s) of this length
Roget thinks that advent means coming
CORRECT
```

**Question 10**
```
concisely | succinctly | powerfully | positively | freely
succinctly (ANSWER) is NOT IN THE INDEX
concisely ADV. [concisely] to powerfully ADV. [powerfully], length = 16,
    6 path(s) of this length
concisely ADV. [concisely] to positively ADV. [positively], length = 12,
    4 path(s) of this length
concisely ADV. to freely VB., length = 12, 6 path(s) of this length
Roget thinks that concisely means freely
INCORRECT
```

**Question 11**
```
salutes | greetings | information | ceremonies | privileges
salutes N. [salutes] to greetings N. [greetings], length = 0, 3 path(s)
    of this length
salutes VB. [salutes] to information N. [information], length = 6, 1
    path(s) of this length
salutes N. [salutes] to ceremonies N. [ceremonies], length = 0, 1 path(s)
    of this length
salutes VB. to privileges N., length = 12, 26 path(s) of this length
Roget thinks that salutes means greetings
TIE BROKEN
CORRECT
```

**Question 12**
solitary | alone | alert | restless | fearless
solitary ADJ. [solitary] to alone ADJ. [alone], length = 0, 6 path(s) of
    this length
solitary N. [solitary] to alert ADJ. [alert], length = 12, 10 path(s) of
    this length
solitary N. [solitary] to restless ADJ. [restless], length = 8, 1 path(s)
    of this length
solitary N. to fearless ADJ., length = 14, 9 path(s) of this length
Roget thinks that solitary means alone
CORRECT

**Question 13**
hasten | accelerate | permit | determine | accompany
hasten VB. [hasten] to accelerate VB. [accelerate], length = 2, 5 path(s)
    of this length
hasten VB. [hasten] to permit N. [permit], length = 10, 6 path(s) of this length
hasten VB. [hasten] to determine VB. [determine], length = 2, 1 path(s)
    of this length
hasten VB. to accompany VB., length = 10, 12 path(s) of this length
Roget thinks that hasten means accelerate
TIE BROKEN
CORRECT

**Question 14**
perseverance | endurance | skill | generosity | disturbance
perseverance N. [perseverance] to endurance N. [endurance], length = 2, 4
    path(s) of this length
perseverance N. [perseverance] to skill N. [skill], length = 10, 16
    path(s) of this length
perseverance N. [perseverance] to generosity N. [generosity], length =
    14, 2 path(s) of this length
perseverance N. to disturbance N., length = 14, 11 path(s) of this length
Roget thinks that perseverance means endurance
CORRECT

**Question 15**
fanciful | imaginative | familiar | apparent | logical
fanciful ADJ. [fanciful] to imaginative ADJ. [imaginative], length = 0, 2
    path(s) of this length
fanciful ADJ. [fanciful] to familiar ADJ. [familiar], length = 10, 10
    path(s) of this length
fanciful ADJ. [fanciful] to apparent ADJ. [apparent], length = 12, 13
    path(s) of this length
fanciful ADJ. to logical ADJ., length = 10, 32 path(s) of this length
Roget thinks that fanciful means imaginative
CORRECT

**Question 16**
showed | demonstrated | published | repeated | postponed
showed VB. [showed] to demonstrated VB. [demonstrated], length = 0, 15
    path(s) of this length
showed VB. [showed] to published VB. [published], length = 0, 7 path(s)
    of this length
showed N. [showed] to repeated N. [repeated], length = 0, 2 path(s) of
    this length
showed VB. to postponed VB., length = 12, 52 path(s) of this length
Roget thinks that showed means demonstrated
TIE BROKEN
CORRECT

**Question 17**
constantly | continually | instantly | rapidly | accidentally
constantly ADV. [constantly] to continually ADV. [continually], length =
    0, 1 path(s) of this length
constantly ADV. [constantly] to instantly ADV. [instantly], length = 8, 1
    path(s) of this length
constantly ADV. [constantly] to rapidly ADV. [rapidly], length = 16, 2
    path(s) of this length
constantly ADV. to accidentally ADV., length = 14, 2 path(s) of this length
Roget thinks that constantly means continually
CORRECT

**Question 18**
issues | subjects | training | salaries | benefits
issues N. [issues] to subjects N. [subjects], length = 0, 1 path(s) of
    this length

issues N. [issues] to training N. [training], length = 10, 149 path(s) of
    this length
issues N. [issues] to salaries N. [salaries], length = 10, 14 path(s) of
    this length
issues VB. to benefits N., length = 10, 62 path(s) of this length
Roget thinks that issues means subjects
CORRECT

**Question 19**
furnish | supply | impress | protect | advise
furnish VB. [furnish] to supply VB. [supply], length = 0, 2 path(s) of
    this length
furnish VB. [furnish] to impress VB. [impress], length = 10, 15 path(s)
    of this length
furnish VB. [furnish] to protect VB. [protect], length = 10, 6 path(s) of
    this length
furnish VB. to advise VB., length = 10, 10 path(s) of this length
Roget thinks that furnish means supply
CORRECT

**Question 20**
costly | expensive | beautiful | popular | complicated
costly ADJ. [costly] to expensive ADJ. [expensive], length = 0, 2 path(s)
    of this length
costly ADJ. [costly] to beautiful ADJ. [beautiful], length = 4, 1 path(s)
    of this length
costly ADJ. [costly] to popular N. [popular], length = 8, 1 path(s) of
    this length
costly ADJ. to complicated VB., length = 12, 2 path(s) of this length
Roget thinks that costly means expensive
CORRECT

**Question 21**
recognized | acknowledged | successful | depicted | welcomed
recognized VB. [recognized] to acknowledged VB. [acknowledged], length =
    0, 5 path(s) of this length
recognized VB. [recognized] to successful ADJ. [successful], length = 10,
    91 path(s) of this length
recognized VB. [recognized] to depicted VB. [depicted], length = 12, 16
    path(s) of this length
recognized VB. to welcomed VB., length = 0, 1 path(s) of this length
Roget thinks that recognized means acknowledged
TIE BROKEN
CORRECT

**Question 22**
spot | location | climate | latitude | sea
spot N. [spot] to location N. [location], length = 2, 1 path(s) of this length
spot VB. [spot] to climate N. [climate], length = 8, 2 path(s) of this length
spot N. [spot] to latitude N. [latitude], length = 4, 1 path(s) of this length
spot ADJ. to sea ADJ., length = 8, 6 path(s) of this length
Roget thinks that spot means location
CORRECT

**Question 23**
make | earn | print | trade | borrow
make VB. [make] to earn VB. [earn], length = 2, 5 path(s) of this length
make VB. [make] to print VB. [print], length = 2, 9 path(s) of this length
make VB. [make] to trade VB. [trade], length = 0, 5 path(s) of this length
make VB. to borrow VB., length = 2, 9 path(s) of this length
Roget thinks that make means trade
INCORRECT

**Question 24**
often | frequently | definitely | chemically | hardly
often ADV. [often] to frequently ADV. [frequently], length = 0, 3 path(s)
    of this length
often ADV. [often] to definitely ADV. [definitely], length = 14, 5
    path(s) of this length
chemically is NOT IN THE INDEX
often ADV. to hardly ADV., length = 2, 1 path(s) of this length
Roget thinks that often means frequently
CORRECT

**Question 25**
easygoing | relaxed | frontier | boring | farming
easygoing ADJ. [easygoing] to relaxed VB. [relaxed], length = 6, 1

```
    path(s) of this length
easygoing ADJ. [easygoing] to frontier N. [frontier], length = 16, 24
    path(s) of this length
easygoing ADJ. [easygoing] to boring ADJ. [boring], length = 8, 1 path(s)
    of this length
easygoing ADJ. to farming N., length = 12, 2 path(s) of this length
Roget thinks that easygoing means relaxed
CORRECT
```

**Question 26**
```
debate | argument | war | election | competition
debate N. [debate] to argument N. [argument], length = 0, 4 path(s) of
    this length
debate N. [debate] to war N. [war], length = 2, 3 path(s) of this length
debate VB. [debate] to election N. [election], length = 10, 11 path(s) of
    this length
debate N. to competition N., length = 2, 2 path(s) of this length
Roget thinks that debate means argument
CORRECT
```

**Question 27**
```
narrow | thin | clear | freezing | poisonous
narrow VB. [narrow] to thin VB. [thin], length = 0, 10 path(s) of this length
narrow N. [narrow] to clear N. [clear], length = 0, 2 path(s) of this length
narrow VB. [narrow] to freezing N. [freezing], length = 6, 4 path(s) of
    this length
narrow N. to poisonous ADJ., length = 6, 2 path(s) of this length
Roget thinks that narrow means thin
TIE BROKEN
CORRECT
```

**Question 28**
```
arranged | planned | explained | studied | discarded
arranged VB. [arranged] to planned VB. [planned], length = 0, 6 path(s)
    of this length
arranged VB. [arranged] to explained VB. [explained], length = 4, 1
    path(s) of this length
arranged VB. [arranged] to studied VB. [studied], length = 4, 1 path(s)
    of this length
arranged ADJ. to discarded N., length = 10, 30 path(s) of this length
Roget thinks that arranged means planned
CORRECT
```

**Question 29**
```
infinite | limitless | relative | unusual | structural
infinite ADJ. [infinite] to limitless ADJ. [limitless], length = 0, 2 path(s) of this length
infinite ADJ. [infinite] to relative ADJ. [relative], length = 12, 6 path(s) of this length
infinite ADJ. [infinite] to unusual ADJ. [unusual], length = 4, 1 path(s) of this length
infinite ADJ. to structural ADJ., length = 12, 2 path(s) of this length
Roget thinks that infinite means limitless
CORRECT
```

**Question 30**
```
showy | striking | prickly | entertaining | incidental
showy ADJ. [showy] to striking ADJ. [striking], length = 2, 1 path(s) of
    this length
showy ADJ. [showy] to prickly N. [prickly], length = 10, 4 path(s) of
    this length
showy ADJ. [showy] to entertaining VB. [entertaining], length = 6, 2
    path(s) of this length
showy ADJ. to incidental N., length = 10, 2 path(s) of this length
Roget thinks that showy means striking
CORRECT
```

**Question 31**
```
levied | imposed | believed | requested | correlated
levied VB. [levied] to imposed VB. [imposed], length = 4, 1 path(s) of
    this length
levied VB. [levied] to believed VB. [believed], length = 12, 48 path(s)
    of this length
levied VB. [levied] to requested VB. [requested], length = 0, 5 path(s)
    of this length
levied VB. to correlated VB., length = 12, 9 path(s) of this length
Roget thinks that levied means requested
INCORRECT
```

**Question 32**

```
deftly | skillfully | prudently | occasionally | humorously
deftly (PROBLEM) not found in the index!!
```

**Question 33**
```
distribute | circulate | commercialize | research | acknowledge
distribute VB. [distribute] to circulate VB. [circulate], length = 0, 2
  path(s) of this length
distribute VB. [distribute] to commercialize VB. [commercialize], length
  = 12, 1 path(s) of this length
distribute VB. [distribute] to research N. [research], length = 12, 13
  path(s) of this length
distribute VB. to acknowledge VB., length = 10, 8 path(s) of this length
Roget thinks that distribute means circulate
CORRECT
```

**Question 34**
```
discrepancies | differences | weights | deposits | wavelengths
discrepancies N. [discrepancies] to differences N. [differences], length
  = 0, 2 path(s) of this length
discrepancies N. [discrepancies] to weights VB. [weights], length = 14,
  14 path(s) of this length
discrepancies N. [discrepancies] to deposits N. [deposits], length = 14,
  4 path(s) of this length
discrepancies N. to wavelengths N., length = 16, 12 path(s) of this length
Roget thinks that discrepancies means differences
CORRECT
```

**Question 35**
```
prolific | productive | serious | capable | promising
prolific ADJ. [prolific] to productive ADJ. [productive], length = 0, 5
  path(s) of this length
prolific ADJ. [prolific] to serious ADJ. [serious], length = 10, 20
  path(s) of this length
prolific ADJ. [prolific] to capable ADJ. [capable], length = 12, 6
  path(s) of this length
prolific ADJ. to promising N., length = 10, 27 path(s) of this length
Roget thinks that prolific means productive
CORRECT
```

**Question 36**
```
unmatched | unequaled | unrecognized | alienated | emulated
unmatched ADJ. [unmatched] to unequaled ADJ. [unequaled], length = 2, 1
  path(s) of this length
unmatched ADJ. [unmatched] to unrecognized ADJ. [unrecognized], length =
  16, 6 path(s) of this length
unmatched ADJ. [unmatched] to alienated VB. [alienated], length = 14, 2
  path(s) of this length
unmatched ADJ. to emulated ADJ., length = 2, 1 path(s) of this length
Roget thinks that unmatched means unequaled
CORRECT
```

**Question 37**
```
peculiarly | uniquely | partly | patriotically | suspiciously
uniquely (ANSWER) is NOT IN THE INDEX
peculiarly ADV. [peculiarly] to partly ADV. [partly], length = 8, 1
  path(s) of this length
patriotically is NOT IN THE INDEX
suspiciously is NOT IN THE INDEX
Roget thinks that peculiarly means partly
INCORRECT
```

**Question 38**
```
hue | color | glare | contrast | scent
hue N. [hue] to color N. [color], length = 2, 4 path(s) of this length
hue N. [hue] to glare VB. [glare], length = 10, 6 path(s) of this length
hue N. [hue] to contrast N. [contrast], length = 10, 1 path(s) of this length
hue N. to scent VB., length = 10, 4 path(s) of this length
Roget thinks that hue means color
CORRECT
```

**Question 39**
```
hind | rear | curved | muscular |hairy
hind ADJ. [hind] to rear ADJ. [rear], length = 2, 1 path(s) of this length
hind N. [hind] to curved N. [curved], length = 4, 1 path(s) of this length
hind N. [hind] to muscular N. [muscular], length = 14, 2 path(s) of this length
hind ADJ. to hairy ADJ., length = 12, 2 path(s) of this length
Roget thinks that hind means rear
```

```
CORRECT
```

**Question 40**
```
highlight | accentuate | alter | imitate | restore
highlight VB. [highlight] to accentuate VB. [accentuate], length = 2, 2
   path(s) of this length
highlight VB. [highlight] to alter N. [alter], length = 12, 10 path(s) of
   this length
highlight VB. [highlight] to imitate VB. [imitate], length = 4, 2 path(s)
   of this length
highlight N. to restore VB., length = 10, 19 path(s) of this length
Roget thinks that highlight means accentuate
CORRECT
```

**Question 41**
```
hastily | hurriedly | shrewdly | habitually | chronologically
hastily ADV. [hastily] to hurriedly ADV. [hurriedly], length = 0, 1
   path(s) of this length
hastily ADV. [hastily] to shrewdly ADV. [shrewdly], length = 14, 2
   path(s) of this length
hastily ADV. [hastily] to habitually ADV. [habitually], length = 14, 2
   path(s) of this length
chronologically is NOT IN THE INDEX
Roget thinks that hastily means hurriedly
CORRECT
```

**Question 42**
```
temperate | mild | cold | short | windy
temperate ADJ. [temperate] to mild ADJ. [mild], length = 0, 3 path(s) of
   this length
temperate ADJ. [temperate] to cold ADJ. [cold], length = 2, 4 path(s) of
   this length
temperate ADJ. [temperate] to short N. [short], length = 2, 1 path(s) of
   this length
temperate VB. to windy ADJ., length = 8, 2 path(s) of this length
Roget thinks that temperate means mild
CORRECT
```

**Question 43**
```
grin | smile | exercise | rest | joke
grin N. [grin] to smile N. [smile], length = 0, 4 path(s) of this length
grin N. [grin] to exercise VB. [exercise], length = 10, 4 path(s) of this length
grin N. [grin] to rest VB. [rest], length = 10, 16 path(s) of this length
grin N. to joke N., length = 2, 2 path(s) of this length
Roget thinks that grin means smile
CORRECT
```

**Question 44**
```
verbally | orally | overtly | fittingly | verbosely
orally (ANSWER) is NOT IN THE INDEX
overtly is NOT IN THE INDEX
verbally ADV. [verbally] to fittingly ADV. [fittingly], length = 16, 4
   path(s) of this length
verbosely is NOT IN THE INDEX
Roget thinks that verbally means fittingly
INCORRECT
```

**Question 45**
```
physician | doctor | chemist | pharmacist | nurse
physician N. [physician] to doctor N. [doctor], length = 0, 3 path(s) of
   this length
physician N. [physician] to chemist N. [chemist], length = 4, 1 path(s)
   of this length
physician N. [physician] to pharmacist N. [pharmacist], length = 4, 1
   path(s) of this length
physician N. to nurse VB., length = 2, 2 path(s) of this length
Roget thinks that physician means doctor
CORRECT
```

**Question 46**
```
essentially | basically | possibly | eagerly | ordinarily
essentially ADV. [essentially] to basically ADV. [basically], length =
   16, 5 path(s) of this length
essentially ADV. [essentially] to possibly ADV. [possibly], length = 2, 1
   path(s) of this length
essentially ADV. [essentially] to eagerly ADV. [eagerly], length = 16, 5
path(s) of this length
```

```
ordinarily is NOT IN THE INDEX
Roget thinks that essentially means possibly
INCORRECT
```

**Question 47**
```
keen | sharp | useful | simple | famous
keen ADJ. [keen] to sharp ADJ. [sharp], length = 0, 12 path(s) of this length
keen ADJ. [keen] to useful N. [useful], length = 10, 6 path(s) of this length
keen ADJ. [keen] to simple ADJ. [simple], length = 10, 43 path(s) of this length
keen ADJ. to famous ADJ., length = 12, 9 path(s) of this length
Roget thinks that keen means sharp
CORRECT
```

**Question 48**
```
situated | positioned | rotating | isolated | emptying
situated ADJ. [situated] to positioned ADJ. [positioned], length = 0, 2
   path(s) of this length
situated VB. [situated] to rotating VB. [rotating], length = 14, 51
   path(s) of this length
situated VB. [situated] to isolated N. [isolated], length = 6, 1 path(s)
   of this length
situated VB. to emptying VB., length = 8, 4 path(s) of this length
Roget thinks that situated means positioned
CORRECT
```

**Question 49**
```
principal | major | most | numerous | exceptional
principal N. [principal] to major N. [major], length = 0, 7 path(s) of
   this length
principal N. [principal] to most ADJ. [most], length = 6, 2 path(s) of
   this length
principal N. [principal] to numerous ADJ. [numerous], length = 14, 4
   path(s) of this length
principal N. to exceptional ADJ., length = 6, 1 path(s) of this length
Roget thinks that principal means major
CORRECT
```

**Question 50**
```
slowly | gradually | rarely | effectively | continuously
slowly ADV. [slowly] to gradually ADV. [gradually], length = 4, 1 path(s)
   of this length
slowly VB. [slowly] to rarely ADV. [rarely], length = 12, 4 path(s) of
   this length
effectively is NOT IN THE INDEX
slowly VB. to continuously ADV., length = 10, 8 path(s) of this length
Roget thinks that slowly means gradually
CORRECT
```

**Question 51**
```
built | constructed | proposed | financed | organized
built VB. [built] to constructed VB. [constructed], length = 0, 5 path(s)
   of this length
built ADJ. [built] to proposed VB. [proposed], length = 10, 22 path(s) of
   this length
built VB. [built] to financed VB. [financed], length = 10, 1 path(s) of
   this length
built VB. to organized VB., length = 0, 2 path(s) of this length
Roget thinks that built means constructed
TIE BROKEN
CORRECT
```

**Question 52**
```
tasks | jobs | customers | materials | shops
tasks N. [tasks] to jobs N. [jobs], length = 0, 9 path(s) of this length
tasks N. [tasks] to customers N. [customers], length = 10, 21 path(s) of
   this length
tasks N. [tasks] to materials N. [materials], length = 10, 60 path(s) of
   this length
tasks N. to shops VB., length = 6, 2 path(s) of this length
Roget thinks that tasks means jobs
CORRECT
```

**Question 53**
```
unlikely | improbable | disagreeable | different | unpopular
unlikely ADJ. [unlikely] to improbable ADJ. [improbable], length = 0, 1
   path(s) of this length
unlikely VB. [unlikely] to disagreeable ADJ. [disagreeable], length = 16,
```

```
    30 path(s) of this length
unlikely VB. [unlikely] to different N. [different], length = 12, 5
    path(s) of this length
unlikely VB. to unpopular ADJ., length = 16, 48 path(s) of this length
Roget thinks that unlikely means improbable
CORRECT
```

**Question 54**
```
halfheartedly | apathetically | customarily | bipartisanly |
    unconventionally
halfheartedly (PROBLEM) not found in the index!!
```

**Question 55**
```
annals | chronicles | homes | trails | songs
annals N. [annals] to chronicles N. [chronicles], length = 0, 3 path(s)
    of this length
annals N. [annals] to homes VB. [homes], length = 12, 36 path(s) of this length
annals N. [annals] to trails N. [trails], length = 4, 1 path(s) of this length
annals N. to songs N., length = 12, 6 path(s) of this length
Roget thinks that annals means chronicles
CORRECT
```

**Question 56**
```
wildly | furiously | distinctively | mysteriously | abruptly
wildly VB. [wildly] to furiously ADV. [furiously], length = 16, 3 path(s)
    of this length
distinctively is NOT IN THE INDEX
mysteriously is NOT IN THE INDEX
wildly VB. to abruptly ADV., length = 12, 1 path(s) of this length
Roget thinks that wildly means abruptly
INCORRECT
```

**Question 57**
```
hailed | acclaimed | judged | remembered | addressed
hailed VB. [hailed] to acclaimed VB. [acclaimed], length = 0, 2 path(s)
    of this length
hailed N. [hailed] to judged VB. [judged], length = 12, 62 path(s) of
    this length
hailed N. [hailed] to remembered VB. [remembered], length = 12, 46
    path(s) of this length
hailed N. to addressed N., length = 0, 2 path(s) of this length
Roget thinks that hailed means acclaimed
CORRECT
```

**Question 58**
```
command | mastery | observation | love | awareness
command N. [command] to mastery N. [mastery], length = 2, 1 path(s) of
    this length
command VB. [command] to observation N. [observation], length = 6, 2
    path(s) of this length
command N. [command] to love VB. [love], length = 2, 7 path(s) of this length
command VB. to awareness N., length = 12, 39 path(s) of this length
Roget thinks that command means love
TIE LOST
INCORRECT
```

**Question 59**
```
concocted | devised | cleaned | requested | supervised
concocted VB. [concocted] to devised VB. [devised], length = 0, 3 path(s)
    of this length
concocted VB. [concocted] to cleaned VB. [cleaned], length = 8, 2 path(s)
    of this length
concocted VB. [concocted] to requested VB. [requested], length = 10, 76
    path(s) of this length
concocted VB. to supervised VB., length = 14, 2 path(s) of this length
Roget thinks that concocted means devised
CORRECT
```

**Question 60**
```
prospective | potential | particular | prudent | prominent
prospective ADJ. [prospective] to potential ADJ. [potential], length = 2,
    1 path(s) of this length
prospective N. [prospective] to particular N. [particular], length = 10,
    14 path(s) of this length
prospective ADJ. [prospective] to prudent ADJ. [prudent], length = 2, 1
    path(s) of this length
prospective ADJ. to prominent ADJ., length = 12, 3 path(s) of this length
```

Roget thinks that prospective means potential
CORRECT

**Question 61**
generally | broadly | descriptively | controversially | accurately
broadly (ANSWER) is NOT IN THE INDEX
descriptively is NOT IN THE INDEX
controversially is NOT IN THE INDEX
generally ADV. to accurately ADV., length = 16, 5 path(s) of this length
Roget thinks that generally means accurately
INCORRECT

**Question 62**
sustained | prolonged | refined | lowered | analyzed
sustained VB. [sustained] to prolonged VB. [prolonged], length = 0, 3
   path(s) of this length
sustained VB. [sustained] to refined N. [refined], length = 6, 1 path(s)
   of this length
sustained VB. [sustained] to lowered VB. [lowered], length = 10, 46
   path(s) of this length
analyzed is NOT IN THE INDEX
Roget thinks that sustained means prolonged
CORRECT

**Question 63**
perilous | dangerous | binding | exciting | offensive
perilous ADJ. [perilous] to dangerous ADJ. [dangerous], length = 0, 1
   path(s) of this length
perilous ADJ. [perilous] to binding VB. [binding], length = 10, 2 path(s)
   of this length
perilous ADJ. [perilous] to exciting VB. [exciting], length = 14, 3
   path(s) of this length
perilous ADJ. to offensive ADJ., length = 10, 1 path(s) of this length
Roget thinks that perilous means dangerous
CORRECT

**Question 64**
tranquillity | peacefulness | harshness | weariness | happiness
tranquillity N. [tranquillity] to peacefulness N. [peacefulness], length
   = 0, 1 path(s) of this length
tranquillity N. [tranquillity] to harshness N. [harshness], length = 8, 1
   path(s) of this length
tranquillity N. [tranquillity] to weariness N. [weariness], length = 8, 1
   path(s) of this length
tranquillity N. to happiness N., length = 10, 8 path(s) of this length
Roget thinks that tranquillity means peacefulness
CORRECT

**Question 65**
dissipate | disperse | isolate | disguise | photograph
dissipate VB. [dissipate] to disperse VB. [disperse], length = 0, 5
   path(s) of this length
dissipate VB. [dissipate] to isolate VB. [isolate], length = 10, 1
   path(s) of this length
dissipate VB. [dissipate] to disguise N. [disguise], length = 10, 1
   path(s) of this length
dissipate VB. to photograph N., length = 14, 4 path(s) of this length
Roget thinks that dissipate means disperse
CORRECT

**Question 66**
primarily | chiefly | occasionally | cautiously | consistently
chiefly (ANSWER) is NOT IN THE INDEX
primarily ADV. [primarily] to occasionally ADV. [occasionally], length =
   10, 1 path(s) of this length
primarily ADV. [primarily] to cautiously ADV. [cautiously], length = 16,
   3 path(s) of this length
primarily ADV. to consistently ADJ., length = 12, 1 path(s) of this length
Roget thinks that primarily means occasionally
INCORRECT

**Question 67**
colloquial | conversational | recorded | misunderstood | incorrect
colloquial ADJ. [colloquial] to conversational ADJ. [conversational],
   length = 12, 8 path(s) of this length
colloquial ADJ. [colloquial] to recorded VB. [recorded], length = 12, 128
   path(s) of this length

```
colloquial ADJ. [colloquial] to misunderstood VB. [misunderstood], length
   = 12, 16 path(s) of this length
colloquial ADJ. to incorrect ADJ., length = 12, 12 path(s) of this length
Roget thinks that colloquial means recorded
INCORRECT
```

**Question 68**
```
resolved | settled | publicized | forgotten | examined
resolved N. [resolved] to settled N. [settled], length = 0, 3 path(s) of
   this length
resolved VB. [resolved] to publicized VB. [publicized], length = 12, 6
   path(s) of this length
resolved VB. [resolved] to forgotten N. [forgotten], length = 10, 24
   path(s) of this length
resolved VB. to examined VB., length = 12, 5 path(s) of this length
Roget thinks that resolved means settled
CORRECT
```

**Question 69**
```
feasible | possible | permitted | equitable | evident
feasible ADJ. [feasible] to possible ADJ. [possible], length = 0, 3
   path(s) of this length
feasible ADJ. [feasible] to permitted ADJ. [permitted], length = 2, 1
   path(s) of this length
feasible ADJ. [feasible] to equitable ADJ. [equitable], length = 16, 9
   path(s) of this length
feasible N. to evident ADJ., length = 12, 6 path(s) of this length
Roget thinks that feasible means possible
CORRECT
```

**Question 70**
```
expeditiously | rapidly | frequently | actually | repeatedly
expeditiously (PROBLEM) not found in the index!!
```

**Question 71**
```
percentage | proportion | volume | sample | profit
percentage N. [percentage] to proportion N. [proportion], length = 2, 2
   path(s) of this length
percentage N. [percentage] to volume N. [volume], length = 4, 1 path(s)
   of this length
percentage N. [percentage] to sample N. [sample], length = 2, 1 path(s)
   of this length
percentage N. to profit N., length = 2, 1 path(s) of this length
Roget thinks that percentage means proportion
TIE BROKEN
CORRECT
```

**Question 72**
```
terminated | ended | posed | postponed | evaluated
terminated VB. [terminated] to ended VB. [ended], length = 0, 6 path(s)
   of this length
terminated VB. [terminated] to posed VB. [posed], length = 14, 15 path(s)
   of this length
terminated VB. [terminated] to postponed VB. [postponed], length = 12, 1
   path(s) of this length
terminated ADJ. to evaluated VB., length = 12, 2 path(s) of this length
Roget thinks that terminated means ended
CORRECT
```

**Question 73**
```
uniform | alike | hard | complex | sharp
uniform ADJ. [uniform] to alike ADJ. [alike], length = 2, 1 path(s) of
   this length
uniform N. [uniform] to hard N. [hard], length = 4, 2 path(s) of this
   length
uniform ADJ. [uniform] to complex N. [complex], length = 6, 5 path(s) of
   this length
uniform ADJ. to sharp N., length = 6, 3 path(s) of this length
Roget thinks that uniform means alike
CORRECT
```

**Question 74**
```
figure | solve | list | divide | express
figure VB. [figure] to solve VB. [solve], length = 12, 10 path(s) of this length
figure N. [figure] to list N. [list], length = 2, 4 path(s) of this length
figure VB. [figure] to divide VB. [divide], length = 2, 1 path(s) of this length
figure N. to express VB., length = 2, 2 path(s) of this length
```

Roget thinks that figure means list
INCORRECT

**Question 75**
sufficient | enough | recent | physiological | valuable
sufficient ADJ. [sufficient] to enough ADJ. [enough], length = 2, 1
  path(s) of this length
sufficient N. [sufficient] to recent ADJ. [recent], length = 14, 15
  path(s) of this length
sufficient ADJ. [sufficient] to physiological ADJ. [physiological],
  length = 16, 8 path(s) of this length
sufficient ADJ. to valuable ADJ., length = 4, 2 path(s) of this length
Roget thinks that sufficient means enough
CORRECT

**Question 76**
fashion | manner | ration | fathom | craze
fashion N. [fashion] to manner N. [manner], length = 0, 5 path(s) of this length
fashion N. [fashion] to ration VB. [ration], length = 10, 11 path(s) of
  this length
fashion N. [fashion] to fathom VB. [fathom], length = 12, 22 path(s) of
  this length
fashion N. to craze N., length = 0, 3 path(s) of this length
Roget thinks that fashion means manner
TIE BROKEN
CORRECT

**Question 77**
marketed | sold | frozen | sweetened | diluted
marketed N. [marketed] to sold N. [sold], length = 0, 5 path(s) of this length
marketed N. [marketed] to frozen N. [frozen], length = 10, 108 path(s) of
  this length
marketed N. [marketed] to sweetened VB. [sweetened], length = 10, 10
  path(s) of this length
marketed N. to diluted ADJ., length = 12, 5 path(s) of this length
Roget thinks that marketed means sold
CORRECT

**Question 78**
bigger | larger | steadier | closer | better
bigger ADJ. [bigger] to larger ADJ. [larger], length = 0, 4 path(s) of
  this length
bigger N. [bigger] to steadier ADJ. [steadier], length = 8, 1 path(s) of
  this length
bigger ADJ. [bigger] to closer VB. [closer], length = 8, 7 path(s) of
  this length
bigger N. to better VB., length = 2, 1 path(s) of this length
Roget thinks that bigger means larger
CORRECT

**Question 79**
roots | origins | rituals | cure | function
roots N. [roots] to origins N. [origins], length = 0, 3 path(s) of this length
roots VB. [roots] to rituals N. [rituals], length = 6, 1 path(s) of this length
roots VB. [roots] to cure VB. [cure], length = 8, 1 path(s) of this length
roots N. to function N., length = 4, 1 path(s) of this length
Roget thinks that roots means origins
CORRECT

**Question 80**
normally | ordinarily | haltingly | permanently | periodically
normally (PROBLEM) not found in the index!!

**Final score: 63/80. 9 ties broken, 2 ties lost.**

Question word not in index: 4 times.
Answer word not in index: 5 times.
Other word not in index: 17 times.

The following question words were not found in Roget: [deftly, halfheartedly, expeditiously,
normally]

The following answer words were not found in Roget: [succinctly, uniquely, orally, broadly,
chiefly]

Other words that were not found in Roget: [uniquely, linearly, typically, tentatively,
chemically, patriotically, suspiciously, chronologically, overtly, verbosely, ordinarily,

effectively, distinctively, mysteriously, descriptively, controversially, analyzed]

# 1.B. 50 *ESL* Questions

**Question 1**
rusty | corroded | black | dirty | painted
rusty ADJ. [rusty] to corroded VB. [corroded], length = 6, 1 path(s) of
    this length
rusty ADJ. [rusty] to black N. [black], length = 6, 9 path(s) of this length
rusty ADJ. [rusty] to dirty ADJ. [dirty], length = 0, 1 path(s) of this length
rusty ADJ. to painted ADJ., length = 2, 1 path(s) of this length
Roget thinks that rusty means dirty
INCORRECT

**Question 2**
brass | metal | wood | stone | plastic
brass N. [brass] to metal N. [metal], length = 0, 5 path(s) of this length
brass N. [brass] to wood N. [wood], length = 4, 1 path(s) of this length
brass N. [brass] to stone N. [stone], length = 2, 5 path(s) of this length
brass N. to plastic N., length = 4, 2 path(s) of this length
Roget thinks that brass means metal
CORRECT

**Question 3**
spin | twirl | ache | sweat | flush
spin VB. [spin] to twirl VB. [twirl], length = 0, 2 path(s) of this length
spin N. [spin] to ache VB. [ache], length = 12, 1 path(s) of this length
spin N. [spin] to sweat N. [sweat], length = 0, 1 path(s) of this length
spin VB. to flush VB., length = 4, 1 path(s) of this length
Roget thinks that spin means twirl
TIE BROKEN
CORRECT

**Question 4**
passage | hallway | ticket | entrance | room
passage N. [passage] to hallway N. [hallway], length = 2, 2 path(s) of
    this length
passage N. [passage] to ticket N. [ticket], length = 4, 2 path(s) of this length
passage N. [passage] to entrance N. [entrance], length = 2, 5 path(s) of
    this length
passage N. to room N., length = 2, 3 path(s) of this length
Roget thinks that passage means entrance
INCORRECT

**Question 5**
yield | submit | challenge | boast | scorn
yield VB. [yield] to submit VB. [submit], length = 0, 9 path(s) of this length
yield VB. [yield] to challenge VB. [challenge], length = 4, 2 path(s) of
    this length
yield VB. [yield] to boast VB. [boast], length = 8, 1 path(s) of this length
yield VB. to scorn VB., length = 10, 11 path(s) of this length
Roget thinks that yield means submit
CORRECT

**Question 6**
lean | rest | scrape | grate | refer
lean VB. [lean] to rest VB. [rest], length = 0, 2 path(s) of this length
lean VB. [lean] to scrape VB. [scrape], length = 2, 1 path(s) of this length
lean N. [lean] to grate VB. [grate], length = 6, 2 path(s) of this length
lean VB. to refer VB., length = 12, 15 path(s) of this length
Roget thinks that lean means rest
CORRECT

**Question 7**
barrel | cask | bottle | box | case
barrel N. [barrel] to cask N. [cask], length = 0, 2 path(s) of this length
barrel N. [barrel] to bottle N. [bottle], length = 4, 3 path(s) of this length
barrel N. [barrel] to box N. [box], length = 0, 9 path(s) of this length
barrel N. to case N., length = 4, 4 path(s) of this length
Roget thinks that barrel means box
TIE LOST
INCORRECT

**Question 8**
nuisance | pest | garbage | relief | troublesome
nuisance N. [nuisance] to pest N. [pest], length = 0, 4 path(s) of this length
nuisance N. [nuisance] to garbage N. [garbage], length = 14, 4 path(s) of

```
     this length
nuisance N. [nuisance] to relief N. [relief], length = 8, 2 path(s) of
     this length
nuisance N. to troublesome ADJ., length = 6, 1 path(s) of this length
Roget thinks that nuisance means pest
CORRECT
```

**Question 9**
```
rug | carpet | sofa | ottoman | hallway
rug N. [rug] to carpet N. [carpet], length = 2, 2 path(s) of this length
rug N. [rug] to sofa N. [sofa], length = 12, 2 path(s) of this length
rug N. [rug] to ottoman N. [ottoman], length = 12, 2 path(s) of this length
rug N. to hallway N., length = 16, 2 path(s) of this length
Roget thinks that rug means carpet
CORRECT
```

**Question 10**
```
tap | drain | boil | knock | rap
tap VB. [tap] to drain VB. [drain], length = 0, 8 path(s) of this length
tap N. [tap] to boil VB. [boil], length = 6, 1 path(s) of this length
tap N. [tap] to knock N. [knock], length = 0, 10 path(s) of this length
tap N. to rap N., length = 0, 5 path(s) of this length
Roget thinks that tap means knock
TIE LOST
INCORRECT
```

**Question 11**
```
split | divided | crushed | grated | bruised
split VB. [split] to divided VB. [divided], length = 2, 8 path(s) of this
     length
split VB. [split] to crushed VB. [crushed], length = 4, 1 path(s) of this
     length
split N. [split] to grated VB. [grated], length = 6, 1 path(s) of this length
split VB. to bruised VB., length = 10, 7 path(s) of this length
Roget thinks that split means divided
CORRECT
```

**Question 12**
```
lump | chunk | stem | trunk | limb
lump N. [lump] to chunk N. [chunk], length = 0, 3 path(s) of this length
lump N. [lump] to stem N. [stem], length = 2, 1 path(s) of this length
lump N. [lump] to trunk N. [trunk], length = 2, 2 path(s) of this length
lump N. to limb N., length = 2, 3 path(s) of this length
Roget thinks that lump means chunk
CORRECT
```

**Question 13**
```
outline | contour | pair | blend | block
outline N. [outline] to contour N. [contour], length = 0, 7 path(s) of
     this length
outline N. [outline] to pair N. [pair], length = 10, 41 path(s) of this length
outline N. [outline] to blend VB. [blend], length = 10, 8 path(s) of this
     length
outline N. to block VB., length = 2, 8 path(s) of this length
Roget thinks that outline means contour
CORRECT
```

**Question 14**
```
swear | vow | explain | think | describe
swear VB. [swear] to vow VB. [vow], length = 0, 4 path(s) of this length
swear VB. [swear] to explain VB. [explain], length = 8, 1 path(s) of this
     length
swear VB. [swear] to think VB. [think], length = 4, 3 path(s) of this length
swear VB. to describe VB., length = 12, 77 path(s) of this length
Roget thinks that swear means vow
CORRECT
```

**Question 15**
```
relieved | rested | sleepy | tired | hasty
relieved VB. [relieved] to rested VB. [rested], length = 0, 6 path(s) of
     this length
relieved VB. [relieved] to sleepy ADJ. [sleepy], length = 8, 1 path(s) of
     this length
relieved VB. [relieved] to tired VB. [tired], length = 8, 3 path(s) of
     this length
relieved VB. to hasty N., length = 2, 1 path(s) of this length
```

```
Roget thinks that relieved means rested
CORRECT
```

**Question 16**
```
deserve | merit | need | want | expect
deserve VB. [deserve] to merit VB. [merit], length = 0, 2 path(s) of this
    length
deserve VB. [deserve] to need VB. [need], length = 10, 16 path(s) of this
    length
deserve VB. [deserve] to want VB. [want], length = 10, 23 path(s) of this
    length
deserve VB. to expect VB., length = 4, 1 path(s) of this length
Roget thinks that deserve means merit
CORRECT
```

**Question 17**
```
haste | a hurry | anger | ear | spite
haste N. [haste] to hurry N. [a hurry], length = 0, 6 path(s) of this length
haste N. [haste] to anger N. [anger], length = 4, 1 path(s) of this length
haste N. [haste] to ear N. [ear], length = 12, 10 path(s) of this length
haste N. to spite VB., length = 12, 5 path(s) of this length
Roget thinks that haste means a hurry
CORRECT
```

**Question 18**
```
stiff | firm | dark | drunk | cooked
stiff N. [stiff] to firm N. [firm], length = 2, 4 path(s) of this length
stiff ADJ. [stiff] to dark ADJ. [dark], length = 2, 1 path(s) of this length
stiff ADJ. [stiff] to drunk ADJ. [drunk], length = 2, 3 path(s) of this length
stiff ADJ. to cooked VB., length = 12, 34 path(s) of this length
Roget thinks that stiff means firm
TIE BROKEN
CORRECT
```

**Question 19**
```
verse | section | weed | twig | branch
verse N. [verse] to section N. [section], length = 2, 4 path(s) of this length
verse N. [verse] to weed VB. [weed], length = 10, 15 path(s) of this length
verse N. [verse] to twig N. [twig], length = 4, 1 path(s) of this length
verse N. to branch N., length = 4, 2 path(s) of this length
Roget thinks that verse means section
CORRECT
```

**Question 20**
```
steep | sheer | bare | rugged | stone
steep ADJ. [steep] to sheer ADJ. [sheer], length = 0, 3 path(s) of this length
steep VB. [steep] to bare ADJ. [bare], length = 8, 1 path(s) of this length
steep ADJ. [steep] to rugged ADJ. [rugged], length = 2, 1 path(s) of this length
steep VB. to stone N., length = 8, 1 path(s) of this length
Roget thinks that steep means sheer
CORRECT
```

**Question 21**
```
envious | jealous | enthusiastic | hurt | relieved
envious ADJ. [envious] to jealous ADJ. [jealous], length = 0, 7 path(s)
    of this length
envious ADJ. [envious] to enthusiastic ADJ. [enthusiastic], length = 12,
    1 path(s) of this length
envious ADJ. [envious] to hurt ADJ. [hurt], length = 2, 1 path(s) of this
    length
envious ADJ. to relieved VB., length = 8, 2 path(s) of this length
Roget thinks that envious means jealous
CORRECT
```

**Question 22**
```
paste | dough | syrup | block | jelly
paste N. [paste] to dough N. [dough], length = 0, 2 path(s) of this length
paste N. [paste] to syrup N. [syrup], length = 2, 1 path(s) of this length
paste N. [paste] to block N. [block], length = 8, 1 path(s) of this length
paste N. to jelly N., length = 2, 1 path(s) of this length
Roget thinks that paste means dough
CORRECT
```

**Question 23**
```
scorn | refuse | enjoy | avoid | plan
scorn VB. [scorn] to refuse VB. [refuse], length = 2, 1 path(s) of this length
scorn VB. [scorn] to enjoy VB. [enjoy], length = 8, 1 path(s) of this length
```

```
scorn N. [scorn] to avoid VB. [avoid], length = 10, 45 path(s) of this length
scorn VB. to plan VB., length = 10, 24 path(s) of this length
Roget thinks that scorn means refuse
CORRECT
```

**Question 24**
```
refer | direct | call | carry | explain
refer VB. [refer] to direct VB. [direct], length = 2, 1 path(s) of this length
refer VB. [refer] to call VB. [call], length = 2, 2 path(s) of this length
refer VB. [refer] to carry VB. [carry], length = 10, 39 path(s) of this length
refer VB. to explain VB., length = 4, 1 path(s) of this length
Roget thinks that refer means call
TIE LOST
INCORRECT
```

**Question 25**
```
limb | branch | bark | trunk | twig
limb N. [limb] to branch N. [branch], length = 0, 8 path(s) of this length
limb N. [limb] to bark N. [bark], length = 10, 10 path(s) of this length
limb N. [limb] to trunk N. [trunk], length = 2, 5 path(s) of this length
limb N. to twig N., length = 0, 7 path(s) of this length
Roget thinks that limb means branch
TIE BROKEN
CORRECT
```

**Question 26**
```
pad | cushion | board | block | tablet
pad VB. [pad] to cushion VB. [cushion], length = 0, 5 path(s) of this length
pad N. [pad] to board N. [board], length = 2, 5 path(s) of this length
pad VB. [pad] to block VB. [block], length = 2, 7 path(s) of this length
pad N. to tablet N., length = 2, 2 path(s) of this length
Roget thinks that pad means cushion
CORRECT
```

**Question 27**
```
boast | brag | yell | complain | explain
boast VB. [boast] to brag VB. [brag], length = 0, 9 path(s) of this length
boast VB. [boast] to yell N. [yell], length = 12, 22 path(s) of this length
boast VB. [boast] to complain VB. [complain], length = 8, 1 path(s) of
    this length
boast VB. to explain VB., length = 10, 24 path(s) of this length
Roget thinks that boast means brag
CORRECT
```

**Question 28**
```
applause | approval | fear | shame | friends
applause N. [applause] to approval N. [approval], length = 0, 2 path(s)
    of this length
applause N. [applause] to fear VB. [fear], length = 6, 1 path(s) of this length
applause N. [applause] to shame VB. [shame], length = 8, 4 path(s) of
    this length
applause N. to friends N., length = 8, 1 path(s) of this length
Roget thinks that applause means approval
CORRECT
```

**Question 29**
```
sheet | leaf | book | block | tap
sheet N. [sheet] to leaf N. [leaf], length = 0, 5 path(s) of this length
sheet N. [sheet] to book N. [book], length = 0, 14 path(s) of this length
sheet N. [sheet] to block N. [block], length = 4, 2 path(s) of this length
sheet VB. to tap VB., length = 4, 1 path(s) of this length
Roget thinks that sheet means book
TIE LOST
INCORRECT
```

**Question 30**
```
stem | stalk | bark | column | trunk
stem N. [stem] to stalk N. [stalk], length = 0, 4 path(s) of this length
stem VB. [stem] to bark N. [bark], length = 10, 14 path(s) of this length
stem N. [stem] to column N. [column], length = 2, 2 path(s) of this length
stem N. to trunk N., length = 0, 4 path(s) of this length
Roget thinks that stem means stalk
CORRECT
```

**Question 31**
```
seize | take | refer | request | yield
seize VB. [seize] to take VB. [take], length = 0, 8 path(s) of this length
```

```
seize VB. [seize] to refer VB. [refer], length = 12, 16 path(s) of this length
seize VB. [seize] to request N. [request], length = 6, 1 path(s) of this length
seize VB. to yield VB., length = 10, 18 path(s) of this length
Roget thinks that seize means take
CORRECT
```

**Question 32**
```
trunk | chest | bag | closet | swing
trunk N. [trunk] to chest N. [chest], length = 2, 1 path(s) of this length
trunk N. [trunk] to bag N. [bag], length = 4, 2 path(s) of this length
trunk N. [trunk] to closet N. [closet], length = 4, 2 path(s) of this length
trunk N. to swing N., length = 4, 1 path(s) of this length
Roget thinks that trunk means chest
CORRECT
```

**Question 33**
```
weed | unwanted plant | cloth | animal | vegetable
weed N. [weed] to plant N. [unwanted plant], length = 0, 4 path(s) of
    this length
weed VB. [weed] to cloth N. [cloth], length = 6, 1 path(s) of this length
weed N. [weed] to animal N. [animal], length = 0, 1 path(s) of this length
weed N. to vegetable N., length = 0, 4 path(s) of this length
Roget thinks that weed means unwanted plant
TIE BROKEN
CORRECT
```

**Question 34**
```
approval | endorsement | gift | statement | confession
approval N. [approval] to endorsement N. [endorsement], length = 2, 1
    path(s) of this length
approval ADJ. [approval] to gift N. [gift], length = 10, 30 path(s) of
    this length
approval ADJ. [approval] to statement N. [statement], length = 10, 30
    path(s) of this length
approval N. to confession N., length = 2, 1 path(s) of this length
Roget thinks that approval means endorsement
CORRECT
```

**Question 35**
```
mass | lump | service | worship | element
mass N. [mass] to lump N. [lump], length = 0, 11 path(s) of this length
mass N. [mass] to service N. [service], length = 0, 14 path(s) of this length
mass N. [mass] to worship N. [worship], length = 2, 3 path(s) of this length
mass N. to element N., length = 4, 2 path(s) of this length
Roget thinks that mass means service
TIE LOST
INCORRECT
```

**Question 36**
```
swing | sway | bounce | break | crash
swing VB. [swing] to sway VB. [sway], length = 0, 5 path(s) of this length
swing VB. [swing] to bounce VB. [bounce], length = 2, 2 path(s) of this length
swing VB. [swing] to break VB. [break], length = 2, 2 path(s) of this length
swing N. to crash N., length = 2, 4 path(s) of this length
Roget thinks that swing means sway
CORRECT
```

**Question 37**
```
sore | painful | red | hot | rough
sore ADJ. [sore] to painful ADJ. [painful], length = 0, 5 path(s) of this
    length
sore VB. [sore] to red VB. [red], length = 0, 3 path(s) of this length
sore N. [sore] to hot N. [hot], length = 0, 3 path(s) of this length
sore ADJ. to rough ADJ., length = 2, 3 path(s) of this length
Roget thinks that sore means painful
TIE BROKEN
CORRECT
```

**Question 38**
```
hinder | block | assist | relieve | yield
hinder VB. [hinder] to block VB. [block], length = 2, 2 path(s) of this length
hinder VB. [hinder] to assist VB. [assist], length = 8, 1 path(s) of this length
hinder VB. [hinder] to relieve VB. [relieve], length = 8, 3 path(s) of
    this length
hinder VB. to yield VB., length = 10, 40 path(s) of this length
Roget thinks that hinder means block
CORRECT
```

**Question 39**
sticky | gooey | smooth | shiny | wet
sticky ADJ. [sticky] to gooey ADJ. [gooey], length = 0, 2 path(s) of this length
sticky N. [sticky] to smooth N. [smooth], length = 2, 1 path(s) of this length
sticky ADJ. [sticky] to shiny ADJ. [shiny], length = 14, 2 path(s) of
   this length
sticky ADJ. to wet VB., length = 10, 34 path(s) of this length
Roget thinks that sticky means gooey
CORRECT

**Question 40**
confession | statement | service | plea | bargain
confession N. [confession] to statement N. [statement], length = 0, 3
   path(s) of this length
confession N. [confession] to service N. [service], length = 4, 4 path(s)
   of this length
confession N. [confession] to plea N. [plea], length = 2, 1 path(s) of
   this length
confession N. to bargain N., length = 4, 1 path(s) of this length
Roget thinks that confession means statement
CORRECT

**Question 41**
weave | intertwine | print | stamp | shake
weave VB. [weave] to intertwine VB. [intertwine], length = 0, 5 path(s)
   of this length
weave VB. [weave] to print VB. [print], length = 2, 2 path(s) of this length
weave VB. [weave] to stamp VB. [stamp], length = 8, 1 path(s) of this length
weave VB. to shake VB., length = 2, 1 path(s) of this length
Roget thinks that weave means intertwine
CORRECT

**Question 42**
saucer | dish | box | frisbee | can
saucer N. [saucer] to dish N. [dish], length = 2, 4 path(s) of this length
saucer N. [saucer] to box N. [box], length = 4, 8 path(s) of this length
frisbee is NOT IN THE INDEX
saucer N. to can N., length = 4, 3 path(s) of this length
Roget thinks that saucer means dish
CORRECT

**Question 43**
substance | thing | posture | level | score
substance N. [substance] to thing N. [thing], length = 2, 13 path(s) of
   this length
substance N. [substance] to posture N. [posture], length = 2, 1 path(s)
   of this length
substance N. [substance] to level N. [level], length = 8, 2 path(s) of
   this length
substance N. to score N., length = 8, 1 path(s) of this length
Roget thinks that substance means thing
TIE BROKEN
CORRECT

**Question 44**
firmly | steadfastly | reluctantly | sadly | hopefully
steadfastly (ANSWER) is NOT IN THE INDEX
firmly VB. [firmly] to reluctantly ADV. [reluctantly], length = 16, 6
   path(s) of this length
firmly ADV. [firmly] to sadly ADV. [sadly], length = 12, 1 path(s) of
   this length
firmly VB. to hopefully ADV., length = 16, 3 path(s) of this length
Roget thinks that firmly means sadly
INCORRECT

**Question 45**
twist | intertwine | clip | fasten | curl
twist VB. [twist] to intertwine VB. [intertwine], length = 0, 2 path(s)
   of this length
twist N. [twist] to clip N. [clip], length = 4, 2 path(s) of this length
twist VB. [twist] to fasten VB. [fasten], length = 4, 3 path(s) of this length
twist N. to curl N., length = 2, 4 path(s) of this length
Roget thinks that twist means intertwine
CORRECT

**Question 46**

```
scrape | grate | chop | mince | slice
scrape VB. [scrape] to grate VB. [grate], length = 0, 4 path(s) of this length
scrape VB. [scrape] to chop VB. [chop], length = 8, 1 path(s) of this length
scrape VB. [scrape] to mince VB. [mince], length = 2, 1 path(s) of this length
scrape VB. to slice N., length = 10, 25 path(s) of this length
Roget thinks that scrape means grate
CORRECT
```

**Question 47**
```
grind | rub | slice | hit | tap
grind VB. [grind] to rub VB. [rub], length = 2, 6 path(s) of this length
grind VB. [grind] to slice VB. [slice], length = 4, 2 path(s) of this length
grind VB. [grind] to hit VB. [hit], length = 10, 37 path(s) of this length
grind VB. to tap ADJ., length = 10, 49 path(s) of this length
Roget thinks that grind means rub
CORRECT
```

**Question 48**
```
swell | enlarge | move | curl | shake
swell VB. [swell] to enlarge VB. [enlarge], length = 0, 4 path(s) of this length
swell N. [swell] to move VB. [move], length = 6, 1 path(s) of this length
swell VB. [swell] to curl VB. [curl], length = 10, 34 path(s) of this length
swell N. to shake N., length = 4, 1 path(s) of this length
Roget thinks that swell means enlarge
CORRECT
```

**Question 49**
```
harvest | intake | stem | lump | split
harvest N. [harvest] to intake N. [intake], length = 4, 1 path(s) of this length
harvest N. [harvest] to stem N. [stem], length = 8, 1 path(s) of this length
harvest N. [harvest] to lump N. [lump], length = 2, 1 path(s) of this length
harvest N. to split N., length = 4, 1 path(s) of this length
Roget thinks that harvest means lump
INCORRECT
```

**Question 50**
```
approve | support | boast | scorn | anger
approve VB. [approve] to support VB. [support], length = 2, 3 path(s) of
   this length
approve VB. [approve] to boast VB. [boast], length = 4, 1 path(s) of this length
approve VB. [approve] to scorn VB. [scorn], length = 2, 1 path(s) of this length
approve VB. to anger N., length = 8, 1 path(s) of this length
Roget thinks that approve means support
TIE BROKEN
CORRECT
```

**Final score: 41/50. 7 ties broken, 5 ties lost.**

```
Question word not in index: 0 times.
Answer word not in index: 1 times.
Other word not in index: 1 times.

The following question words were not found in Roget: []
The following answer words were not found in Roget: [steadfastly]
Other words that were not found in Roget: [frisbee]
```

# 1.C.  20 *RDWP* Questions – January 2000: Nature

**Question 1**
```
eddy | whirlpool | current | wave | wind
eddy N. [eddy] to whirlpool N. [whirlpool], length = 0, 4 path(s) of this length
eddy N. [eddy] to current N. [current], length = 2, 2 path(s) of this length
eddy N. [eddy] to wave N. [wave], length = 2, 3 path(s) of this length
eddy N. to wind N., length = 2, 5 path(s) of this length
Roget thinks that eddy means whirlpool
CORRECT
```

**Question 2**
```
bough | branch | barricade | shaded area | fallen tree
bough N. [bough] to branch N. [branch], length = 0, 5 path(s) of this length
bough N. [bough] to barricade N. [barricade], length = 12, 1 path(s) of this length
bough N. [bough] to shaded N. [shaded area], length = 10, 31 path(s) of this length
bough N. [bough] to tree N. [fallen tree], length = 2, 2 path(s) of this length
```

Roget thinks that bough means branch
CORRECT

**Question 3**
heath | overgrown open land | burned-over area | thin forest | pasture
heath N. [heath] to land N. [overgrown open land], length = 0, 2 path(s) of this length
heath N. [heath] to area N. [burned-over area], length = 4, 1 path(s) of this length
heath N. [heath] to forest N. [thin forest], length = 0, 2 path(s) of this length
heath N. to pasture N., length = 4, 2 path(s) of this length
Roget thinks that heath means overgrown open land
CORRECT

**Question 4**
scud | run straight | move slowly | falter | vaporize
scud VB. [scud] to run VB. [run straight], length = 2, 5 path(s) of this length
scud VB. [scud] to move slowly VB. [move slowly], length = 8, 1 path(s) of this length
scud VB. [scud] to falter VB. [falter], length = 8, 2 path(s) of this length
scud N. to vaporize VB., length = 6, 1 path(s) of this length
Roget thinks that scud means run straight
CORRECT

**Question 5**
williwaw | sudden windblast | rainsquall | songbird | meadow flower
williwaw N. [williwaw] to sudden N. [sudden windblast], length = 14, 2 path(s) of this length
rainsquall is NOT IN THE INDEX
williwaw N. [williwaw] to songbird N. [songbird], length = 14, 2 path(s) of this length
williwaw N. [williwaw] to flower N. [meadow flower], length = 4, 1 path(s) of this length
Roget thinks that williwaw means meadow flower
INCORRECT

**Question 6**
verge | brink | middle path | bare ground | vantage point
verge N. [verge] to brink N. [brink], length = 0, 3 path(s) of this length
verge N. [verge] to path N. [middle path], length = 4, 2 path(s) of this length
verge VB. [verge] to ground VB. [bare ground], length = 10, 77 path(s) of this length
verge N. to vantage point N., length = 16, 9 path(s) of this length
Roget thinks that verge means brink
CORRECT

**Question 7**
dale | valley | retreat | shelter | plain
dale N. [dale] to valley N. [valley], length = 2, 1 path(s) of this length
dale N. [dale] to retreat VB. [retreat], length = 16, 36 path(s) of this length
dale N. [dale] to shelter N. [shelter], length = 14, 2 path(s) of this length
dale N. to plain N., length = 2, 1 path(s) of this length
Roget thinks that dale means valley
CORRECT

**Question 8**
limpid | clear | still | flat | luminous
limpid ADJ. [limpid] to clear ADJ. [clear], length = 0, 3 path(s) of this length
limpid ADJ. [limpid] to still VB. [still], length = 10, 23 path(s) of this length
limpid ADJ. [limpid] to flat ADJ. [flat], length = 10, 31 path(s) of this length
limpid ADJ. to luminous ADJ., length = 2, 1 path(s) of this length
Roget thinks that limpid means clear
CORRECT

**Question 9**
floe | floating ice | frozen stream | lump | driftwood
floe N. [floe] to ice N. [floating ice], length = 0, 4 path(s) of this length
floe N. [floe] to frozen N. [frozen stream], length = 2, 2 path(s) of this length
floe N. [floe] to lump N. [lump], length = 12, 4 path(s) of this length
floe N. to driftwood N., length = 16, 6 path(s) of this length
Roget thinks that floe means floating ice
CORRECT

**Question 10**
cascade | waterfall | thunder | swift current | edge
cascade N. [cascade] to waterfall N. [waterfall], length = 0, 2 path(s) of this length
cascade VB. [cascade] to thunder VB. [thunder], length = 12, 2 path(s) of this length
cascade N. [cascade] to current N. [swift current], length = 4, 1 path(s) of this length
cascade VB. to edge VB., length = 10, 8 path(s) of this length
Roget thinks that cascade means waterfall
CORRECT

**Question 11**
undulation | rise and fall | faint motion | ebb and flow | quivering

undulation N. [undulation] to rise and fall VB. [rise and fall], length = 10, 1 path(s) of this
length
undulation N. [undulation] to motion N. [faint motion], length = 2, 2 path(s) of this length
undulation N. [undulation] to ebb and flow N. [ebb and flow], length = 4, 2 path(s) of this
length
undulation N. to quivering VB., length = 6, 1 path(s) of this length
Roget thinks that undulation means faint motion
INCORRECT

**Question 12**
crag | steep rock | headland | barren hill | niche
crag N. [crag] to steep N. [steep rock], length = 0, 1 path(s) of this length
crag N. [crag] to headland N. [headland], length = 2, 1 path(s) of this length
crag N. [crag] to hill N. [barren hill], length = 2, 2 path(s) of this length
crag N. to niche N., length = 10, 3 path(s) of this length
Roget thinks that crag means steep rock
CORRECT

**Question 13**
truss | cluster of flowers | main stem | bunch of grass | fallen petals
truss VB. [truss] to cluster VB. [cluster of flowers], length = 4, 3 path(s) of this length
truss N. [truss] to stem N. [main stem], length = 2, 2 path(s) of this length
truss VB. [truss] to bunch VB. [bunch of grass], length = 2, 5 path(s) of this length
truss VB. [truss] to fallen VB. [fallen petals], length = 8, 3 path(s) of this length
Roget thinks that truss means bunch of grass
INCORRECT

**Question 14**
slough | deep mire | quicksand | shower | erosion
slough N. [slough] to mire N. [deep mire], length = 0, 1 path(s) of this length
slough N. [slough] to quicksand N. [quicksand], length = 2, 1 path(s) of this length
slough VB. [slough] to shower N. [shower], length = 6, 5 path(s) of this length
slough N. to erosion N., length = 10, 16 path(s) of this length
Roget thinks that slough means deep mire
CORRECT

**Question 15**
lee | shelter | cove | grassland | riverbank
lee N. [lee] to shelter N. [shelter], length = 2, 2 path(s) of this length
lee N. [lee] to cove N. [cove], length = 14, 1 path(s) of this length
lee N. [lee] to grassland N. [grassland], length = 14, 1 path(s) of this length
riverbank is NOT IN THE INDEX
Roget thinks that lee means shelter
CORRECT

**Question 16**
brackish | salty | dirty | rough | noisy
brackish ADJ. [brackish] to salty ADJ. [salty], length = 0, 1 path(s) of this length
brackish ADJ. [brackish] to dirty VB. [dirty], length = 10, 6 path(s) of this length
brackish ADJ. [brackish] to rough ADJ. [rough], length = 4, 1 path(s) of this length
brackish ADJ. to noisy ADJ., length = 10, 1 path(s) of this length
Roget thinks that brackish means salty
CORRECT

**Question 17**
precipice | vertical rockface | wide gap | broken path | descent
precipice N. [precipice] to vertical N. [vertical rockface], length = 2, 2 path(s) of this length
precipice N. [precipice] to wide ADJ. [wide gap], length = 10, 22 path(s) of this length
precipice N. [precipice] to broken VB. [broken path], length = 8, 1 path(s) of this length
precipice N. to descent N., length = 2, 2 path(s) of this length
Roget thinks that precipice means descent
INCORRECT

**Question 18**
chasm | deep fissure | wide opening | mountain pass | series of falls
chasm N. [chasm] to fissure N. [deep fissure], length = 0, 2 path(s) of this length
chasm N. [chasm] to wide opening ADJ. [wide opening], length = 10, 3 path(s) of this length
chasm N. [chasm] to pass N. [mountain pass], length = 2, 1 path(s) of this length
chasm N. [chasm] to falls N. [series of falls], length = 2, 1 path(s) of this length
Roget thinks that chasm means deep fissure
CORRECT

**Question 19**
sediment | settles to the bottom | floats | holds together | covers rocks
sediment N. [sediment] to the N. [settles to the bottom], length = 10, 24 path(s) of this length
sediment N. [sediment] to floats VB. [floats], length = 10, 9 path(s) of this length
sediment N. [sediment] to together VB. [holds together], length = 6, 1 path(s) of this length

```
sediment N. [sediment] to rocks N. [covers rocks], length = 2, 1 path(s) of this length
Roget thinks that sediment means covers rocks
INCORRECT
```

**Question 20**
```
torrent | violent flow | drift | swell | deep sound
torrent N. [torrent] to violent N. [violent flow], length = 4, 1 path(s) of this length
torrent N. [torrent] to drift VB. [drift], length = 8, 1 path(s) of this length
torrent N. [torrent] to swell N. [swell], length = 4, 1 path(s) of this length
torrent N. [torrent] to sound ADJ. [deep sound], length = 6, 2 path(s) of this length
Roget thinks that torrent means violent flow
CORRECT
```

**Final score: 15/20. 0 ties broken, 0 ties lost.**

The answer was not in the index 2 times.

The question was not in the index 0 times.

```
-- NEW STATS --
Question word not in index: 0 times.
Answer word not in index: 0 times.
Other word not in index: 2 times.
```

The following question words were not found in Roget: []
The following answer words were not found in Roget: []
Other words that were not found in Roget: [rainsquall, riverbank]

# 2 Semantic Distance measured using the Hirst and St-Onge *WordNet*-based meaure

## 2.A. 80 *TOEFL* Questions

**Question 1**
```
enormously | tremendously | appropriately | uniquely | decidedly
enormously   tremendously  16
enormously   appropriately  0
enormously   uniquely  0
enormously   decidedly  0
WordNet thinks that the answer is tremendously
CORRECT
```

**Question 2**
```
provisions | stipulations | interrelations | jurisdictions | interpretations
provisions   stipulations  0
provisions   interrelations  0
provisions   jurisdictions  0
provisions   interpretations  0
WordNet thinks that the answer is stipulations
4 answers tied [score = 0.25]
```

**Question 3**
```
haphazardly | randomly | dangerously | densely | linearly
haphazardly   randomly  16
haphazardly   dangerously  0
haphazardly   densely  0
haphazardly   linearly  0
WordNet thinks that the answer is randomly
CORRECT
```

**Question 4**
```
prominent | conspicuous | battered | ancient | mysterious
prominent   conspicuous  16
prominent   battered  0
prominent   ancient  0
prominent   mysterious  0
WordNet thinks that the answer is conspicuous
CORRECT
```

**Question 5**
```
zenith | pinnacle | completion | outset | decline
zenith  pinnacle   2
zenith  completion  0
zenith  outset  0
zenith  decline  0
WordNet thinks that the answer is pinnacle
CORRECT
```

**Question 6**
```
flawed | imperfect | tiny | lustrous | crude
flawed  imperfect  0
flawed  tiny  0
flawed  lustrous  0
flawed  crude  0
WordNet thinks that the answer is imperfect
4 answers tied [score = 0.25]
```

**Question 7**
```
urgently | desperately | typically | conceivably | tentatively
urgently  desperately  16
urgently  typically  0
urgently  conceivably  0
urgently  tentatively  0
WordNet thinks that the answer is desperately
CORRECT
```

**Question 8**
```
consumed | eaten | bred | caught | supplied
consumed  eaten  16
consumed  bred  0
consumed  caught  0
consumed  supplied  3
WordNet thinks that the answer is eaten
CORRECT
```

**Question 9**
```
advent | coming | arrest | financing | stability
advent  coming  16
advent  arrest  0
advent  financing  0
advent  stability  0
WordNet thinks that the answer is coming
CORRECT
```

**Question 10**
```
concisely | succinctly | powerfully | positively | freely
concisely  succinctly  0
concisely  powerfully  0
concisely  positively  0
concisely  freely  0
WordNet thinks that the answer is succinctly
4 answers tied [score = 0.25]
```

**Question 11**
```
salutes | greetings | information | ceremonies | privileges
salutes  greetings  4
salutes  information  3
['ceremonies' not in WordNet.] salutes  ceremonies
salutes  privileges  0
WordNet thinks that the answer is greetings
CORRECT
```

**Question 12**
```
solitary | alone | alert | restless | fearless
solitary  alone  16
solitary  alert  0
solitary  restless  0
solitary  fearless  0
WordNet thinks that the answer is alone
CORRECT
```

**Question 13**
```
hasten | accelerate | permit | determine | accompany
hasten  accelerate  0
hasten  permit  0
hasten  determine  4
```

```
hasten  accompany  5
WordNet thinks that the answer is accompany
INCORRECT
```

**Question 14**
```
perseverance | endurance | skill | generosity | disturbance
perseverance  endurance  0
perseverance  skill  0
perseverance  generosity  0
perseverance  disturbance  4
WordNet thinks that the answer is disturbance
INCORRECT
```

**Question 15**
```
fanciful | imaginative | familiar | apparent | logical
fanciful  imaginative  4
fanciful  familiar  0
fanciful  apparent  0
fanciful  logical  0
WordNet thinks that the answer is imaginative
CORRECT
```

**Question 16**
```
showed | demonstrated | published | repeated | postponed
showed  demonstrated  16
showed  published  3
showed  repeated  4
showed  postponed  0
WordNet thinks that the answer is demonstrated
CORRECT
```

**Question 17**
```
constantly | continually | instantly | rapidly | accidentally
constantly  continually  0
constantly  instantly  0
constantly  rapidly  0
constantly  accidentally  0
WordNet thinks that the answer is continually
4 answers tied [score = 0.25]
```

**Question 18**
```
issues | subjects | training | salaries | benefits
issues  subjects  16
issues  training  4
['salaries' not in WordNet.] issues  salaries
issues  benefits  0
WordNet thinks that the answer is subjects
CORRECT
```

**Question 19**
```
furnish | supply | impress | protect | advise
furnish  supply  16
furnish  impress  0
furnish  protect  0
furnish  advise  0
WordNet thinks that the answer is supply
CORRECT
```

**Question 20**
```
costly | expensive | beautiful | popular | complicated
costly  expensive  16
costly  beautiful  0
costly  popular  0
costly  complicated  0
WordNet thinks that the answer is expensive
CORRECT
```

**Question 21**
```
recognized | acknowledged | successful | depicted | welcomed
recognized  acknowledged  16
recognized  successful  0
recognized  depicted  0
recognized  welcomed  4
WordNet thinks that the answer is acknowledged
CORRECT
```

**Question 22**

```
spot | location | climate | latitude | sea
spot  location  6
spot  climate   0
spot  latitude  3
spot  sea  3
WordNet thinks that the answer is location
CORRECT
```

**Question 23**
```
make | earn | print | trade | borrow
make  earn   16
make  print  6
make  trade  4
make  borrow  5
WordNet thinks that the answer is earn
CORRECT
```

**Question 24**
```
often | frequently | definitely | chemically | hardly
often  frequently  16
often  definitely  0
often  chemically  0
often  hardly  0
WordNet thinks that the answer is frequently
CORRECT
```

**Question 25**
```
easygoing | relaxed | frontier | boring | farming
easygoing  relaxed   0
easygoing  frontier  0
easygoing  boring    0
easygoing  farming   0
WordNet thinks that the answer is relaxed
4 answers tied [score = 0.25]
```

**Question 26**
```
debate | argument | war | election | competition
debate  argument     16
debate  war          0
debate  election     0
debate  competition  0
WordNet thinks that the answer is argument
CORRECT
```

**Question 27**
```
narrow | thin | clear | freezing | poisonous
narrow  thin       16
narrow  clear      0
narrow  freezing   0
narrow  poisonous  0
WordNet thinks that the answer is thin
CORRECT
```

**Question 28**
```
arranged | planned | explained | studied | discarded
arranged  planned    3
arranged  explained  0
arranged  studied    0
arranged  discarded  0
WordNet thinks that the answer is planned
CORRECT
```

**Question 29**
```
infinite | limitless | relative | unusual | structural
infinite  limitless   16
infinite  relative    0
infinite  unusual     0
infinite  structural  0
WordNet thinks that the answer is limitless
CORRECT
```

**Question 30**
```
showy | striking | prickly | entertaining | incidental
showy  striking      0
showy  prickly       0
showy  entertaining  0
showy  incidental    0
```

```
WordNet thinks that the answer is striking
4 answers tied [score = 0.25]
```

**Question 31**
```
levied | imposed | believed | requested | correlated
levied   imposed   16
levied   believed   0
levied   requested   0
levied   correlated   0
WordNet thinks that the answer is imposed
CORRECT
```

**Question 32**
```
deftly | skillfully | prudently | occasionally | humorously
deftly   skillfully   0
deftly   prudently   0
deftly   occasionally   0
deftly   humorously   0
WordNet thinks that the answer is skillfully
4 answers tied [score = 0.25]
```

**Question 33**
```
distribute | circulate | commercialize | research | acknowledge
distribute   circulate   16
distribute   commercialize   0
distribute   research   0
distribute   acknowledge   2
WordNet thinks that the answer is circulate
CORRECT
```

**Question 34**
```
discrepancies | differences | weights | deposits | wavelengths
['discrepancies' not in WordNet.] discrepancies   differences
['discrepancies' not in WordNet.] discrepancies   weights
['discrepancies' not in WordNet.] discrepancies   deposits
['discrepancies' not in WordNet.] discrepancies   wavelengths
NO ANSWER FOUND
```

**Question 35**
```
prolific | productive | serious | capable | promising
prolific   productive   16
prolific   serious   0
prolific   capable   0
prolific   promising   0
WordNet thinks that the answer is productive
CORRECT
```

**Question 36**
```
unmatched | unequaled | unrecognized | alienated | emulated
unmatched   unequaled   4
unmatched   unrecognized   0
unmatched   alienated   0
unmatched   emulated   0
WordNet thinks that the answer is unequaled
CORRECT
```

**Question 37**
```
peculiarly | uniquely | partly | patriotically | suspiciously
peculiarly   uniquely   0
peculiarly   partly   0
peculiarly   patriotically   0
peculiarly   suspiciously   0
WordNet thinks that the answer is uniquely
4 answers tied [score = 0.25]
```

**Question 38**
```
hue | color | glare | contrast | scent
hue   color   6
hue   glare   2
hue   contrast   0
hue   scent   3
WordNet thinks that the answer is color
CORRECT
```

**Question 39**
```
hind | rear | curved | muscular |hairy
hind   rear   3
```

```
hind  curved  0
hind  muscular  0
hind  hairy  0
WordNet thinks that the answer is rear
CORRECT
```

**Question 40**
```
highlight | accentuate | alter | imitate | restore
highlight  accentuate  6
highlight  alter  0
highlight  imitate  0
highlight  restore  0
WordNet thinks that the answer is accentuate
CORRECT
```

**Question 41**
```
hastily | hurriedly | shrewdly | habitually | chronologically
hastily  hurriedly  16
hastily  shrewdly  0
hastily  habitually  0
hastily  chronologically  0
WordNet thinks that the answer is hurriedly
CORRECT
```

**Question 42**
```
temperate | mild | cold | short | windy
temperate  mild  16
temperate  cold  4
temperate  short  0
temperate  windy  0
WordNet thinks that the answer is mild
CORRECT
```

**Question 43**
```
grin | smile | exercise | rest | joke
grin  smile  16
grin  exercise  0
grin  rest  0
grin  joke  3
WordNet thinks that the answer is smile
CORRECT
```

**Question 44**
```
verbally | orally | overtly | fittingly | verbosely
verbally  orally  0
verbally  overtly  0
verbally  fittingly  0
verbally  verbosely  0
WordNet thinks that the answer is orally
4 answers tied [score = 0.25]
```

**Question 45**
```
physician | doctor | chemist | pharmacist | nurse
physician  doctor  16
physician  chemist  4
physician  pharmacist  4
physician  nurse  4
WordNet thinks that the answer is doctor
CORRECT
```

**Question 46**
```
essentially | basically | possibly | eagerly | ordinarily
essentially  basically  16
essentially  possibly  0
essentially  eagerly  0
essentially  ordinarily  0
WordNet thinks that the answer is basically
CORRECT
```

**Question 47**
```
keen | sharp | useful | simple | famous
keen  sharp  16
keen  useful  0
keen  simple  4
keen  famous  0
WordNet thinks that the answer is sharp
CORRECT
```

**Question 48**
```
situated | positioned | rotating | isolated | emptying
situated  positioned  5
situated  rotating  2
situated  isolated  3
situated  emptying  0
WordNet thinks that the answer is positioned
CORRECT
```

**Question 49**
```
principal | major | most | numerous | exceptional
principal  major  0
principal  most  0
principal  numerous  0
principal  exceptional  0
WordNet thinks that the answer is major
4 answers tied [score = 0.25]
```

**Question 50**
```
slowly | gradually | rarely | effectively | continuously
slowly  gradually  0
slowly  rarely  0
slowly  effectively  0
slowly  continuously  0
WordNet thinks that the answer is gradually
4 answers tied [score = 0.25]
```

**Question 51**
```
built | constructed | proposed | financed | organized
built  constructed  16
built  proposed  0
built  financed  0
built  organized  5
WordNet thinks that the answer is constructed
CORRECT
```

**Question 52**
```
tasks | jobs | customers | materials | shops
tasks  jobs  16
tasks  customers  0
tasks  materials  0
tasks  shops  0
WordNet thinks that the answer is jobs
CORRECT
```

**Question 53**
```
unlikely | improbable | disagreeable | different | unpopular
unlikely  improbable  16
unlikely  disagreeable  0
unlikely  different  0
unlikely  unpopular  0
WordNet thinks that the answer is improbable
CORRECT
```

**Question 54**
```
halfheartedly | apathetically | customarily | bipartisanly | unconventionally
['halfheartedly' not in WordNet.] halfheartedly  apathetically
['halfheartedly' not in WordNet.] halfheartedly  customarily
['halfheartedly' not in WordNet.] halfheartedly  bipartisanly
['halfheartedly' not in WordNet.] halfheartedly  unconventionally
NO ANSWER FOUND
```

**Question 55**
```
annals | chronicles | homes | trails | songs
annals  chronicles  4
annals  homes  0
annals  trails  0
annals  songs  0
WordNet thinks that the answer is chronicles
CORRECT
```

**Question 56**
```
wildly | furiously | distinctively | mysteriously | abruptly
wildly  furiously  0
```

```
wildly   distinctively  0
wildly   mysteriously  0
wildly   abruptly  0
WordNet thinks that the answer is furiously
4 answers tied [score = 0.25]
```

**Question 57**
```
hailed | acclaimed | judged | remembered | addressed
hailed   acclaimed  16
hailed   judged  4
hailed   remembered  0
hailed   addressed  6
WordNet thinks that the answer is acclaimed
CORRECT
```

**Question 58**
```
command | mastery | observation | love | awareness
command  mastery  16
command  observation  2
command  love  0
command  awareness  2
WordNet thinks that the answer is mastery
CORRECT
```

**Question 59**
```
concocted | devised | cleaned | requested | supervised
concocted  devised  5
concocted  cleaned  4
concocted  requested  0
concocted  supervised  0
WordNet thinks that the answer is devised
CORRECT
```

**Question 60**
```
prospective | potential | particular | prudent | prominent
prospective  potential  16
prospective  particular  0
prospective  prudent  0
prospective  prominent  0
WordNet thinks that the answer is potential
CORRECT
```

**Question 61**
```
generally | broadly | descriptively | controversially | accurately
generally  broadly  16
generally  descriptively  0
generally  controversially  0
generally  accurately  0
WordNet thinks that the answer is broadly
CORRECT
```

**Question 62**
```
sustained | prolonged | refined | lowered | analyzed
sustained  prolonged  16
sustained  refined  0
sustained  lowered  2
sustained  analyzed  0
WordNet thinks that the answer is prolonged
CORRECT
```

**Question 63**
```
perilous | dangerous | binding | exciting | offensive
perilous  dangerous  16
perilous  binding  0
perilous  exciting  0
perilous  offensive  0
WordNet thinks that the answer is dangerous
CORRECT
```

**Question 64**
```
tranquillity | peacefulness | harshness | weariness | happiness
tranquillity  peacefulness  4
tranquillity  harshness  2
tranquillity  weariness  3
tranquillity  happiness  4
WordNet thinks that the answer is peacefulness
2 answers tied [score = 0.5]
```

**Question 65**
```
dissipate | disperse | isolate | disguise | photograph
dissipate   disperse   16
dissipate   isolate    0
dissipate   disguise   0
dissipate   photograph 0
WordNet thinks that the answer is disperse
CORRECT
```

**Question 66**
```
primarily | chiefly | occasionally | cautiously | consistently
primarily   chiefly      16
primarily   occasionally 0
primarily   cautiously   0
primarily   consistently 0
WordNet thinks that the answer is chiefly
CORRECT
```

**Question 67**
```
colloquial | conversational | recorded | misunderstood | incorrect
colloquial   conversational 16
colloquial   recorded       0
colloquial   misunderstood  0
colloquial   incorrect      0
WordNet thinks that the answer is conversational
CORRECT
```

**Question 68**
```
resolved | settled | publicized | forgotten | examined
resolved   settled    16
resolved   publicized 0
resolved   forgotten  0
resolved   examined   0
WordNet thinks that the answer is settled
CORRECT
```

**Question 69**
```
feasible | possible | permitted | equitable | evident
feasible   possible  16
feasible   permitted 0
feasible   equitable 0
feasible   evident   0
WordNet thinks that the answer is possible
CORRECT
```

**Question 70**
```
expeditiously | rapidly | frequently | actually | repeatedly
expeditiously   rapidly    0
expeditiously   frequently 0
expeditiously   actually   0
expeditiously   repeatedly 0
WordNet thinks that the answer is rapidly
4 answers tied [score = 0.25]
```

**Question 71**
```
percentage | proportion | volume | sample | profit
percentage   proportion 4
percentage   volume     0
percentage   sample     0
percentage   profit     4
WordNet thinks that the answer is proportion
2 answers tied [score = 0.5]
```

**Question 72**
```
terminated | ended | posed | postponed | evaluated
terminated   ended     16
terminated   posed     4
terminated   postponed 0
terminated   evaluated 0
WordNet thinks that the answer is ended
CORRECT
```

**Question 73**
```
uniform | alike | hard | complex | sharp
uniform   alike 0
uniform   hard  0
```

```
uniform  complex  2
uniform  sharp  0
WordNet thinks that the answer is complex
INCORRECT
```

**Question 74**
```
figure | solve | list | divide | express
figure   solve  4
figure   list  0
figure   divide  4
figure   express  0
WordNet thinks that the answer is solve
2 answers tied [score = 0.5]
```

**Question 75**
```
sufficient | enough | recent | physiological | valuable
sufficient   enough  16
sufficient   recent  0
sufficient   physiological  0
sufficient   valuable  0
WordNet thinks that the answer is enough
CORRECT
```

**Question 76**
```
fashion | manner | ration | fathom | craze
fashion   manner  16
fashion   ration  0
fashion   fathom  0
fashion   craze  4
WordNet thinks that the answer is manner
CORRECT
```

**Question 77**
```
marketed | sold | frozen | sweetened | diluted
marketed   sold  5
marketed   frozen  5
marketed   sweetened  5
marketed   diluted  4
WordNet thinks that the answer is sold
3 answers tied [score = 0.333333333333333]
```

**Question 78**
```
bigger | larger | steadier | closer | better
bigger   larger  16
bigger   steadier  0
bigger   closer  0
bigger   better  0
WordNet thinks that the answer is larger
CORRECT
```

**Question 79**
```
roots | origins | rituals | cure | function
roots   origins  0
roots   rituals  0
roots   cure  0
roots   function  0
WordNet thinks that the answer is origins
4 answers tied [score = 0.25]
```

**Question 80**
```
normally | ordinarily | haltingly | permanently | periodically
normally   ordinarily  16
normally   haltingly  0
normally   permanently  0
normally   periodically  0
WordNet thinks that the answer is ordinarily
CORRECT
```

**Total questions = 80, score = 62.3333333333333, correct = 57, ties = 18**
```
Number of problem words not found in WordNet: 2
Number of other words not found in WordNet: 2
Problem words not in WordNet: halfheartedly discrepancies
Other words not in WordNet: ceremonies salaries
```

## 2.B.  50 *ESL* Questions

**Question 1**
```
rusty | corroded | black | dirty | painted
rusty  corroded  0
rusty  black  3
rusty  dirty  0
rusty  painted  0
WordNet thinks that the answer is black
INCORRECT
```

**Question 2**
```
brass | metal | wood | stone | plastic
brass  metal  6
brass  wood  2
brass  stone  3
brass  plastic  0
WordNet thinks that the answer is metal
CORRECT
```

**Question 3**
```
spin | twirl | ache | sweat | flush
spin  twirl  16
spin  ache  0
spin  sweat  0
spin  flush  4
WordNet thinks that the answer is twirl
CORRECT
```

**Question 4**
```
passage | hallway | ticket | entrance | room
passage  hallway  5
passage  ticket  2
passage  entrance  4
passage  room  2
WordNet thinks that the answer is hallway
CORRECT
```

**Question 5**
```
yield | submit | challenge | boast | scorn
yield  submit  5
yield  challenge  0
yield  boast  0
yield  scorn  0
WordNet thinks that the answer is submit
CORRECT
```

**Question 6**
```
lean | rest | scrape | grate | refer
lean  rest  5
lean  scrape  2
lean  grate  0
lean  refer  3
WordNet thinks that the answer is rest
CORRECT
```

**Question 7**
```
barrel | cask | bottle | box | case
barrel  cask  16
barrel  bottle  5
barrel  box  5
barrel  case  5
WordNet thinks that the answer is cask
CORRECT
```

**Question 8**
```
nuisance | pest | garbage | relief | troublesome
nuisance  pest  3
nuisance  garbage  0
nuisance  relief  2
nuisance  troublesome  0
WordNet thinks that the answer is pest
CORRECT
```

**Question 9**
```
rug | carpet | sofa | ottoman | hallway
rug  carpet  16
```

```
rug  sofa  3
rug  ottoman  3
rug  hallway  0
WordNet thinks that the answer is carpet
CORRECT
```

**Question 10**
```
tap | drain | boil | knock | rap
tap  drain  3
tap  boil  4
tap  knock  16
tap  rap  16
WordNet thinks that the answer is knock
INCORRECT
```

**Question 11**
```
split | divided | crushed | grated | bruised
split  divided  16
split  crushed  5
split  grated  4
split  bruised  3
WordNet thinks that the answer is divided
CORRECT
```

**Question 12**
```
lump | chunk | stem | trunk | limb
lump  chunk  16
lump  stem  0
lump  trunk  3
lump  limb  2
WordNet thinks that the answer is chunk
CORRECT
```

**Question 13**
```
outline | contour | pair | blend | block
outline  contour  4
outline  pair  0
outline  blend  3
outline  block  3
WordNet thinks that the answer is contour
CORRECT
```

**Question 14**
```
swear | vow | explain | think | describe
swear  vow  4
swear  explain  4
swear  think  3
swear  describe  0
WordNet thinks that the answer is vow
2 answers tied [score = 0.5]
```

**Question 15**
```
relieved | rested | sleepy | tired | hasty
relieved  rested  0
relieved  sleepy  0
relieved  tired  3
relieved  hasty  0
WordNet thinks that the answer is tired
INCORRECT
```

**Question 16**
```
deserve | merit | need | want | expect
deserve  merit  16
deserve  need  5
deserve  want  5
deserve  expect  0
WordNet thinks that the answer is merit
CORRECT
```

**Question 17**
```
haste | a hurry | anger | ear | spite
['a hurry' not in WordNet.] haste  a hurry
haste  anger  0
haste  ear  0
haste  spite  0
WordNet thinks that the answer is anger
INCORRECT
```

K - 33

```
Question 18
stiff | firm | dark | drunk | cooked
stiff  firm   3
stiff  dark   0
stiff  drunk  16
stiff  cooked 0
WordNet thinks that the answer is drunk
INCORRECT

Question 19
verse | section | weed | twig | branch
verse  section  4
verse  weed  0
verse  twig  0
verse  branch  0
WordNet thinks that the answer is section
CORRECT

Question 20
steep | sheer | bare | rugged | stone
steep  sheer  16
steep  bare  0
steep  rugged  0
steep  stone  2
WordNet thinks that the answer is sheer
CORRECT

Question 21
envious | jealous | enthusiastic  | hurt | relieved
envious  jealous  16
envious  enthusiastic   0
envious  hurt  0
envious  relieved  0
WordNet thinks that the answer is jealous
CORRECT

Question 22
paste | dough | syrup | block | jelly
paste  dough  0
paste  syrup  3
paste  block  0
paste  jelly  3
WordNet thinks that the answer is syrup
INCORRECT

Question 23
scorn | refuse | enjoy | avoid | plan
scorn  refuse  0
scorn  enjoy  0
scorn  avoid  0
scorn  plan  0
WordNet thinks that the answer is refuse
4 answers tied [score = 0.25]

Question 24
refer | direct | call | carry | explain
refer  direct  4
refer  call  2
refer  carry  4
refer  explain  2
WordNet thinks that the answer is direct
2 answers tied [score = 0.5]

Question 25
limb | branch | bark | trunk | twig
limb  branch  16
limb  bark  6
limb  trunk  4
limb  twig  5
WordNet thinks that the answer is branch
CORRECT

Question 26
pad | cushion | board | block | tablet
pad  cushion  5
pad  board  3
```

```
pad   block  5
pad   tablet  16
WordNet thinks that the answer is tablet
INCORRECT
```

**Question 27**
```
boast | brag | yell | complain | explain
boast  brag  16
boast  yell  0
boast  complain  0
boast  explain  3
WordNet thinks that the answer is brag
CORRECT
```

**Question 28**
```
applause | approval | fear | shame | friends
applause  approval  4
applause  fear  0
applause  shame  0
applause  friends  0
WordNet thinks that the answer is approval
CORRECT
```

**Question 29**
```
sheet | leaf | book | block | tap
sheet  leaf  4
sheet  book  3
sheet  block  5
sheet  tap  6
WordNet thinks that the answer is tap
INCORRECT
```

**Question 30**
```
stem | stalk | bark | column | trunk
stem  stalk  16
stem  bark  5
stem  column  5
stem  trunk  5
WordNet thinks that the answer is stalk
CORRECT
```

**Question 31**
```
seize | take | refer | request | yield
seize  take  4
seize  refer  5
seize  request  0
seize  yield  3
WordNet thinks that the answer is refer
INCORRECT
```

**Question 32**
```
trunk | chest | bag | closet | swing
trunk  chest  4
trunk  bag  5
trunk  closet  4
trunk  swing  0
WordNet thinks that the answer is bag
INCORRECT
```

**Question 33**
```
weed | unwanted plant | cloth | animal | vegetable
['unwanted plant' not in WordNet.] weed  unwanted plant
weed  cloth  0
weed  animal  3
weed  vegetable  4
WordNet thinks that the answer is vegetable
INCORRECT
```

**Question 34**
```
approval | endorsement | gift | statement | confession
approval  endorsement  5
approval  gift  0
approval  statement  5
approval  confession  3
WordNet thinks that the answer is endorsement
2 answers tied [score = 0.5]
```

**Question 35**
```
mass | lump | service | worship | element
mass  lump     4
mass  service  5
mass  worship  3
mass  element  2
WordNet thinks that the answer is service
INCORRECT
```

**Question 36**
```
swing | sway | bounce | break | crash
swing  sway    16
swing  bounce   5
swing  break    5
swing  crash    4
WordNet thinks that the answer is sway
CORRECT
```

**Question 37**
```
sore | painful | red | hot | rough
sore  painful  16
sore  red       0
sore  hot       4
sore  rough     4
WordNet thinks that the answer is painful
CORRECT
```

**Question 38**
```
hinder | block | assist | relieve | yield
hinder  block    16
hinder  assist    0
hinder  relieve   0
hinder  yield     0
WordNet thinks that the answer is block
CORRECT
```

**Question 39**
```
sticky | gooey | smooth | shiny | wet
sticky  gooey    4
sticky  smooth   0
sticky  shiny    0
sticky  wet     16
WordNet thinks that the answer is wet
INCORRECT
```

**Question 40**
```
confession | statement | service | plea | bargain
confession  statement  4
confession  service    3
confession  plea       0
confession  bargain    0
WordNet thinks that the answer is statement
CORRECT
```

**Question 41**
```
weave | intertwine | print | stamp | shake
weave  intertwine  5
weave  print       4
weave  stamp       4
weave  shake       4
WordNet thinks that the answer is intertwine
CORRECT
```

**Question 42**
```
saucer | dish | box | frisbee | can
saucer  dish    16
saucer  box      4
saucer  frisbee  5
saucer  can      4
WordNet thinks that the answer is dish
CORRECT
```

**Question 43**
```
substance | thing | posture | level | score
substance  thing    6
substance  posture  3
substance  level    3
```

```
substance  score  2
WordNet thinks that the answer is thing
CORRECT
```

**Question 44**
```
firmly | steadfastly | reluctantly | sadly | hopefully
firmly  steadfastly  16
firmly  reluctantly  0
firmly  sadly  0
firmly  hopefully  0
WordNet thinks that the answer is steadfastly
CORRECT
```

**Question 45**
```
twist | intertwine | clip | fasten | curl
twist  intertwine  4
twist  clip  3
twist  fasten  4
twist  curl  6
WordNet thinks that the answer is curl
INCORRECT
```

**Question 46**
```
scrape | grate | chop | mince | slice
scrape  grate  16
scrape  chop  4
scrape  mince  3
scrape  slice  5
WordNet thinks that the answer is grate
CORRECT
```

**Question 47**
```
grind | rub | slice | hit | tap
grind  rub  4
grind  slice  0
grind  hit  5
grind  tap  5
WordNet thinks that the answer is hit
INCORRECT
```

**Question 48**
```
swell | enlarge | move | curl | shake
swell  enlarge  4
swell  move  5
swell  curl  2
swell  shake  0
WordNet thinks that the answer is move
INCORRECT
```

**Question 49**
```
harvest | intake | stem | lump | split
harvest  intake  0
harvest  stem  0
harvest  lump  0
harvest  split  0
WordNet thinks that the answer is intake
4 answers tied [score = 0.25]
```

**Question 50**
```
approve | support | boast | scorn | anger
approve  support  4
approve  boast  0
approve  scorn  0
approve  anger  0
WordNet thinks that the answer is support
CORRECT
```

**Total questions = 50, score = 31, correct = 29, ties = 5**
```
Number of problem words not found in WordNet: 0
Number of other words not found in WordNet: 2
Problem words not in WordNet:
Other words not in WordNet: a hurry unwanted plant
```

## 2.C.  20 *RDWP* Questions – January 2000: Nature

```
Question 1
eddy | whirlpool | current | wave | wind
eddy   whirlpool  16
eddy   current   4
eddy   wave   4
eddy   wind    0
WordNet thinks that the answer is whirlpool
CORRECT

Question 2
bough | branch | barricade | shaded area | fallen tree
bough   branch  6
bough   barricade   0
['shaded area' not in WordNet.] bough   shaded area
['fallen tree ' not in WordNet.] bough   fallen tree
WordNet thinks that the answer is branch
CORRECT

Question 3
heath | overgrown open land | burned-over area | thin forest | pasture
['overgrown open land' not in WordNet.] heath  overgrown open land
['burned-over area' not in WordNet.] heath  burned-over area
['thin forest' not in WordNet.] heath  thin forest
heath   pasture   0
WordNet thinks that the answer is pasture
INCORRECT

Question 4
scud | run straight | move slowly | falter | vaporize
['run straight' not in WordNet.] scud   run straight
['move slowly' not in WordNet.] scud   move slowly
scud   falter  3
scud   vaporize   0
WordNet thinks that the answer is falter
INCORRECT

Question 5
williwaw | sudden windblast | rainsquall | songbird | meadow flower
['williwaw' not in WordNet.] williwaw  sudden windblast
['williwaw' not in WordNet.] williwaw  rainsquall
['williwaw' not in WordNet.] williwaw  songbird
['williwaw' not in WordNet.] williwaw  meadow flower
NO ANSWER FOUND

Question 6
verge | brink | middle path | bare ground | vantage point
verge   brink  16
['middle path' not in WordNet.] verge  middle path
['bare ground' not in WordNet.] verge  bare ground
verge   vantage point   0
WordNet thinks that the answer is brink
CORRECT

Question 7
dale | valley | retreat | shelter | plain
dale   valley  4
dale   retreat  0
dale   shelter  0
dale   plain  0
WordNet thinks that the answer is valley
CORRECT

Question 8
limpid | clear | still | flat | luminous
limpid   clear  16
limpid   still  0
limpid   flat  0
limpid   luminous   0
WordNet thinks that the answer is clear
CORRECT

Question 9
floe | floating ice | frozen stream | lump | driftwood
['floating ice' not in WordNet.] floe  floating ice
```

```
['frozen stream' not in WordNet.] floe  frozen stream
floe  lump  0
floe  driftwood  0
WordNet thinks that the answer is lump
INCORRECT
```

**Question 10**
```
cascade | waterfall | thunder | swift current | edge
cascade  waterfall  4
cascade  thunder  2
['swift current' not in WordNet.] cascade  swift current
cascade  edge  3
WordNet thinks that the answer is waterfall
CORRECT
```

**Question 11**
```
undulation | rise and fall | faint motion | ebb and flow | quivering
['rise and fall' not in WordNet.] undulation  rise and fall
['faint motion' not in WordNet.] undulation  faint motion
['ebb and flow' not in WordNet.] undulation  ebb and flow
undulation  quivering  0
WordNet thinks that the answer is quivering
INCORRECT
```

**Question 12**
```
crag | steep rock | headland | barren hill | niche
['steep rock' not in WordNet.] crag  steep rock
crag  headland  2
['barren hill' not in WordNet.] crag  barren hill
crag  niche  0
WordNet thinks that the answer is headland
INCORRECT
```

**Question 13**
```
truss | cluster of flowers | main stem | bunch of grass | fallen petals
['cluster of flowers' not in WordNet.] truss  cluster of flowers
['main stem' not in WordNet.] truss  main stem
['bunch of grass' not in WordNet.] truss  bunch of grass
['fallen petals ' not in WordNet.] truss  fallen petals
NO ANSWER FOUND
```

**Question 14**
```
slough | deep mire | quicksand | shower | erosion
['deep mire' not in WordNet.] slough  deep mire
slough  quicksand  0
slough  shower  0
slough  erosion  0
WordNet thinks that the answer is quicksand
INCORRECT
```

**Question 15**
```
lee | shelter | cove | grassland | riverbank
lee  shelter  0
lee  cove  0
lee  grassland  0
lee  riverbank  0
WordNet thinks that the answer is shelter
4 answers tied [score = 0.25]
```

**Question 16**
```
brackish | salty | dirty | rough | noisy
brackish  salty  4
brackish  dirty  0
brackish  rough  0
brackish  noisy  0
WordNet thinks that the answer is salty
CORRECT
```

**Question 17**
```
precipice | vertical rockface | wide gap | broken path | descent
['vertical rockface' not in WordNet.] precipice  vertical rockface
['wide gap' not in WordNet.] precipice  wide gap
['broken path' not in WordNet.] precipice  broken path
precipice  descent  3
WordNet thinks that the answer is descent
INCORRECT
```

**Question 18**
chasm | deep fissure | wide opening | mountain pass | series of falls
['deep fissure' not in WordNet.] chasm   deep fissure
['wide opening' not in WordNet.] chasm   wide opening
chasm  mountain pass  3
['series of falls ' not in WordNet.] chasm   series of falls
WordNet thinks that the answer is mountain pass
INCORRECT

**Question 19**
sediment | settles to the bottom | floats | holds together | covers rocks
['settles to the bottom' not in WordNet.] sediment   settles to the bottom
sediment  floats  2
['holds together' not in WordNet.] sediment   holds together
['covers rocks ' not in WordNet.] sediment   covers rocks
WordNet thinks that the answer is floats
INCORRECT

**Question 20**
torrent | violent flow | drift | swell | deep sound
['violent flow' not in WordNet.] torrent   violent flow
torrent   drift  0
torrent   swell  0
['deep sound' not in WordNet.] torrent   deep sound
WordNet thinks that the answer is drift
INCORRECT

**Total questions = 20, score = 7.25, correct = 7, ties = 1**
Number of problem words not found in WordNet: 1
Number of other words not found in WordNet: 33

Problem words not in WordNet: williwaw

Other words not in WordNet: covers rocks, wide opening, ebb and flow, settles to the bottom,
vertical rockface, fallen petals, violent flow, burned-over area, barren hill, bunch of grass,
deep fissure, main stem, bare ground, thin forest, wide gap, faint motion, overgrown open land,
rise and fall, floating ice, cluster of flowers, deep mire, middle path, frozen stream, steep
rock, run straight, broken path, holds together, deep sound, swift current, move slowly, series
of falls, fallen tree, shaded area

# Appendix L: A Lexical Chain Building Example

This appendix shows the step-by-step output of my lexical chain building program that uses the *ELKB*. St-Onge (1995) and Ellman (2000) also use this text, attributed to Einstein, to demonstrate their lexical chain building systems.

**Step 1:** *Choose a Set of Thesaural Relations*

Chapter 5 presents the thesaural relations used by the *ELKB*.

**Step 2:** *Select a Set of Candidate Words*

> We **suppose** a very long **train travelling** along the **rails** with a **constant velocity** v and in the **direction** indicated in **Figure 1**. **People travelling** in this **train** will with **advantage** use the **train** as a rigid **reference-body**; they **regard** all **events** in **reference** to the **train**. Then every **event** which **takes** place along the **line** also **takes** place at a particular point of the **train**. Also, the **definition** of **simultaneity** can be given **relative** to the **train** in exactly the same way as with **respect** to the **embankment**.

**Step 3:** *Build All Proto-Chains for Each Candidate Word*

suppose, takes [score: 2.0, sense: 480, line: 1]
suppose, takes, takes [score: 3.0, sense: 480, line: 1]
suppose, regard [score: 2.0, sense: 485, line: 1]
suppose, regard, takes [score: 3.0, sense: 485, line: 1]
suppose, regard, takes, takes [score: 4.0, sense: 485, line: 1]
suppose, takes [score: 2.0, sense: 512, line: 1]
suppose, takes, takes [score: 3.0, sense: 512, line: 1]
suppose, takes [score: 2.0, sense: 510, line: 1]
suppose, takes, takes [score: 3.0, sense: 510, line: 1]

train, train [score: 2.0, sense: 534, line: 1]
train, train, train [score: 3.0, sense: 534, line: 1]
train, train, train, train [score: 4.0, sense: 534, line: 1]
train, train, train, train, takes [score: 5.0, sense: 534, line: 1]
train, train, train, train, takes, takes [score: 6.0, sense: 534, line: 1]
train, train, train, train, takes, takes, train [score: 7.0, sense: 534, line: 1]
train, train, train, train, takes, takes, train, train [score: 8.0, sense: 534, line: 1]
train, train [score: 2.0, sense: 536, line: 1]
train, train, train [score: 3.0, sense: 536, line: 1]
train, train, train, train [score: 4.0, sense: 536, line: 1]
train, train, train, train, takes [score: 5.0, sense: 536, line: 1]
train, train, train, train, takes, takes [score: 6.0, sense: 536, line: 1]
train, train, train, train, takes, takes, train [score: 7.0, sense: 536, line: 1]
train, train, train, train, takes, takes, train, train [score: 8.0, sense: 536, line: 1]
train, train [score: 2.0, sense: 284, line: 1]
train, train, train [score: 3.0, sense: 284, line: 1]

train, train, train, train [score: 4.0, sense: 284, line: 1]
train, train, train, train, train [score: 5.0, sense: 284, line: 1]
train, train, train, train, train, train [score: 6.0, sense: 284, line: 1]
train, train [score: 2.0, sense: 217, line: 1]
train, train, train [score: 3.0, sense: 217, line: 1]
train, train, train, train [score: 4.0, sense: 217, line: 1]
train, train, train, train, train [score: 5.0, sense: 217, line: 1]
train, train, train, train, train, train [score: 6.0, sense: 217, line: 1]
train, train [score: 2.0, sense: 267, line: 1]
train, train, train [score: 3.0, sense: 267, line: 1]
train, train, train, train [score: 4.0, sense: 267, line: 1]
train, train, train, train, train [score: 5.0, sense: 267, line: 1]
train, train, train, train, train, train [score: 6.0, sense: 267, line: 1]
train, rails [score: 2.0, sense: 274, line: 1]
train, rails, train [score: 3.0, sense: 274, line: 1]
train, rails, train, train [score: 4.0, sense: 274, line: 1]
train, rails, train, train, train [score: 5.0, sense: 274, line: 1]
train, rails, train, train, train, train [score: 6.0, sense: 274, line: 1]
train, rails, train, train, train, train, train [score: 7.0, sense: 274, line: 1]
train, train [score: 2.0, sense: 837, line: 1]
train, train, train [score: 3.0, sense: 837, line: 1]
train, train, train, train [score: 4.0, sense: 837, line: 1]
train, train, train, train, train [score: 5.0, sense: 837, line: 1]
train, train, train, train, train, train [score: 6.0, sense: 837, line: 1]
train, train [score: 2.0, sense: 268, line: 1]
train, train, train [score: 3.0, sense: 268, line: 1]
train, train, train, train [score: 4.0, sense: 268, line: 1]
train, train, train, train, train [score: 5.0, sense: 268, line: 1]
train, train, train, train, train, train [score: 6.0, sense: 268, line: 1]
train, train [score: 2.0, sense: 238, line: 1]
train, train, train [score: 3.0, sense: 238, line: 1]
train, train, train, train [score: 4.0, sense: 238, line: 1]
train, train, train, train, train [score: 5.0, sense: 238, line: 1]
train, train, train, train, train, train [score: 6.0, sense: 238, line: 1]
train, train [score: 2.0, sense: 357, line: 1]
train, train, train [score: 3.0, sense: 357, line: 1]
train, train, train, train [score: 4.0, sense: 357, line: 1]
train, train, train, train, train [score: 5.0, sense: 357, line: 1]
train, train, train, train, train, train [score: 6.0, sense: 357, line: 1]
train, train [score: 2.0, sense: 72, line: 1]
train, train, train [score: 3.0, sense: 72, line: 1]
train, train, train, train [score: 4.0, sense: 72, line: 1]
train, train, train, train, line [score: 5.0, sense: 72, line: 1]
train, train, train, train, line, train [score: 6.0, sense: 72, line: 1]
train, train, train, train, line, train, train [score: 7.0, sense: 72, line: 1]
train, train [score: 2.0, sense: 658, line: 1]
train, train, train [score: 3.0, sense: 658, line: 1]
train, train, train, train [score: 4.0, sense: 658, line: 1]
train, train, train, train, train [score: 5.0, sense: 658, line: 1]
train, train, train, train, train, train [score: 6.0, sense: 658, line: 1]
train, train [score: 2.0, sense: 461, line: 1]
train, train, train [score: 3.0, sense: 461, line: 1]
train, train, train, train [score: 4.0, sense: 461, line: 1]
train, train, train, train, train [score: 5.0, sense: 461, line: 1]

train, train, train, train, train, train [score: 6.0, sense: 461, line: 1]
train, train [score: 2.0, sense: 277, line: 1]
train, train, train [score: 3.0, sense: 277, line: 1]
train, train, train, train [score: 4.0, sense: 277, line: 1]
train, train, train, train, train [score: 5.0, sense: 277, line: 1]
train, train, train, train, train, train [score: 6.0, sense: 277, line: 1]
train, train [score: 2.0, sense: 742, line: 1]
train, train, train [score: 3.0, sense: 742, line: 1]
train, train, train, train [score: 4.0, sense: 742, line: 1]
train, train, train, train, train [score: 5.0, sense: 742, line: 1]
train, train, train, train, train, train [score: 6.0, sense: 742, line: 1]
train, train [score: 2.0, sense: 71, line: 1]
train, train, train [score: 3.0, sense: 71, line: 1]
train, train, train, train [score: 4.0, sense: 71, line: 1]
train, train, train, train, line [score: 5.0, sense: 71, line: 1]
train, train, train, train, line, train [score: 6.0, sense: 71, line: 1]
train, train, train, train, line, train, train [score: 7.0, sense: 71, line: 1]
train, train [score: 2.0, sense: 228, line: 1]
train, train, train [score: 3.0, sense: 228, line: 1]
train, train, train, train [score: 4.0, sense: 228, line: 1]
train, train, train, train, train [score: 5.0, sense: 228, line: 1]
train, train, train, train, train, train [score: 6.0, sense: 228, line: 1]
train, train [score: 2.0, sense: 273, line: 1]
train, train, train [score: 3.0, sense: 273, line: 1]
train, train, train, train [score: 4.0, sense: 273, line: 1]
train, train, train, train, train [score: 5.0, sense: 273, line: 1]
train, train, train, train, train, train [score: 6.0, sense: 273, line: 1]
train, train [score: 2.0, sense: 362, line: 1]
train, train, train [score: 3.0, sense: 362, line: 1]
train, train, train, train [score: 4.0, sense: 362, line: 1]
train, train, train, train, train [score: 5.0, sense: 362, line: 1]
train, train, train, train, train, train [score: 6.0, sense: 362, line: 1]
train, train [score: 2.0, sense: 441, line: 1]
train, train, train [score: 3.0, sense: 441, line: 1]
train, train, train, train [score: 4.0, sense: 441, line: 1]
train, train, train, train, train [score: 5.0, sense: 441, line: 1]
train, train, train, train, train, train [score: 6.0, sense: 441, line: 1]
train, rails [score: 2.0, sense: 624, line: 1]
train, rails, train [score: 3.0, sense: 624, line: 1]
train, rails, train, train [score: 4.0, sense: 624, line: 1]
train, rails, train, train, train [score: 5.0, sense: 624, line: 1]
train, rails, train, train, train, line [score: 6.0, sense: 624, line: 1]
train, rails, train, train, train, line, train [score: 7.0, sense: 624, line: 1]
train, rails, train, train, train, line, train, train [score: 8.0, sense: 624, line: 1]
train, rails, train, train, train, line, train, train, embankment [score: 9.0, sense: 624, line: 1]
train, train [score: 2.0, sense: 669, line: 1]
train, train, train [score: 3.0, sense: 669, line: 1]
train, train, train, train [score: 4.0, sense: 669, line: 1]
train, train, train, train, train [score: 5.0, sense: 669, line: 1]
train, train, train, train, train, train [score: 6.0, sense: 669, line: 1]
train, train [score: 2.0, sense: 67, line: 1]
train, train, train [score: 3.0, sense: 67, line: 1]
train, train, train, train [score: 4.0, sense: 67, line: 1]
train, train, train, train, train [score: 5.0, sense: 67, line: 1]

train, train, train, train, train, train [score: 6.0, sense: 67, line: 1]
train, train [score: 2.0, sense: 278, line: 1]
train, train, train [score: 3.0, sense: 278, line: 1]
train, train, train, train [score: 4.0, sense: 278, line: 1]
train, train, train, train, train [score: 5.0, sense: 278, line: 1]
train, train, train, train, train, train [score: 6.0, sense: 278, line: 1]
train, train [score: 2.0, sense: 288, line: 1]
train, train, train [score: 3.0, sense: 288, line: 1]
train, train, train, train [score: 4.0, sense: 288, line: 1]
train, train, train, train, train [score: 5.0, sense: 288, line: 1]
train, train, train, train, train, train [score: 6.0, sense: 288, line: 1]
train, train [score: 2.0, sense: 40, line: 1]
train, train, train [score: 3.0, sense: 40, line: 1]
train, train, train, train [score: 4.0, sense: 40, line: 1]
train, train, train, train, train [score: 5.0, sense: 40, line: 1]
train, train, train, train, train, train [score: 6.0, sense: 40, line: 1]
train, train [score: 2.0, sense: 369, line: 1]
train, train, train [score: 3.0, sense: 369, line: 1]
train, train, train, train [score: 4.0, sense: 369, line: 1]
train, train, train, train, train [score: 5.0, sense: 369, line: 1]
train, train, train, train, train, train [score: 6.0, sense: 369, line: 1]
train, train [score: 2.0, sense: 610, line: 1]
train, train, train [score: 3.0, sense: 610, line: 1]
train, train, train, train [score: 4.0, sense: 610, line: 1]
train, train, train, train, takes [score: 5.0, sense: 610, line: 1]
train, train, train, train, takes, takes [score: 6.0, sense: 610, line: 1]
train, train, train, train, takes, takes, train [score: 7.0, sense: 610, line: 1]
train, train, train, train, takes, takes, train, train [score: 8.0, sense: 610, line: 1]
train, train [score: 2.0, sense: 83, line: 1]
train, train, train [score: 3.0, sense: 83, line: 1]
train, train, train, train [score: 4.0, sense: 83, line: 1]
train, train, train, train, train [score: 5.0, sense: 83, line: 1]
train, train, train, train, train, train [score: 6.0, sense: 83, line: 1]
train, train [score: 2.0, sense: 689, line: 1]
train, train, train [score: 3.0, sense: 689, line: 1]
train, train, train, train [score: 4.0, sense: 689, line: 1]
train, train, train, train, train [score: 5.0, sense: 689, line: 1]
train, train, train, train, train, train [score: 6.0, sense: 689, line: 1]
train, train [score: 2.0, sense: 164, line: 1]
train, train, train [score: 3.0, sense: 164, line: 1]
train, train, train, train [score: 4.0, sense: 164, line: 1]
train, train, train, train, train [score: 5.0, sense: 164, line: 1]
train, train, train, train, train, train [score: 6.0, sense: 164, line: 1]

travelling, travelling [score: 2.0, sense: 265, line: 1]
travelling, travelling [score: 2.0, sense: 589, line: 1]
travelling, travelling [score: 2.0, sense: 981, line: 1]
travelling, travelling, takes [score: 3.0, sense: 981, line: 1]
travelling, travelling, takes, takes [score: 4.0, sense: 981, line: 1]
travelling, travelling [score: 2.0, sense: 75, line: 1]
travelling, direction [score: 2.0, sense: 271, line: 1]
travelling, direction, travelling [score: 3.0, sense: 271, line: 1]
travelling, travelling [score: 2.0, sense: 276, line: 1]
travelling, travelling [score: 2.0, sense: 295, line: 1]

travelling, travelling [score: 2.0, sense: 793, line: 1]
travelling, travelling [score: 2.0, sense: 152, line: 1]
travelling, travelling [score: 2.0, sense: 618, line: 1]
travelling, travelling [score: 2.0, sense: 145, line: 1]
travelling, travelling [score: 2.0, sense: 117, line: 1]
travelling, travelling [score: 2.0, sense: 59, line: 1]
travelling, travelling [score: 2.0, sense: 282, line: 1]
travelling, velocity [score: 2.0, sense: 465, line: 1]
travelling, velocity, travelling [score: 3.0, sense: 465, line: 1]
travelling, travelling [score: 2.0, sense: 269, line: 1]
travelling, travelling [score: 2.0, sense: 266, line: 1]
travelling, travelling [score: 2.0, sense: 298, line: 1]
travelling, travelling [score: 2.0, sense: 453, line: 1]
travelling, travelling [score: 2.0, sense: 194, line: 1]
travelling, travelling [score: 2.0, sense: 314, line: 1]
travelling, travelling [score: 2.0, sense: 84, line: 1]
travelling, travelling, rigid [score: 3.0, sense: 84, line: 1]
travelling, travelling [score: 2.0, sense: 305, line: 1]
travelling, travelling [score: 2.0, sense: 744, line: 1]

rails, respect [score: 1.75, sense: 924, line: 1]

constant, rigid [score: 2.0, sense: 494, line: 1]
constant, line [score: 2.0, sense: 16, line: 1]

direction, line [score: 2.0, sense: 693, line: 1]
direction, regard [score: 2.0, sense: 9, line: 1]
direction, regard, reference [score: 3.0, sense: 9, line: 1]
direction, regard, reference, respect [score: 4.0, sense: 9, line: 1]
direction, line [score: 2.0, sense: 281, line: 1]
direction, line [score: 2.0, sense: 220, line: 1]
direction, line [score: 2.0, sense: 547, line: 1]

advantage, line [score: 2.0, sense: 640, line: 2]
advantage, takes [score: 2.0, sense: 916, line: 2]
advantage, takes, takes [score: 3.0, sense: 916, line: 2]

regard, respect [score: 2.0, sense: 920, line: 2]
regard, respect [score: 2.0, sense: 887, line: 2]
regard, respect [score: 2.0, sense: 880, line: 2]
regard, respect [score: 2.0, sense: 768, line: 2]
regard, takes [score: 2.0, sense: 438, line: 2]
regard, takes, takes [score: 3.0, sense: 438, line: 2]
regard, reference [score: 2.0, sense: 10, line: 2]
regard, reference, respect [score: 3.0, sense: 10, line: 2]
regard, reference [score: 2.0, sense: 923, line: 2]
regard, reference, respect [score: 3.0, sense: 923, line: 2]
regard, respect [score: 2.0, sense: 866, line: 2]

events, event [score: 2.0, sense: 725, line: 2]
events, event [score: 2.0, sense: 526, line: 2]
events, event [score: 2.0, sense: 1, line: 2]
events, event [score: 2.0, sense: 124, line: 2]
events, event [score: 2.0, sense: 157, line: 2]

events, event [score: 2.0, sense: 590, line: 2]
events, event [score: 2.0, sense: 716, line: 2]
events, event [score: 2.0, sense: 167, line: 2]
events, event [score: 2.0, sense: 137, line: 2]
events, event [score: 2.0, sense: 8, line: 2]
events, event [score: 2.0, sense: 474, line: 2]
events, event [score: 2.0, sense: 616, line: 2]
events, event [score: 2.0, sense: 596, line: 2]
events, event [score: 2.0, sense: 154, line: 2]
events, event [score: 2.0, sense: 473, line: 2]

takes, takes [score: 2.0, sense: 18, line: 3]
takes, takes [score: 2.0, sense: 761, line: 3]
takes, takes [score: 2.0, sense: 583, line: 3]
takes, takes [score: 2.0, sense: 808, line: 3]
takes, takes [score: 2.0, sense: 86, line: 3]
takes, takes [score: 2.0, sense: 498, line: 3]
takes, takes [score: 2.0, sense: 74, line: 3]
takes, takes [score: 2.0, sense: 36, line: 3]
takes, takes [score: 2.0, sense: 714, line: 3]
takes, takes [score: 2.0, sense: 851, line: 3]
takes, takes, respect [score: 3.0, sense: 851, line: 3]
takes, takes [score: 2.0, sense: 825, line: 3]
takes, takes [score: 2.0, sense: 490, line: 3]
takes, takes [score: 2.0, sense: 828, line: 3]
takes, takes [score: 2.0, sense: 622, line: 3]
takes, takes [score: 2.0, sense: 148, line: 3]
takes, takes [score: 2.0, sense: 791, line: 3]
takes, takes [score: 2.0, sense: 859, line: 3]
takes, takes [score: 2.0, sense: 173, line: 3]
takes, takes [score: 2.0, sense: 20, line: 3]
takes, takes [score: 2.0, sense: 672, line: 3]
takes, takes [score: 2.0, sense: 959, line: 3]
takes, takes [score: 2.0, sense: 833, line: 3]
takes, takes [score: 2.0, sense: 963, line: 3]
takes, takes [score: 2.0, sense: 955, line: 3]
takes, takes [score: 2.0, sense: 296, line: 3]
takes, takes [score: 2.0, sense: 458, line: 3]
takes, takes [score: 2.0, sense: 673, line: 3]
takes, takes [score: 2.0, sense: 57, line: 3]
takes, takes [score: 2.0, sense: 638, line: 3]
takes, takes, respect [score: 3.0, sense: 638, line: 3]
takes, takes [score: 2.0, sense: 551, line: 3]
takes, takes [score: 2.0, sense: 37, line: 3]
takes, takes [score: 2.0, sense: 740, line: 3]
takes, takes [score: 2.0, sense: 660, line: 3]
takes, takes [score: 2.0, sense: 810, line: 3]
takes, takes [score: 2.0, sense: 198, line: 3]
takes, takes [score: 2.0, sense: 468, line: 3]
takes, takes [score: 2.0, sense: 204, line: 3]
takes, takes [score: 2.0, sense: 39, line: 3]
takes, takes [score: 2.0, sense: 163, line: 3]
takes, takes [score: 2.0, sense: 459, line: 3]
takes, takes [score: 2.0, sense: 706, line: 3]

takes, takes [score: 2.0, sense: 310, line: 3]
takes, takes [score: 2.0, sense: 619, line: 3]
takes, takes [score: 2.0, sense: 852, line: 3]
takes, takes [score: 2.0, sense: 671, line: 3]
takes, takes [score: 2.0, sense: 712, line: 3]
takes, takes [score: 2.0, sense: 900, line: 3]
takes, takes [score: 2.0, sense: 187, line: 3]
takes, takes [score: 2.0, sense: 708, line: 3]
takes, takes [score: 2.0, sense: 788, line: 3]
takes, takes [score: 2.0, sense: 786, line: 3]
takes, takes [score: 2.0, sense: 831, line: 3]
takes, takes [score: 2.0, sense: 188, line: 3]
takes, takes [score: 2.0, sense: 65, line: 3]
takes, takes [score: 2.0, sense: 508, line: 3]
takes, takes [score: 2.0, sense: 525, line: 3]
takes, takes [score: 2.0, sense: 542, line: 3]
takes, takes [score: 2.0, sense: 46, line: 3]
takes, takes [score: 2.0, sense: 745, line: 3]
takes, takes [score: 2.0, sense: 189, line: 3]
takes, takes [score: 2.0, sense: 823, line: 3]
takes, takes [score: 2.0, sense: 108, line: 3]
takes, takes [score: 2.0, sense: 192, line: 3]
takes, takes [score: 2.0, sense: 144, line: 3]
takes, takes [score: 2.0, sense: 721, line: 3]
takes, takes [score: 2.0, sense: 627, line: 3]
takes, takes [score: 2.0, sense: 682, line: 3]
takes, takes [score: 2.0, sense: 516, line: 3]
takes, takes [score: 2.0, sense: 915, line: 3]
takes, takes [score: 2.0, sense: 603, line: 3]
takes, takes [score: 2.0, sense: 891, line: 3]
takes, takes [score: 2.0, sense: 584, line: 3]
takes, takes [score: 2.0, sense: 78, line: 3]
takes, takes [score: 2.0, sense: 457, line: 3]
takes, takes [score: 2.0, sense: 308, line: 3]
takes, takes [score: 2.0, sense: 829, line: 3]
takes, takes [score: 2.0, sense: 304, line: 3]
takes, takes [score: 2.0, sense: 917, line: 3]
takes, takes [score: 2.0, sense: 858, line: 3]
takes, takes [score: 2.0, sense: 165, line: 3]
takes, takes [score: 2.0, sense: 910, line: 3]
takes, takes [score: 2.0, sense: 802, line: 3]
takes, takes [score: 2.0, sense: 172, line: 3]
takes, takes [score: 2.0, sense: 767, line: 3]
takes, takes [score: 2.0, sense: 370, line: 3]
takes, takes [score: 2.0, sense: 662, line: 3]
takes, takes [score: 2.0, sense: 311, line: 3]
takes, takes [score: 2.0, sense: 881, line: 3]
takes, takes [score: 2.0, sense: 773, line: 3]
takes, takes [score: 2.0, sense: 979, line: 3]
takes, takes [score: 2.0, sense: 704, line: 3]
takes, takes [score: 2.0, sense: 854, line: 3]
takes, takes, respect [score: 3.0, sense: 854, line: 3]
takes, takes [score: 2.0, sense: 586, line: 3]
takes, takes [score: 2.0, sense: 481, line: 3]

takes, takes [score: 2.0, sense: 664, line: 3]
takes, takes [score: 2.0, sense: 985, line: 3]
takes, takes [score: 2.0, sense: 211, line: 3]
takes, takes [score: 2.0, sense: 836, line: 3]
takes, takes [score: 2.0, sense: 895, line: 3]
takes, takes [score: 2.0, sense: 550, line: 3]
takes, takes [score: 2.0, sense: 986, line: 3]
takes, takes [score: 2.0, sense: 807, line: 3]
takes, takes [score: 2.0, sense: 855, line: 3]
takes, takes [score: 2.0, sense: 31, line: 3]
takes, takes [score: 2.0, sense: 178, line: 3]
takes, takes [score: 2.0, sense: 388, line: 3]
takes, line [score: 2.0, sense: 193, line: 3]
takes, line, takes [score: 3.0, sense: 193, line: 3]
takes, takes [score: 2.0, sense: 166, line: 3]
takes, takes [score: 2.0, sense: 667, line: 3]
takes, takes [score: 2.0, sense: 548, line: 3]
takes, takes [score: 2.0, sense: 229, line: 3]
takes, takes [score: 2.0, sense: 919, line: 3]
takes, takes [score: 2.0, sense: 988, line: 3]
takes, takes [score: 2.0, sense: 299, line: 3]
takes, takes [score: 2.0, sense: 38, line: 3]
takes, takes [score: 2.0, sense: 50, line: 3]
takes, takes [score: 2.0, sense: 56, line: 3]
takes, takes [score: 2.0, sense: 301, line: 3]
takes, takes [score: 2.0, sense: 885, line: 3]
takes, takes, respect [score: 3.0, sense: 885, line: 3]
takes, takes [score: 2.0, sense: 634, line: 3]
takes, takes [score: 2.0, sense: 958, line: 3]
takes, takes [score: 2.0, sense: 782, line: 3]
takes, takes [score: 2.0, sense: 882, line: 3]
takes, takes [score: 2.0, sense: 605, line: 3]
takes, takes [score: 2.0, sense: 889, line: 3]
takes, takes [score: 2.0, sense: 676, line: 3]

line, relative [score: 2.0, sense: 27, line: 3]
line, definition [score: 2.0, sense: 236, line: 3]

## Step 4: *Select the Best Proto-chain for Each Candidate Word*

train, rails, train, train, train, line, train, train, embankment [score: 9.0, sense: 624, line: 1]
direction, regard, reference, respect [score: 4.0, sense: 9, line: 1]
travelling, travelling, takes, takes [score: 4.0, sense: 981, line: 1]
suppose, regard, takes, takes [score: 4.0, sense: 485, line: 1]
regard, takes, takes [score: 3.0, sense: 438, line: 2]
advantage, takes, takes [score: 3.0, sense: 916, line: 2]
takes, takes, respect [score: 3.0, sense: 851, line: 3]
constant, rigid [score: 2.0, sense: 494, line: 1]
events, event [score: 2.0, sense: 725, line: 2]
line, relative [score: 2.0, sense: 27, line: 3]
rails, respect [score: 1.75, sense: 924, line: 1]

**Step 5:** *Select the Lexical Chains*

train, rails, train, train, train, line, train, train, embankment [score: 9.0, sense: 624, line: 1]
suppose, regard, takes, takes [score: 4.0, sense: 485, line: 1]
direction, reference, respect [score: 3.0, sense: 9, line: 1]
travelling, travelling [score: 2.0, sense: 981, line: 1]
constant, rigid [score: 2.0, sense: 494, line: 1]
events, event [score: 2.0, sense: 725, line: 2]

# Appendix M: The First Two Levels of the *WordNet 1.7.1* Noun Hierarchy

This appendix presents the 9 unique beginners of *WordNet 1.7.1* and the 161 first level hyponyms.

**entity, physical thing**
- ⇒ thing
- ⇒ causal agent, cause, causal agency
- ⇒ object, physical object
- ⇒ substance, matter
- ⇒ location
- ⇒ subject, content, depicted object
- ⇒ thing
- ⇒ imaginary place
- ⇒ anticipation
- ⇒ body of water, water
- ⇒ enclosure, natural enclosure
- ⇒ expanse
- ⇒ inessential, nonessential
- ⇒ necessity, essential, requirement, requisite, necessary
- ⇒ part, piece
- ⇒ sky
- ⇒ unit, building block
- ⇒ variable

**psychological feature**
- ⇒ cognition, knowledge, noesis
- ⇒ motivation, motive, need
- ⇒ feeling

**abstraction**
- ⇒ time
- ⇒ space
- ⇒ attribute
- ⇒ relation
- ⇒ measure, quantity, amount, quantum
- ⇒ set

**state**
- ⇒ skillfulness
- ⇒ cognitive state, state of mind
- ⇒ cleavage
- ⇒ medium
- ⇒ ornamentation
- ⇒ condition
- ⇒ condition, status
- ⇒ conditionality
- ⇒ situation, state of affairs

⇒ relationship
⇒ relationship
⇒ tribalism
⇒ utopia
⇒ dystopia
⇒ wild, natural state, state of nature
⇒ isomerism
⇒ degree, level, stage, point
⇒ office, power
⇒ status, position
⇒ being, beingness, existence
⇒ nonbeing
⇒ death
⇒ employment, employ
⇒ unemployment
⇒ order
⇒ disorder
⇒ hostility, enmity, antagonism
⇒ conflict
⇒ illumination
⇒ freedom
⇒ representation, delegacy, agency
⇒ dependence, dependance, dependency
⇒ motion
⇒ motionlessness, stillness
⇒ dead letter, non-issue
⇒ action, activity, activeness
⇒ inaction, inactivity, inactiveness
⇒ temporary state
⇒ imminence, imminency, impendence, impendency, forthcomingness
⇒ readiness, preparedness, preparation
⇒ physiological state, physiological condition
⇒ kalemia
⇒ union, unification
⇒ maturity, matureness
⇒ immaturity, immatureness
⇒ grace, saving grace, state of grace
⇒ damnation, eternal damnation
⇒ omniscience
⇒ omnipotence
⇒ perfection, flawlessness, ne plus ultra
⇒ integrity, unity, wholeness
⇒ imperfection, imperfectness
⇒ receivership
⇒ ownership
⇒ obligation
⇒ end, destruction, death
⇒ revocation, annulment
⇒ sale
⇒ turgor
⇒ homozygosity

⇒ heterozygosity
⇒ polyvalence, polyvalency, multivalence, mulltivalency
⇒ utilization

**event**
⇒ might-have-been
⇒ nonevent
⇒ happening, occurrence, natural event
⇒ social event
⇒ miracle
⇒ migration
⇒ Fall

**act, human action, human activity**
⇒ action
⇒ nonaccomplishment, nonachievement
⇒ leaning
⇒ motivation, motivating
⇒ assumption
⇒ rejection
⇒ forfeit, forfeiture, sacrifice
⇒ activity
⇒ wear, wearing
⇒ judgment, judgement, assessment
⇒ production
⇒ stay
⇒ residency, residence, abidance
⇒ inactivity
⇒ hindrance, interference
⇒ stop, stoppage
⇒ group action
⇒ distribution
⇒ legitimation
⇒ waste, permissive waste
⇒ proclamation, promulgation
⇒ communication, communicating
⇒ speech act

**group, grouping**
⇒ arrangement
⇒ straggle
⇒ kingdom
⇒ biological group
⇒ community, biotic community
⇒ world, human race, humanity, humankind, human beings, humans, mankind, man
⇒ people
⇒ social group
⇒ collection, aggregation, accumulation, assemblage
⇒ edition
⇒ electron shell
⇒ ethnic group, ethnos

⇒ race
⇒ association
⇒ subgroup
⇒ sainthood
⇒ citizenry, people
⇒ population
⇒ multitude, masses, mass, hoi polloi, people
⇒ circuit
⇒ system
⇒ series
⇒ actinoid, actinide, actinon
⇒ rare earth, rare-earth element, lanthanoid, lanthanide, lanthanon
⇒ halogen

**possession**
⇒ property, belongings, holding, material possession
⇒ territory, dominion, territorial dominion, province
⇒ white elephant
⇒ transferred property, transferred possession
⇒ circumstances
⇒ assets
⇒ treasure
⇒ liabilities

**phenomenon**
⇒ natural phenomenon
⇒ levitation
⇒ metempsychosis, rebirth
⇒ consequence, effect, outcome, result, event, issue, upshot
⇒ luck, fortune, chance, hazard
⇒ luck, fortune
⇒ pulsation
⇒ process